\documentclass[pdflatex,sn-mathphys-num]{sn-jnl}

\usepackage{url} 
\usepackage{hyperref}
\usepackage{graphicx}%
\usepackage{multirow}%
\usepackage{amsmath,amssymb,amsfonts}%
\usepackage{amsthm}%
\usepackage{mathrsfs}%
\usepackage[title]{appendix}%
\usepackage{xcolor}%
\usepackage{textcomp}%
\usepackage{manyfoot}%
\usepackage{booktabs}%
\usepackage{subfig}%
\usepackage{colortbl}
\usepackage{algorithm}%
\usepackage{algorithmicx}%
\usepackage{algpseudocode}%
\usepackage{listings}%
\usepackage{array}

\theoremstyle{thmstyleone}%
\theoremstyle{thmstyletwo}%

\theoremstyle{thmstylethree}%

\begin{document}

\title[Article Title]{AQUA-Net: Adaptive Frequency Fusion and Illumination Aware Network for Underwater Image Enhancement}

\author*[1]{\fnm{Munsif} \sur{Ali}}
\email{ali.munsif@szn.it}
\author*[2]{\fnm{Najmul} \sur{Hassan}}
\email{nhassan@u-aizu.ac.jp}
\author[1]{\fnm{Lucia} \sur{Ventura}}
\email{lucia.ventura@szn.it}
\author[1]{\fnm{Davide} \sur{Di Bari}}
\email{davide.dibari@szn.it}
\author[1]{\fnm{Simonepietro} \sur{Canese}}
\email{simonepietro.canese@szn.it}
\author[2]{\fnm{Jungpil} \sur{Shin}}
\email{jpshin@u-aizu.ac.jp}
\affil*[1]{%
\orgname{Stazione Zoologica Anton Dohrn},
\orgaddress{
\street{Villa Comunale},
\city{Napoli},
\postcode{80121},
\country{Italy}
}}

\affil[2]{%
\orgdiv{School of Computer Science and Engineering},
\orgname{The University of Aizu},
\orgaddress{
\city{Aizuwakamatsu},
\state{Fukushima},
\country{Japan}
}}


\abstract{Underwater images often suffer from severe color distortion, low contrast, and a hazy appearance due to wavelength-dependent light absorption and scattering. Simultaneously, existing deep learning models exhibit high computational complexity and require a substantial number of parameters, which limits their practical deployment for real-time underwater applications. To address these challenges, this paper presents a novel underwater image enhancement model, called \textcolor{red}{\textbf{A}}daptive Fre\textcolor{red}{\textbf{q}}uency Fusion and Ill\textcolor{red}{\textbf{u}}mination \textcolor{red}{\textbf{A}}ware \textcolor{red}{\textbf{Net}}work (AQUA-Net). It integrates a hierarchical residual encoder–decoder with dual auxiliary branches, which operate in the frequency and illumination domains. The frequency fusion encoder enriches spatial representations with frequency cues from the Fourier domain and preserves fine textures and structural details. 
Inspired by Retinex theory, the illumination-aware branch explicitly learns a spatial illumination map under paired-reference supervision and refines it through gamma correction before incorporating it into the decoder for spatially adaptive reconstruction. This joint spatial, frequency, and illumination design enables the model to effectively restore color balance, visual contrast, and perceptual realism under diverse underwater lighting conditions. Additionally, we present a high-resolution, real-world underwater video-derived dataset from the Mediterranean Sea, which captures challenging deep-sea conditions with realistic visual degradations to enable robust evaluation and development of deep learning models.
Extensive experiments on multiple benchmark datasets show that AQUA-Net performs on par with state-of-the-art methods in both qualitative and quantitative evaluations while using less number of parameters. Ablation studies further confirm that the frequency and illumination branches provide complementary contributions that improve visibility and color representation. Overall, the proposed model shows strong generalization capability and robustness, and it provides an effective solution for real-world underwater imaging applications. The code for the proposed model is available at: 
\href{https://munsifali11.github.io/AQUA-Net_Project/}{AQUA-Net}.}




\maketitle

\section{Introduction}\label{sec1}

underwater images (UWIs) are crucial for observing marine life and exploring complex ocean ecosystems. However, images captured in these environments often suffer significant image degradation. As light travels through water, it undergoes wavelength and distance-dependent absorption and scattering, leading to color degradation, reduced contrast, and the loss of important visual details\cite{akkaynak2017space, ullah2024diverse}. The suspended particles, varying water conditions, and irregular optical properties introduce color shifts, reduced contrast, and worsen visibility. These effects vary with water conditions and the irregular optical properties of the underwater environment, making underwater image enhancement (UIE)  a challenging task. Obtaining clean, visually reliable UWIs is crucial for improving image quality, visibility, and enabling accurate observation and analysis. To address these challenges, many researchers developed different UIE models, such as the physical basis model and physically based free models \cite{huang2018shallow,hassan2021retinex}. Physics-based methods mainly aim to accurately estimate the medium transmission and other imaging parameters, such as background light, to reconstruct a clean image by inverting the underwater image formation model \cite{akkaynak2019sea}. Although these approaches can work well under certain conditions, their performance often becomes unstable and highly sensitive when dealing with complex or challenging underwater scenes. This difficulty arises because accurately estimating the medium transmission is essential, yet challenging. This is because the UWIs vary widely and are classified into ten classes based on the Jerlov water type \cite{jerlov1964optical,ullah2024diverse}, each with different optical properties. As a result, estimating underwater imaging parameters accurately becomes complicated for traditional-based methods, including the physics-based model and physics-based free model.

Recently, advanced deep neural networks have demonstrated remarkable performance on UIE and improved both quantitative metrics and perceptual quality \cite{peng2023u,qi2021underwater,li2019underwater,tang2022autoenhancer,fabbri2018enhancing,li2017watergan}.
Despite these gains, several of these approaches \cite{li2019underwater,fu2022uncertainty,peng2023u,wang2025optimized} are computationally complex and require a significantly large number of parameters and Floating Point Operations (FLOPs), which limit their practicality for real-world deployment. Additionally, existing architectures rely on generic encoder–decoder structures originally developed for natural-image tasks rather than underwater environments \cite{wang2025optimized,zhu2025new}. These models struggle to fully account for the unique spectral distortions, frequency-dependent degradation, and non-uniform illumination patterns found in underwater scenes. As a result, they often enhance images globally but remain limited in recovering fine textures, suppressing low-frequency haze, or reconstructing spatially consistent color distributions. This mismatch between model design and underwater imaging physics restricts their generalization capability and leads to inconsistent restoration across diverse water types.

\begin{figure}[!htp]
\centering
\subfloat[Raw input][\scriptsize Raw input]{%
    \includegraphics[width=1.5in,height=1.1in]{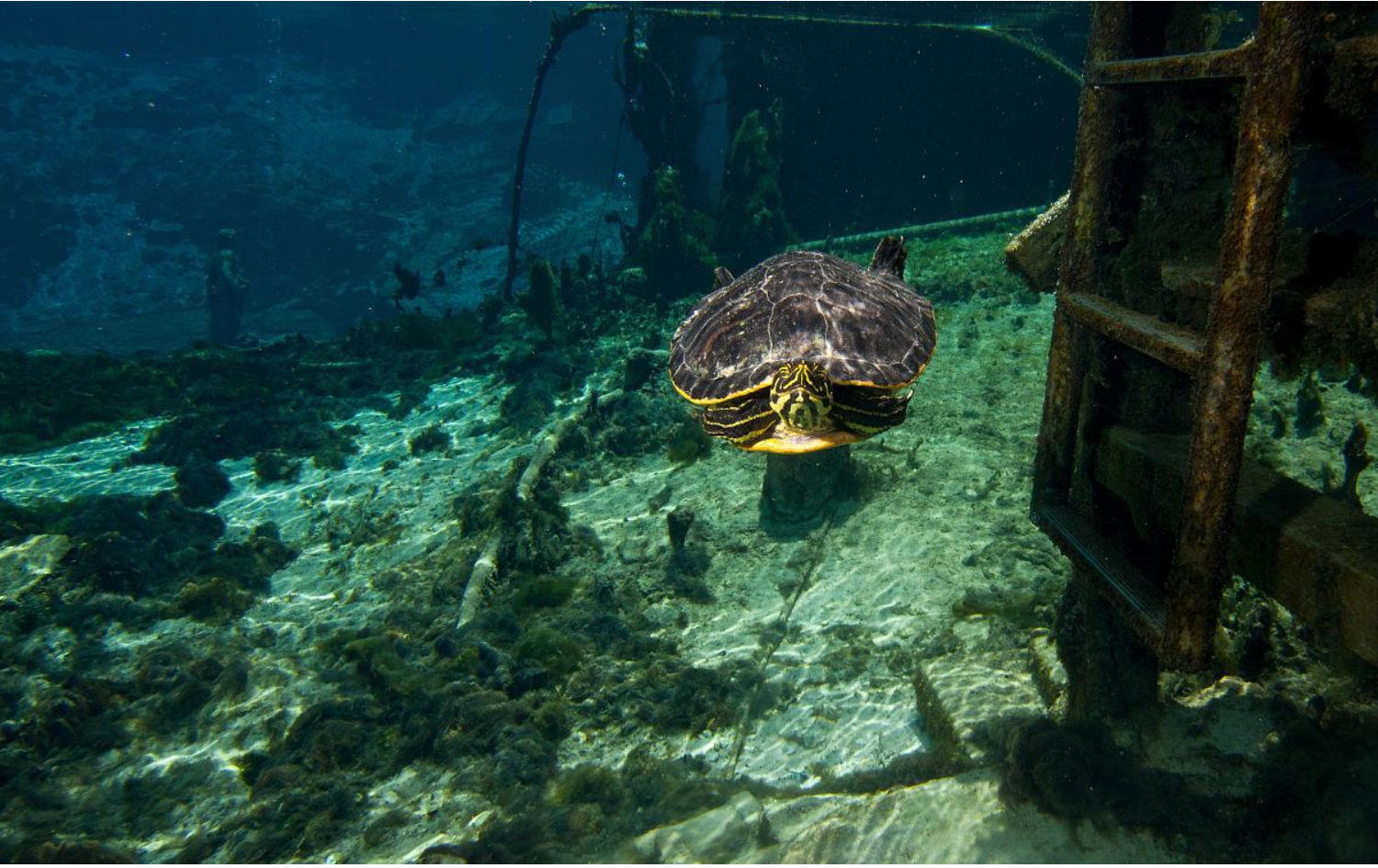} \label{fig:image1}
}\hspace{-0.05em}%
\subfloat[Encoder Decoder][\scriptsize Encoder Decoder]{%
    \includegraphics[width=1.5in,height=1.1in]{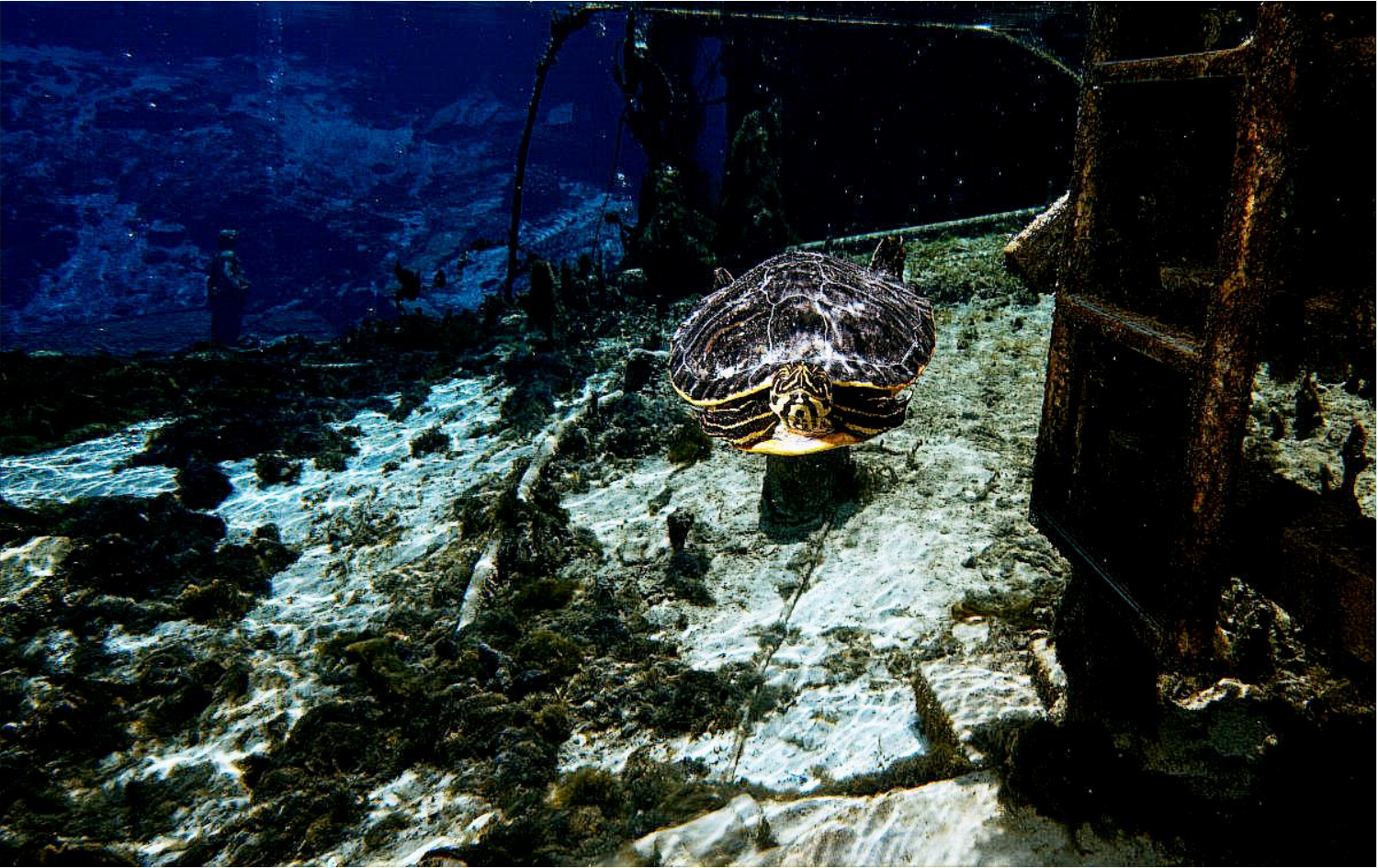} \label{fig:image2}
}\hspace{-0.05em}\\%
\subfloat[Frequency][\scriptsize Frequency]{%
    \includegraphics[width=1.5in,height=1.1in]{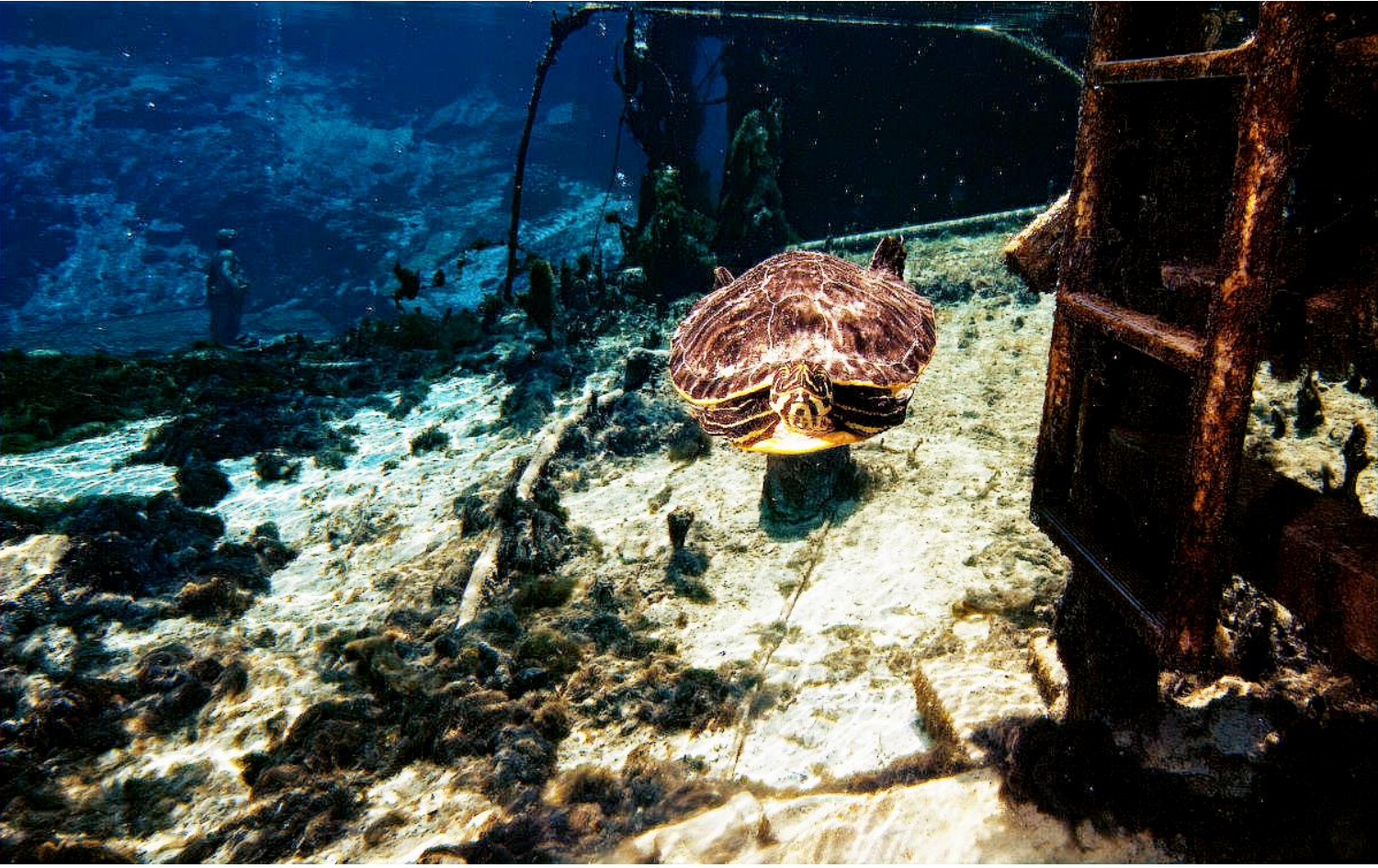} \label{fig:image3}
}\hspace{-0.05em}%
\subfloat[AQUA-Net][\scriptsize AQUA-Net]{%
    \includegraphics[width=1.5in,height=1.1in]{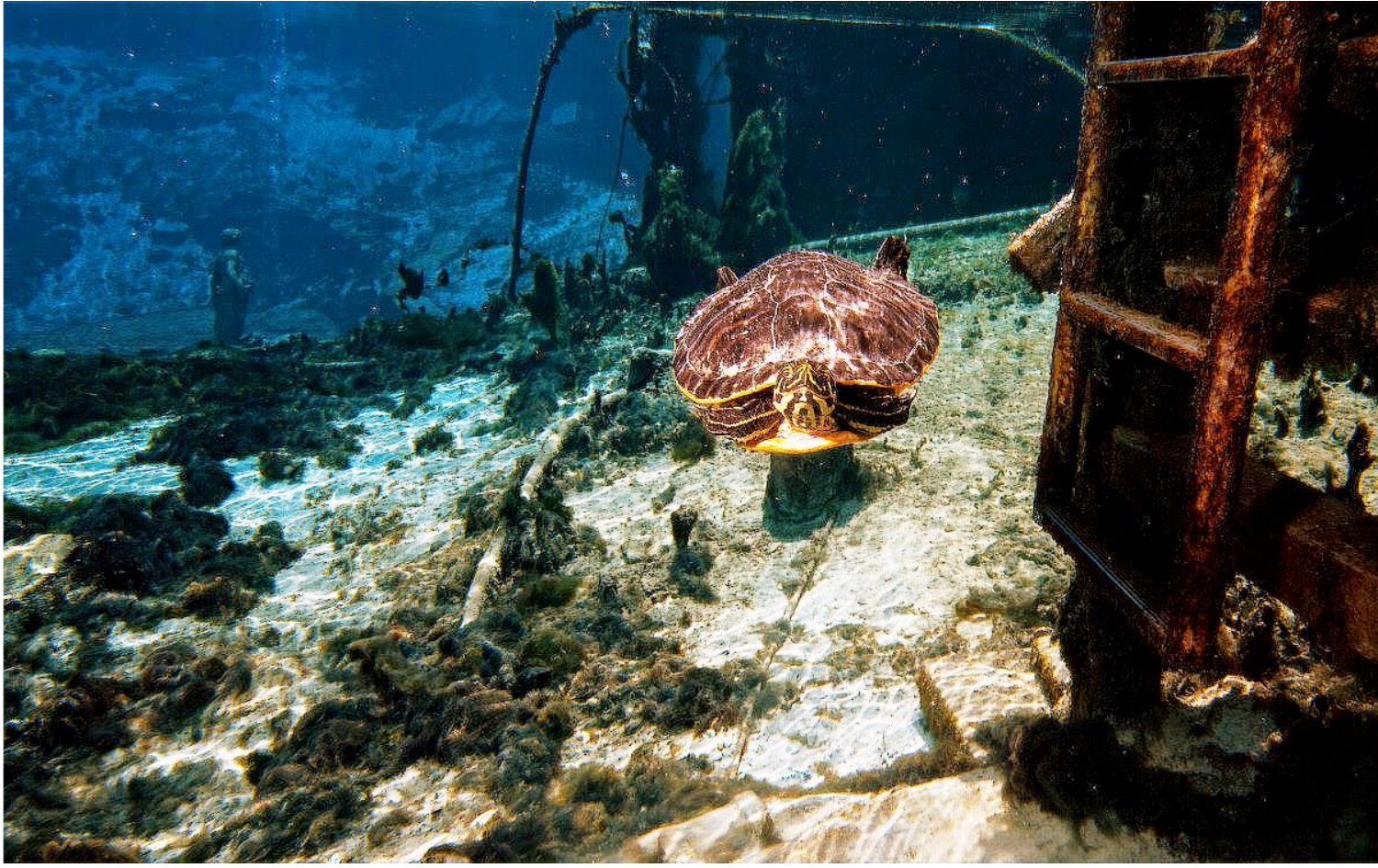} \label{fig:image4}
}%
\caption{Component-wise evaluation of AQUA-Net on an underwater image: (a) raw input, (b) encoder–decoder baseline, (c) encoder–decoder with an added frequency branch, and (d) full AQUA-Net combining global correction, frequency-guided enhancement, and illumination estimation.}
\label{fig: block_results}
\end{figure}

Underwater images suffer from degradation caused by wavelength-dependent light absorption and scattering, leading to illumination imbalance, color distortion, and loss of structural details \cite{li2017watergan}. Although existing deep learning and frequency-domain UIE methods can recover global appearance and fine textures \cite{wei2022uhd,cheng2024fdce}, they do not explicitly constrain spatially varying illumination, which can result in inconsistent brightness across regions with different depths and exposure conditions. This motivates us to introduce an illumination-aware branch into AQUA-Net. Inspired by Retinex-based illumination modeling and gamma correction \cite{li2018lightennet}, the branch explicitly estimates and corrects the illumination component, while paired reference images provide additional supervision for learning a meaningful illumination representation \cite{yan2025igdnet}. The corrected illumination is then incorporated into the decoder to provide spatial guidance during reconstruction. Thus, the proposed design complements frequency-based feature restoration by explicitly addressing illumination imbalance, enabling more consistent brightness, color recovery, and structural preservation in complex underwater scenes. To better understand these limitations, Figure~\ref{fig: block_results} presents a component-wise evaluation of our framework. The raw underwater inputs exhibit severe wavelength-dependent attenuation, color imbalance, and substantial loss of structural detail, as shown in Figure~\ref{fig:image1}. A conventional encoder–decoder network recovers part of the global illumination but remains ineffective at resolving complex color shifts or suppressing low-frequency scattering, resulting in visually inconsistent reconstructions as shown in Figure~\ref{fig:image2}. Incorporating a frequency decomposition branch improves edge sharpness and restores suppressed textures, yet it lacks the contextual awareness required to regulate low-frequency haze and stabilize global color correction, as shown in Figure~\ref{fig:image3}. Our complete architecture, AQUA-Net, unifies these complementary cues by combining a refined encoder–decoder backbone for global correction, a frequency-guided enhancement block to recover fine-scale structures, and an illumination estimation branch that stabilizes brightness across depth-varying regions, as depicted in Figure~\ref{fig:image4}. This coordinated design produces a more coherent and visually accurate reconstruction, recovering balanced colors, restoring scene contrast, and preserving high-frequency texture across diverse underwater conditions.

Moreover, underwater image and video analysis play a critical role in deep-sea exploration, ecological monitoring, robotic operations, and the evaluation of deep learning models. Existing benchmark datasets \cite{li2019underwater, islam2020fast, liu2020real} have enabled significant advances in UIE and restoration. However, they mainly focus on shallow-water or laboratory conditions and do not adequately represent the challenging environments encountered in real deep-sea operations. Many existing datasets provide limited coverage of depth, water conditions, and complex visual degradations, including variations in illumination, turbidity, low light, color attenuation, and back-scatter, making them insufficient to fully capture the diversity of real underwater environments \cite{folkman2025data, jian2024underwater, purnima2025devising}. To address this gap, we present a novel deep-sea video-derived dataset collected in the Mediterranean Sea, spanning depths from 108 m to 760 m across three locations. The dataset comprises high-resolution frames that capture realistic visual conditions and provides a challenging, ecologically valid testbed for evaluating underwater image analysis algorithms, including enhancement, denoising, and real-time models. The main contributions of this study are as follows:

\begin{itemize}
    \item An illumination-aware enhancement branch is introduced to estimate a spatially adaptive illumination map that guides the decoder, which enables effective correction of non-uniform lighting and depth-dependent color attenuation. Moreover, a frequency-guided enhancement module is developed to operate in the Fourier domain, which recovers frequency textures and injects frequency-refined features into the encoder to improve edge sharpness and structural clarity.
    \item A lightweight encoder–decoder architecture is constructed to fuse spatial, illumination, and frequency-domain cues through multi-scale residual modules and illumination-guided skip connections, providing robust enhancement across diverse underwater degradation conditions.
    \item This work introduces the DeepSea dataset, a high-resolution underwater dataset that captures real deep-sea conditions with realistic visual degradations. It serves as a testbed for evaluating deep learning models for real underwater image analysis.
    \item AQUA-Net’s performance is validated on multiple UIE benchmarks as well as on our own dataset through quality analyses and quantitative metrics. It shows comparable results to state-of-the-art (SOTA) approaches with less computational complexity.
\end{itemize}

\section{Literature Review}
The UIE methods are generally divided into two categories.\\
\textbf{(1) Traditional UIE Methods}:
The traditional UIE methods aimed to improve visual quality by directly adjusting pixel values. These methods typically focus on enhancing one or more visual properties such as contrast, brightness, or color balance \cite{hassan2021retinex,ancuti2017color,zhang2022underwater}. These UIE methods include dynamic range stretching, pixel distribution adjustment, histogram equalization, contrast enhancement, and white balance correction \cite{ghani2015underwater,iqbal2010enhancing}. Ancuti et al. \cite{ancuti2012enhancing}  first generated color-corrected and contrast-enhanced versions of UIWs, computed corresponding weight maps, and fused these results to combine the advantages of both versions. Later, Ancuti et al. \cite{ancuti2017color} improved this fusion-based strategy using a multiscale fusion approach, blending two image versions derived from a white-balancing algorithm. Further, Ancuti et al. \cite{ancuti2019color} introduced a color channel compensation preprocessing method to address severe color degradation under challenging conditions, such as underwater or hazy environments. The 3C operator restores lost color information in at least one channel, thereby improving the performance of traditional restoration methods. Additionally, other studies, such as Hitam et al. \cite{hitam2013mixture}, utilized Adaptive Histogram Equalization (AHE) and contrast adjustment in RGB and HSV color spaces to enhance contrast and reduce noise. Retinex-based approaches have also been explored; Fu et al. \cite{fu2014retinex} proposed a retinex-based model involving color correction, layer decomposition, and enhancement. 
While Hassan et. al \cite{hassan2021retinex} further improved the Retinex-based model, including Contrast-Limited-AHE (CLAHE) and a Retinex-based algorithm to correct color distortions by decomposing the image into reflectance and illumination components for color restoration. Finally, bilateral filtering is applied as post-processing to smooth noise and preserve edges. Zhang et al. \cite{zhang2025underwater} propose a hybrid UIE method that fuses spatial and frequency domain processing to restore color, enhance contrast, and achieve good results.

In addition to these UIE methods, physical model–based techniques attempt to address the inverse problem of underwater image degradation. They aim to model how clear images become distorted underwater by simulating light absorption and scattering processes \cite{xie2021variational,ullah2024diverse}. 
Peng et al. \cite{peng2018generalization} estimated ambient light and scene transmission by analyzing the difference between the observed intensity and ambient light, considering depth-related color shifts.
Samiullah et al. \cite {ullah2024diverse} propose an improved physical model called the Diverse Underwater Image Formation Model (DUIFM) to UIE  by better accounting for variations in optical properties across different water types. 

Despite significant progress, traditional and physical model–based UIE methods still face key challenges. Most of these methods depend on manually tuned parameters and handcrafted priors, making them sensitive to variations in lighting, depth, and water clarity. Physical-based models often oversimplify underwater light transmission, leading to incomplete color recovery and detail loss in turbid or low-visibility conditions.

\textbf{(2) Deep Learing UIE Methods}:
The advancement of deep learning-based methods shows a remarkable performance in the UIE \cite{li2019underwater,peng2023u,li2021underwater}. These methods use different strategies, including the Convolutional Neural Network (CNN) based approaches proposed by Li et al. \cite{li2019underwater}, such as Water-Net, a CNN-based UIE framework. They used three pre-processed steps of each input image obtained through white balance correction, gamma adjustment, and histogram equalization that are fused by a CNN that learns confidence maps to produce the final enhanced result. Similarly, Wang et al. \cite{wang2017deep}  proposed UIE-Net, an end-to-end CNN framework for UIE that jointly performs color correction and haze removal.
Li et al. \cite{li2020underwater} trained  UWCNN models, each tailored to specific underwater scenes, enabling real-time enhancement of underwater videos due to their lightweight design. Li et al. \cite{li2021underwater} developed the UColor network, which integrates multi-color space embedding. They employ an inverse transmission map as an attention mechanism, guiding the network to focus more on severely degraded regions for improved restoration quality. Fu et al. \cite{fu2022uncertainty} developed PUIE, a probabilistic network combining a variational autoencoder with a consensus process. Their method effectively handles reference map ambiguity and bias, resulting in robust enhancement performance comparable with existing methods. Similarly, Guo et al. \cite{guo2023underwater} introduced URanker, a ranking-based underwater image quality assessment model built on a convolutional attentional Transformer. They use histogram priors and cross-scale correspondence to assess global and local degradation, providing perceptual ranking supervision that significantly improves the performance of U-shaped UIE networks. For UIE tasks involving object detection, Liu et al. \cite{liu2022twin} introduced an object-guided twin adversarial contrastive learning method, which improves both image quality and detection accuracy in raw underwater scenes. Zhang et al. \cite{zhang2025underwater} later proposed a cascaded contrastive learning framework that progressively refines image quality through multi-level representation learning, achieving more consistent color and structure restoration than conventional single-stage networks. 

In addition, Wang et al. \cite{wang2025optimized} introduced OUNet-JL, an optimized UNet framework that integrates a multi-residual module, spatial multi-scale feature extraction with channel attention, and a strengthen-operate-subtract reconstruction module, supervised by a joint loss combining structural, perceptual, and total variation terms.
Most of the existing DL-based UIE methods still suffer from limited color correction, detail loss, and high computational cost, whereas our proposed frequency-guided encoder-decoder achieves more efficient and balanced enhancement results


\section{Proposed Method}
\begin{figure*}[t!]
\centering
\includegraphics[width=0.7\linewidth]{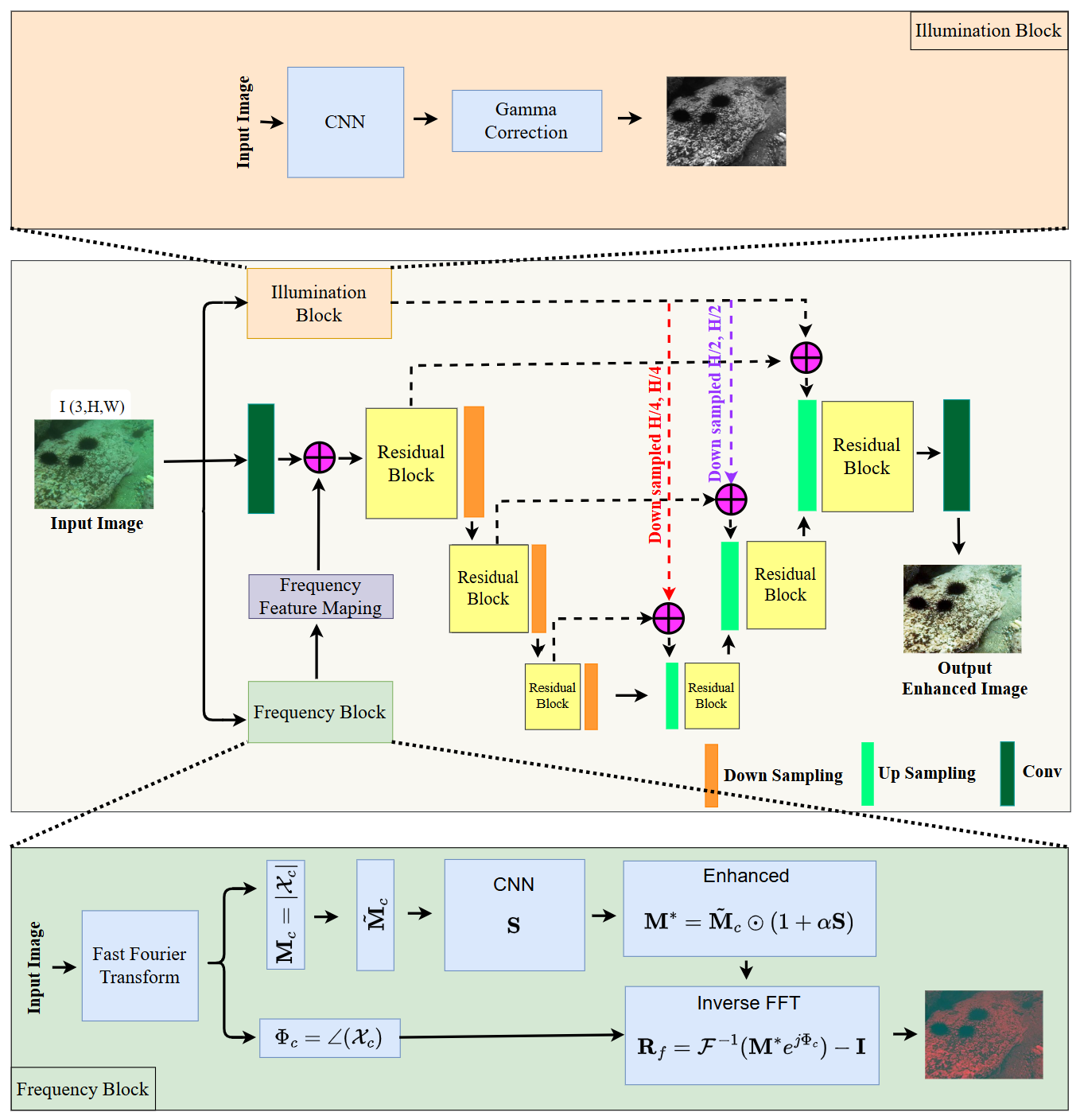}
\caption{\textbf{AQUA-Net architecture.} The network combines a frequency enhancement block, an illumination branch, and a multi-level encoder–decoder backbone with Residual Enhancement Modules (REMs). The frequency block output fuses with the encoder input. Skip connections integrate illumination information to enhance feature refinement and restore underwater image quality.}
\label{fig: model}
\end{figure*}

This section presents an overview of the proposed AQUA-Net model.
The proposed model employs a hierarchical residual encoder–decoder backbone and integrates two auxiliary modules: a frequency enhancement and an illumination block, as illustrated in Figure \ref{fig: model}. The network enhances underwater degraded images to improve textural details and correct illumination imbalances.
 
 The encoder, frequency, and illumination block concurrently process the input image. In the encoder, each stage employs a REM built from depthwise separable convolution \cite{howard2017mobilenets}, followed by down-sampling operations to extract multi-scale hierarchical features efficiently. The frequency branch transforms the input into the Fourier domain using the Fast Fourier Transform (FFT). It normalizes the magnitude component and adaptively refines it through convolution layers, while the phase attribute remains unchanged. It is because the phase spectrum preserves the overall semantic structure of the image \cite{yang2020fda, xu2021fourier, huang2021fsdr}.
The resulting frequency correction map, obtained by subtracting the inverse FFT from the input image, projects into the feature space and fuses with the encoder input to preserve fine textures and high-frequency details.
Simultaneously, the illumination block predicts a spatially varying illumination map that guides the decoder. During decoding, the illumination features are interpolated to match the resolution of the corresponding skip connections and combined with the encoder features, which enables adaptive compensation of lighting non-uniformity. The decoder progressively up-samples and refines the feature maps through residual enhancement modules. The final convolution layer reconstructs the enhanced images with improved clarity and contrast. 



\subsection{Residual Enhancement Module (REM)}
\begin{figure}[t!]
\centering
\includegraphics[width=0.6\linewidth]{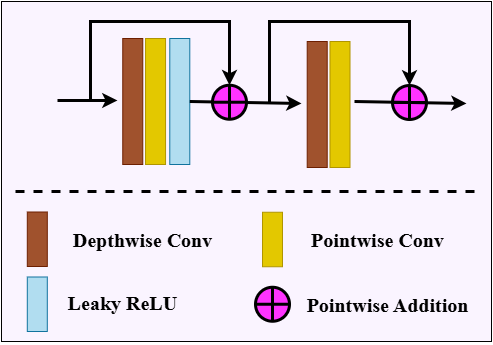}
\caption{Residual Enhancement Module (REM) structure. The REM consists of two depthwise separable convolution blocks followed by a point-wise convolution with Leaky ReLU activation only in the first stage. Each block performs element-wise addition with its input to form a residual connection.}
\label{fig: REM}
\end{figure}

The REM serves as the core computational unit of the encoder–decoder backbone, designed to refine features efficiently and maintain computational complexity. As shown in Fig. \ref{fig: REM}, each REM employs depthwise separable convolutions to decompose a standard convolution into spatial and channel-wise operations, which reduces parameters and computational cost \cite{howard2017mobilenets}. A Leaky ReLU activation introduces non-linearity and enhances the representation of subtle intensity variations common in underwater scenes. The inclusion of a residual connection enables the module to learn residual mappings, facilitates stable gradient flow, and preserves essential low-level details \cite{he2016deep}. Overall, the REM enhances feature refinement and texture preservation, and contributes to both the efficiency and performance of the proposed AQUA-Net model.

\subsection{Frequency Fusion Encoder}
The frequency fusion encoder integrates spatial and frequency-domain representations to enhance structural details and textural richness in the early stages of the network. Unlike \cite{cheng2024fdce}, the proposed frequency block is lightweight and adaptively enhances image features, and injects the frequency correction map at the start of the model. The input image first passes through a frequency enhancement block, which operates in the Fourier domain to amplify frequency components such as edges and fine details that often degrade in underwater conditions \cite{ata2021underwater, ata2022underwater, hou2009simple}. The resulting frequency correction map, obtained from the frequency block, projects into the feature space and fuses with the encoder input, so the network can exploit both spatial context and frequency-domain sharpness. Figure \ref{fig: model} shows the frequency block in the lower part of the architecture.

 Formally, the input image $\mathbf{I} \in \mathbb{R}^{B \times C \times H \times W}$ is transformed into the frequency domain  using the two-dimensional Fast Fourier Transform (FFT) \cite{xu2021fourier}:
\begin{equation}
\mathcal{X}_c(u,v) =
\frac{1}{\sqrt{HW}}
\sum_{h=0}^{H-1}\sum_{w=0}^{W-1}
\mathbf{I}_c(h,w)
e^{-j2\pi \left(\frac{uh}{H} + \frac{vw}{W}\right)},
\label{eq:fft}
\end{equation}
where $\mathcal{X}_c(u,v)$ denotes the complex spectrum of the $c$-th channel, $(H, W)$ are the spatial dimensions of the image, and $B$ is the batch size. The complex-valued frequency representation is decomposed into its magnitude and phase components as: $\mathbf{M}_c = |\mathcal{X}_c|$ and $\Phi_c = \angle(\mathcal{X}_c)$.
To stabilize spectral learning, the magnitude spectrum is normalized as:
\begin{equation}
    \tilde{\mathbf{M}}_c = \frac{\mathbf{M}_c}{\mu(\mathbf{M}_c) + \epsilon}, \qquad
    \mu(\mathbf{M}_c) = \frac{1}{HW}\sum_{h=0}^{H-1}\sum_{w=0}^{W-1} \mathbf{M}_c(h,w),
\end{equation}
where $\mu(\cdot)$ computes the spatial mean for each channel and $\epsilon$ is a small constant to prevent numerical instability.  
The normalized spectrum $\tilde{\mathbf{M}}$ is processed by a lightweight convolution network with weights $\mathbf{W}_1$ and $\mathbf{W}_2$ to generate an adaptive modulation map:
\begin{equation}
\mathbf{S} = 
\sigma_2\!\left(
\mathbf{W}_2 * 
\sigma_1\!\left(
\mathbf{W}_1 * \tilde{\mathbf{M}}_c
\right)\right),
\label{eq:freq_conv}
\end{equation}
where $*$ denotes convolution, and $\sigma_1(\cdot)$ and $\sigma_2(\cdot)$ represent non-linear activation functions. The enhanced magnitude spectrum is then computed as:
\begin{equation}
    \mathbf{M}^* = \tilde{\mathbf{M}}_c \odot (1 + \alpha \mathbf{S}),
\end{equation}
where $\alpha$ is a learnable scaling coefficient and $\odot$ denotes element-wise multiplication.  
At this stage, the modulation operation plays a crucial role. It is applied to the frequency magnitude because underwater degradation disproportionately suppresses the high-frequency magnitude, which reduces edge sharpness and texture contrast \cite{ata2021underwater, ata2022underwater, hou2009simple}. Although the phase spectrum encodes the spatial arrangement of structural details, the magnitude controls the strength and visibility of these details \cite{hansen2007structural, yin2019fourier,cheng2024fdce}. 
The CNN-generated modulation map provides a learnable, adaptive mechanism that determines where and by how much the magnitude should be enhanced, while the additive formulation preserves the original spectral baseline to prevent distortion. The controlled scaling factor limits excessive amplification, avoids ringing or noise, and enables stable enhancement of degraded high-frequency components. Therefore, spatial geometry is preserved during reconstruction, and visual consistency is maintained because the phase remains unchanged. Overall, the modulation operation acts as a targeted frequency-domain sharpener that improves clarity, maintains robustness, and preserves visual realism in the restored image.

The inverse FFT reconstructs the enhanced image in the spatial domain and computes the high-frequency correction map, which is given as: 
\begin{equation}
    \mathbf{R}_f = \mathcal{F}^{-1}(\mathbf{M}^* e^{j\Phi_c}) - \mathbf{I},
\end{equation}
where $\mathcal{F}^{-1}(\cdot)$ denotes the inverse Fourier Transform.
This frequency correction map is projected into the latent feature space via a $3\times3$ convolution $\phi_p(\cdot)$ and fused with the initial encoder feature map:
\begin{equation}
    \mathbf{X}_0 = \phi_p(\mathbf{R}_f) + \phi_0(\mathbf{I}),
\end{equation}
where $\phi_0(\cdot)$ represents the initial convolutional projection.  
This fusion enables the encoder to simultaneously leverage frequency-driven textural cues and spatially rich contextual information, thereby improving both local contrast and global structure.

 The enhanced spatial–frequency representation $\mathbf{X}_0$ obtained from the fusion stage serves as the input to the encoder hierarchy for further feature abstraction. 
The encoder comprises three sequential encoder blocks, each composed of a REM followed by a down-sampling layer. At each stage, the spatial resolution is reduced by a factor of two, while the channel dimension is doubled, which enables progressive extraction of multi-scale hierarchical features. This hierarchical encoding structure allows the network to jointly capture fine-grained local textures and global contextual cues, thereby improving the representation for UIE.

\subsection{Illumination-Aware Decoder}

The illumination-aware decoder is designed to reconstruct the enhanced underwater image by jointly exploiting hierarchical encoder features and spatially adaptive illumination information. Underwater images commonly exhibit spatially non-uniform illumination, color attenuation, and contrast degradation caused by wavelength-dependent light absorption and scattering \cite{li2017watergan}. These effects can vary substantially across a scene, which makes spatially uniform enhancement insufficient to recover natural brightness and local details.

Motivated by the Retinex theory, which represents an observed image as the element-wise product of reflectance and illumination, $\mathbf{I} = \mathbf{R} \odot \mathbf{L}$, where $\mathbf{R}$ denotes the scene reflectance and $\mathbf{L}$ represents the illumination component \cite{fu2014retinex}, we introduce an illumination estimation branch to explicitly extract spatial illumination information from the underwater input. Unlike conventional Retinex approaches that estimate illumination using hand-crafted filtering or multi-scale image processing, the proposed illumination estimator learns a pixel-wise illumination representation directly from the input image in a data-driven manner \cite{li2018lightennet}.

The illumination estimator $\phi_l(\cdot)$ consists of a lightweight convolutional network that directly predicts a single-channel illumination map:
\begin{equation}
    \mathbf{L}_{p} = \sigma\left(\phi_l(\mathbf{I}_u)\right),
\end{equation}
where $\mathbf{I}_u$ denotes the underwater input image, $\phi_l(\cdot)$ represents the illumination estimation network, $\mathbf{L}_{p}$ is the predicted illumination map, and $\sigma(\cdot)$ denotes the sigmoid activation function. The sigmoid constrains the estimated illumination to the range $[0,1]$, providing a bounded and spatially continuous representation of scene illumination.

To further correct the estimated illumination, we introduce a gamma transformation:
\begin{equation}
    \mathbf{L}_{\gamma}
    =
    \left(
    \operatorname{clip}(\mathbf{L}_{p},\epsilon,1)
    \right)^{\gamma},
\end{equation}
where $\gamma$ is the gamma correction parameter and $\epsilon$ is a small positive constant introduced for numerical stability. In our implementation, $\gamma=0.7$. The resulting $\mathbf{L}_{\gamma}$ provides a nonlinear adjustment of the estimated illumination while preserving its spatial structure. In particular, gamma correction enables the illumination representation to be adaptively shifted toward a more suitable brightness distribution before being incorporated into the reconstruction process.

To make the predicted illumination map physically meaningful rather than relying solely on the final image reconstruction loss, the illumination estimator is additionally supervised using the paired reference image. Given the corresponding high-quality reference image $\mathbf{I}_g$, its illumination target is estimated using the maximum RGB response \cite{yan2025igdnet}:
\begin{equation}
    \mathbf{L}_{g}
    =
    \max_{c \in \{R,G,B\}}
    \mathbf{I}_{g}^{c},
\end{equation}
where $\mathbf{I}_{g}^{c}$ denotes the $c$-th color channel of the reference image. The illumination supervision is then defined as an L1 reconstruction objective:
\begin{equation}
    \mathcal{L}_{\mathrm{illum}}
    =
    \left\|
    \mathbf{L}_{p}
    -
    \mathbf{L}_{g}
    \right\|_{1}.
\end{equation}
This auxiliary supervision encourages the illumination estimator to learn a meaningful spatial illumination representation from the paired underwater reference training samples.

During decoding, feature maps $\mathbf{E}_k$ at level $k$ are upsampled using $\psi_u(\cdot)$ and fused with skip connections $\mathbf{S}_k$ from the encoder. The gamma-corrected illumination map $\mathbf{L}_{\gamma}$ is interpolated to match the spatial resolution of each decoder stage and guides the fusion as

\begin{equation}
\mathbf{D}_k =
\psi_u(\mathbf{E}_{k+1})
+
\mathbf{S}_k
+
\mathbf{L}_{\gamma,k}.
\end{equation}

where $\odot$ denotes element-wise multiplication and $\mathbf{L}_{\gamma,k}$ is the rescaled gamma-corrected illumination map at level $k$. 

Finally, the enhanced image is reconstructed from the final decoder representation using a convolutional reconstruction layer:
\begin{equation}
    \hat{\mathbf{I}}
    =
    \sigma
    \left(
    \phi_r(\mathbf{D}_1)
    \right),
\end{equation}
where $\phi_r(\cdot)$ denotes the final reconstruction convolution and $\sigma(\cdot)$ constrains the reconstructed image to the normalized intensity range $[0,1]$.

The complete training objective combines the original image reconstruction loss with the proposed illumination supervision:
\begin{equation}
    \mathcal{L}_{\mathrm{total}}
    =
    \mathcal{L}_{\mathrm{AQUA}}
    +
    \lambda_{\mathrm{illum}}
    \mathcal{L}_{\mathrm{illum}},
\end{equation}
where $\mathcal{L}_{\mathrm{AQUA}}$ denotes the original image reconstruction loss, $\mathcal{L}_{\mathrm{illum}}$ denotes the illumination supervision loss, and $\lambda_{\mathrm{illum}}$ controls the contribution of the illumination constraint. In our implementation, the image reconstruction objective is based on the L1 distance between the predicted enhanced image and its paired reference image.

By explicitly supervising illumination and subsequently applying gamma correction before multi-scale decoder fusion, the proposed illumination-aware decoder provides a physically motivated mechanism for handling spatially varying brightness degradation. The resulting design preserves the original hierarchical reconstruction pathway while providing additional illumination guidance for recovering locally consistent brightness, contrast, and structural details in underwater scenes.

\subsection{Dataset Acquisition }
The real underwater videos come from multiple campaigns in the Mediterranean Sea, including three key locations: the Strait of Sicily (depths 138–760 m), off the coast of Bari (470 m), and off the coast of Oristano (108–258 m). From these videos, we extracted high-quality frames, of which 80 representative images were selected for testing and named DeepSea-T80. All images have high resolution and are captured using a 6K cinema-quality camera system (ZCAM E2-F6) mounted on the ROV Tomahawk Light Work Class with a Canon EF 16–35 mm f/2.8L III USM lens. The frames include a wide range of underwater scenes, covering marine life, seabed features, and varied aquatic landscapes, and are recorded under diverse environmental conditions, including varying water clarity and different depths, ensuring a comprehensive representation of real deep-sea visual challenges. Due to the deep-sea environment, natural sunlight at depths greater than 130–150 m is negligible, and the ROV’s artificial lighting serves as the primary illumination source. The combination of high-resolution capture, varying illumination, and diverse substrates produces realistic visual degradations, including color attenuation, low light, back-scatter, and turbidity, making the dataset suitable for evaluating deep learning models for UIE under challenging deep-sea conditions.
\section{Experiments}
\subsection{Implementation}
The proposed AQUA-Net model is implemented in the PyTorch framework. Initially, a base encoder–decoder architecture is developed. Subsequently, frequency and illumination enhancement modules are incorporated into the base network in a fusion manner to improve feature representation, considering UWIs conditions. The input images are resized to $128 \times 128$ pixels with a batch size of 8. Model training is performed using the Stochastic Gradient Descent (SGD) optimizer with a learning rate of 0.001. The model parameters are optimized using the L1 loss function over 100 training epochs. 

\subsubsection{Datasets} 
The AQUA-Net model is trained on the UEIB dataset \cite{li2019underwater}, which consists of 890 images, including corresponding reference images. For evaluation, we utilize a subset of 90 images from this dataset, referred to as UEIB-T90, while the remaining images are used for training. The dataset also includes a challenging set of 60 images without reference images, known as UEIB-C60 \cite{li2019underwater}, on which we further evaluate our model. Additionally, we evaluate the model's performance on the EUVP-T515 \cite{islam2020fast}, RUIE-T78 \cite{liu2020real}, and DeepSea-T80 datasets.

\begin{itemize}
    \item 
            The UEIB \cite{li2019underwater} dataset is compiled from various online sources that contain UIWs captured in real-world aquatic environments. In addition to the raw UIWs, it includes carefully curated reference images that are processed through multiple SOTA enhancement algorithms and then selected based on the most visually and quantitatively superior results.
    \item 
            The authors in \cite{islam2020fast} present a dataset that comprises a total of 20,000 UIWs of low and good quality. The dataset contains 12,000 paired images and 8,000 unpaired images. Similar to \cite{liu2024underwater, liu2025toward}, we use 515 images from this dataset for the evaluation of our AQUA-Net model, which we denote as EUVP-T515.
    \item 
            The RUIE dataset \cite{song2020enhancement} contains real-world UIWs in three folders: UCCS (color distortions), UIQS (various water types and degradation levels), and UHTS (for object detection). 
            A total of 78 images are selected from these sub-folders, denoted as RUIE-T78 \cite{liu2024underwater, liu2025toward}.
    \item 
            From our own captured videos, we extracted 1,533 high-quality frames and selected 80 representative images for evaluation, which we refer to as DeepSea-T80. All images have a high resolution of 1920 × 1080 pixels and were captured using a camera mounted on a remotely operated vehicle (ROV).
            
\end{itemize}

\subsubsection{Baseline Models}
The effectiveness of the AQUA-Net model is evaluated through comparison with several SOTA models. These models fall into two categories: physics-based and data-driven deep learning models. From the first category, Fusion \cite{ancuti2017color}, SMBL \cite{song2020enhancement}, MLLE \cite{zhang2022underwater}, WWPF \cite{zhang2023underwater}, and WFAC \cite{zhang2025underwater1} are considered. From the second category, UWCNN \cite{li2020underwater}, WaterNet \cite{li2019underwater}, UColor \cite{li2021underwater}, PUIE \cite{fu2022uncertainty}, TACL \cite{liu2022twin}, NU2Net \cite{guo2023underwater}, CCL-Net \cite{liu2024underwater}, and OUNet-JL \cite{wang2025optimized} are included. In total, thirteen models are included for comparison.
\subsubsection{Metrics}
The performance of the aforementioned UIE models is evaluated using both reference-based and non-reference metrics. The reference metrics, PSNR and SSIM \cite{wang2004image}, quantify the similarity between enhanced images and ground-truth references in terms of pixel-level accuracy and structural fidelity. The non-reference metrics, UIQM  \cite{panetta2015human} and UCIQE \cite{yang2015underwater}, evaluate perceptual quality by measuring attributes such as colorfulness, contrast, and sharpness, even in the absence of reference images. In all cases, higher metric values indicate the best performance. 

\subsection{Evaluation}
\subsubsection{Quantitative Results} 
The quantitative results presented in Table \ref{tab:main_comparison} highlight the performance of the proposed AQUA-Net model across four widely used evaluation metrics. As shown in Table \ref{tab:main_comparison}, AQUA-Net achieves the highest UIQM score, which indicates that the enhanced images generated by the model exhibit good visual quality in terms of colorfulness, contrast, and sharpness. This result demonstrates the effectiveness of AQUA-Net in restoring perceptual quality under challenging underwater conditions. Moreover, the SSIM value obtained by the model ranks third among the compared SOTA models, which reflects its ability to preserve structural information and maintain similarity with reference images. Although the remaining metrics, such as PSNR and UCIQE, do not reach the top position, their values remain highly competitive, which shows that AQUA-Net performs consistently well across different quantitative measures. Overall, these results show that AQUA-Net gains a balanced performance between objective fidelity and perceptual quality and outperforms or closely matches existing UIE models.
\begin{table}[t!]
\centering
\fontsize{12.0pt}{14.0pt}\selectfont
\caption{Quantitative evaluation on the UEIB dataset using PSNR, SSIM, UIQM, and UCIQE metrics. The top three results are shown in \textbf{\textcolor{red}{red}}, \textbf{\textcolor{green}{\underline{green}}}, and \textbf{\textit{\textcolor{blue}{{blue}}}}.}
\label{tab:main_comparison}
\begin{tabular}{l|cccc}
\hline
\rowcolor[HTML]{ADD8E6} \textbf{Methods} & \textbf{PSNR$\uparrow$} & \textbf{SSIM$\uparrow$} & \textbf{UIQM$\uparrow$} & \textbf{UCIQE$\uparrow$} \\ 
\hline
Raw & 16.134 & 0.748 & 2.346 & 0.362 \\
Fusion(TIP’17)~\cite{ancuti2017color} & 18.033 & 0.861 & 2.684 & 0.406 \\
SMBL(TB’20)~\cite{song2020enhancement} & 16.513 & 0.781 & 2.167 & \textbf{\textit{\textcolor{blue}{0.455}}} \\
MLLE(TIP’22)~\cite{zhang2022underwater} & 18.727 & 0.790 & 2.305 & \textbf{\textcolor{red}{0.468}} \\
UWCNN(PR’20)~\cite{li2020underwater} & 18.147 & 0.847 & 2.878 & 0.357 \\
WaterNet(TIP’19)~\cite{li2019underwater} & 19.914 & 0.859 & 2.846 & 0.410 \\
PUIE(ECCV’22)~\cite{fu2022uncertainty} & \textbf{\textit{\textcolor{blue}{22.023}}} & \textbf{\textcolor{red}{0.893}} & 2.849 & 0.396 \\
TACL(TIP’22)~\cite{liu2022twin} & \textbf{\textcolor{green}{\underline{22.735}}} & 0.864 & 3.016 & \textbf{\textcolor{green}{\underline{0.445}}} \\
WWPF~\cite{zhang2023underwater} & 19.260 & 0.805 & 2.566 & 0.408 \\
WFAC~\cite{zhang2025underwater} & 17.540 & 0.748 & 2.581 & 0.372 \\

NU2Net(AAAI’23)~\cite{guo2023underwater} & \textbf{\textcolor{red}{22.820}} & \textbf{\textcolor{red}{0.893}} & 2.902 & 0.422 \\
CCL-Net(TMM’24)~\cite{liu2024underwater} & 20.181 & 0.866 & \textbf{\textit{\textcolor{blue}{3.021}}} & \textbf{\textcolor{red}{0.464}} \\ 
OUNet-JL(Sci Rep’25)~\cite{wang2025optimized} & 19.541 & 0.829 & \textbf{\textcolor{red}{3.340}} & 0.442 \\ 
\hline
\textbf{AQUA-Net} & 21.482 & \textbf{\textit{\textcolor{blue}{0.890}}} & \textbf{\textcolor{green}{\underline{3.250}}} & 0.397 \\ 
\bottomrule
\end{tabular}
\end{table}

Further evaluation of the proposed model is conducted on the UIEB-C60, EUVP-T515, RUIE-T78, and DeepSea-T80 datasets as shown in the Table \ref{tab:non_ref_1}. On UIEB-C60, the model achieves the second-highest UIQM score, while on EUVP-T515, it attains the third-highest UCIQE score. These results indicate that AQUA-Net also performs well on non-reference datasets and demonstrates good generalization capability across diverse underwater conditions.

\begin{table}[t!]
\centering
\fontsize{10.0pt}{12.0pt}\selectfont
\caption{Quantitative comparison on UIEB-C60, EUVP-T515, RUIE-T78, and DeepSea-T80 in terms of UIQM and UCIQE. The top three scores are marked in \textbf{\textcolor{red}{red}}, \textbf{\textcolor{green}{\underline{green}}}, and 
\textcolor{blue}{\textbf{\textit{blue}}}.}
\renewcommand{\arraystretch}{1.1}
\label{tab:non_ref_1}
\setlength{\tabcolsep}{4pt}
\begin{tabular}{l|cc|cc|cc|cc}
\hline
\rowcolor[HTML]{ADD8E6} \textbf{Methods} 
& \multicolumn{2}{c|}{\textbf{UIEB-C60}} 
& \multicolumn{2}{c|}{\textbf{EUVP-T515}} 
& \multicolumn{2}{c|}{\textbf{RUIE-T78}}
& \multicolumn{2}{c}{\textbf{DeapSea-T80}} \\
& UIQM$\uparrow$ & UCIQE$\uparrow$ 
& UIQM$\uparrow$ & UCIQE$\uparrow$
& UIQM$\uparrow$ & UCIQE$\uparrow$
& UIQM$\uparrow$ & UCIQE$\uparrow$ \\
\hline
Raw & 1.856 & 0.359 & 2.217 & 0.417 & 2.437 & 0.321 & 2.035 & 0.247 \\
Fusion \cite{ancuti2017color} & 2.163 & 0.378 & 2.636 & 0.430 & 2.772 & 0.366 & 2.698 &  0.393 \\
SMBL \cite{song2020enhancement} & 1.724 & \textcolor{green}{\textbf{\underline{0.439}}} & 1.857 & \textcolor{red}{\textbf{0.513}} & 2.459 & \textcolor{blue}{\textbf{\textit{0.431}}} & 2.407 & 0.382 \\
MLLE \cite{zhang2022underwater} & 1.956 & \textcolor{red}{\textbf{0.464}} & 2.354 & \textcolor{green}{\underline{\textbf{0.461}}} & 2.798 & \textcolor{green}{\underline{\textbf{0.441}}} & 2.843 & \textbf{\textcolor{red}{0.436}} \\
WWPF~\cite{zhang2023underwater}& 2.1304 & 0.3943& 2.3941 & 0.4695& 2.7388 & 0.3726&2.485& 0.386\\
WFAC~\cite{zhang2025underwater1}& 2.0680 & 0.3514& 2.6169 & 0.3889& 2.7829 & 0.3607&1.946&0.257 \\
UWCNN \cite{li2020underwater} & 2.433 & 0.340 & \textcolor{green}{\underline{\textbf{2.822}}} & 0.369 & 3.053 & 0.314 & 2.407 & 0.327 \\
WaterNet \cite{li2019underwater} & 2.468 & 0.364 & 2.680 & 0.412 & \textcolor{blue}{\textbf{\textit{3.115}}} & 0.403 & 2.836 &  \textcolor{blue}{\textbf{\textit{0.411}}} \\
PUIE \cite{fu2022uncertainty}  & 2.379 & 0.375 & 2.748 & 0.407 & 2.989 & 0.379 & 2.793 & 0.314 \\
TACL \cite{liu2022twin} & \textcolor{red}{\textbf{2.854}} & 0.424 & \textcolor{blue}{\textbf{\textit{2.837}}} & 0.435 & \textcolor{red}{\textbf{3.237}} & 0.422 & \textcolor{blue}{\textbf{\textit{2.968}}} & \textcolor{green}{\underline{\textbf{0.423}}}\\

NU2Net  \cite{guo2023underwater} & \textcolor{blue}{\textbf{\textit{2.508}}} & 0.402 & 2.767 & \textcolor{blue}{\textbf{\textit{0.422}}} & 3.061 & 0.389 & \textbf{\textcolor{red}{3.071}} & 0.359 \\
CCL-Net \cite{liu2024underwater} & \textcolor{green}{\textbf{\underline{2.622}}} & \textcolor{blue}{\textbf{\textit{0.434}}} & \textcolor{red}{\textbf{2.936}} & \textcolor{green}{\underline{\textbf{0.456}}} & \textcolor{green}{\underline{\textbf{3.168}}} & \textcolor{red}{\textbf{0.447}} & 2.858 & 0.380 \\
OUNet-JL \cite{wang2025optimized} & 2.842 & 0.426 & 3.107 & 0.436 & 3.212 & 0.425 & \textcolor{green}{\underline{\textbf{3.022}}} &  0.388 \\
\hline
\textbf{AQUA-Net} & 2.313 & 0.427 & 2.353 & \textcolor{blue}{\textbf{\textit{0.470}}} & 3.027 & 0.370 & 2.538 & 0.351 \\
\bottomrule
\end{tabular}
\end{table}
\textbf{Parameter Efficiency}:
The AQUA-Net model demonstrates strong computational efficiency compared to other UIE models. As shown in Table \ref{tab:flops_params}, it achieves the second-best performance in terms of both the number of parameters and FLOPs, requiring only 0.333 M parameters and 20.86 G FLOPs. Although UWCNN \cite{li2020underwater} reports the lowest values for both parameters and FLOPs, its qualitative and quantitative results are significantly inferior to those of AQUA-Net. These findings indicate that AQUA-Net gains an excellent balance between computational efficiency and enhancement performance, which makes it suitable for real-time UIE applications.

\begin{table}[t!]
\centering
\fontsize{12.0pt}{14.0pt}\selectfont
\caption{Comparison of different UIE models in terms of FLOPs (G) and parameter counts (M). The three lowest results are highlighted in
\textbf{\textcolor{red}{red}},
\textbf{\textcolor{green}{\underline{green}}},
and \textcolor{blue}{\textbf{\textit{blue}}}, respectively.}
\label{tab:flops_params}

\setlength{\tabcolsep}{4pt}

\begin{tabular}{l|cc}
\toprule
\rowcolor[HTML]{ADD8E6}
\textbf{Methods} &
\textbf{FLOPs (G)$\downarrow$} &
\textbf{Parameters (M)$\downarrow$} \\
\midrule

Fusion (TIP'18) \cite{ancuti2017color} & -- & -- \\
SMBL (TB'20) \cite{song2020enhancement} & -- & -- \\
MLLE (TIP'22) \cite{zhang2022underwater} & -- & -- \\

UWCNN (PR'20) \cite{li2020underwater}
& \textcolor{red}{\textbf{11.36}}
& \textcolor{red}{\textbf{0.04}} \\

WaterNet (TIP'19) \cite{li2019underwater}
& 310.82 & 1.09 \\

PUIE (ECCV'22) \cite{fu2022uncertainty}
& 2073.2 & 0.83 \\

TACL (TIP'22) \cite{liu2022twin}
& 247.46 & 11.37 \\

NU2Net (AAAI'23) \cite{guo2023underwater}
& \textcolor{blue}{\textbf{\textit{46.33}}}
& 3.15 \\

CCL-Net (TMM'24) \cite{liu2024underwater}
& 470.62
& \textcolor{blue}{\textbf{\textit{0.55}}} \\

OUNet-JL (Sci Rep'25) \cite{wang2025optimized}
& 134.04 & 7.12 \\

\midrule

\textbf{AQUA-Net}
& \textcolor{green}{\textbf{\underline{20.86}}}
& \textcolor{green}{\textbf{\underline{0.333}}} \\

\bottomrule
\end{tabular}

\end{table}

\subsubsection{Visual Results}
\begin{figure*}[t!]
\centering

\setlength{\tabcolsep}{1.2pt}
\renewcommand{\arraystretch}{0.35}

\newcommand{\imgwidth}{0.13\textwidth}
\newcommand{\imgheight}{0.085\textwidth}

\begin{tabular}{ccccccc}

\textbf{\tiny Raw} &
\textbf{\tiny Fusion \cite{ancuti2017color}} &
\textbf{\tiny SMBL \cite{song2020enhancement}} &
\textbf{\tiny MLLE \cite{zhang2022underwater}} &
\textbf{\tiny WWPF \cite{zhang2023underwater}} &
\textbf{\tiny WFAC \cite{zhang2025underwater1}} &
\textbf{\tiny WaterNet \cite{li2019underwater}} \\[3pt]

\includegraphics[width=\imgwidth,height=\imgheight]{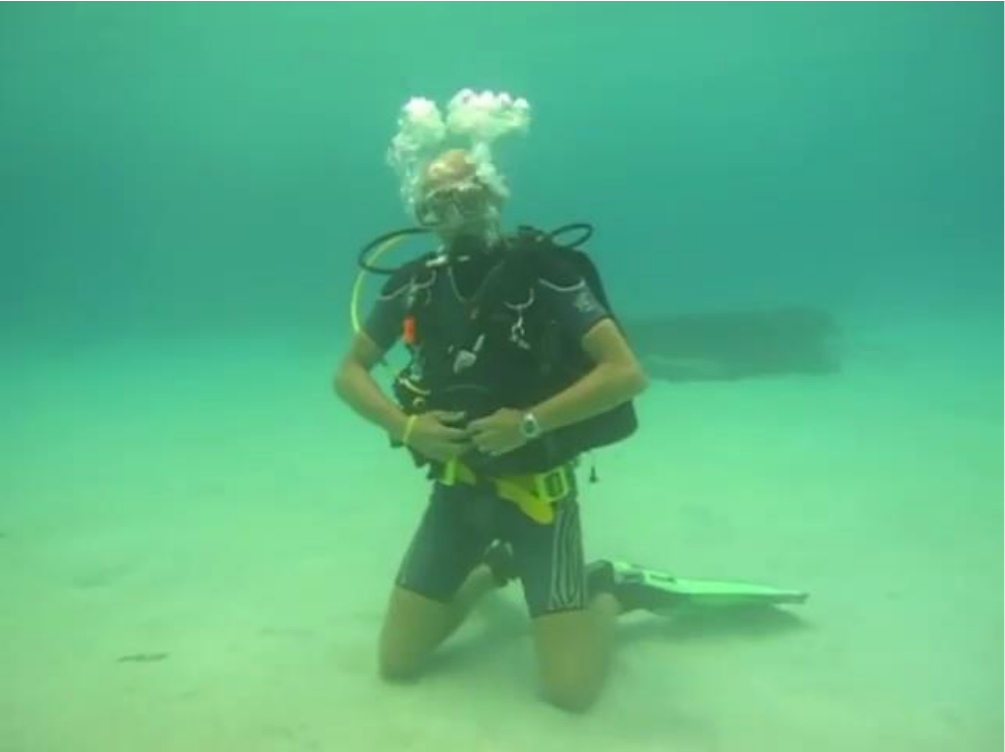} &
\includegraphics[width=\imgwidth,height=\imgheight]{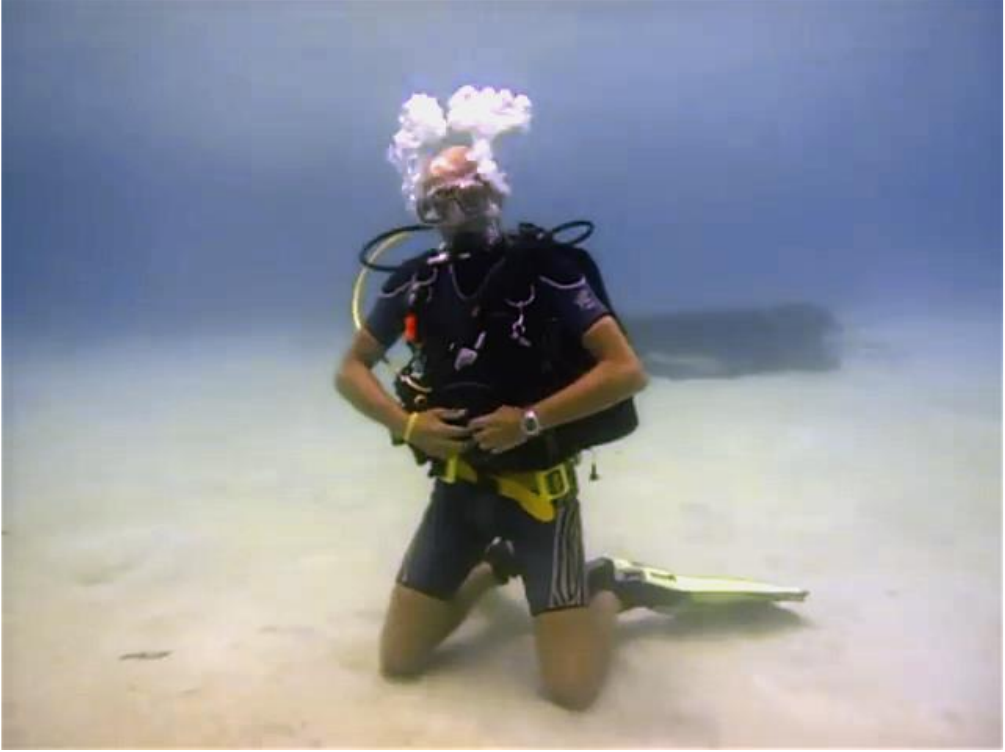} &
\includegraphics[width=\imgwidth,height=\imgheight]{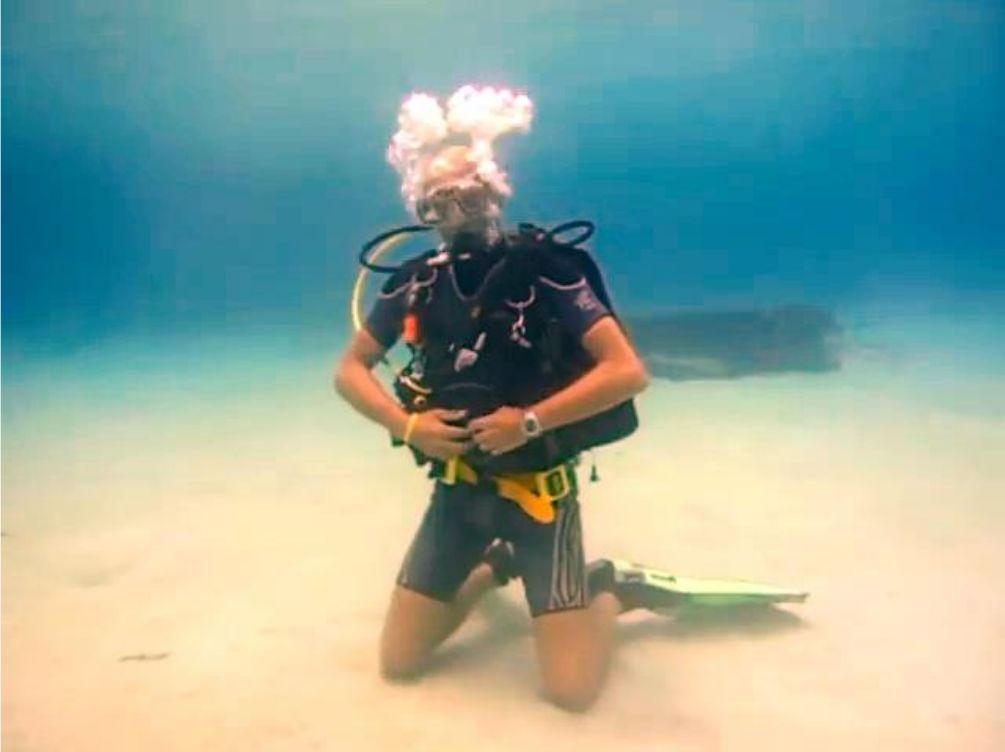} &
\includegraphics[width=\imgwidth,height=\imgheight]{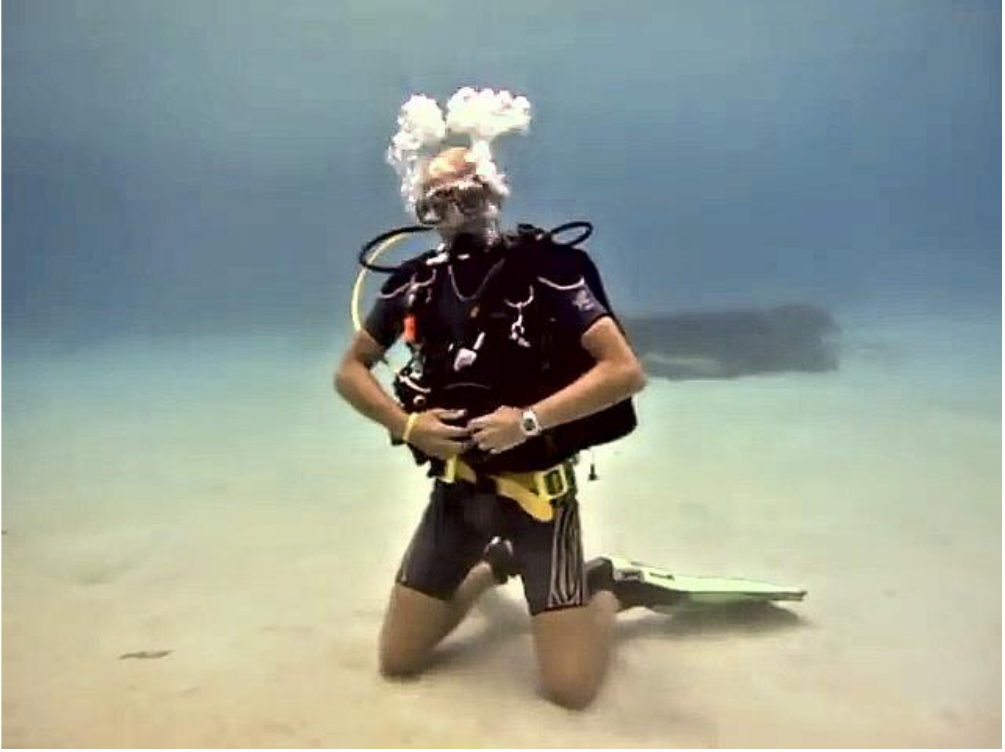} &
\includegraphics[width=\imgwidth,height=\imgheight]{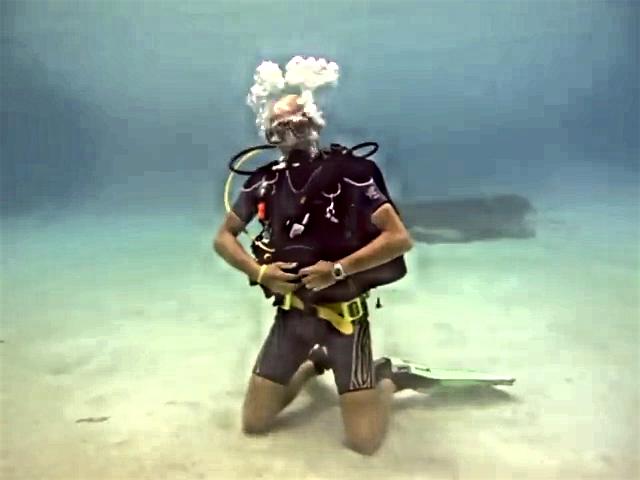} &
\includegraphics[width=\imgwidth,height=\imgheight]{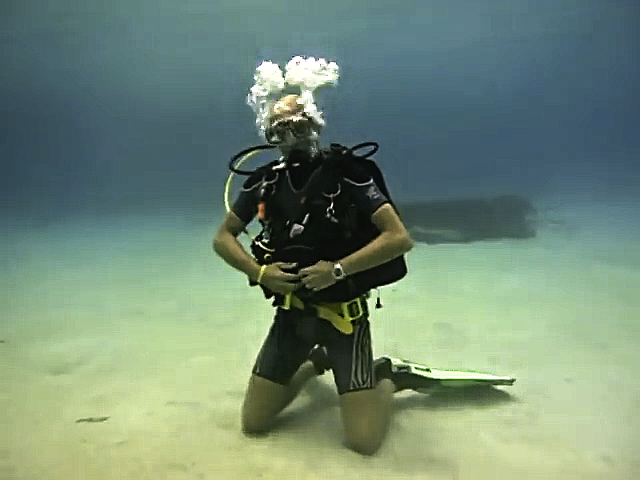} &
\includegraphics[width=\imgwidth,height=\imgheight]{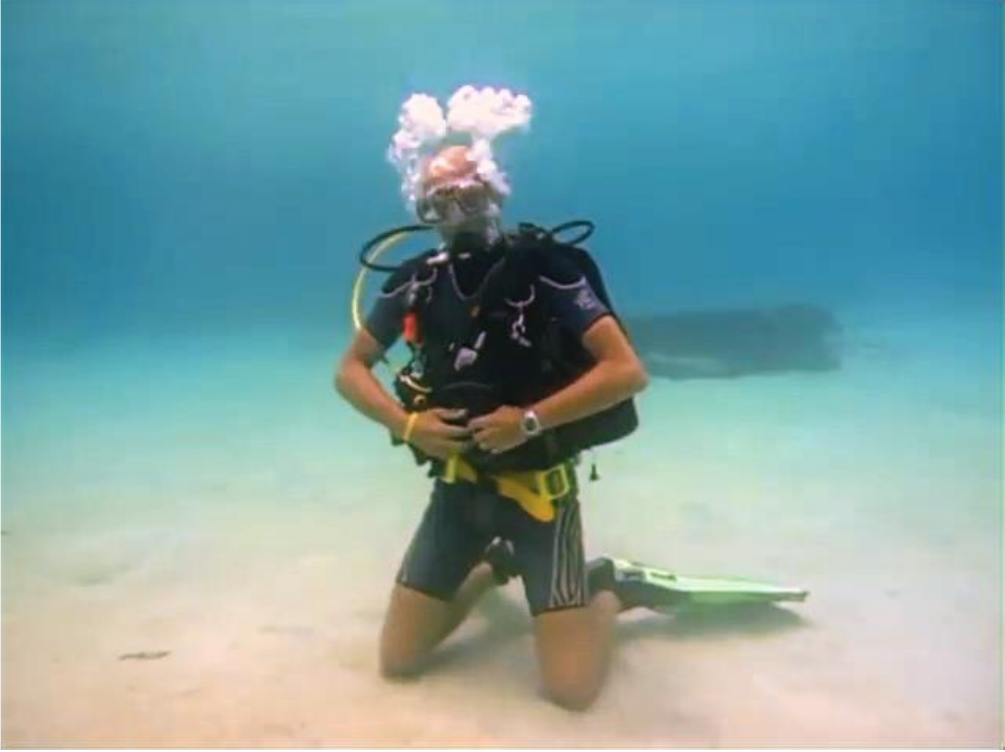} \\

\includegraphics[width=\imgwidth,height=\imgheight]{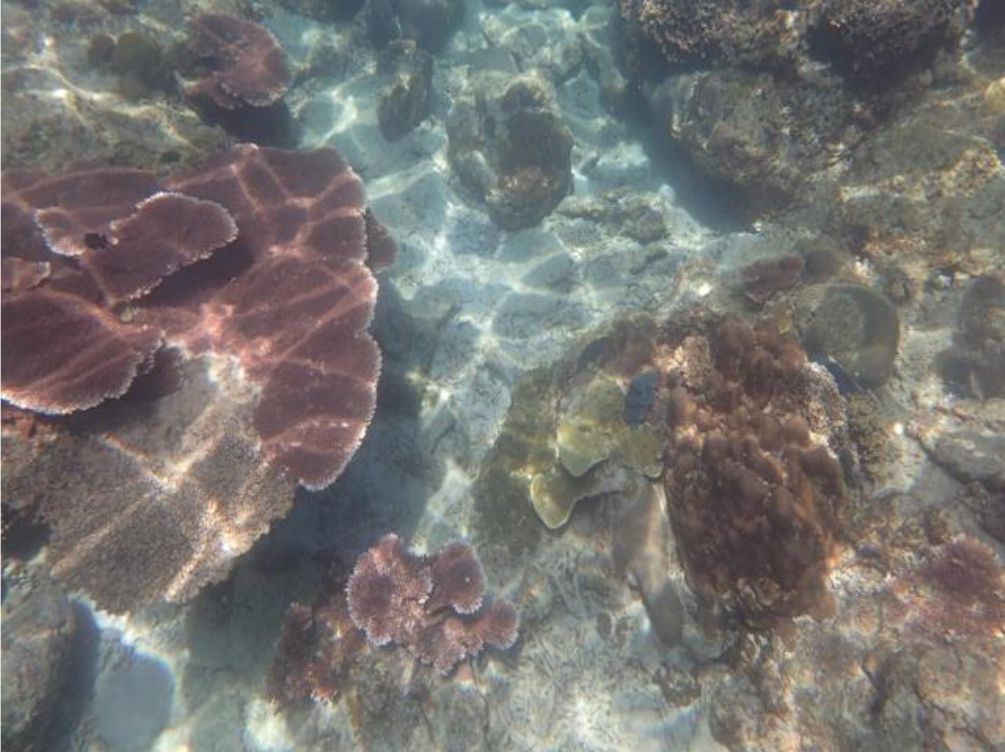} &
\includegraphics[width=\imgwidth,height=\imgheight]{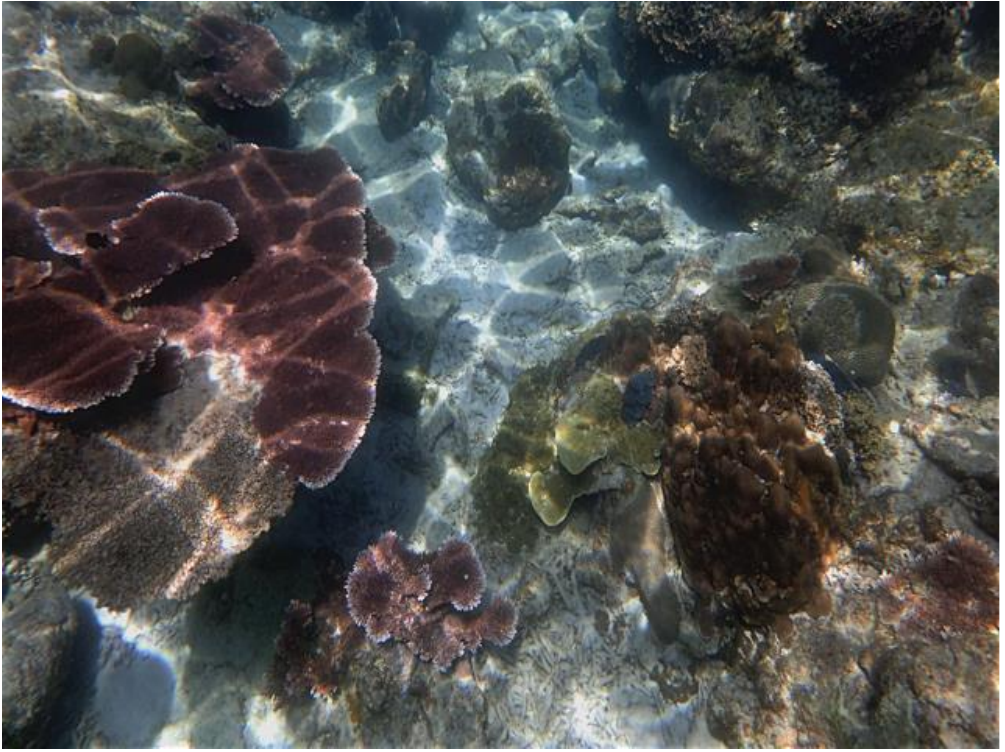} &
\includegraphics[width=\imgwidth,height=\imgheight]{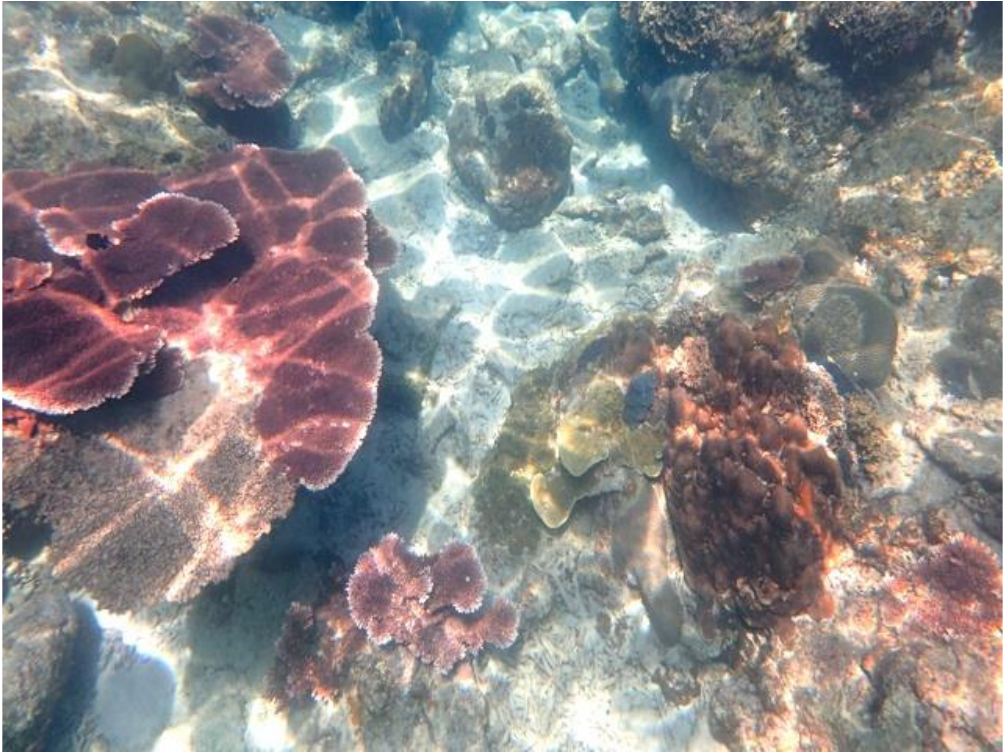} &
\includegraphics[width=\imgwidth,height=\imgheight]{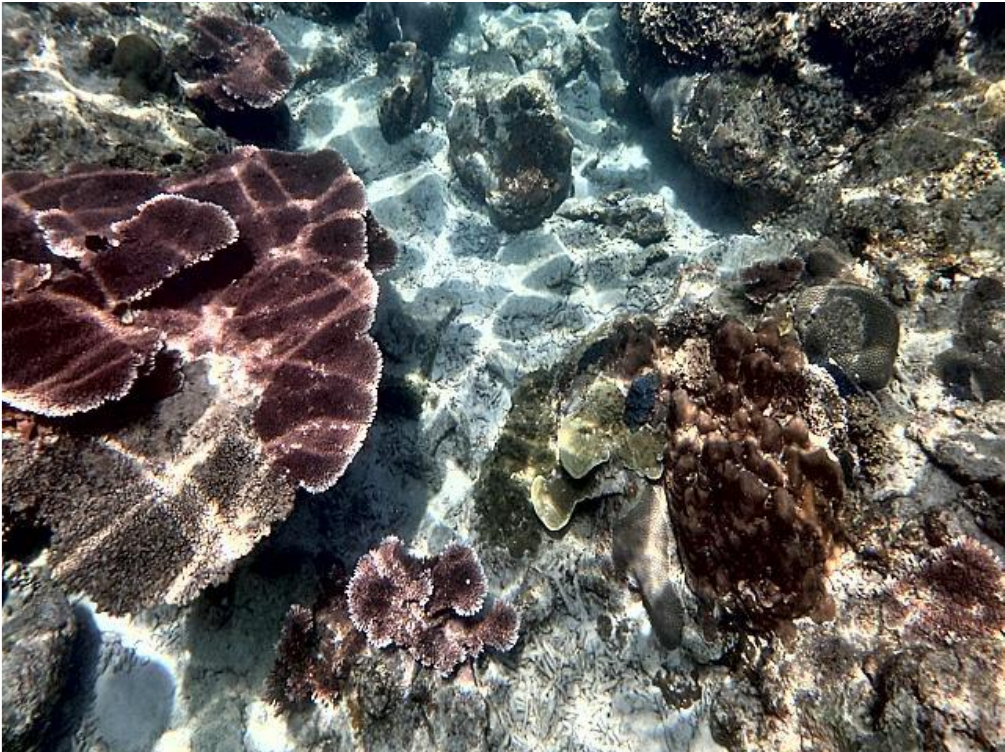} &
\includegraphics[width=\imgwidth,height=\imgheight]{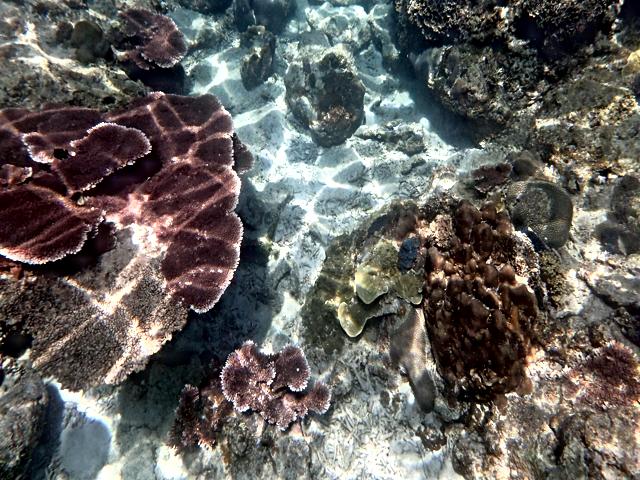} &
\includegraphics[width=\imgwidth,height=\imgheight]{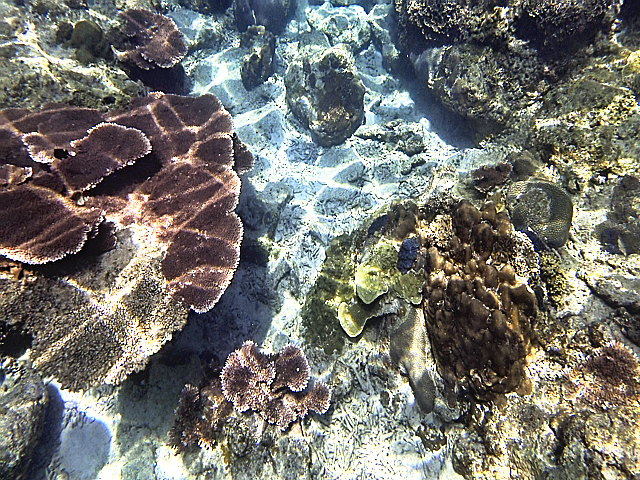} &
\includegraphics[width=\imgwidth,height=\imgheight]{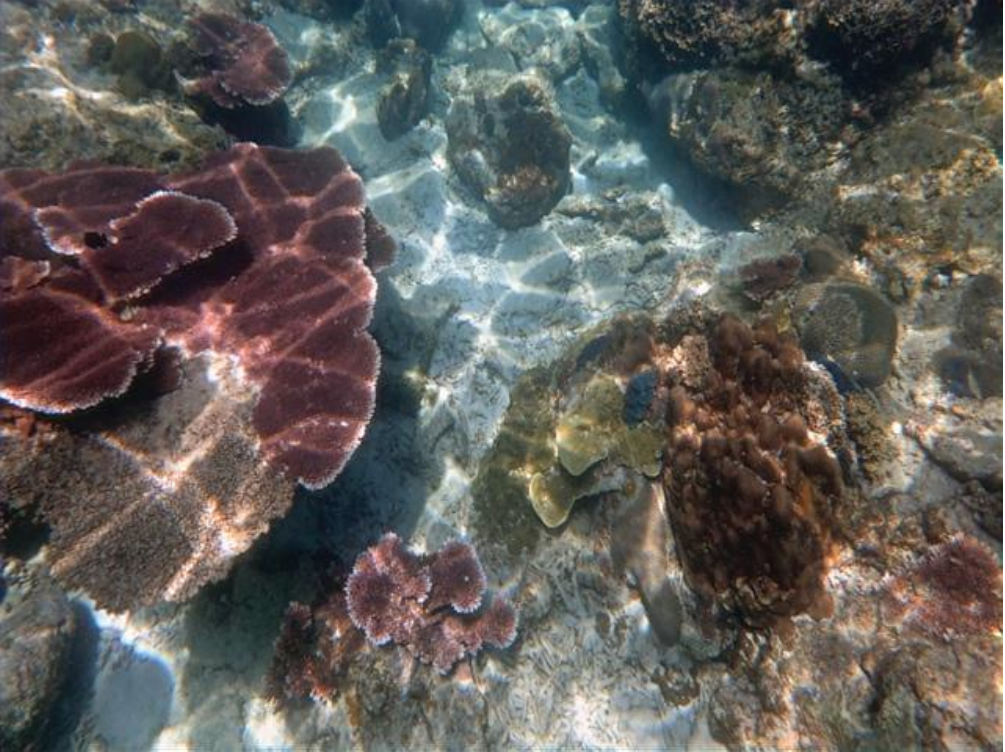} \\[5pt]

\textbf{\tiny UWCNN \cite{li2020underwater}} &
\textbf{\tiny PUIE \cite{fu2022uncertainty}} &
\textbf{\tiny TACL \cite{liu2022twin}} &
\textbf{\tiny NU2Net \cite{guo2023underwater}} &
\textbf{\tiny CCL-Net \cite{liu2024underwater}} &
\textbf{\tiny OUNet-JL \cite{wang2025optimized}} &
\textbf{\tiny AQUA-Net} \\[3pt]

\includegraphics[width=\imgwidth,height=\imgheight]{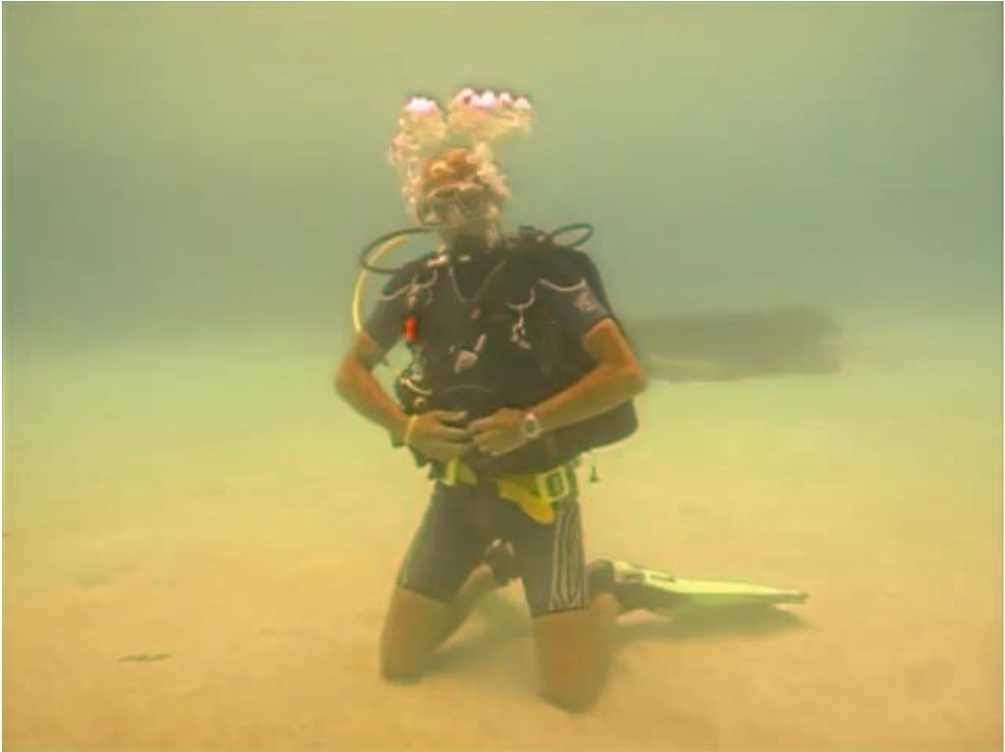} &
\includegraphics[width=\imgwidth,height=\imgheight]{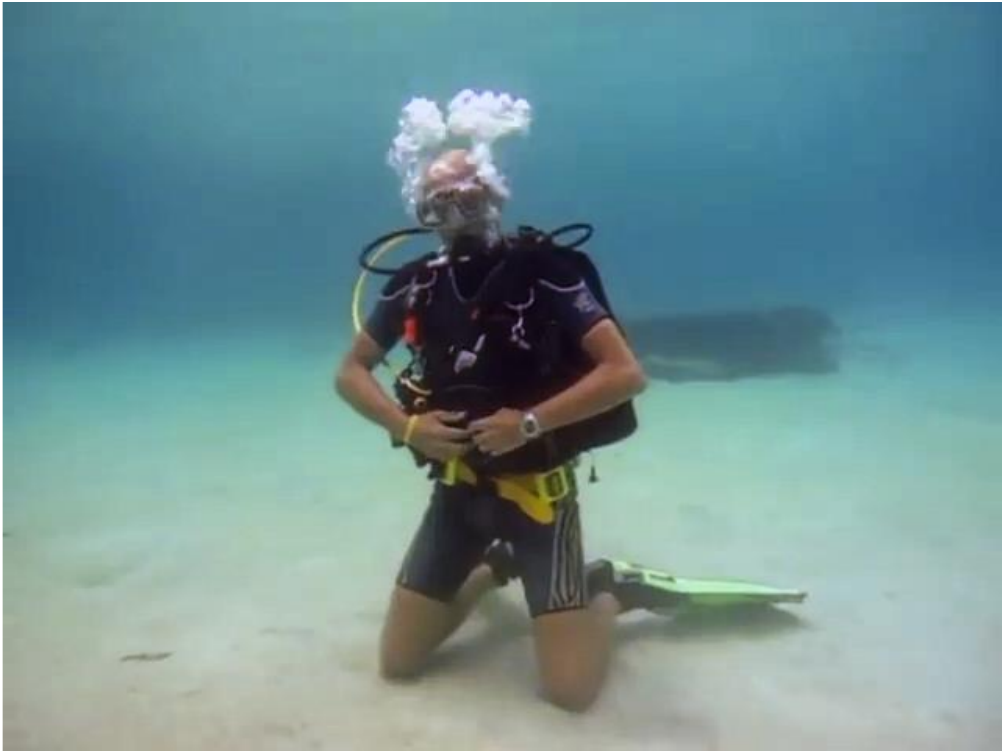} &
\includegraphics[width=\imgwidth,height=\imgheight]{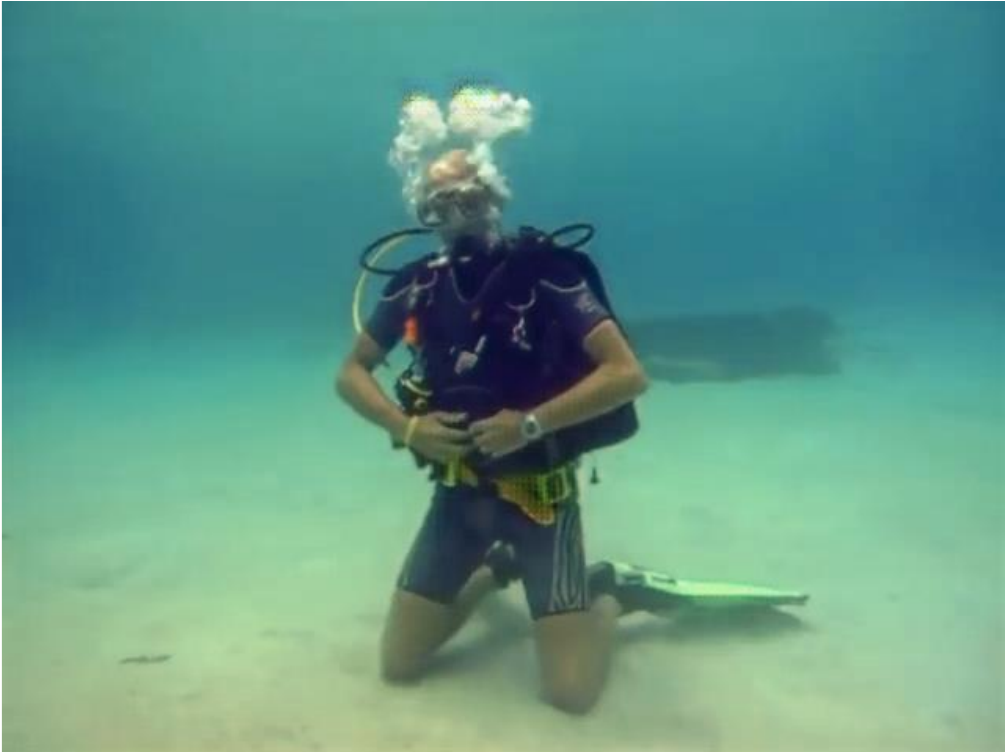} &
\includegraphics[width=\imgwidth,height=\imgheight]{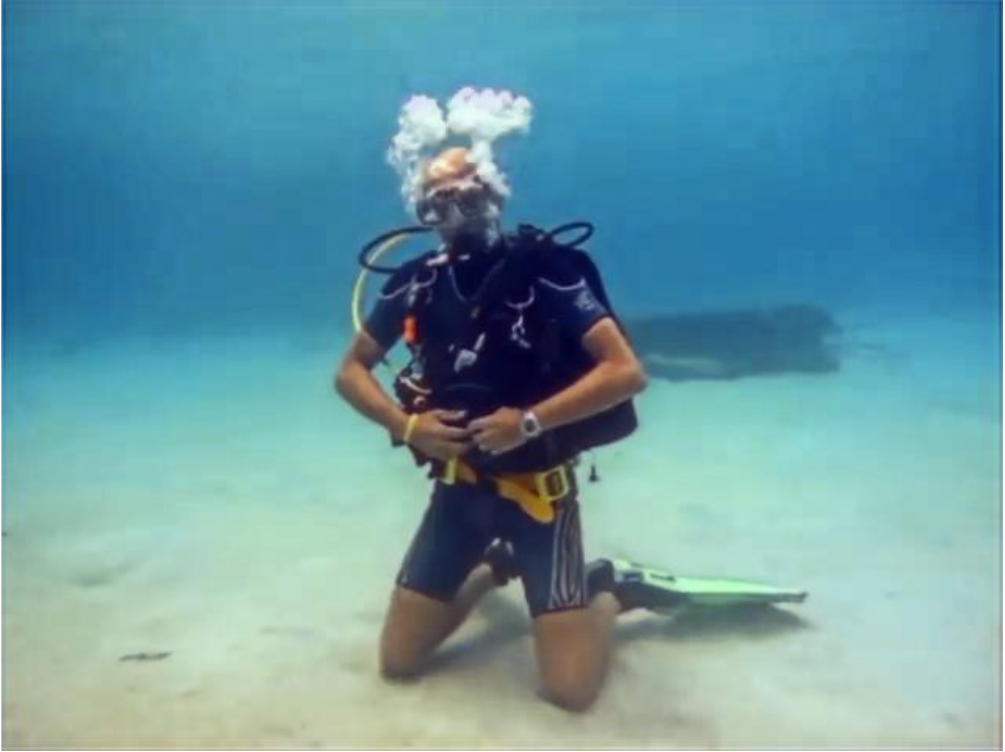} &
\includegraphics[width=\imgwidth,height=\imgheight]{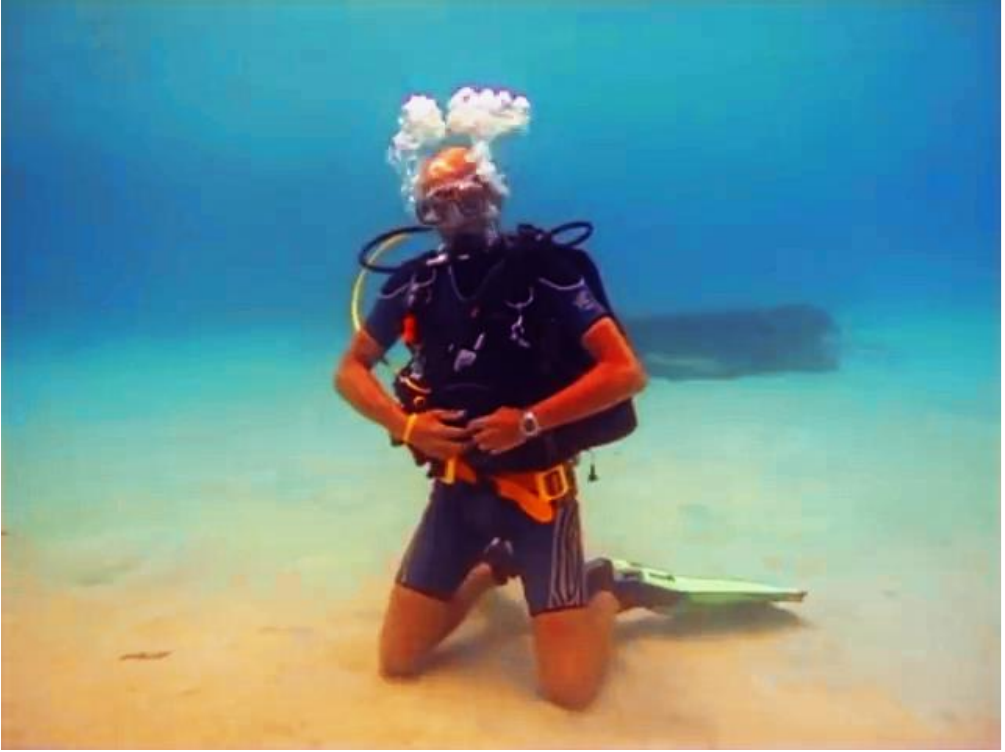} &
\includegraphics[width=\imgwidth,height=\imgheight]{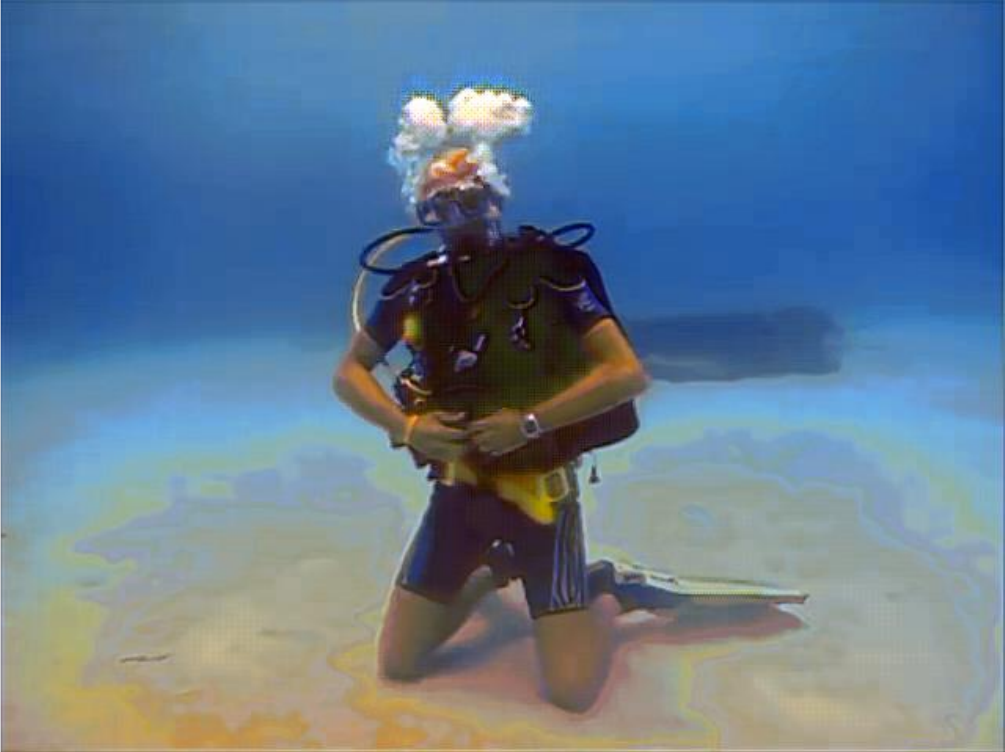} &
\includegraphics[width=\imgwidth,height=\imgheight]{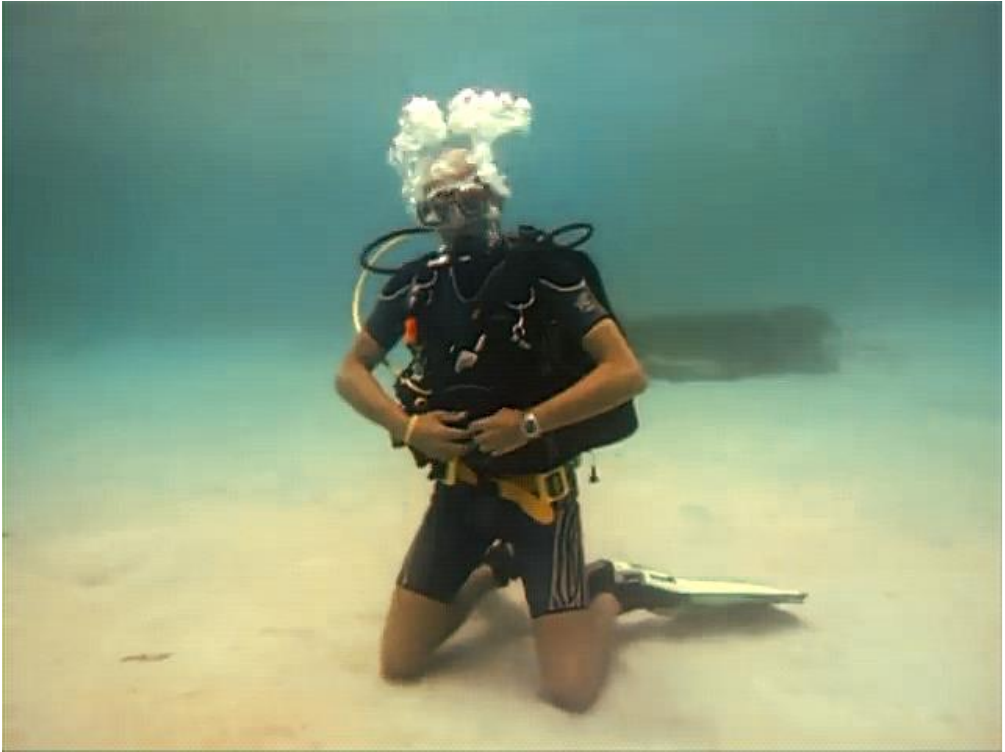} \\

\includegraphics[width=\imgwidth,height=\imgheight]{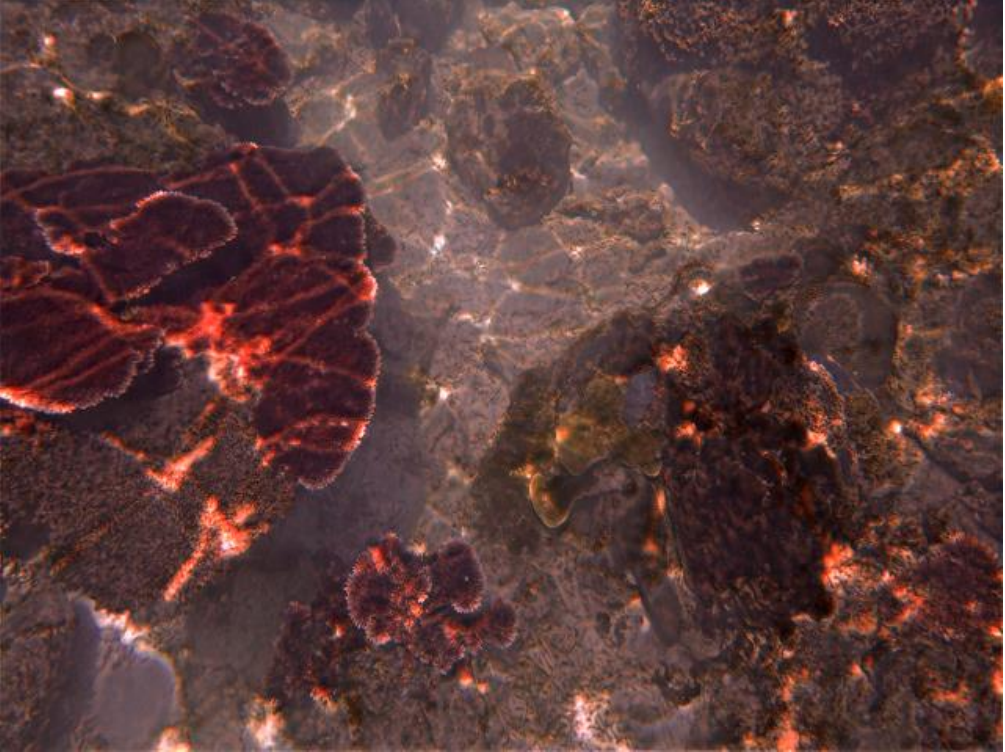} &
\includegraphics[width=\imgwidth,height=\imgheight]{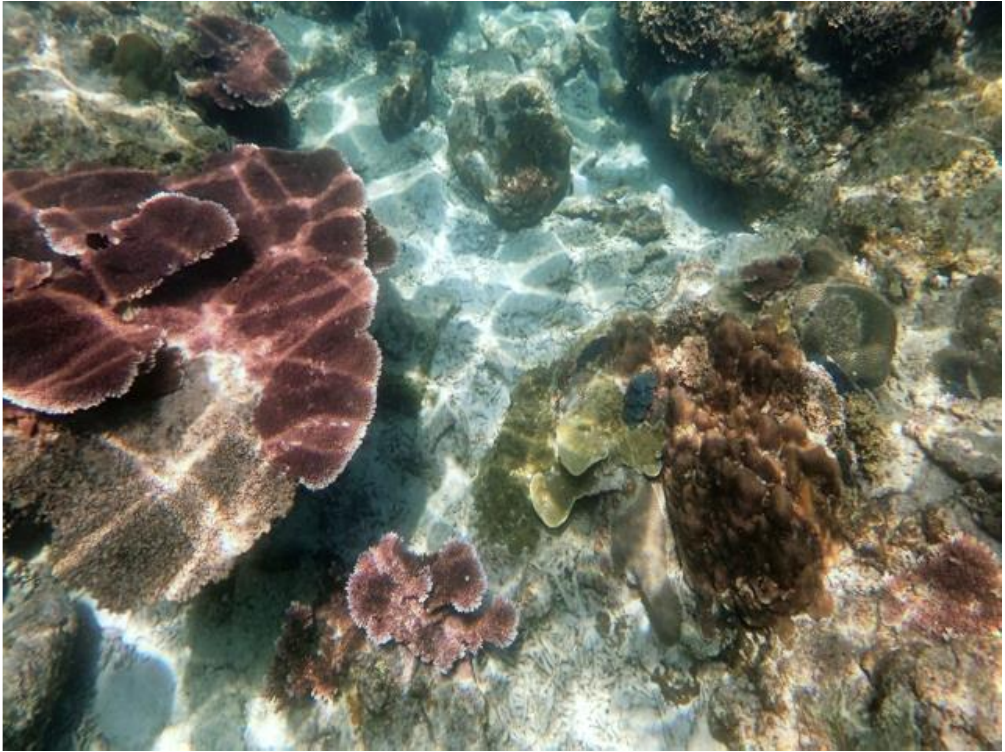} &
\includegraphics[width=\imgwidth,height=\imgheight]{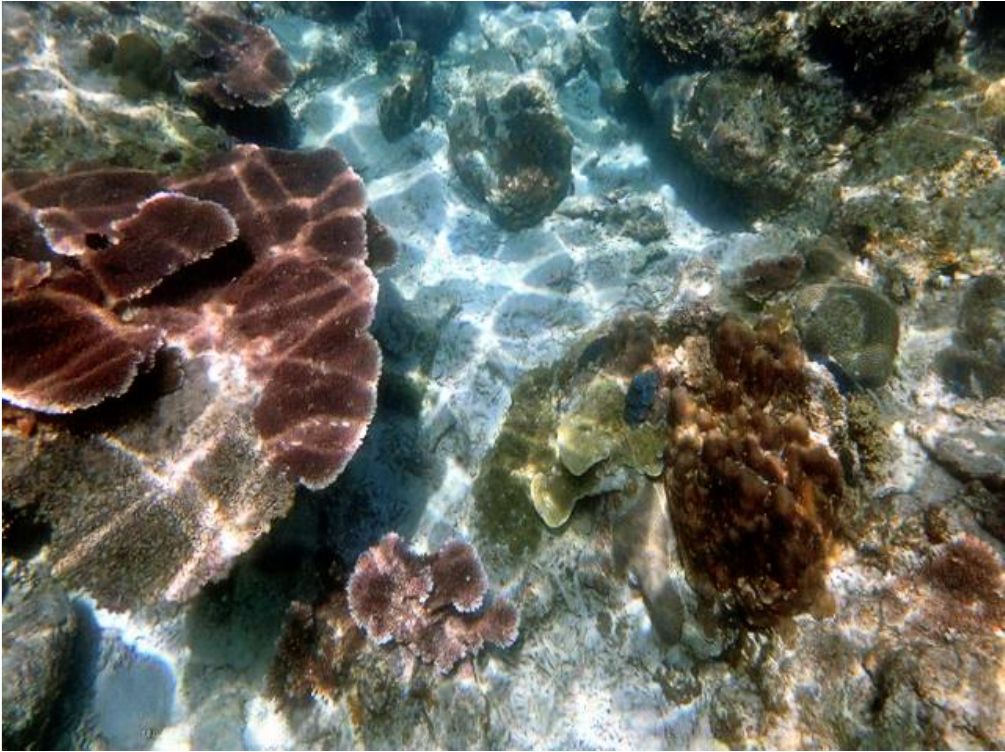} &
\includegraphics[width=\imgwidth,height=\imgheight]{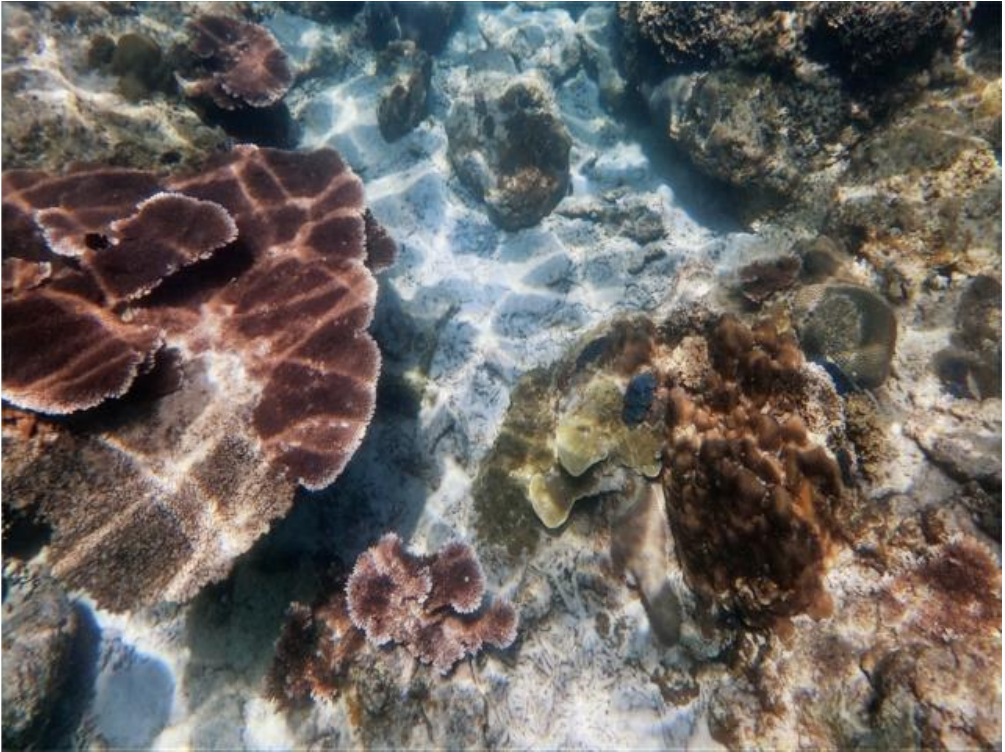} &
\includegraphics[width=\imgwidth,height=\imgheight]{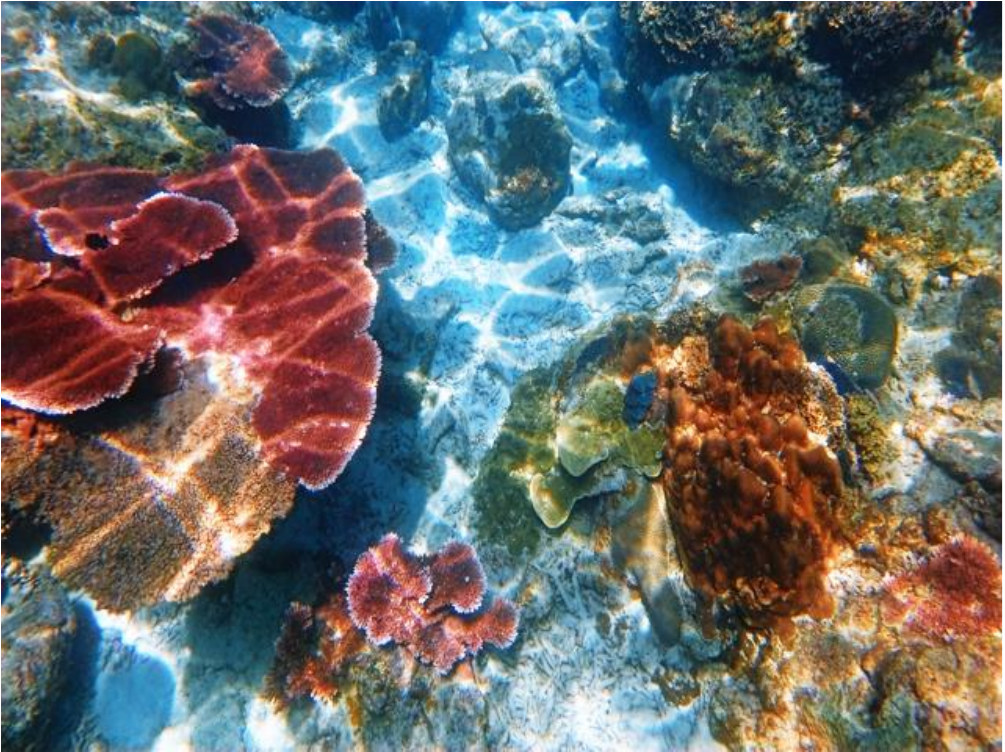} &
\includegraphics[width=\imgwidth,height=\imgheight]{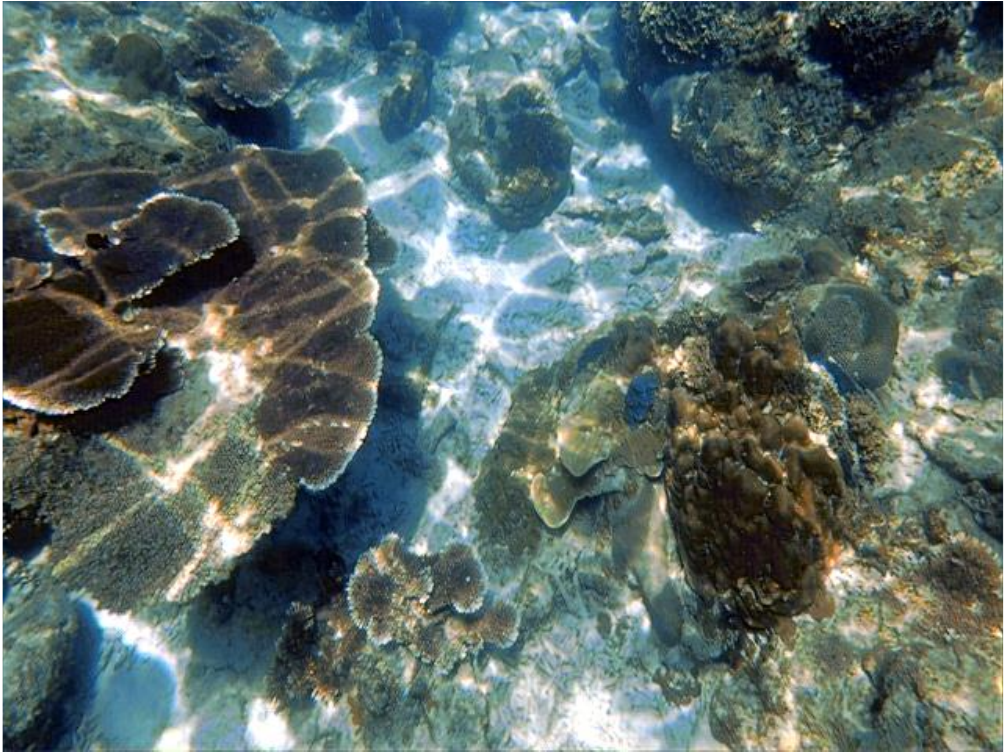} &
\includegraphics[width=\imgwidth,height=\imgheight]{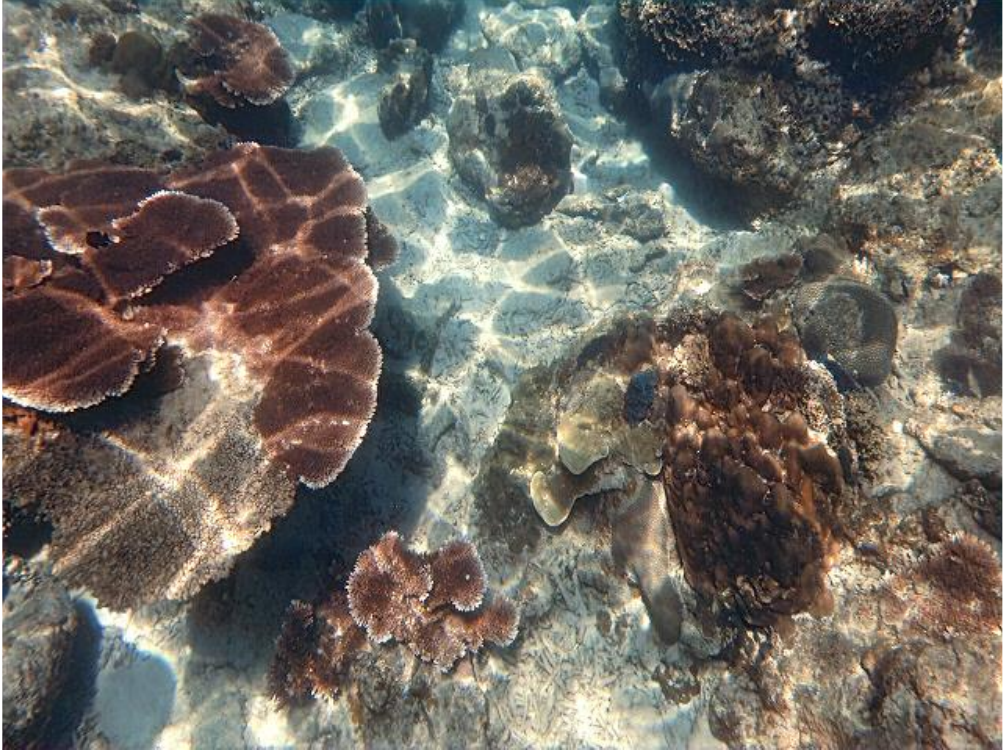} \\

\end{tabular}

\caption{Comparison on the UEIB-T90 dataset showing two samples with greenish and bluish color casts. AQUA-Net effectively corrects the color casts while preserving structural details and natural appearance.}
\label{fig: UEIB-T90}
\end{figure*}

This section presents a visual comparison across diverse datasets to assess the effectiveness of the AQUA-Net model in restoring natural color appearance. Figure \ref{fig: UEIB-T90} shows two representative samples, one with a greenish tint and another with a bluish tint, from the UEIB-T90 dataset. The proposed model effectively suppresses undesired color casts, preserves structural details, and maintains natural color balance. In contrast, SOTA methods, particularly CCL-Net and OUNet-JL, introduce noticeable color artifacts and artificial tones, which reduce visual fidelity. These observations demonstrate that the AQUA-Net model exhibits superior robustness and generalization capability under challenging color conditions.

\begin{figure*}[htbp!]
\centering

\setlength{\tabcolsep}{1.2pt}
\renewcommand{\arraystretch}{0.35}

\newcommand{\csixtyimgwidth}{0.13\textwidth}
\newcommand{\csixtyimgheight}{0.085\textwidth}

\begin{tabular}{ccccccc}

\textbf{\tiny Raw} &
\textbf{\tiny Fusion \cite{ancuti2017color}} &
\textbf{\tiny SMBL \cite{song2020enhancement}} &
\textbf{\tiny MLLE \cite{zhang2022underwater}} &
\textbf{\tiny WWPF \cite{zhang2023underwater}} &
\textbf{\tiny WFAC \cite{zhang2025underwater1}} &
\textbf{\tiny WaterNet \cite{li2019underwater}} \\[3pt]

\includegraphics[width=\csixtyimgwidth,height=\csixtyimgheight]{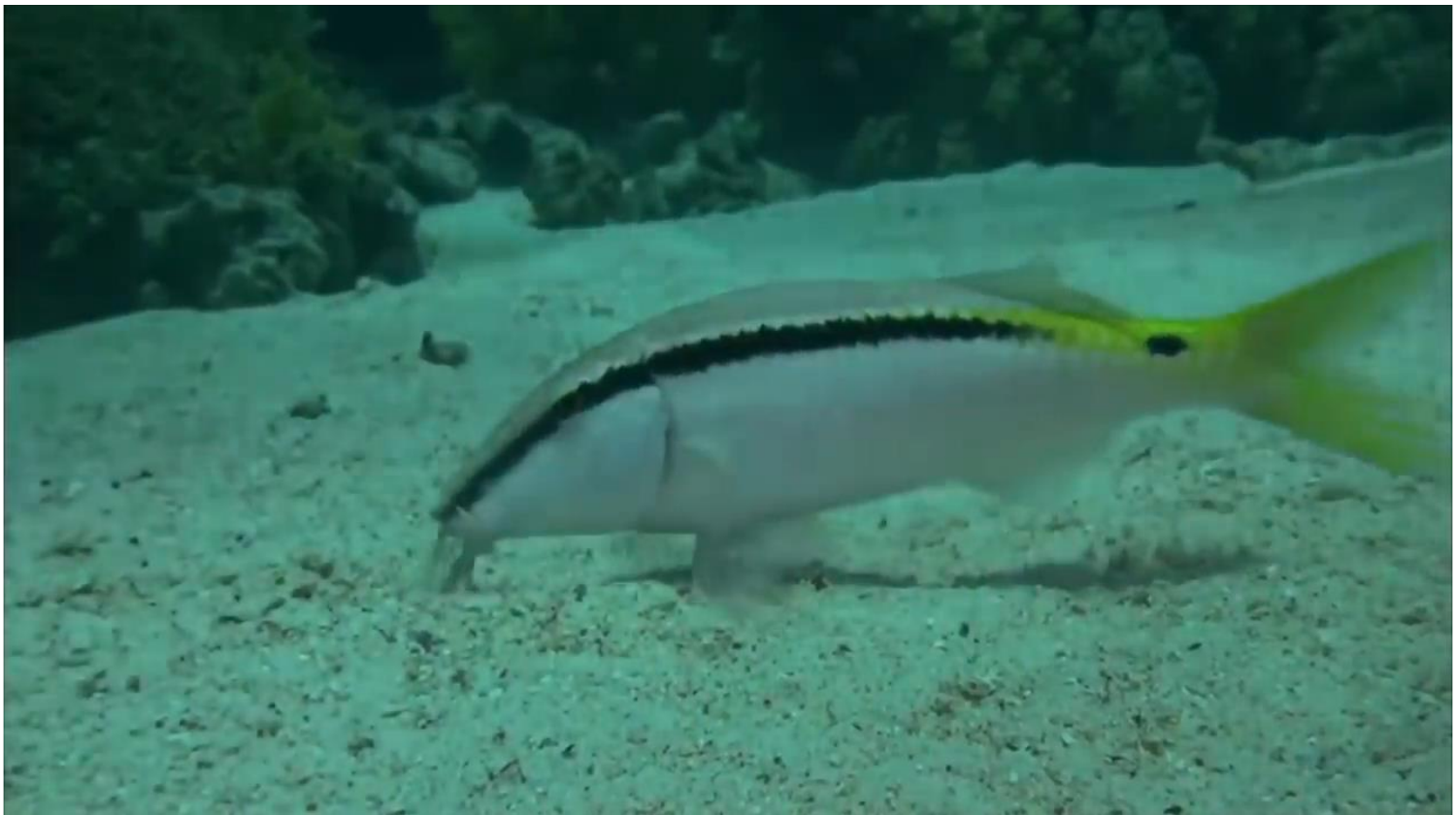} &
\includegraphics[width=\csixtyimgwidth,height=\csixtyimgheight]{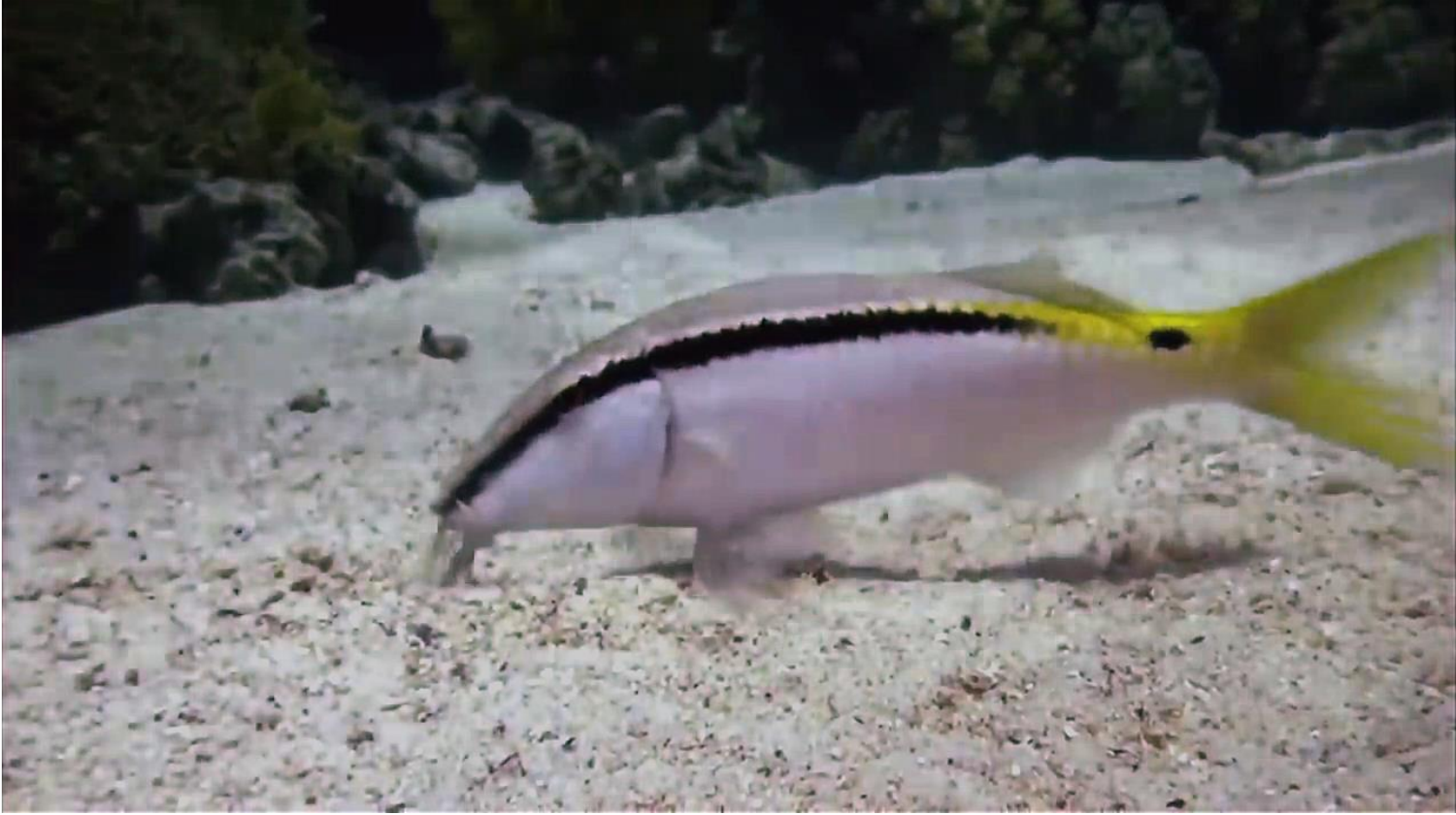} &
\includegraphics[width=\csixtyimgwidth,height=\csixtyimgheight]{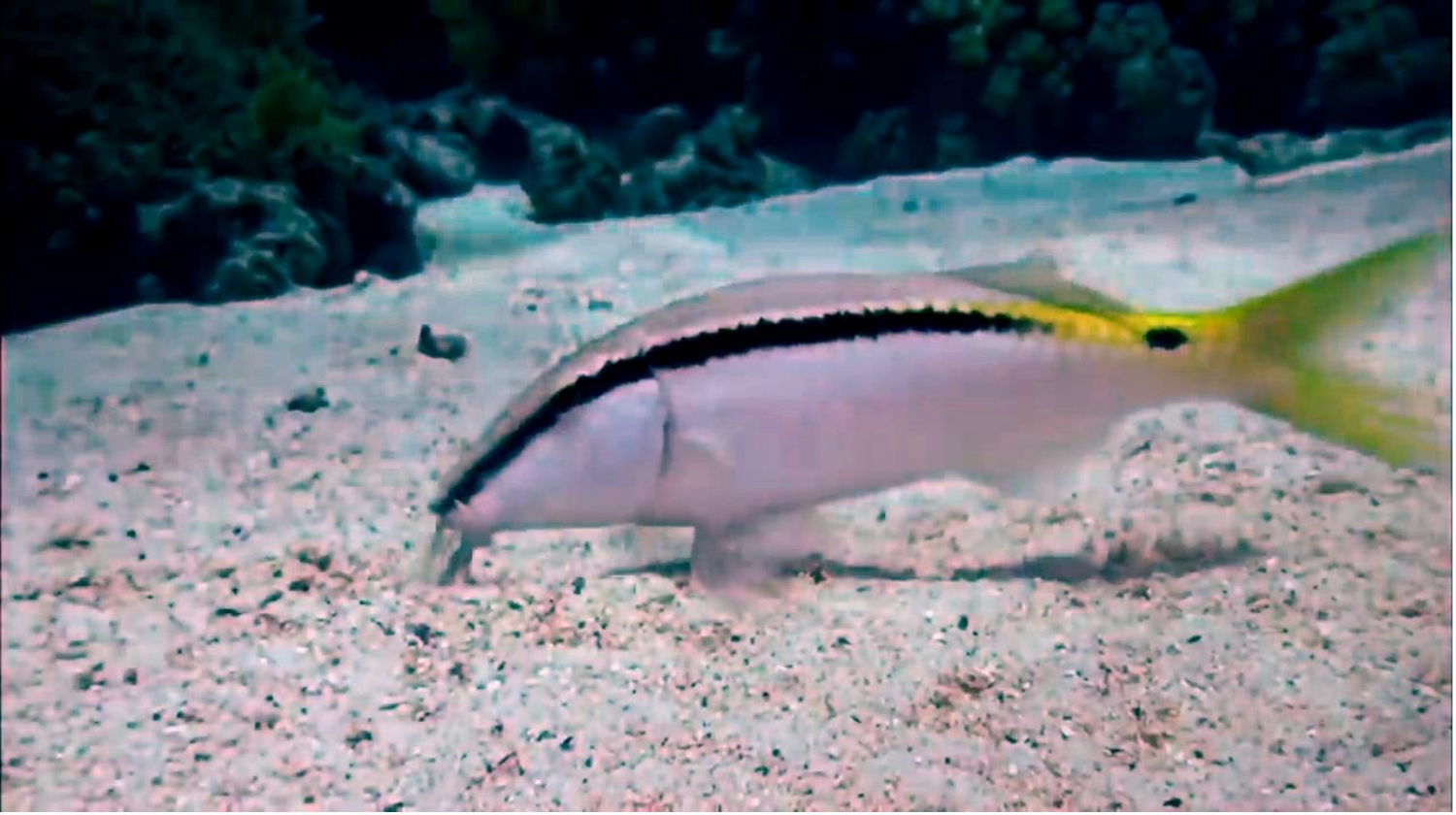} &
\includegraphics[width=\csixtyimgwidth,height=\csixtyimgheight]{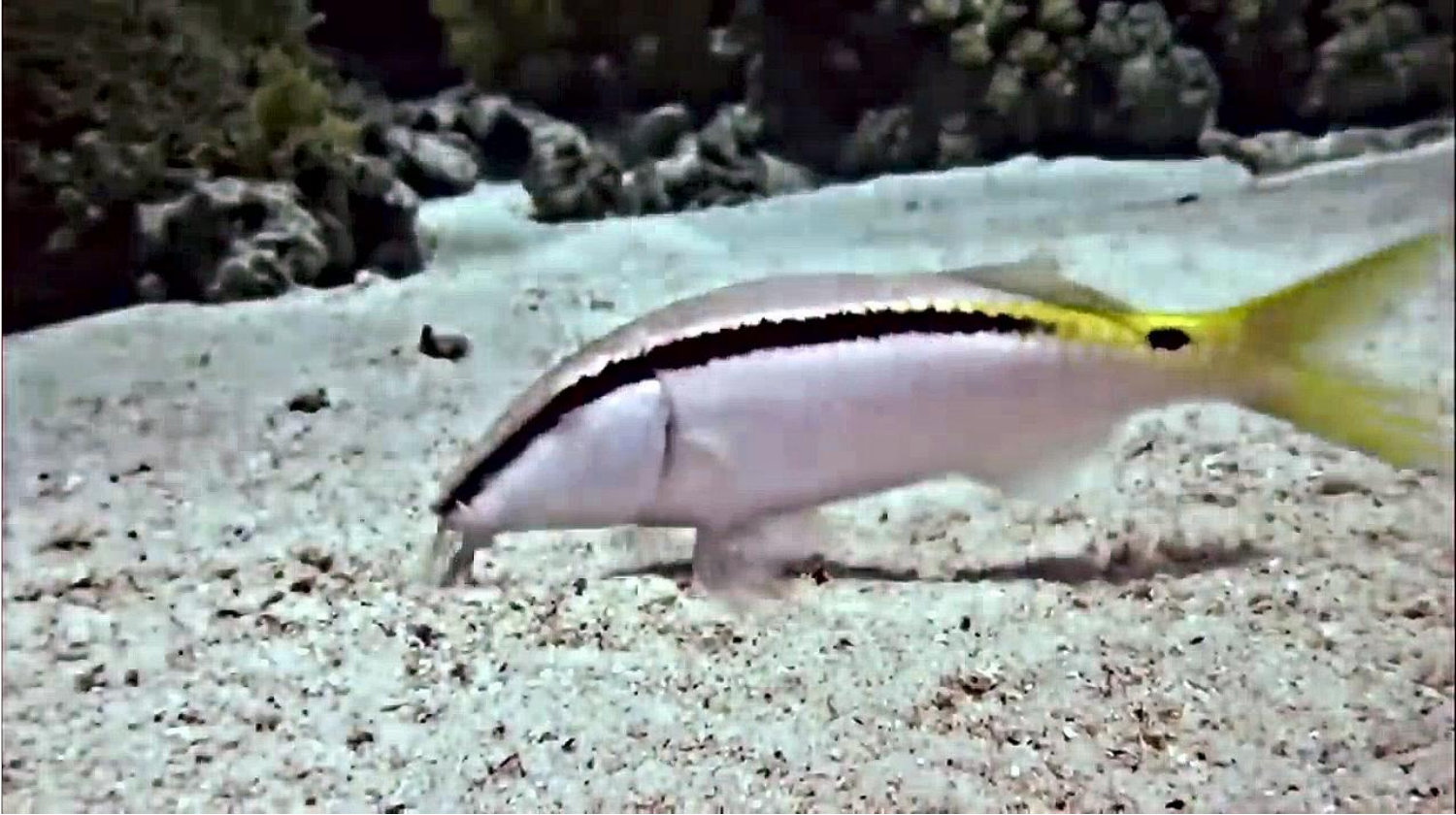} &
\includegraphics[width=\csixtyimgwidth,height=\csixtyimgheight]{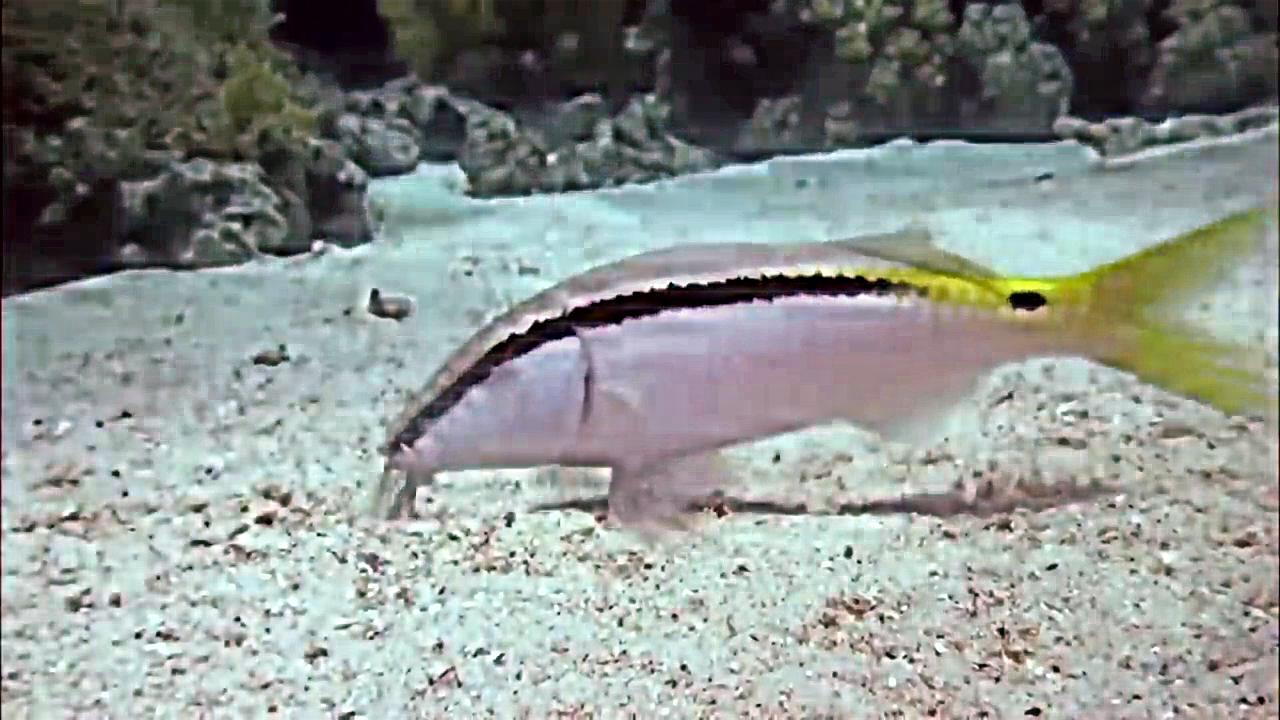} &
\includegraphics[width=\csixtyimgwidth,height=\csixtyimgheight]{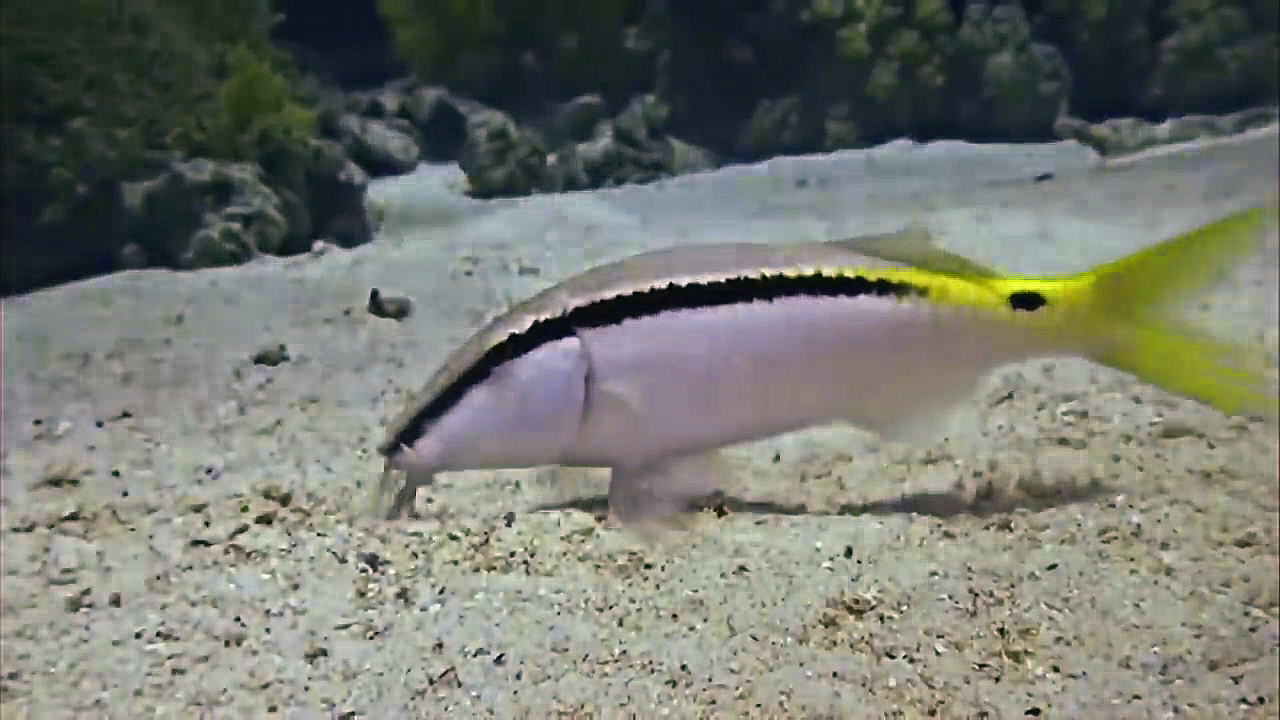} &
\includegraphics[width=\csixtyimgwidth,height=\csixtyimgheight]{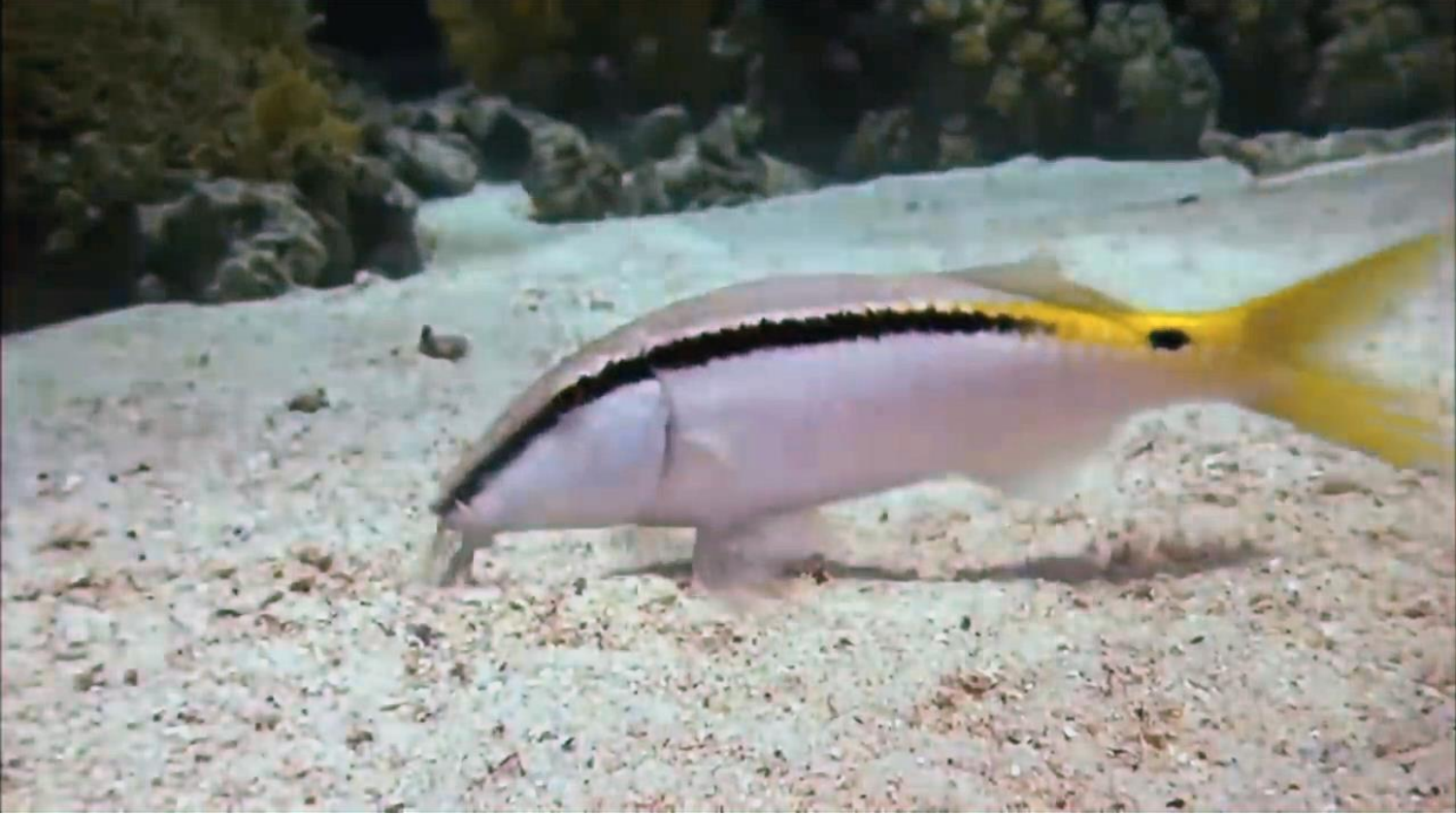} \\

\includegraphics[width=\csixtyimgwidth,height=\csixtyimgheight]{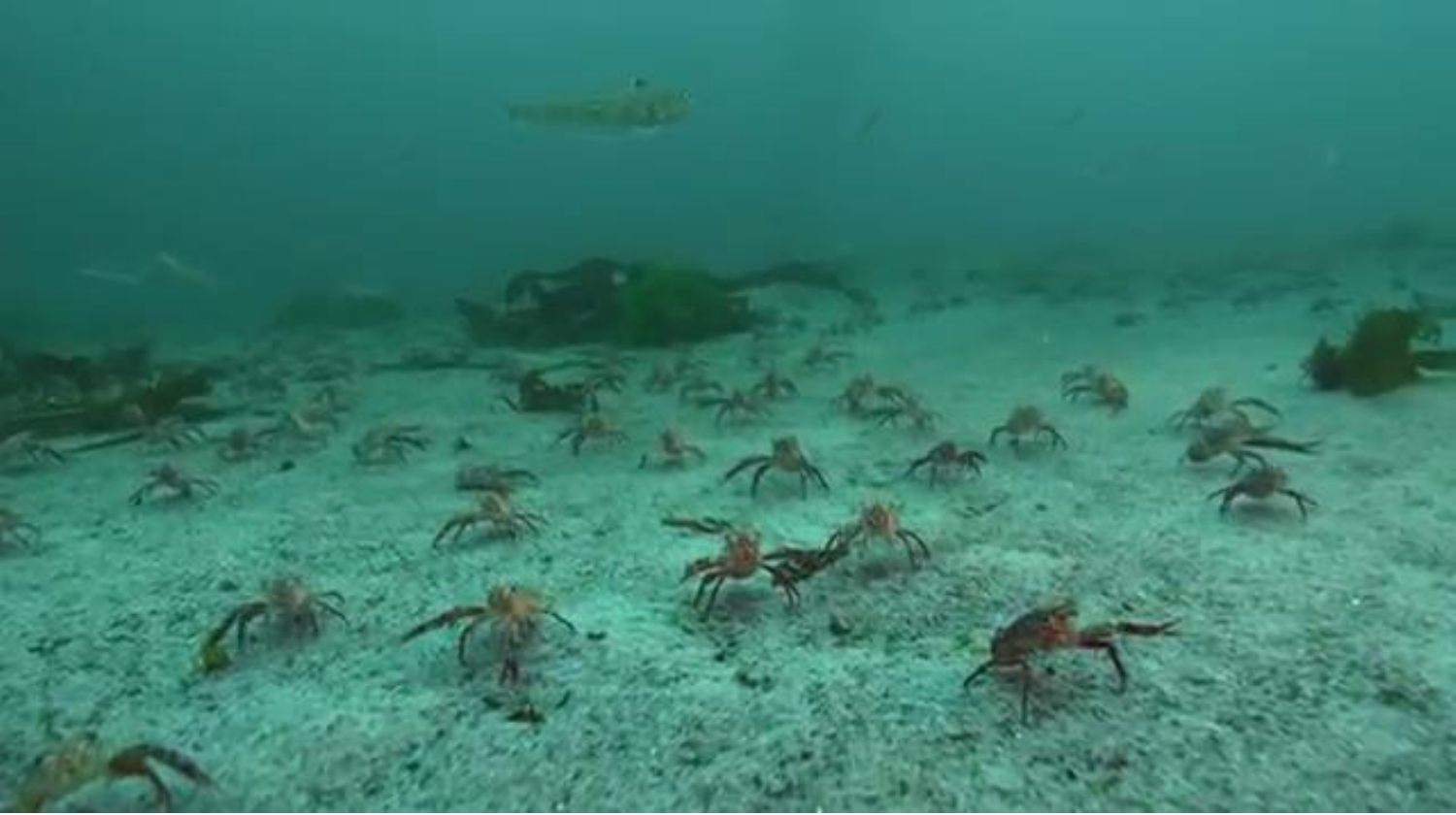} &
\includegraphics[width=\csixtyimgwidth,height=\csixtyimgheight]{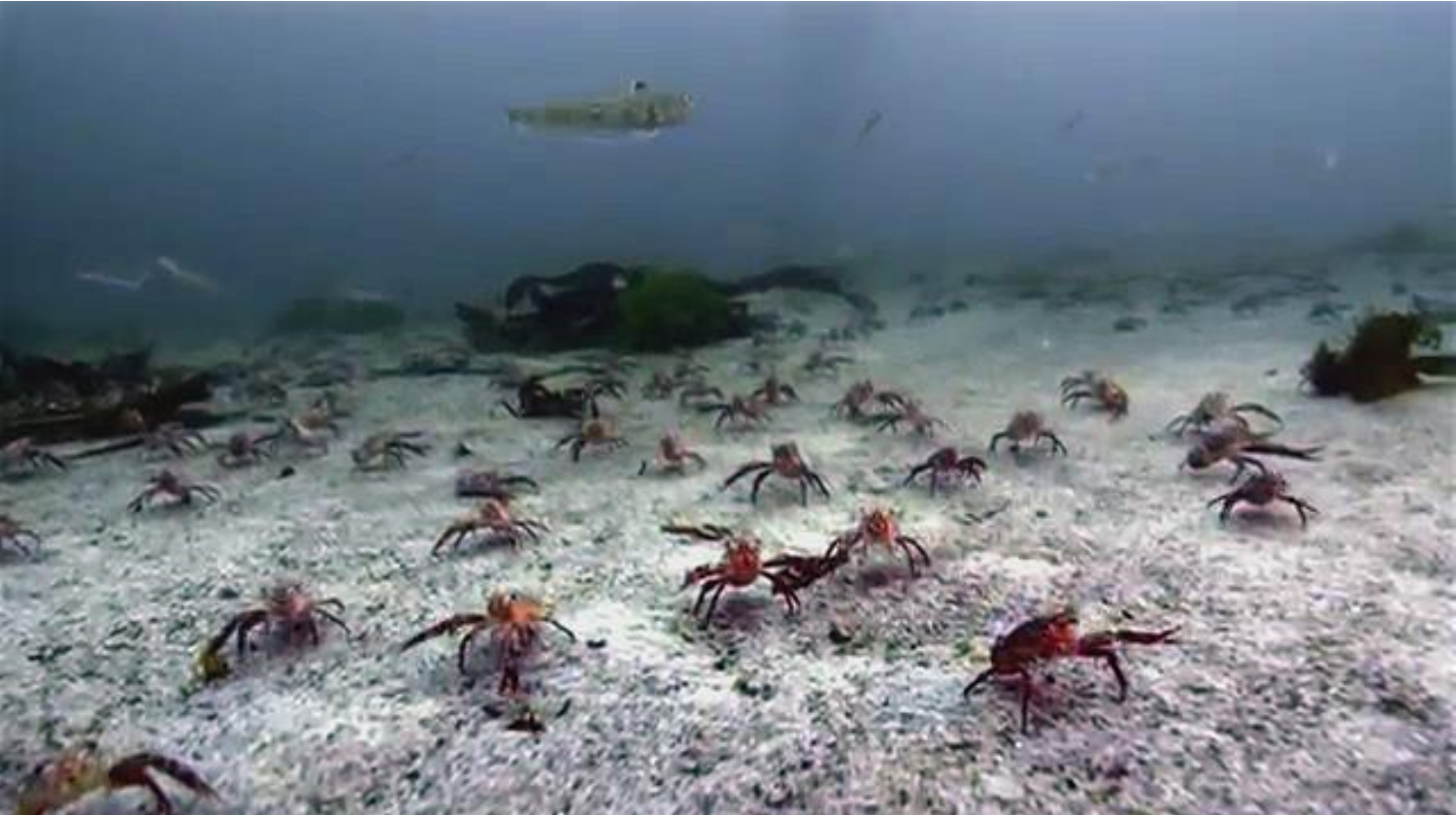} &
\includegraphics[width=\csixtyimgwidth,height=\csixtyimgheight]{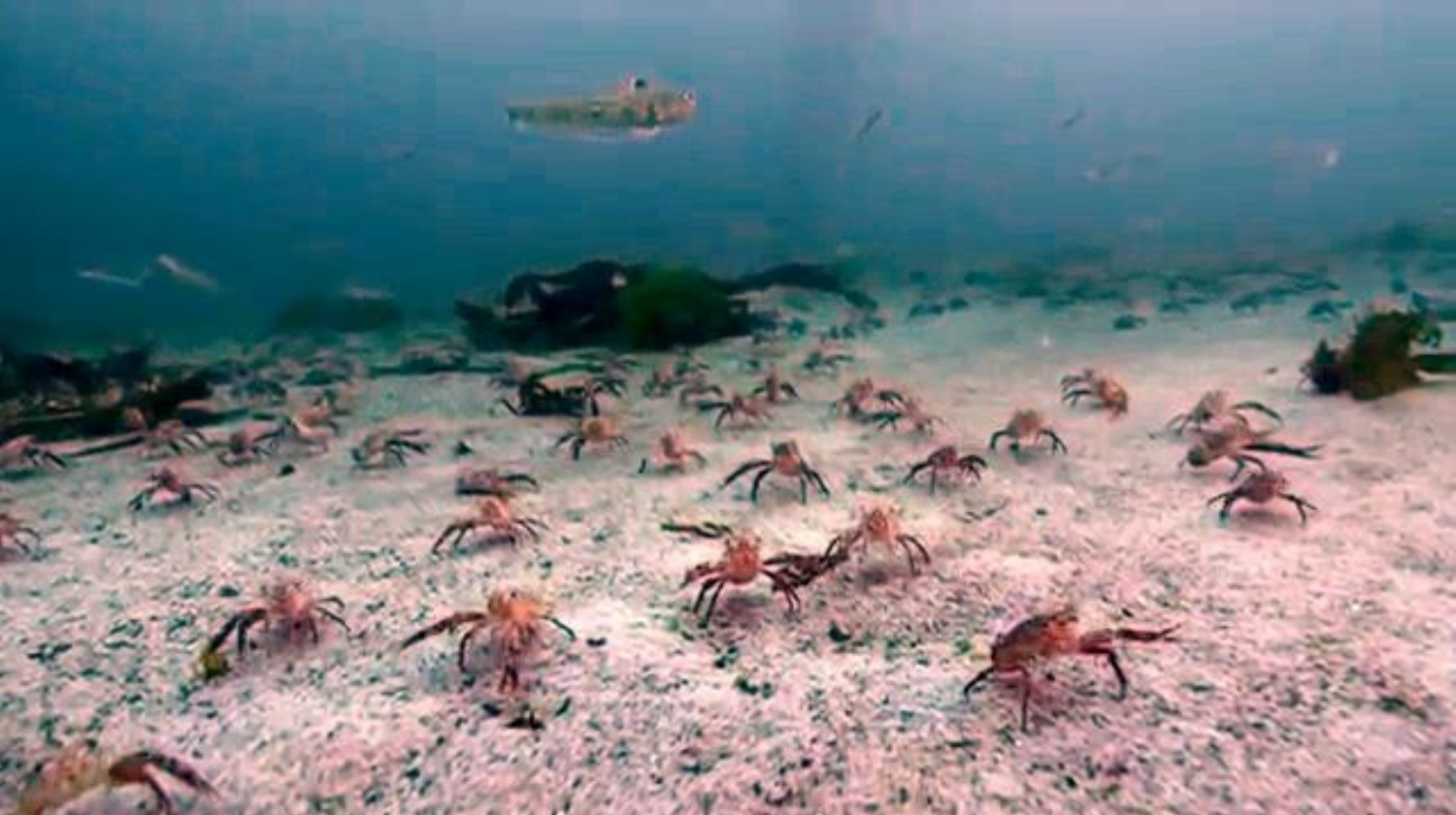} &
\includegraphics[width=\csixtyimgwidth,height=\csixtyimgheight]{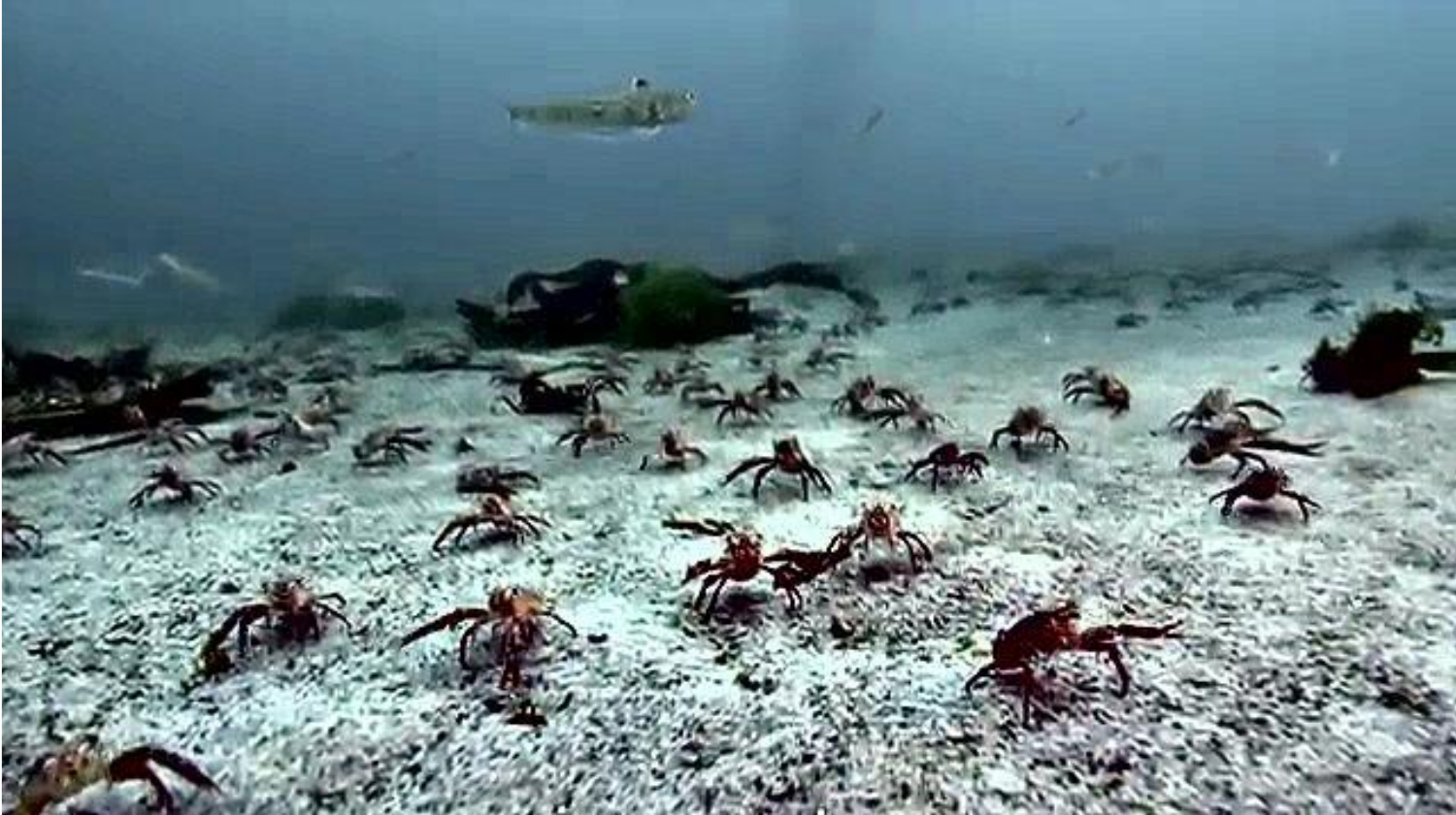} &
\includegraphics[width=\csixtyimgwidth,height=\csixtyimgheight]{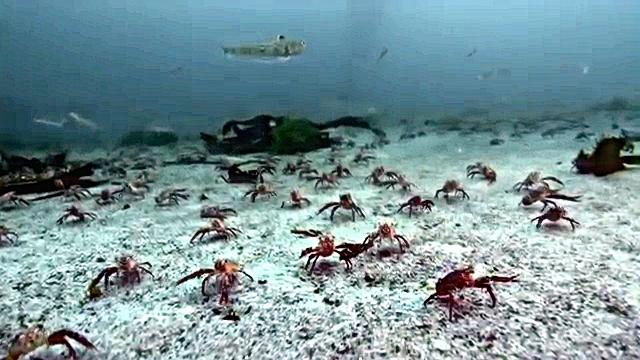} &
\includegraphics[width=\csixtyimgwidth,height=\csixtyimgheight]{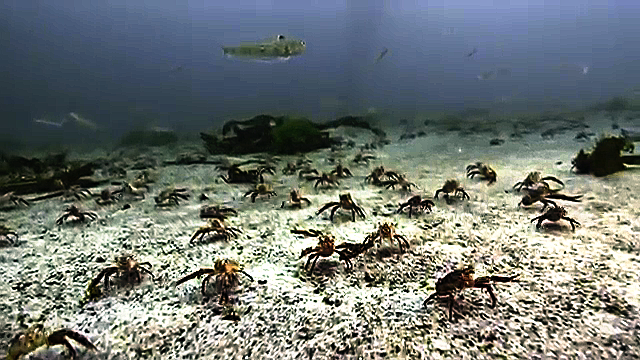} &
\includegraphics[width=\csixtyimgwidth,height=\csixtyimgheight]{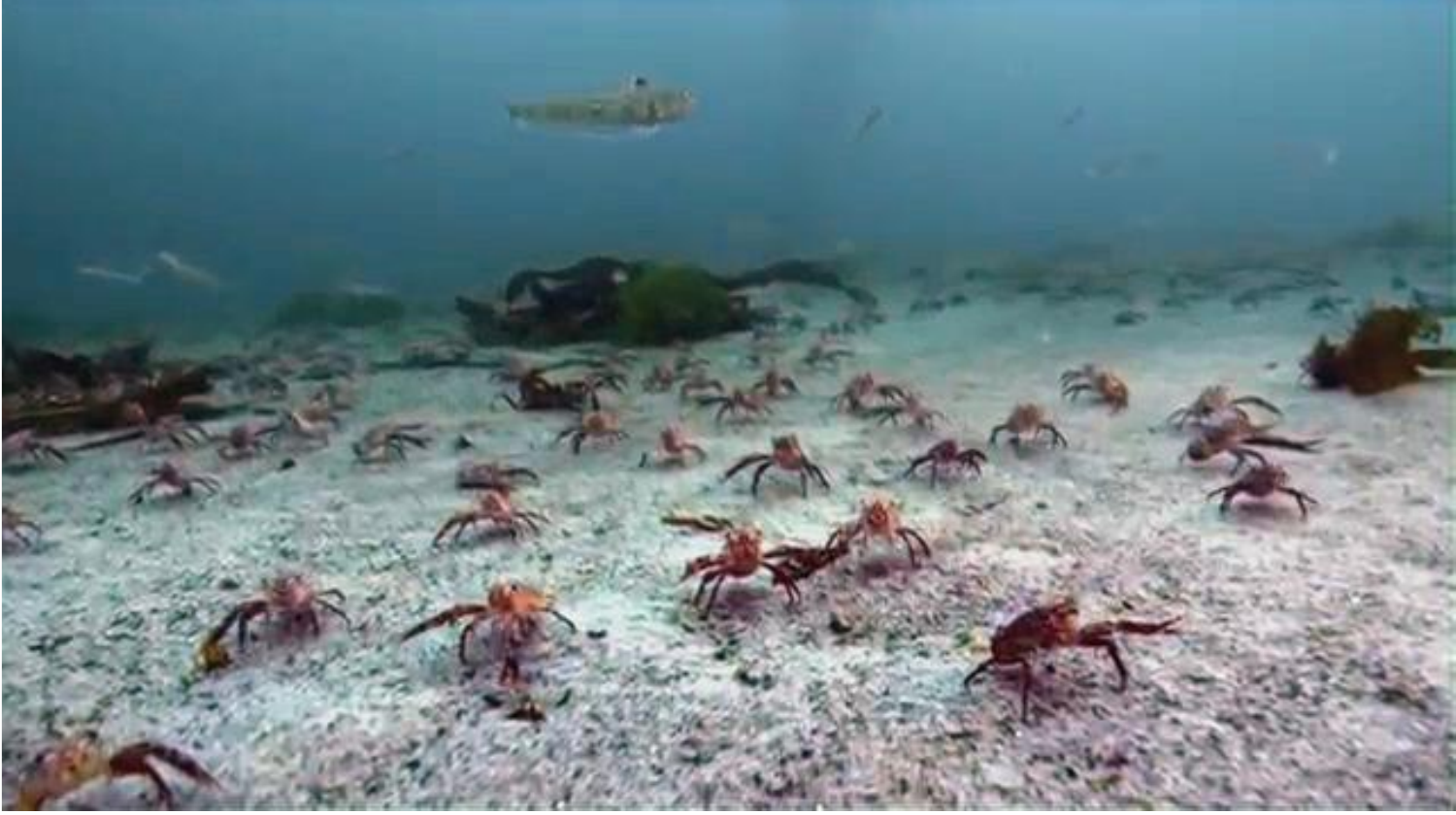} \\[5pt]

\textbf{\tiny UWCNN \cite{li2020underwater}} &
\textbf{\tiny PUIE \cite{fu2022uncertainty}} &
\textbf{\tiny TACL \cite{liu2022twin}} &
\textbf{\tiny NU2Net \cite{guo2023underwater}} &
\textbf{\tiny CCL-Net \cite{liu2024underwater}} &
\textbf{\tiny OUNet-JL \cite{wang2025optimized}} &
\textbf{\tiny AQUA-Net} \\[3pt]

\includegraphics[width=\csixtyimgwidth,height=\csixtyimgheight]{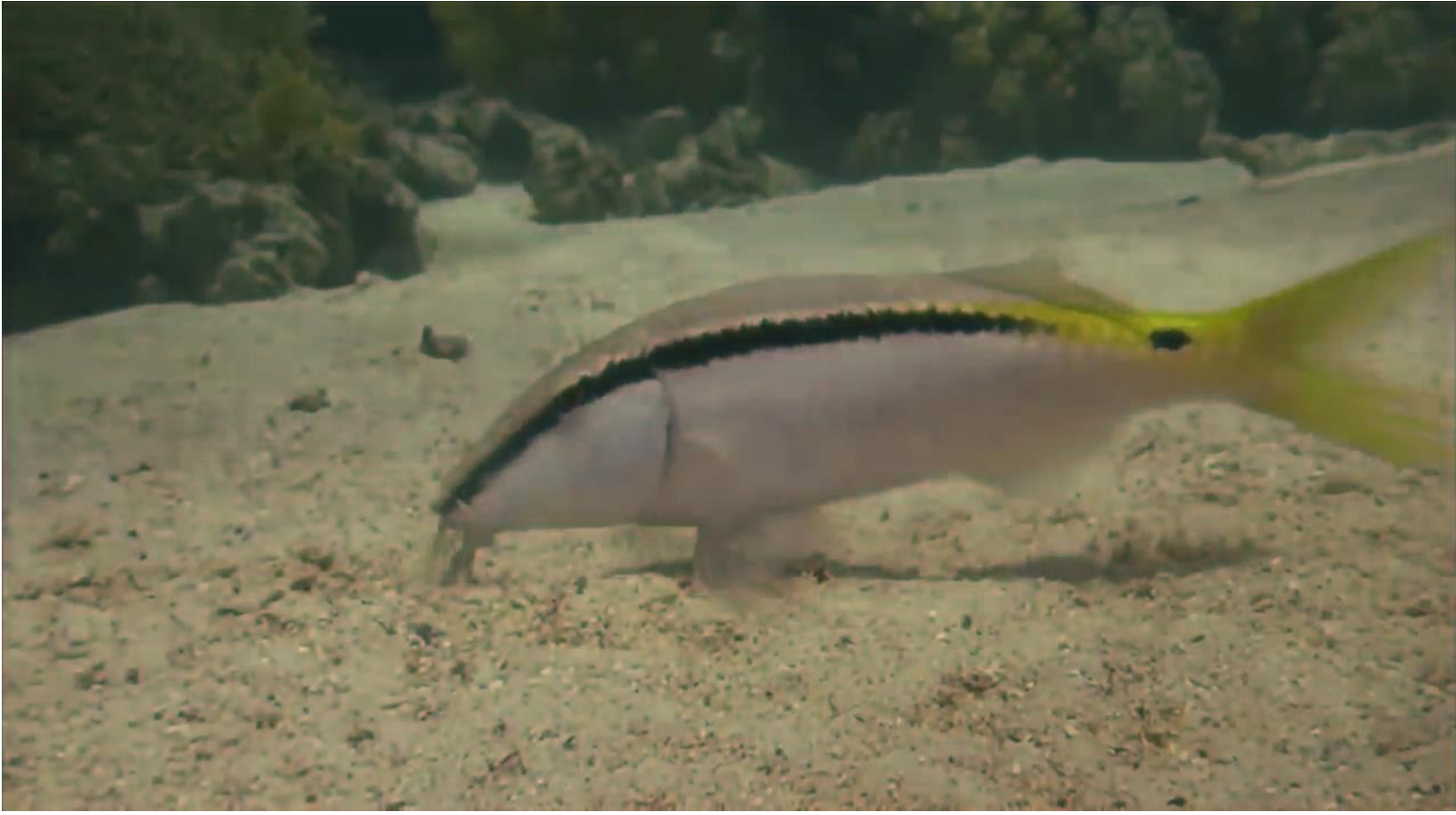} &
\includegraphics[width=\csixtyimgwidth,height=\csixtyimgheight]{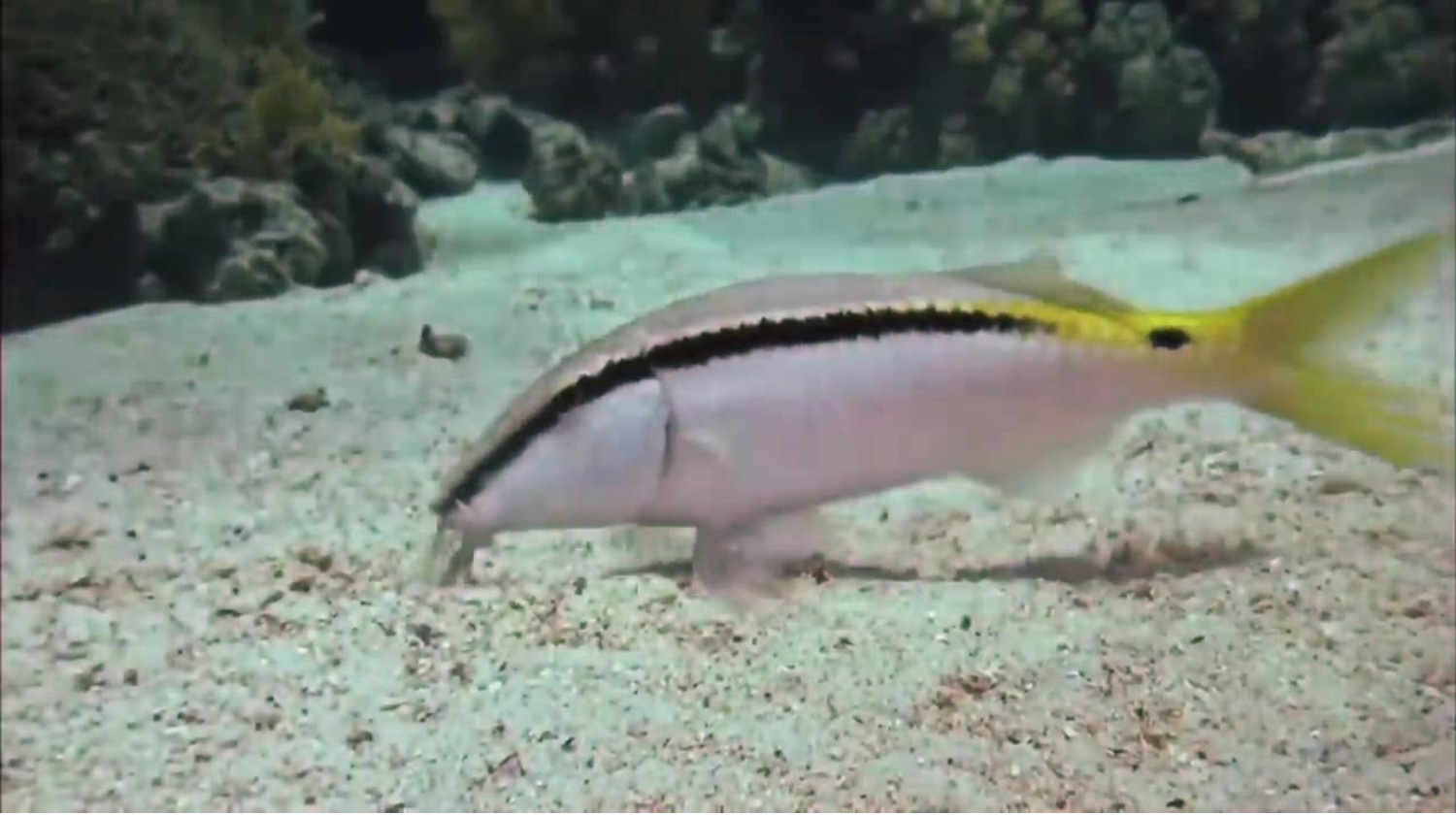} &
\includegraphics[width=\csixtyimgwidth,height=\csixtyimgheight]{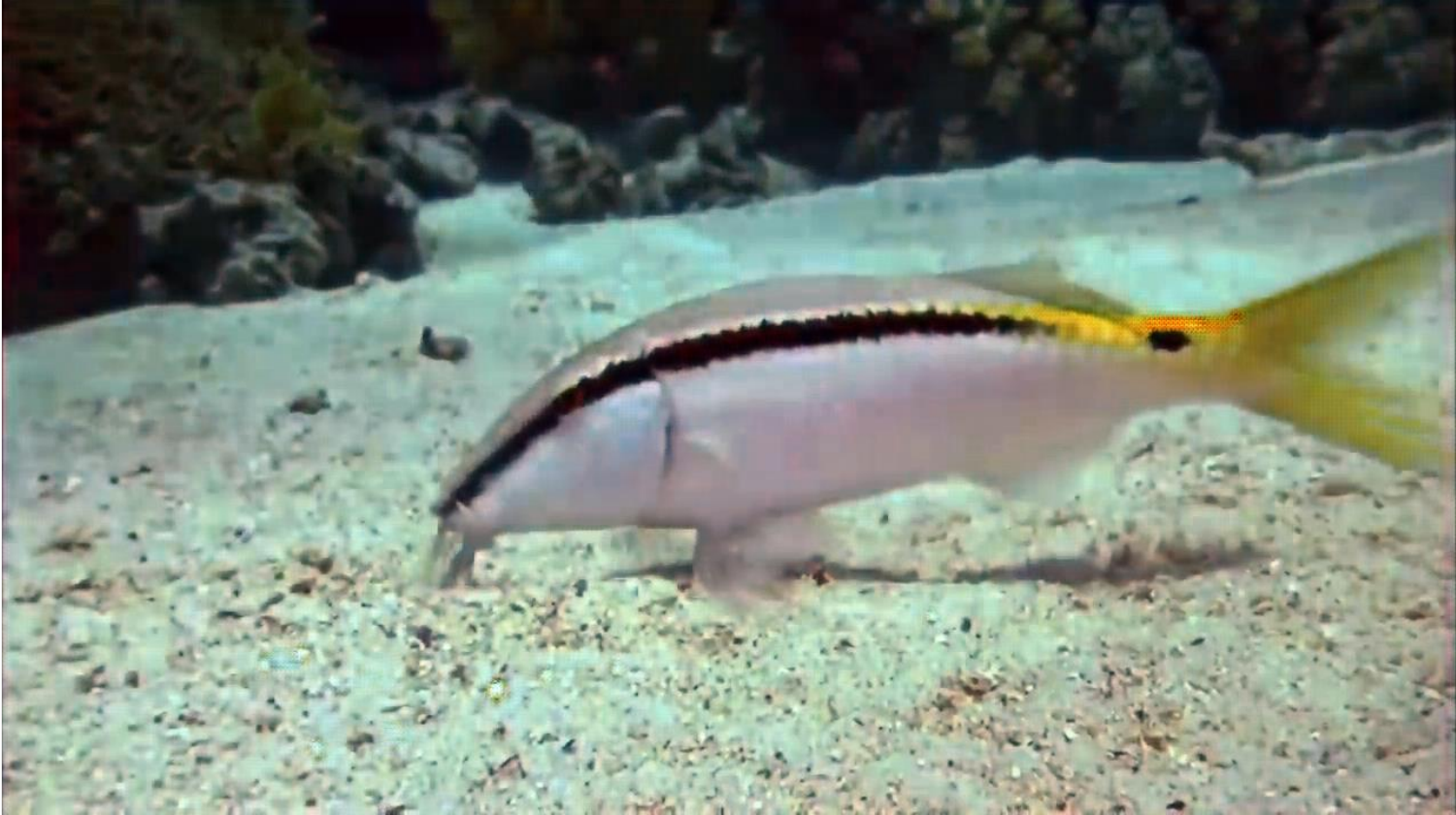} &
\includegraphics[width=\csixtyimgwidth,height=\csixtyimgheight]{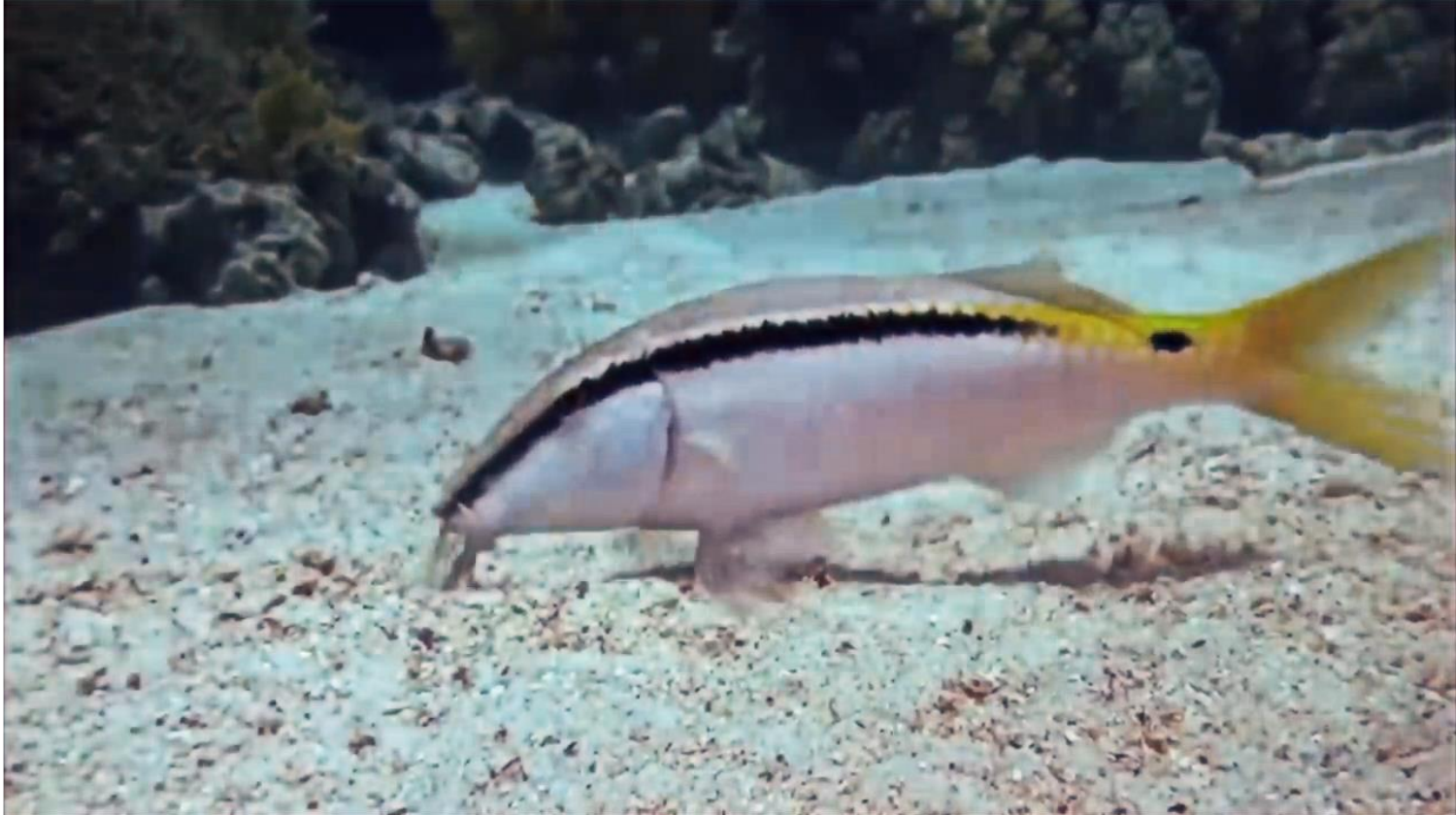} &
\includegraphics[width=\csixtyimgwidth,height=\csixtyimgheight]{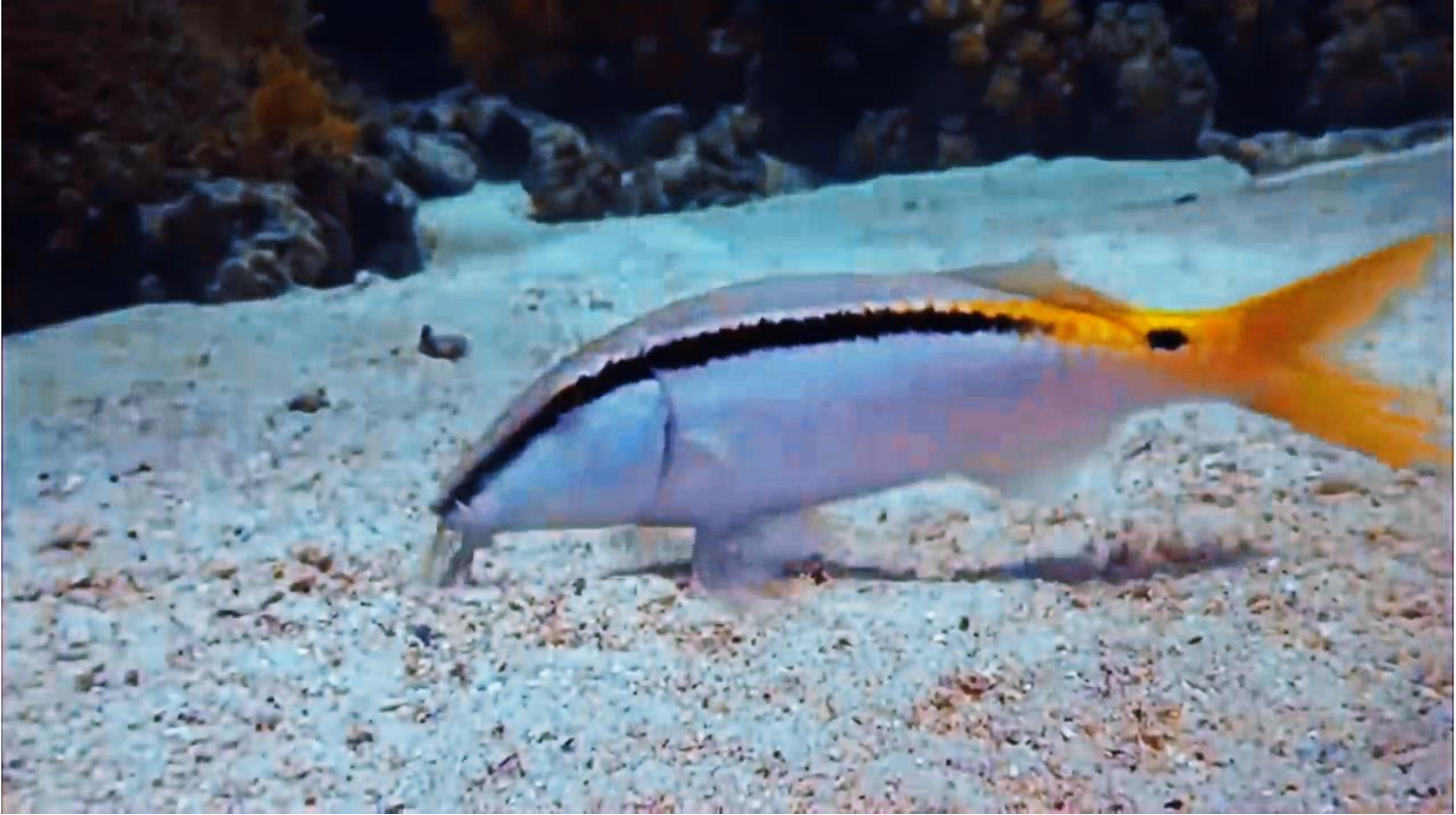} &
\includegraphics[width=\csixtyimgwidth,height=\csixtyimgheight]{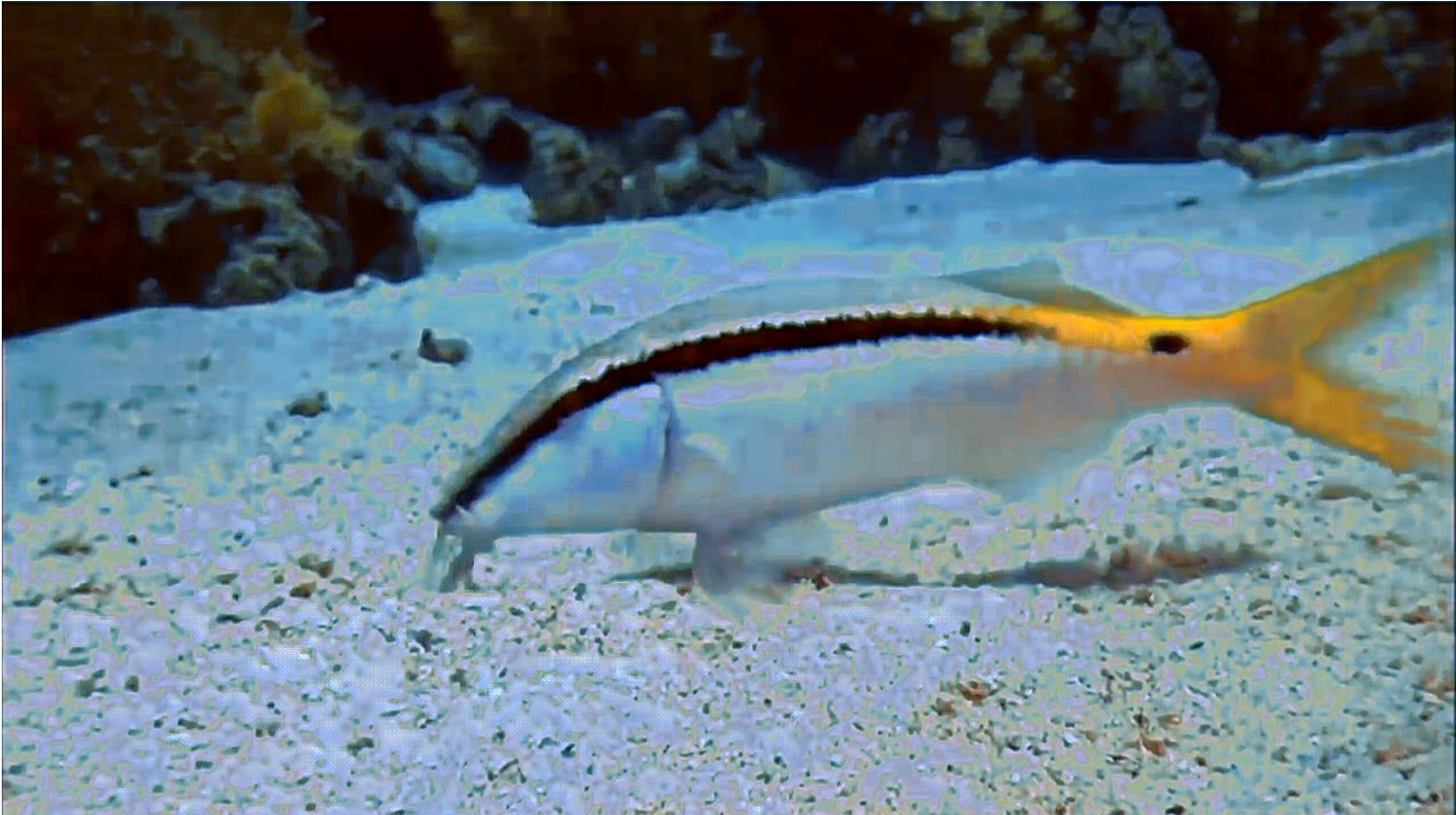} &
\includegraphics[width=\csixtyimgwidth,height=\csixtyimgheight]{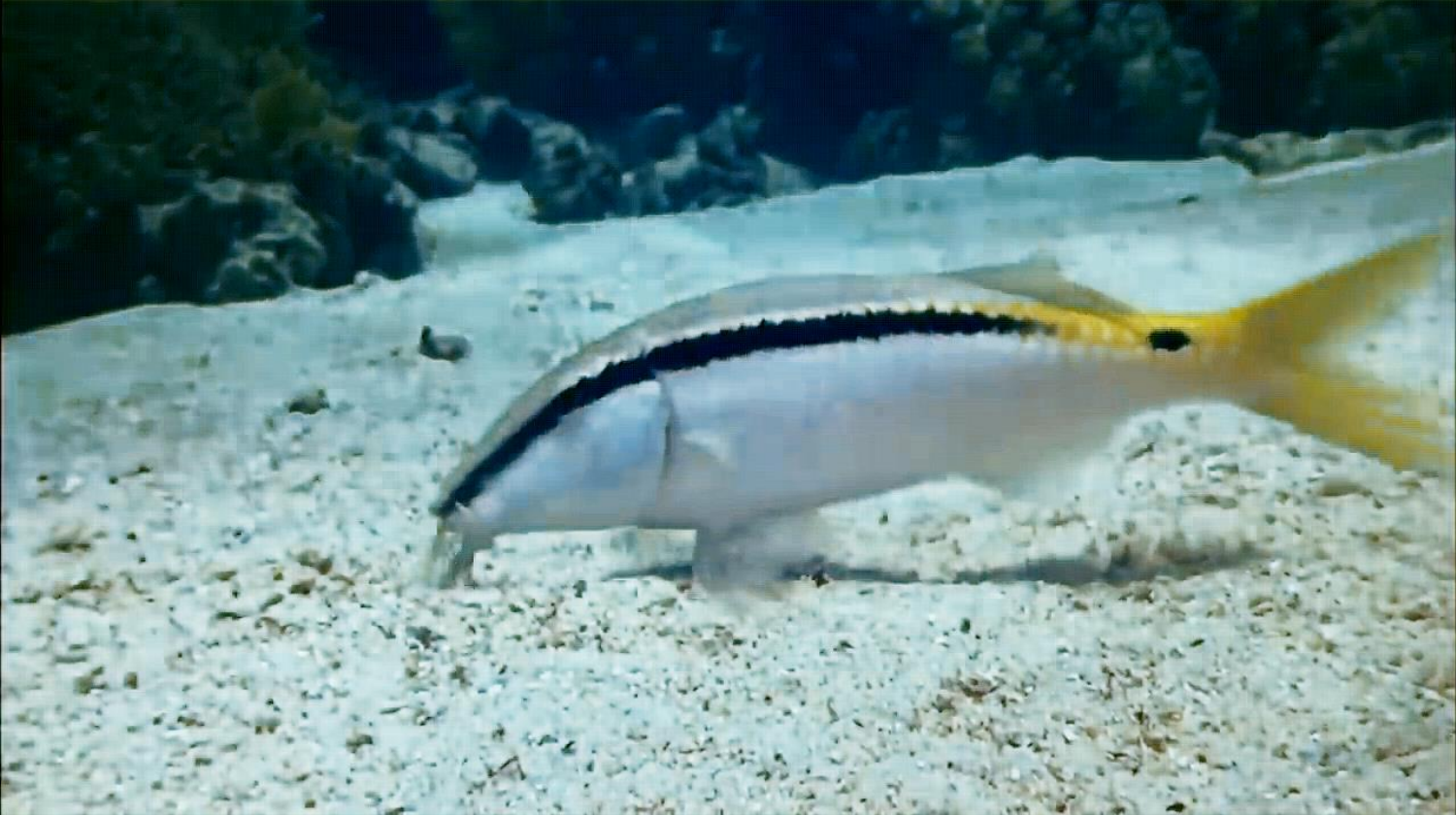} \\

\includegraphics[width=\csixtyimgwidth,height=\csixtyimgheight]{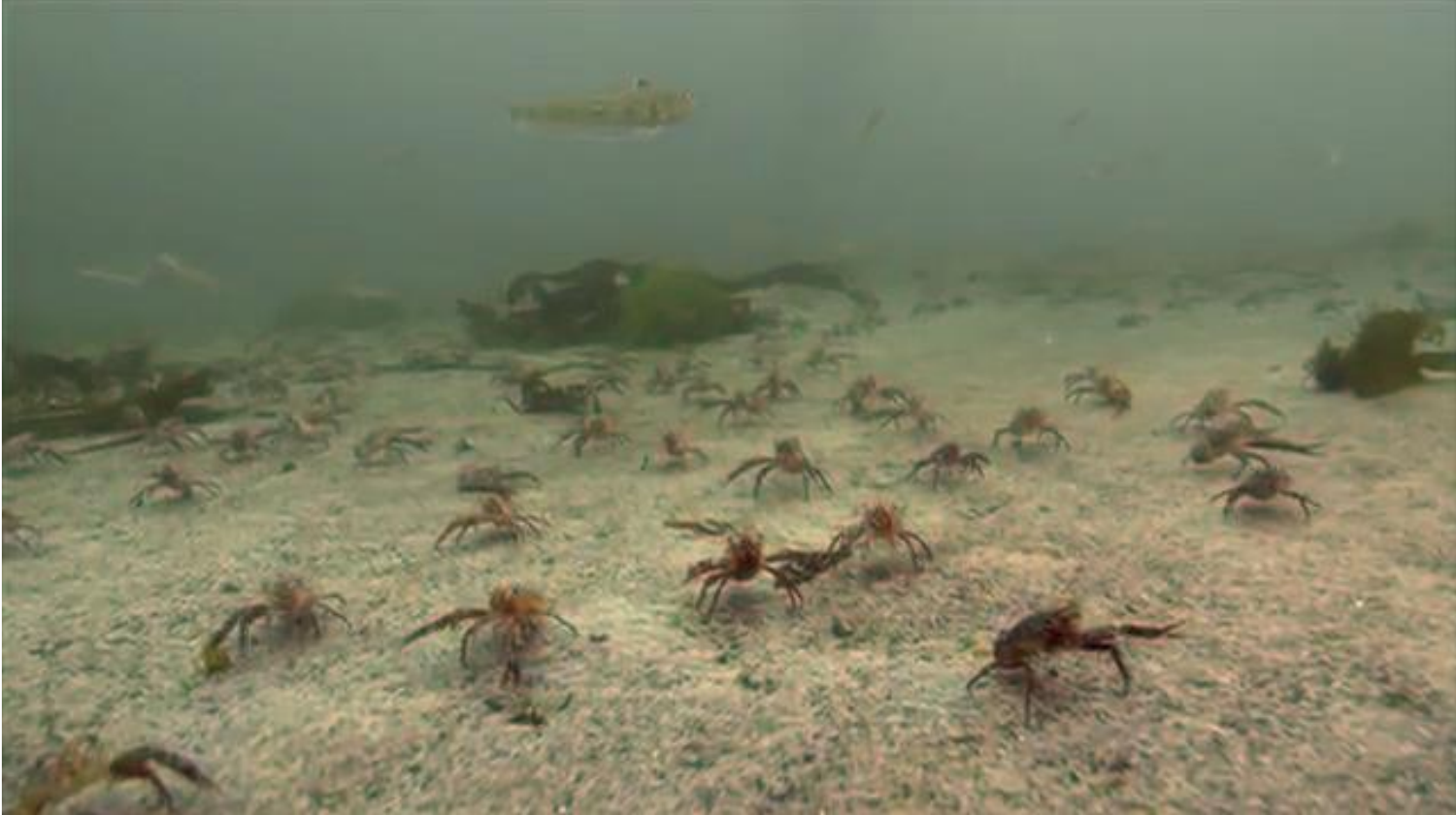} &
\includegraphics[width=\csixtyimgwidth,height=\csixtyimgheight]{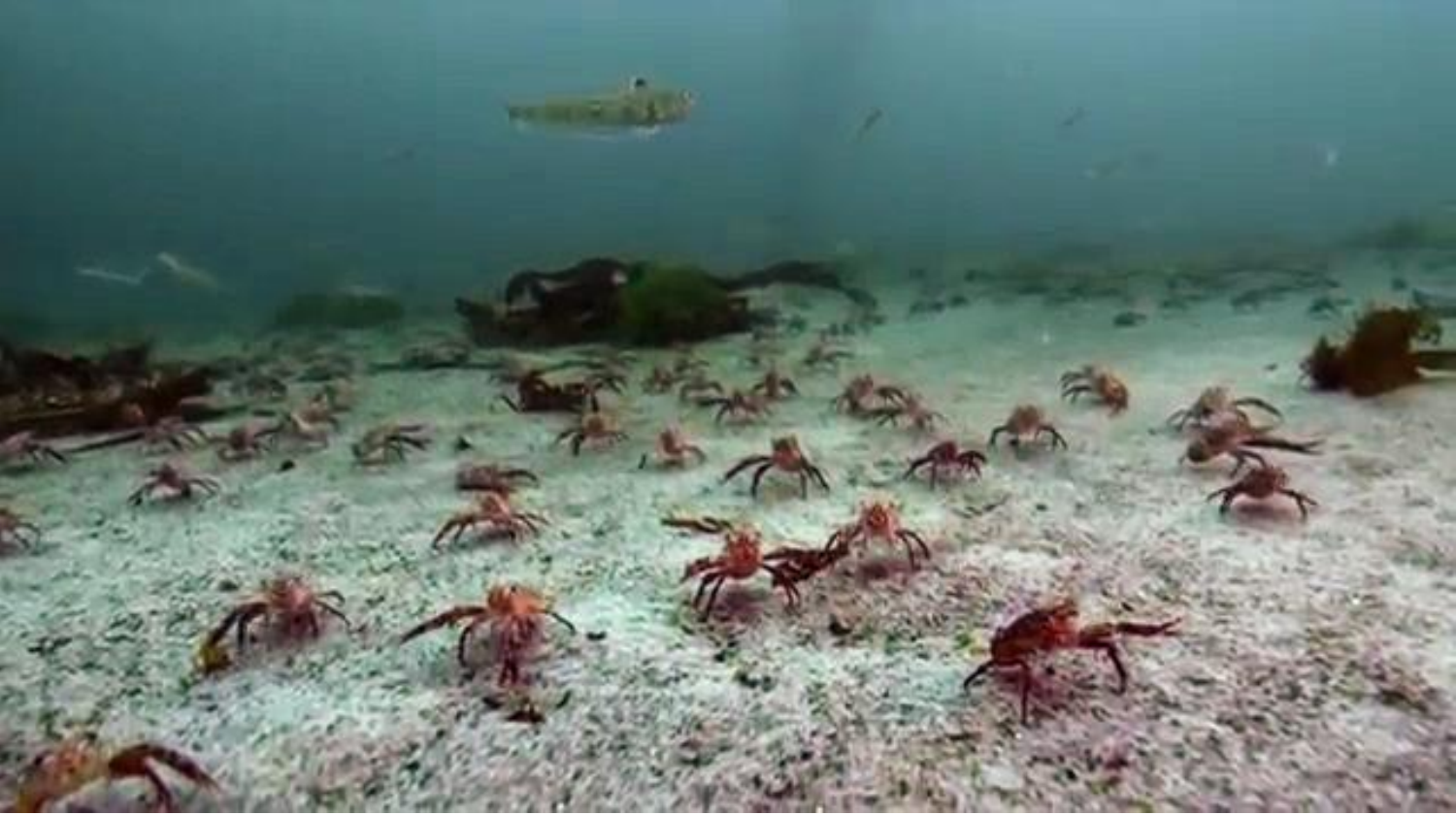} &
\includegraphics[width=\csixtyimgwidth,height=\csixtyimgheight]{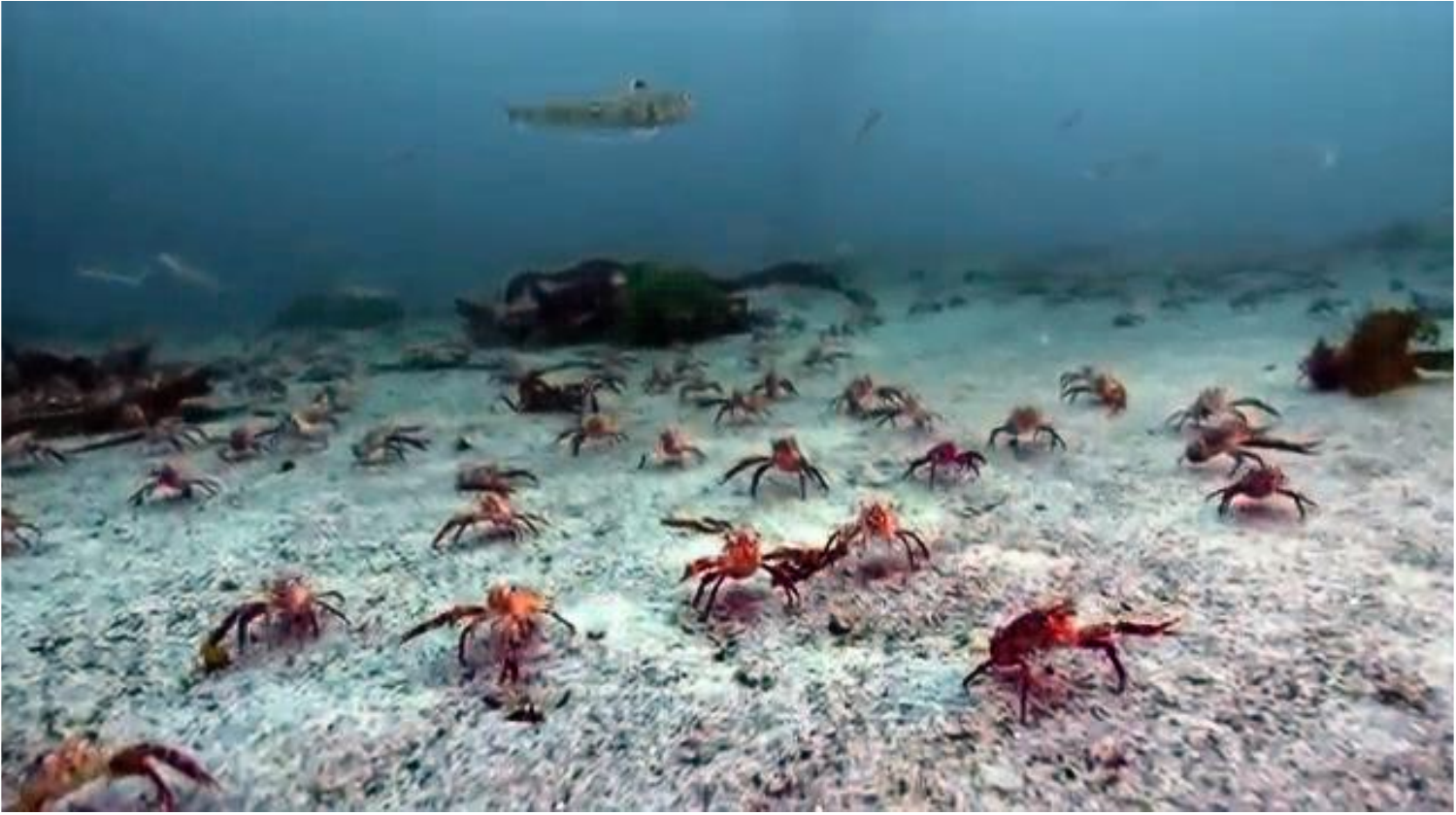} &
\includegraphics[width=\csixtyimgwidth,height=\csixtyimgheight]{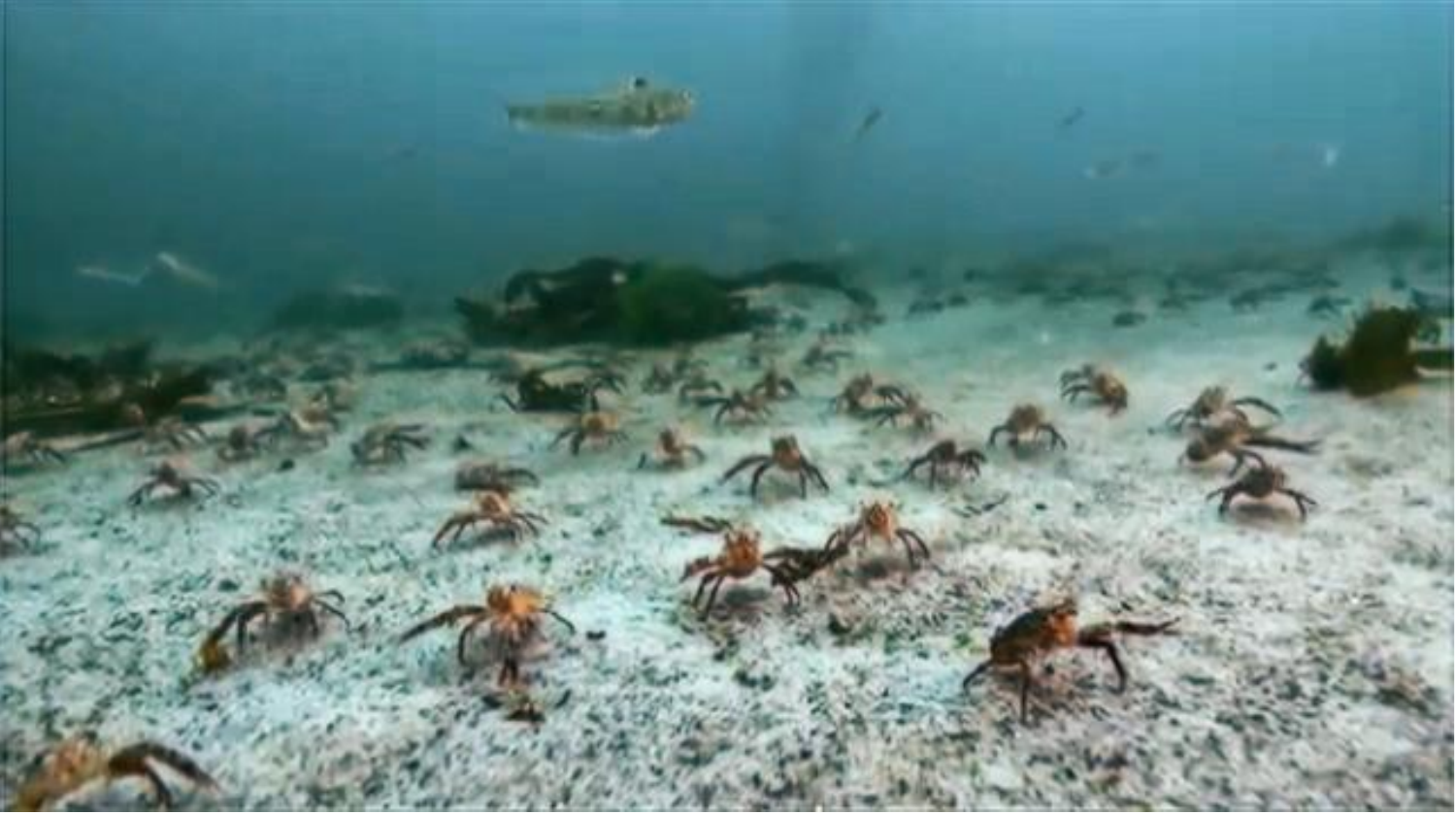} &
\includegraphics[width=\csixtyimgwidth,height=\csixtyimgheight]{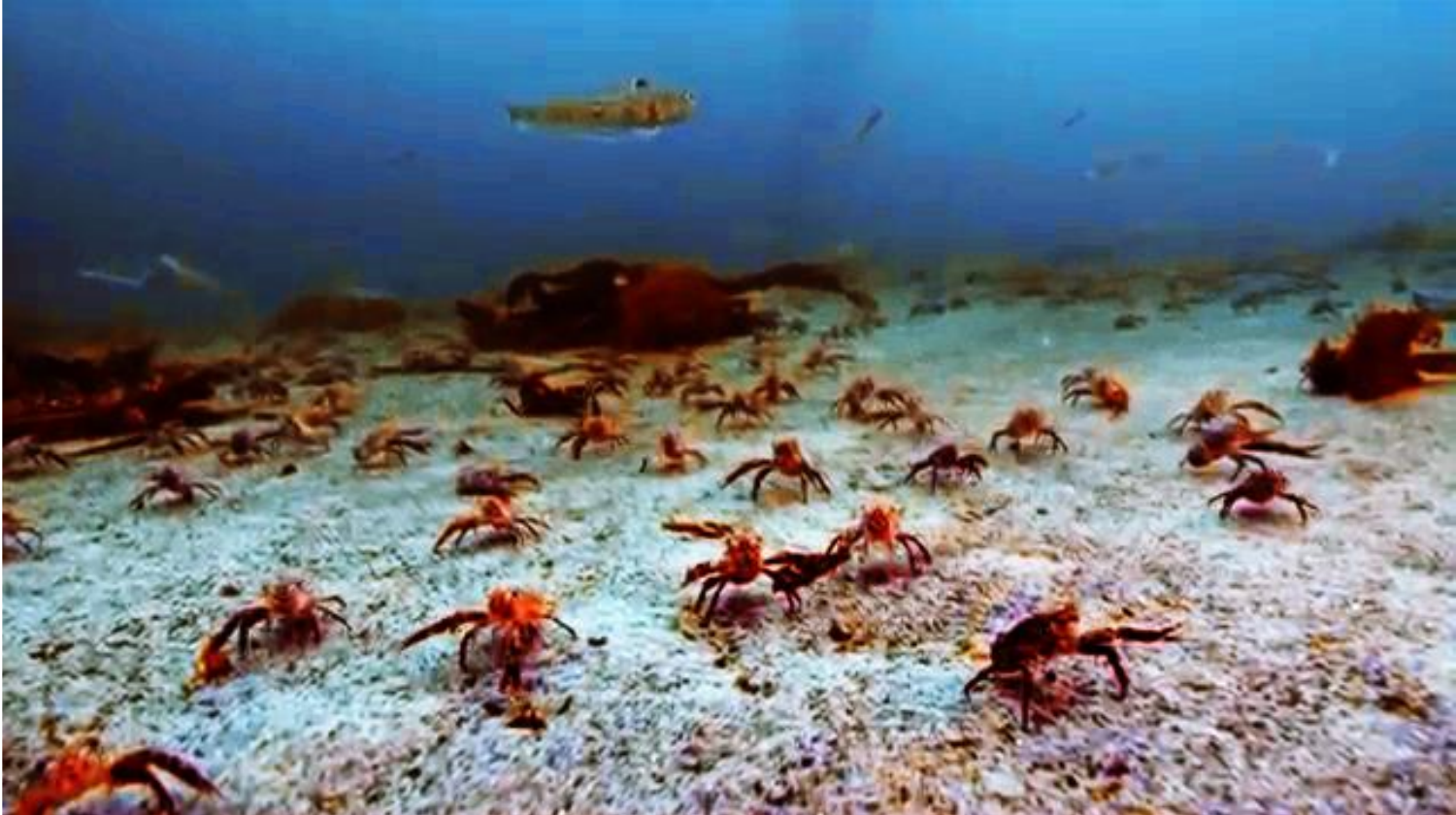} &
\includegraphics[width=\csixtyimgwidth,height=\csixtyimgheight]{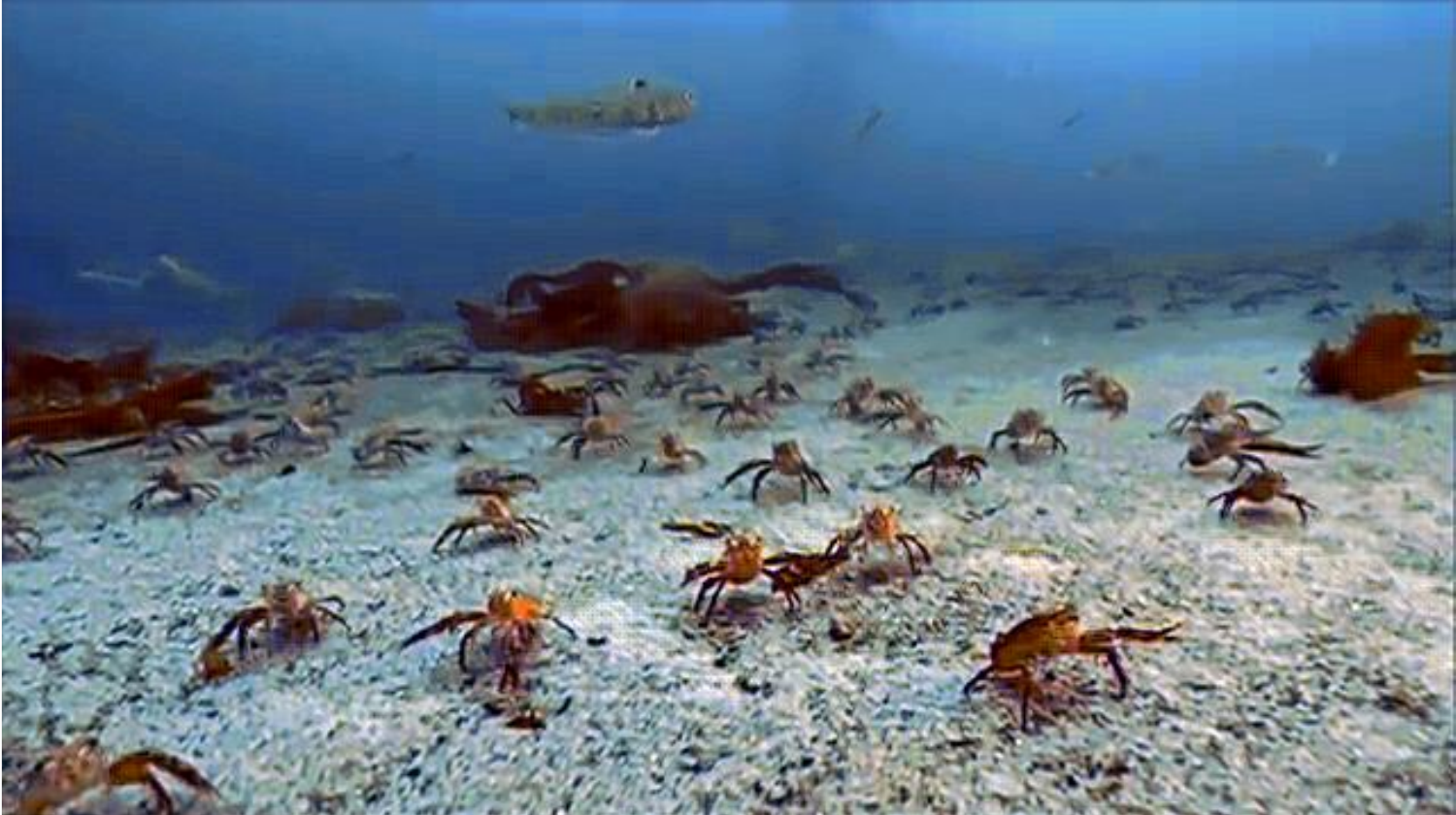} &
\includegraphics[width=\csixtyimgwidth,height=\csixtyimgheight]{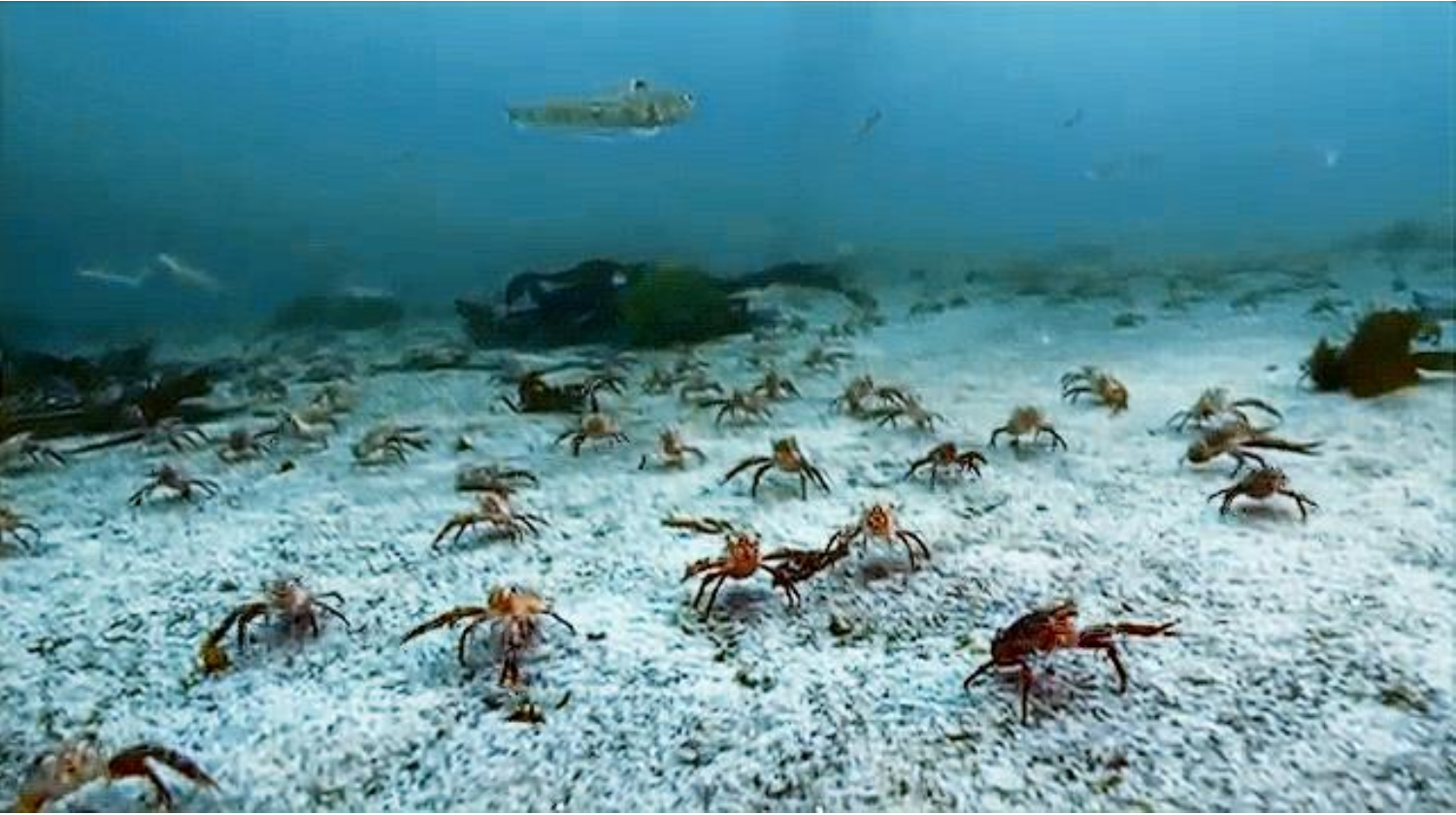} \\

\end{tabular}

\caption{Comparison of the UEIB C60 challenging UIWs. AQUA-Net restores colors and maintains visual consistency.}
\label{fig:UIEB-C60}
\end{figure*}

\textbf{UEIB-C60:} We evaluate our model on a more challenging dataset characterized by strong back scattering and significant color deviations, as shown in Figure \ref{fig:UIEB-C60}. The proposed model achieves superior results compared to SOTA models. In particular, CCL-Net and OUNet-JL introduce a noticeably more yellowish tint in the tail of the fish. In contrast, our model effectively restores natural colors, preserves structural details, and maintains a visually consistent appearance across the entire image, demonstrating robustness under difficult imaging conditions.

\begin{figure*}[t!]
\centering

\setlength{\tabcolsep}{1.2pt}
\renewcommand{\arraystretch}{0.35}

\newcommand{\ruieimgwidth}{0.13\textwidth}
\newcommand{\ruieimgheight}{0.085\textwidth}

\begin{tabular}{ccccccc}

\textbf{\tiny Raw} &
\textbf{\tiny Fusion \cite{ancuti2017color}} &
\textbf{\tiny SMBL \cite{song2020enhancement}} &
\textbf{\tiny MLLE \cite{zhang2022underwater}} &
\textbf{\tiny WWPF \cite{zhang2023underwater}} &
\textbf{\tiny WFAC \cite{zhang2025underwater1}} &
\textbf{\tiny WaterNet \cite{li2019underwater}} \\[3pt]

\includegraphics[width=\ruieimgwidth,height=\ruieimgheight]{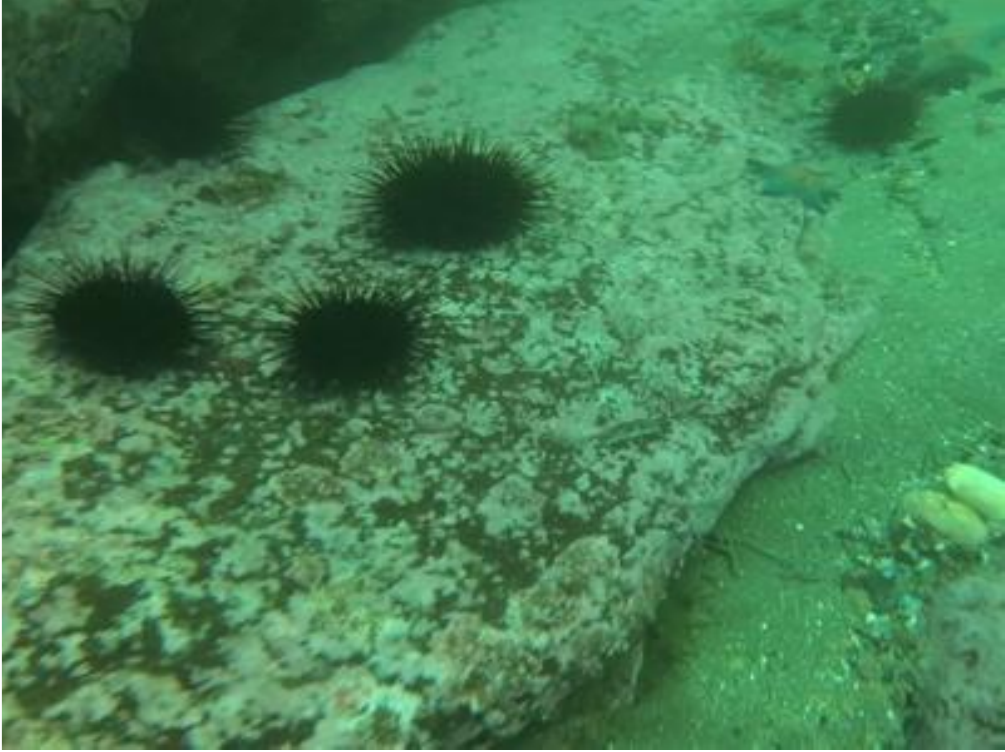} &
\includegraphics[width=\ruieimgwidth,height=\ruieimgheight]{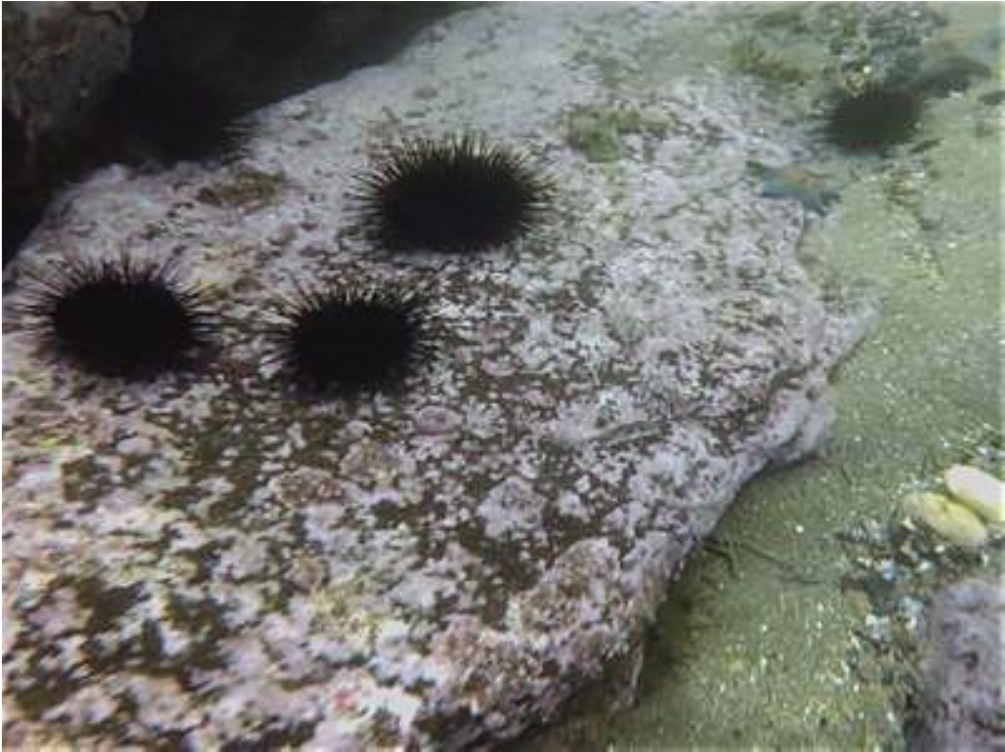} &
\includegraphics[width=\ruieimgwidth,height=\ruieimgheight]{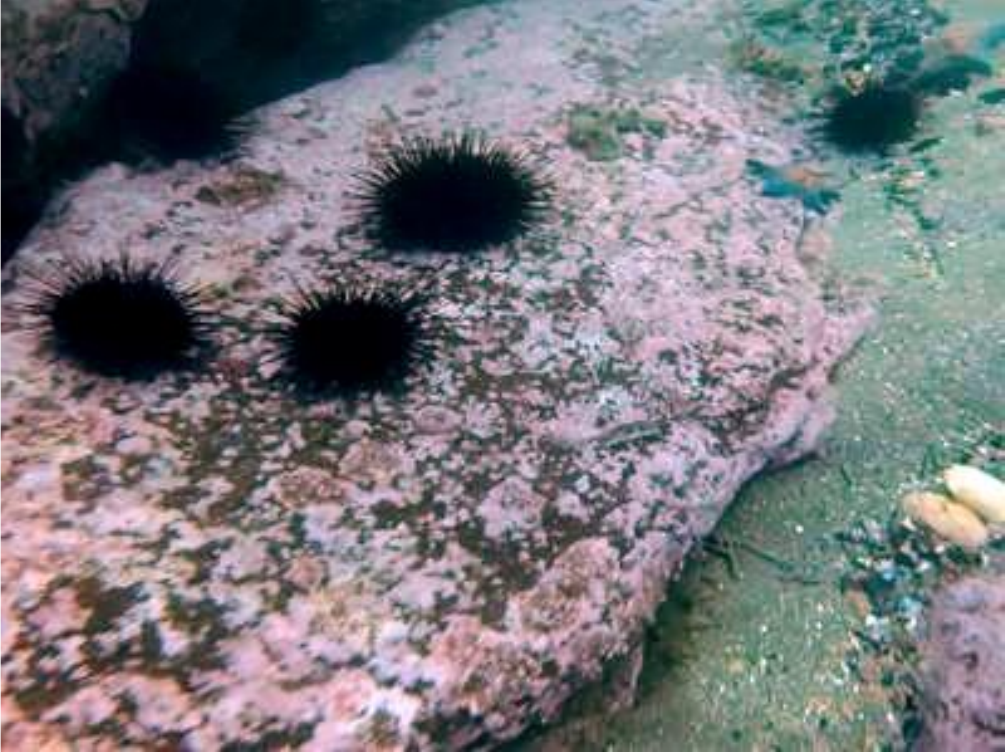} &
\includegraphics[width=\ruieimgwidth,height=\ruieimgheight]{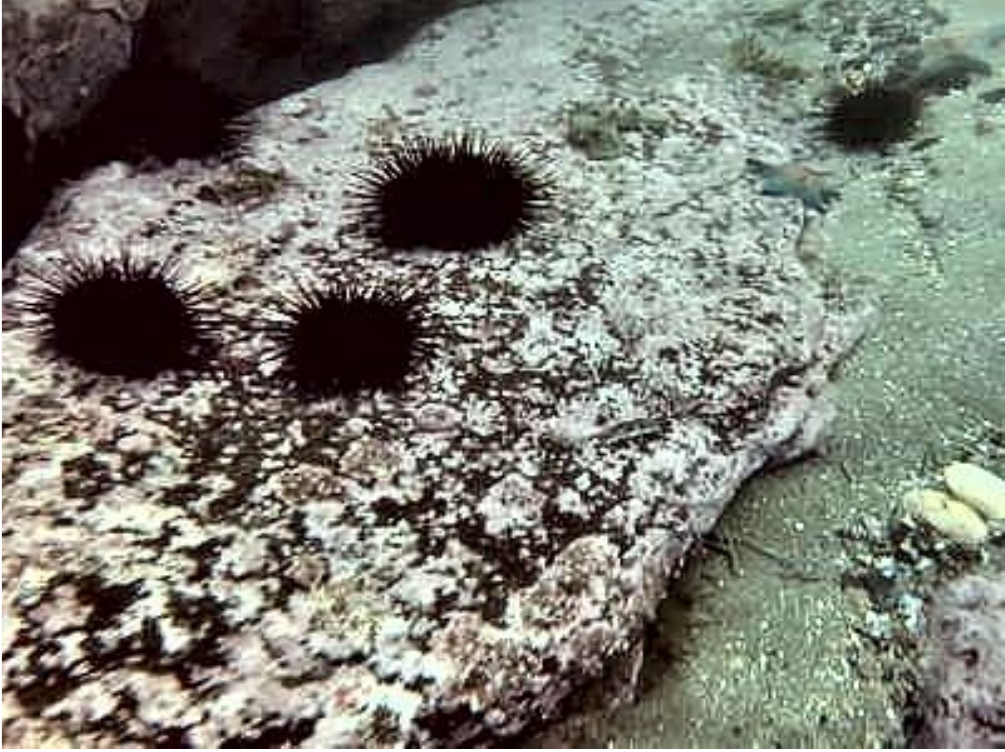} &
\includegraphics[width=\ruieimgwidth,height=\ruieimgheight]{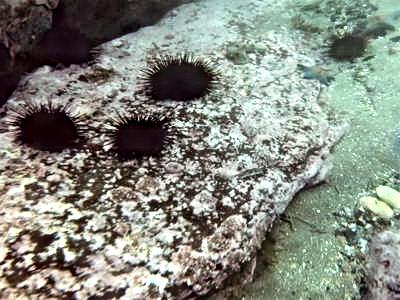} &
\includegraphics[width=\ruieimgwidth,height=\ruieimgheight]{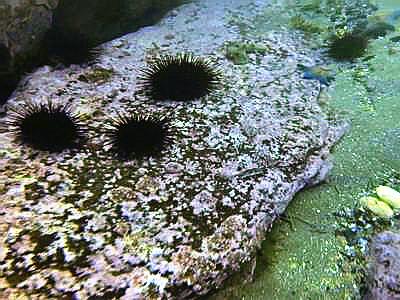} &
\includegraphics[width=\ruieimgwidth,height=\ruieimgheight]{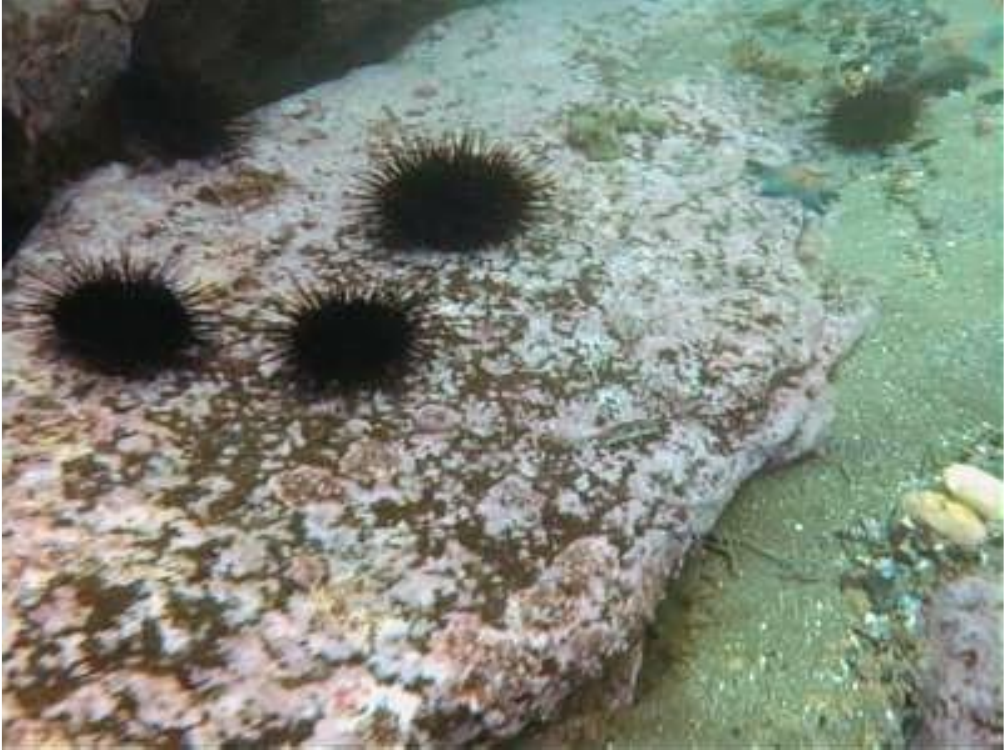} \\

\includegraphics[width=\ruieimgwidth,height=\ruieimgheight]{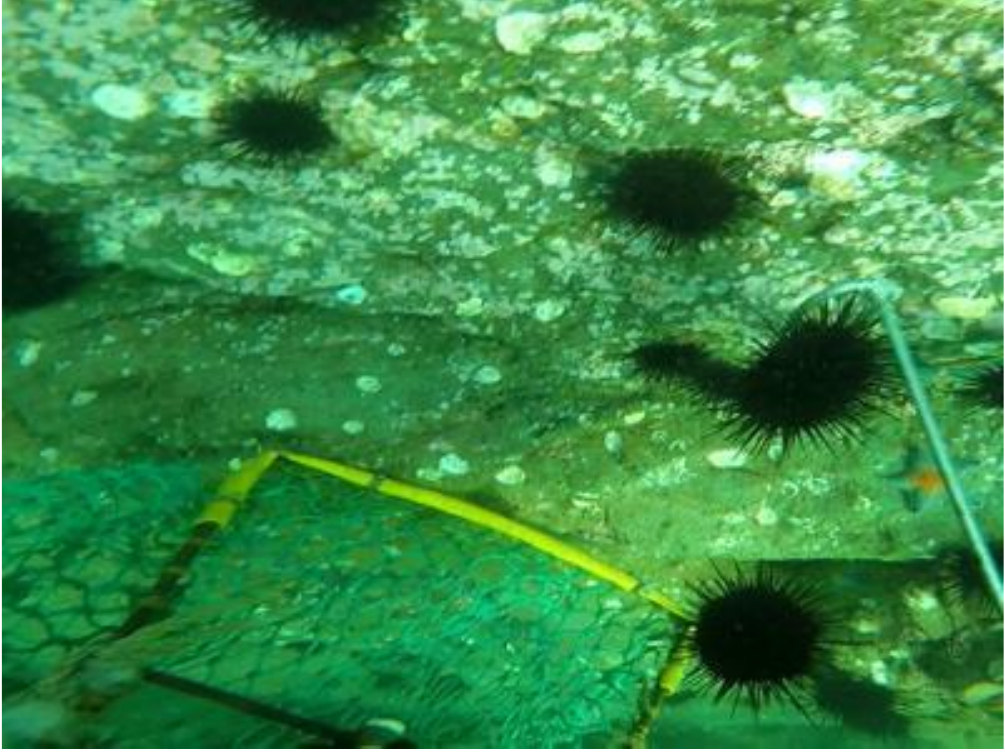} &
\includegraphics[width=\ruieimgwidth,height=\ruieimgheight]{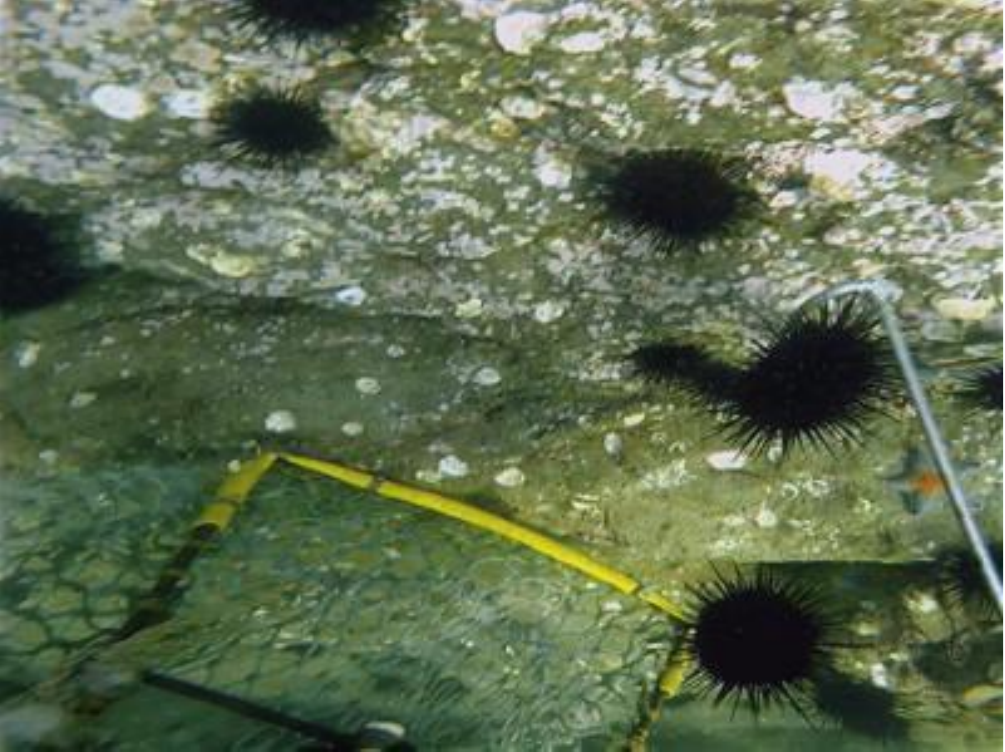} &
\includegraphics[width=\ruieimgwidth,height=\ruieimgheight]{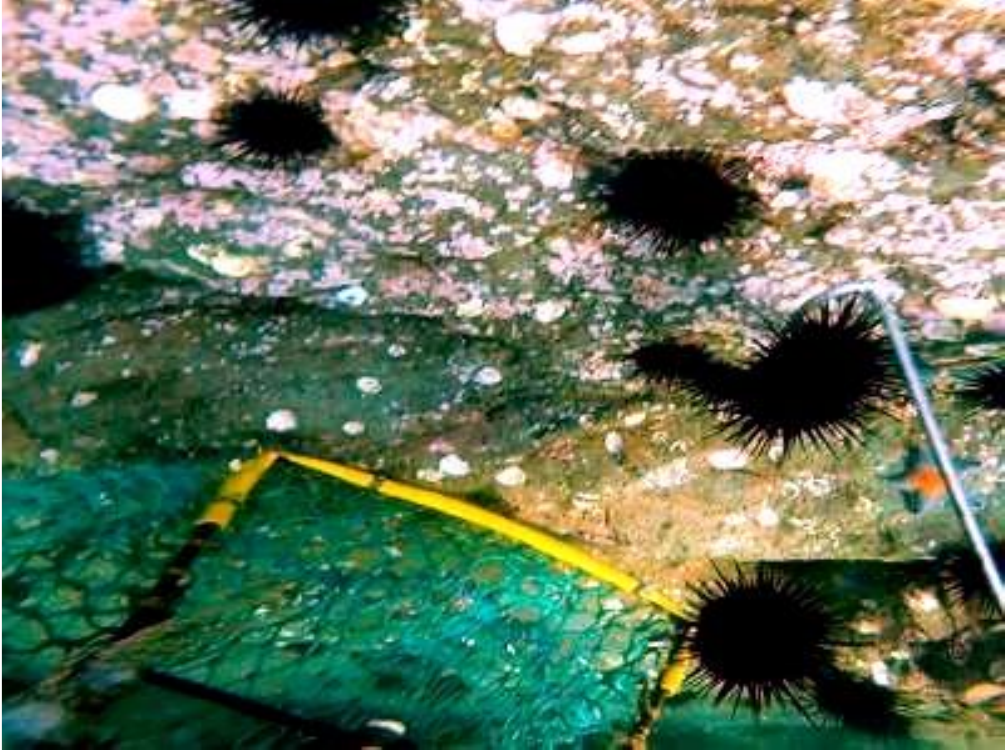} &
\includegraphics[width=\ruieimgwidth,height=\ruieimgheight]{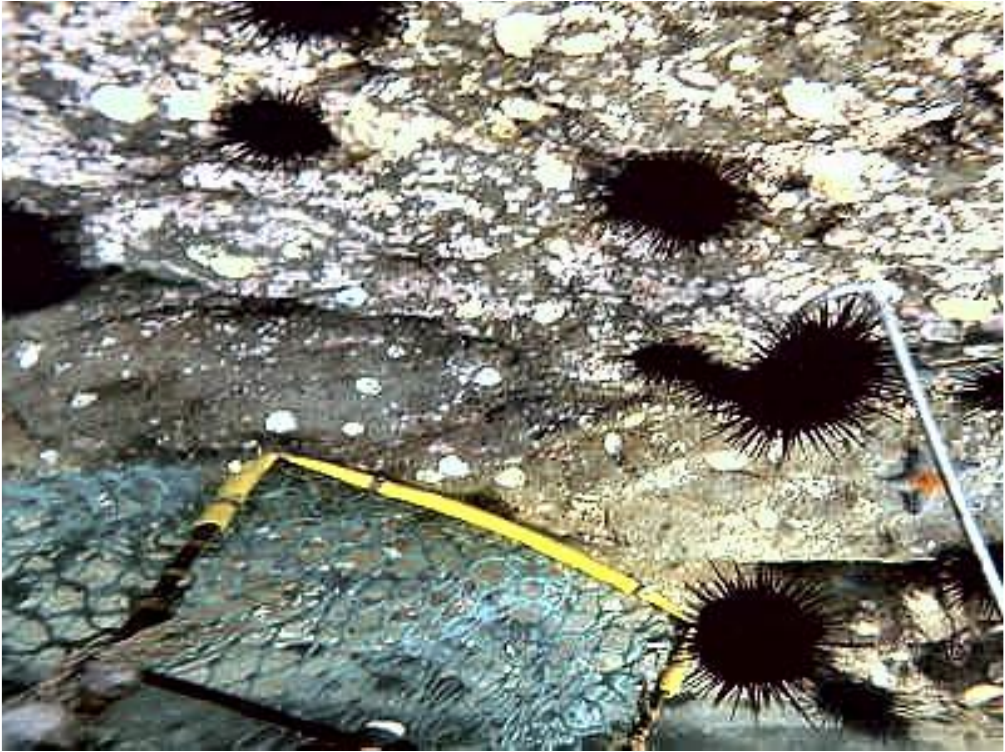} &
\includegraphics[width=\ruieimgwidth,height=\ruieimgheight]{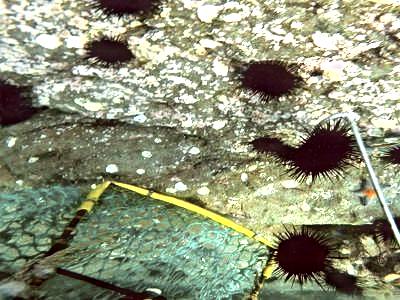} &
\includegraphics[width=\ruieimgwidth,height=\ruieimgheight]{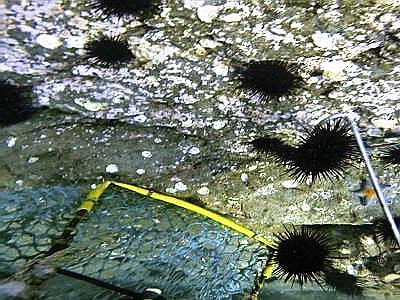} &
\includegraphics[width=\ruieimgwidth,height=\ruieimgheight]{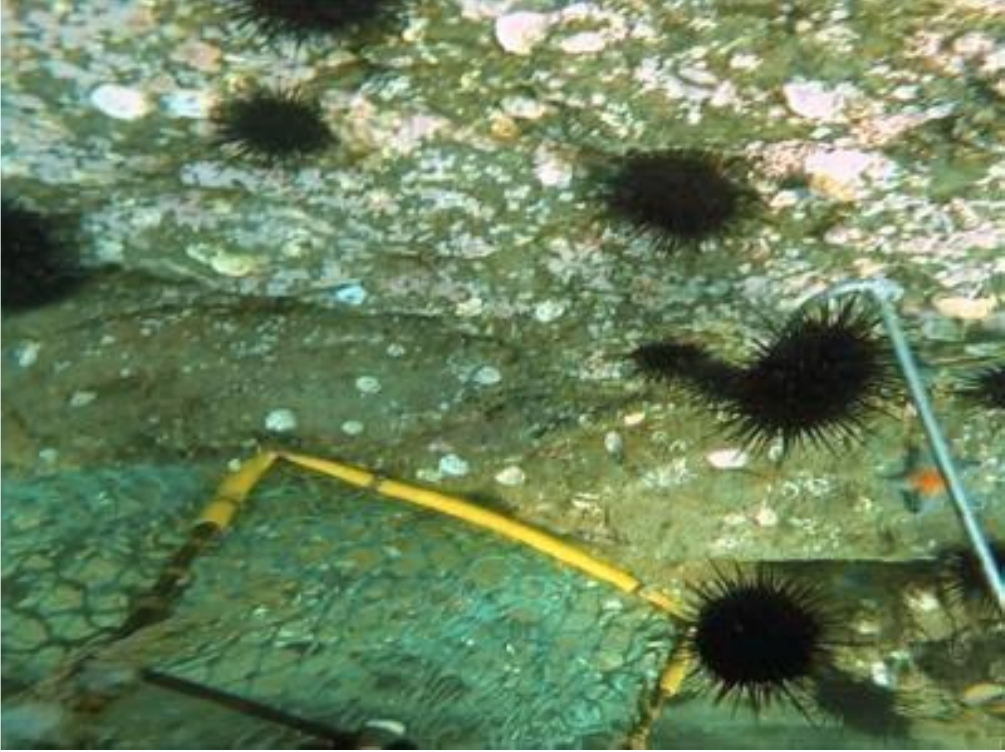} \\[5pt]

\textbf{\tiny UWCNN \cite{li2020underwater}} &
\textbf{\tiny PUIE \cite{fu2022uncertainty}} &
\textbf{\tiny TACL \cite{liu2022twin}} &
\textbf{\tiny NU2Net \cite{guo2023underwater}} &
\textbf{\tiny CCL-Net \cite{liu2024underwater}} &
\textbf{\tiny OUNet-JL \cite{wang2025optimized}} &
\textbf{\tiny AQUA-Net} \\[3pt]

\includegraphics[width=\ruieimgwidth,height=\ruieimgheight]{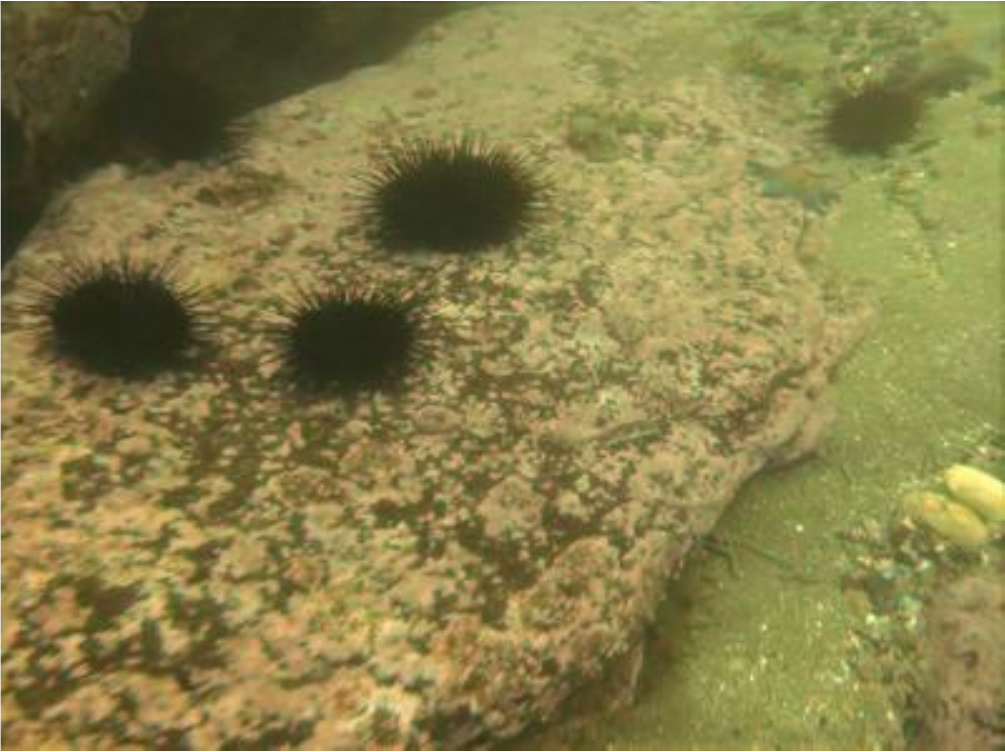} &
\includegraphics[width=\ruieimgwidth,height=\ruieimgheight]{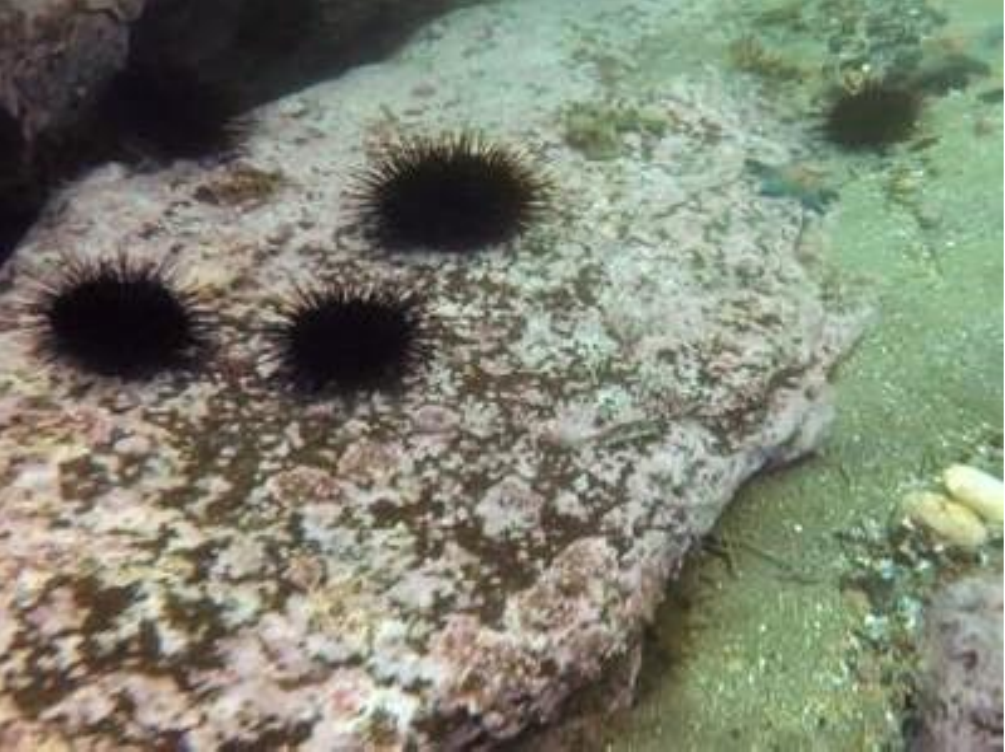} &
\includegraphics[width=\ruieimgwidth,height=\ruieimgheight]{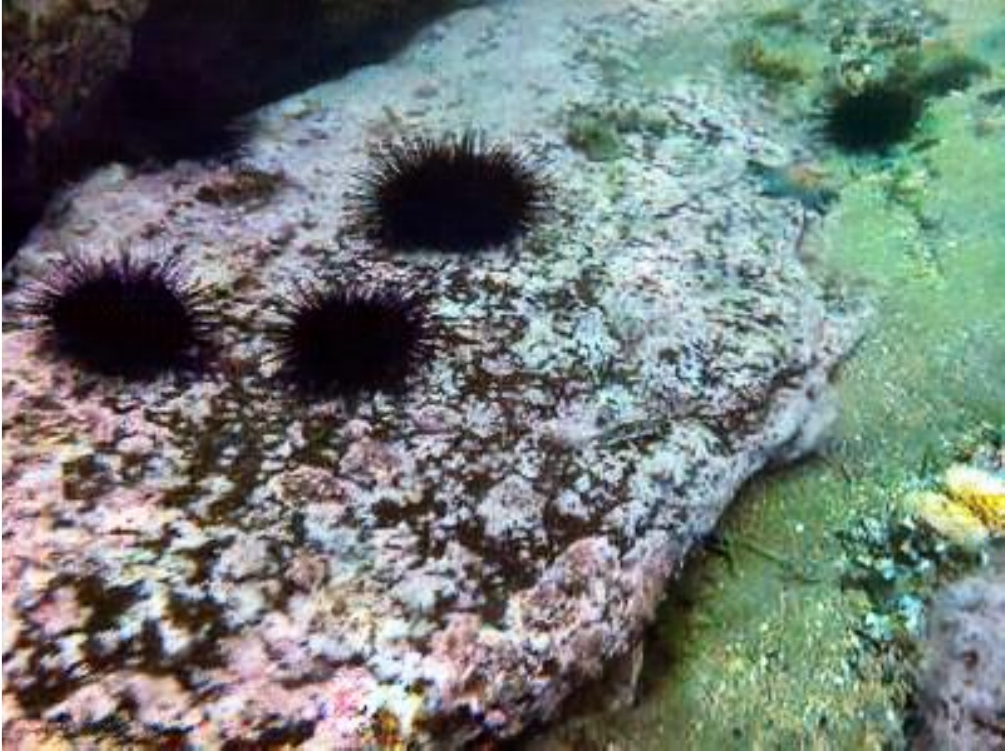} &
\includegraphics[width=\ruieimgwidth,height=\ruieimgheight]{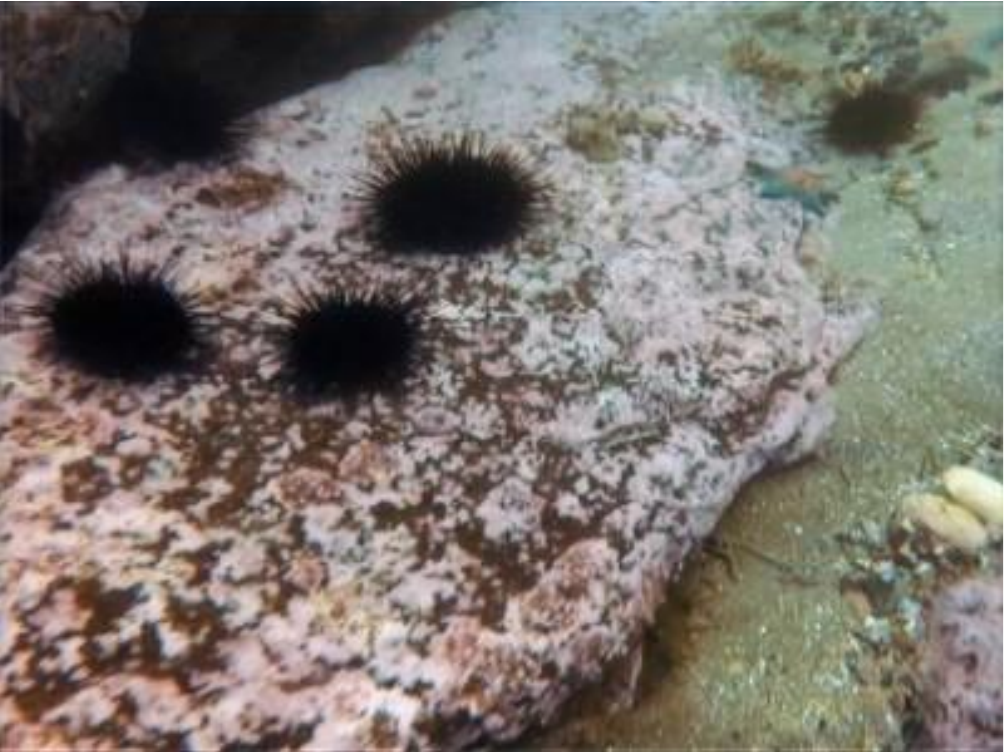} &
\includegraphics[width=\ruieimgwidth,height=\ruieimgheight]{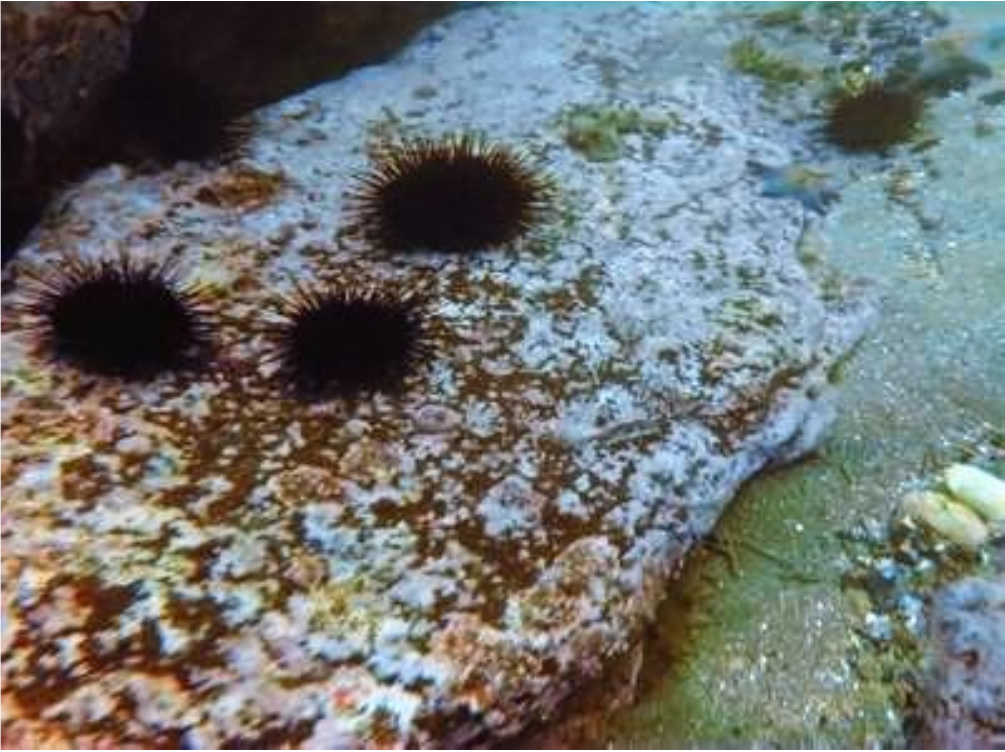} &
\includegraphics[width=\ruieimgwidth,height=\ruieimgheight]{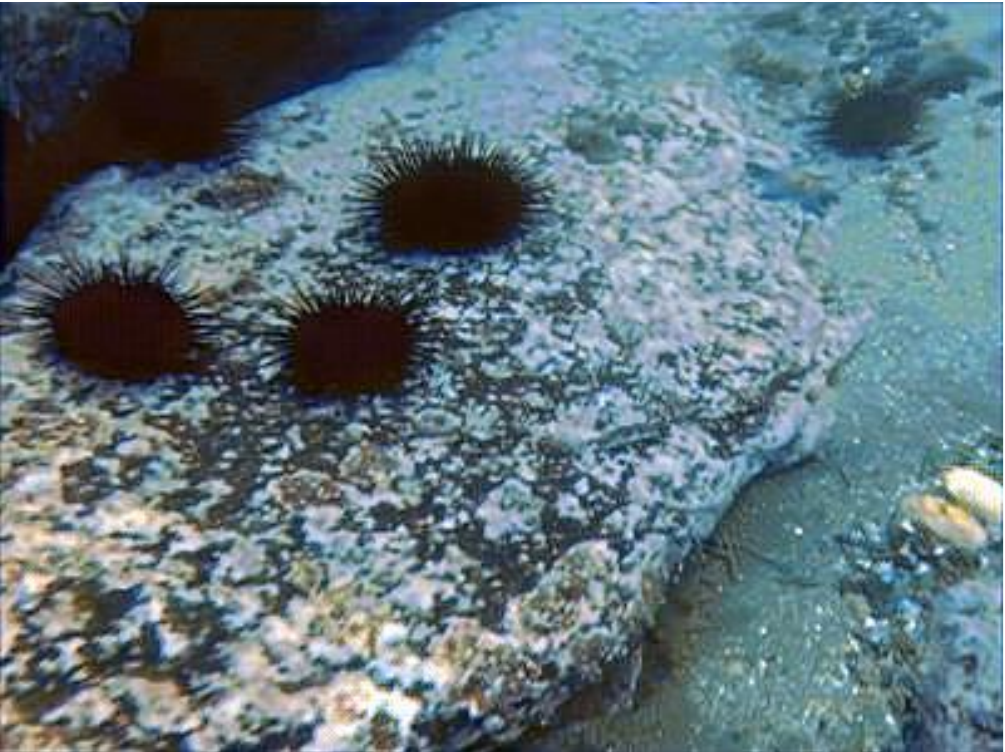} &
\includegraphics[width=\ruieimgwidth,height=\ruieimgheight]{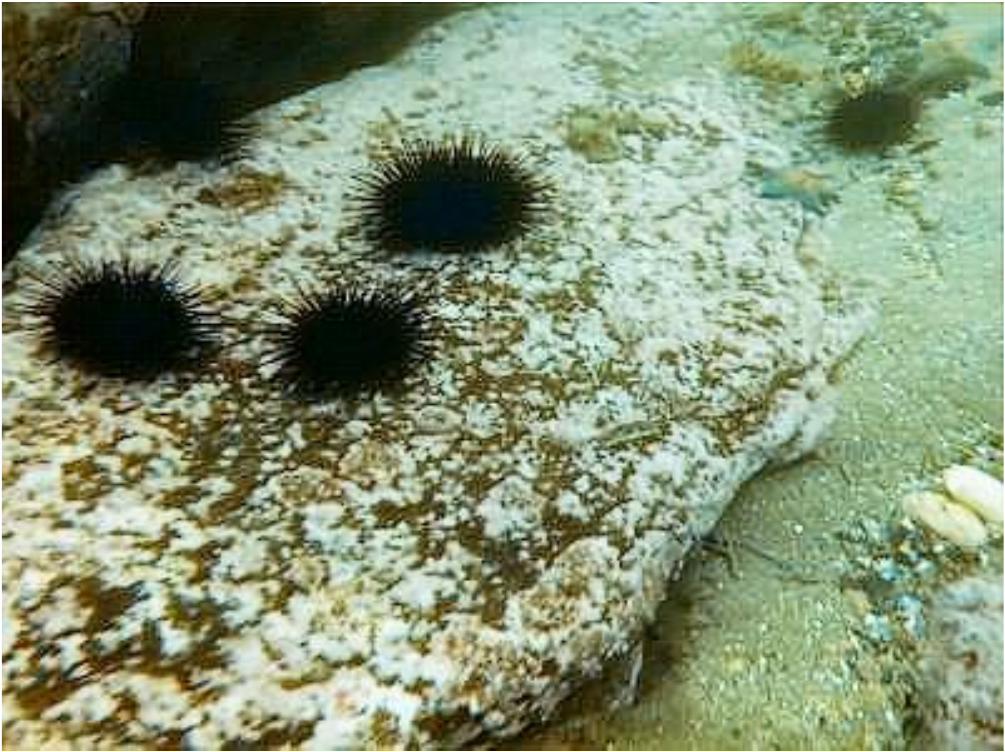} \\

\includegraphics[width=\ruieimgwidth,height=\ruieimgheight]{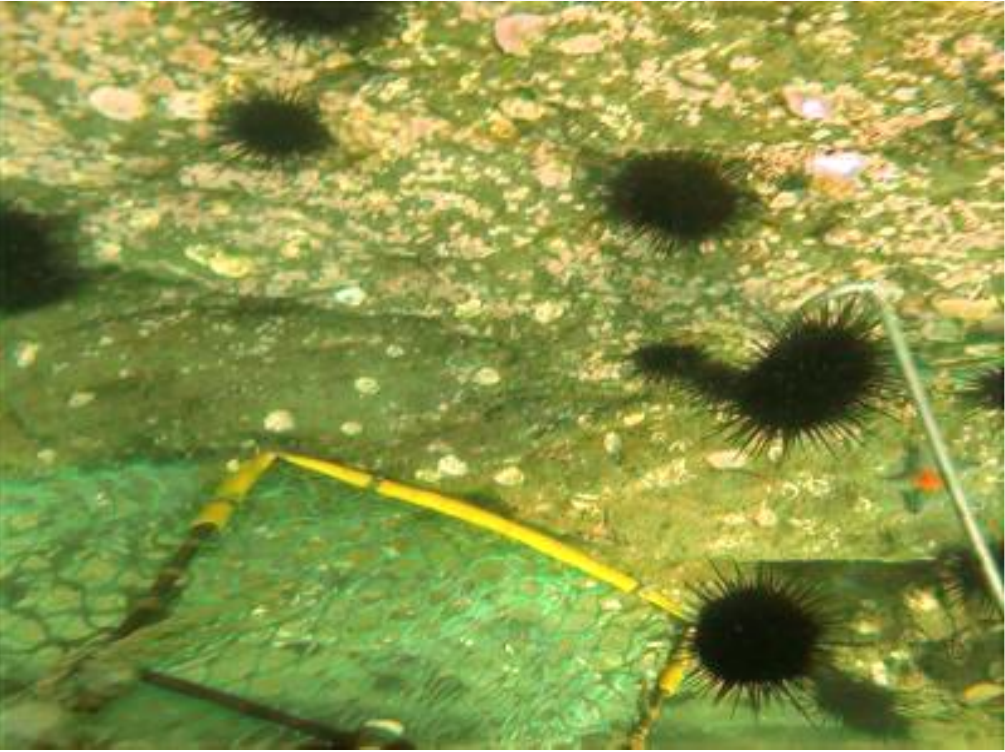} &
\includegraphics[width=\ruieimgwidth,height=\ruieimgheight]{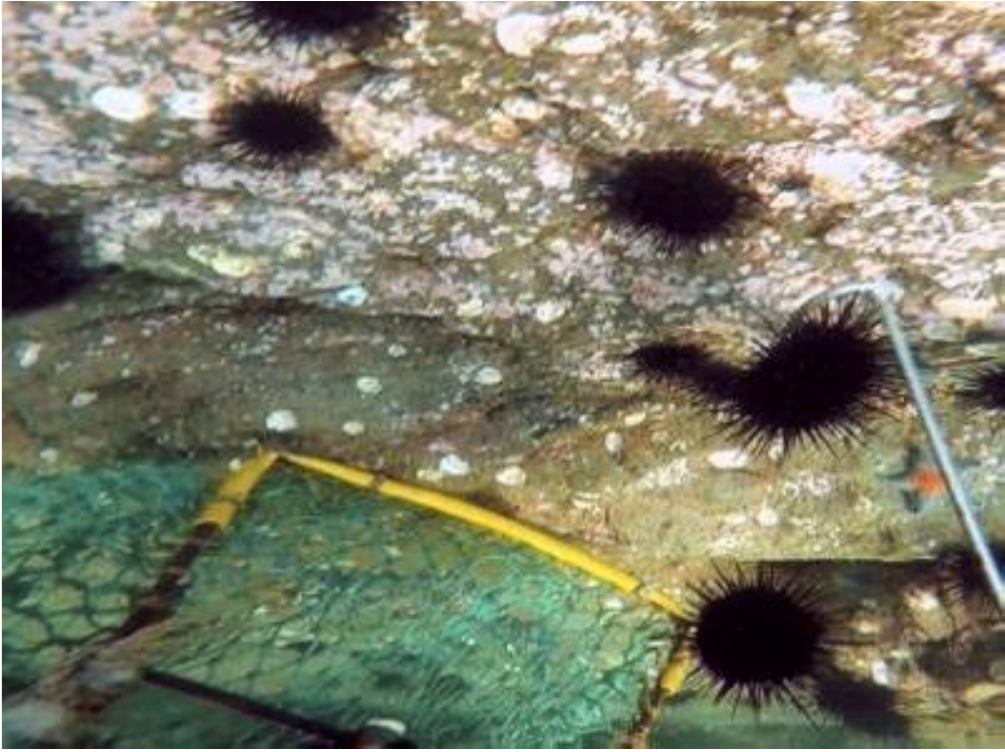} &
\includegraphics[width=\ruieimgwidth,height=\ruieimgheight]{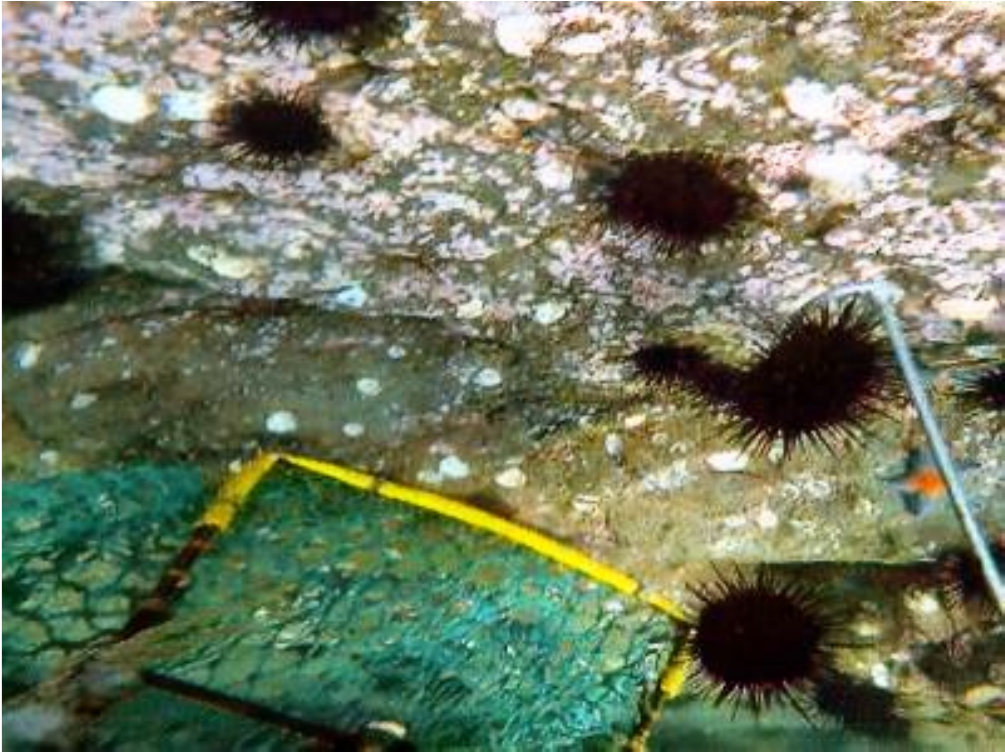} &
\includegraphics[width=\ruieimgwidth,height=\ruieimgheight]{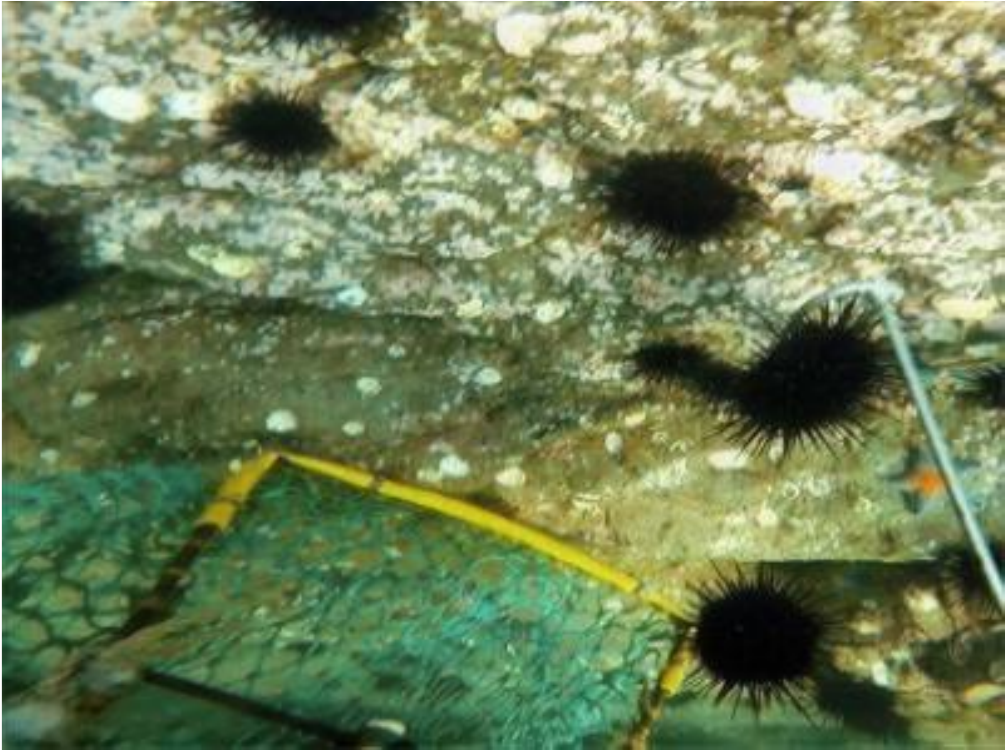} &
\includegraphics[width=\ruieimgwidth,height=\ruieimgheight]{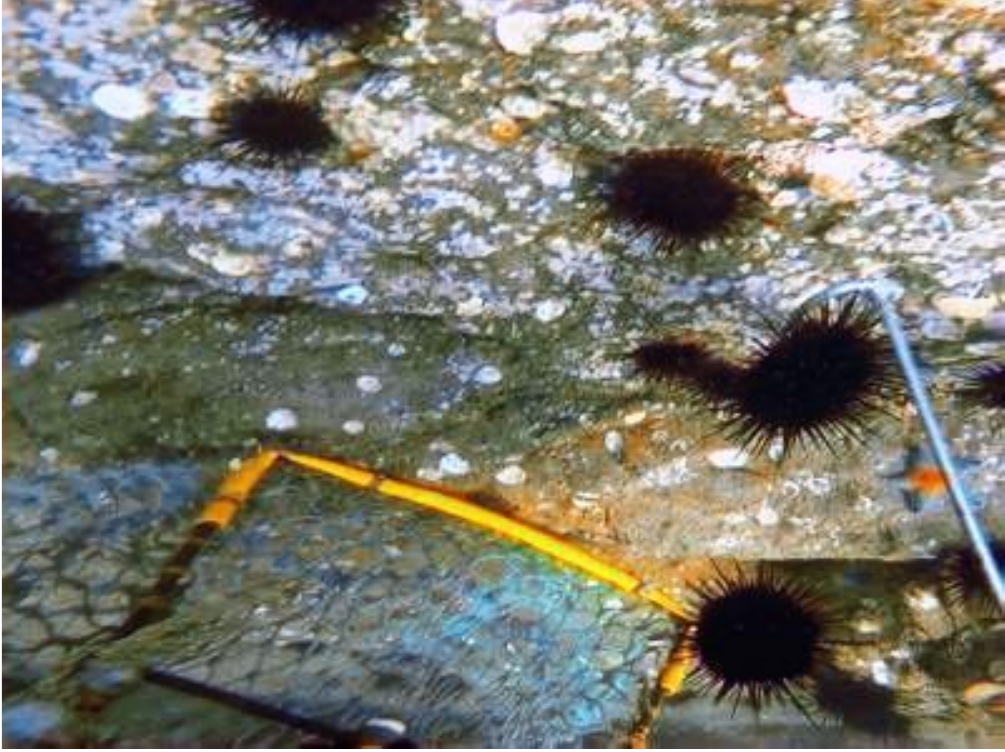} &
\includegraphics[width=\ruieimgwidth,height=\ruieimgheight]{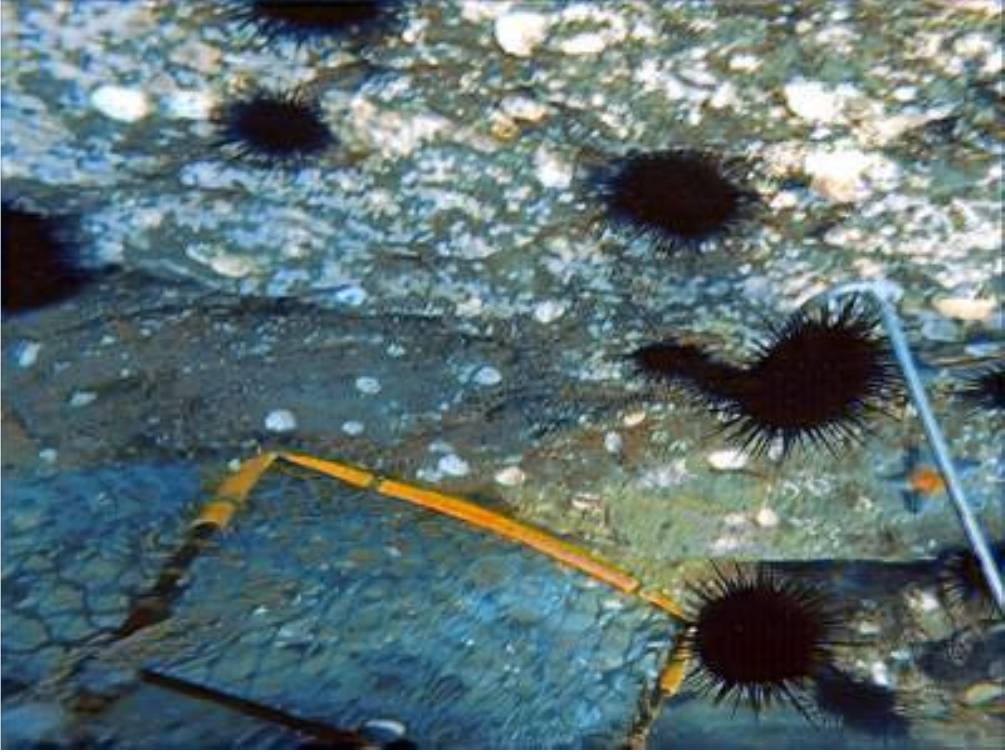} &
\includegraphics[width=\ruieimgwidth,height=\ruieimgheight]{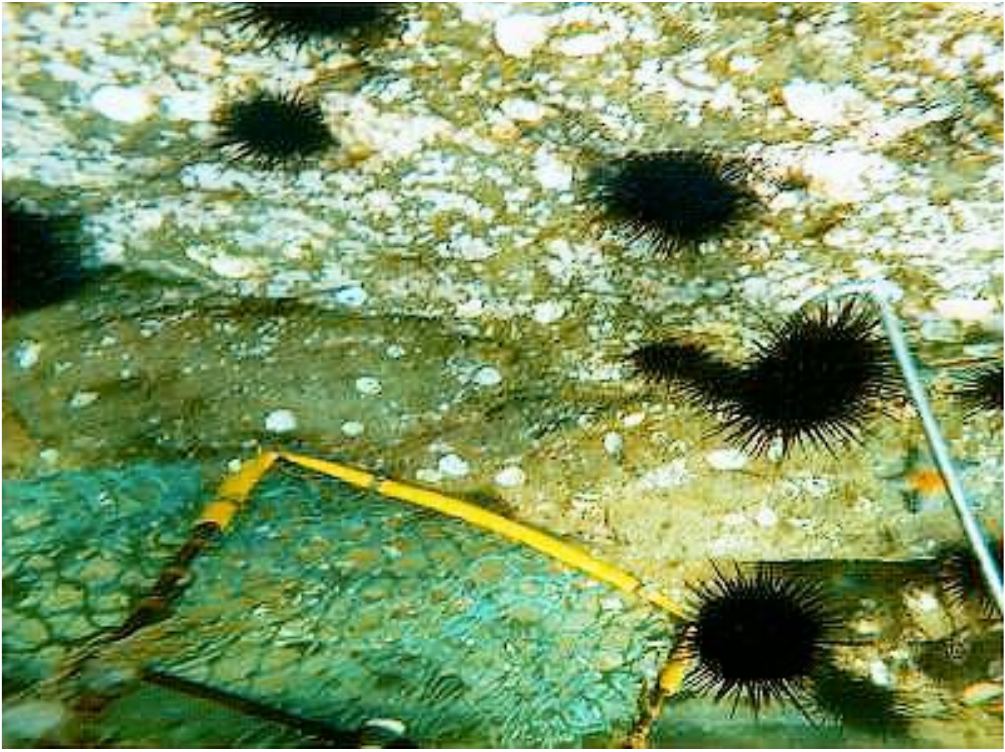} \\

\end{tabular}

\caption{Comparison on the RUIE-T78 underwater image subset with strong greenish degradation. AQUA-Net removes the greenish color cast and clearly restores object visibility, including the black sea urchin.}
\label{fig:RUIE-T78}
\end{figure*}

\textbf{RUIE-T78:} Another challenging two images from the RUIE-T78 show a strong greenish color, which reduces the visibility of objects in the UIWs. Figure \ref{fig:RUIE-T78} shows two representative samples. Our model removes the greenish tint and clearly reveals the black sea urchin, showing its ability to restore natural colors and improve visibility in difficult underwater conditions.  

\begin{figure*}[t!]
\centering

\setlength{\tabcolsep}{1.2pt}
\renewcommand{\arraystretch}{0.35}

\newcommand{\deepimgwidth}{0.13\textwidth}
\newcommand{\deepimgheight}{0.085\textwidth}

\begin{tabular}{ccccccc}

\textbf{\tiny Raw} &
\textbf{\tiny Fusion \cite{ancuti2017color}} &
\textbf{\tiny SMBL \cite{song2020enhancement}} &
\textbf{\tiny MLLE \cite{zhang2022underwater}} &
\textbf{\tiny WWPF \cite{zhang2023underwater}} &
\textbf{\tiny WFAC \cite{zhang2025underwater1}} &
\textbf{\tiny WaterNet \cite{li2019underwater}} \\[3pt]

\includegraphics[width=\deepimgwidth,height=\deepimgheight]{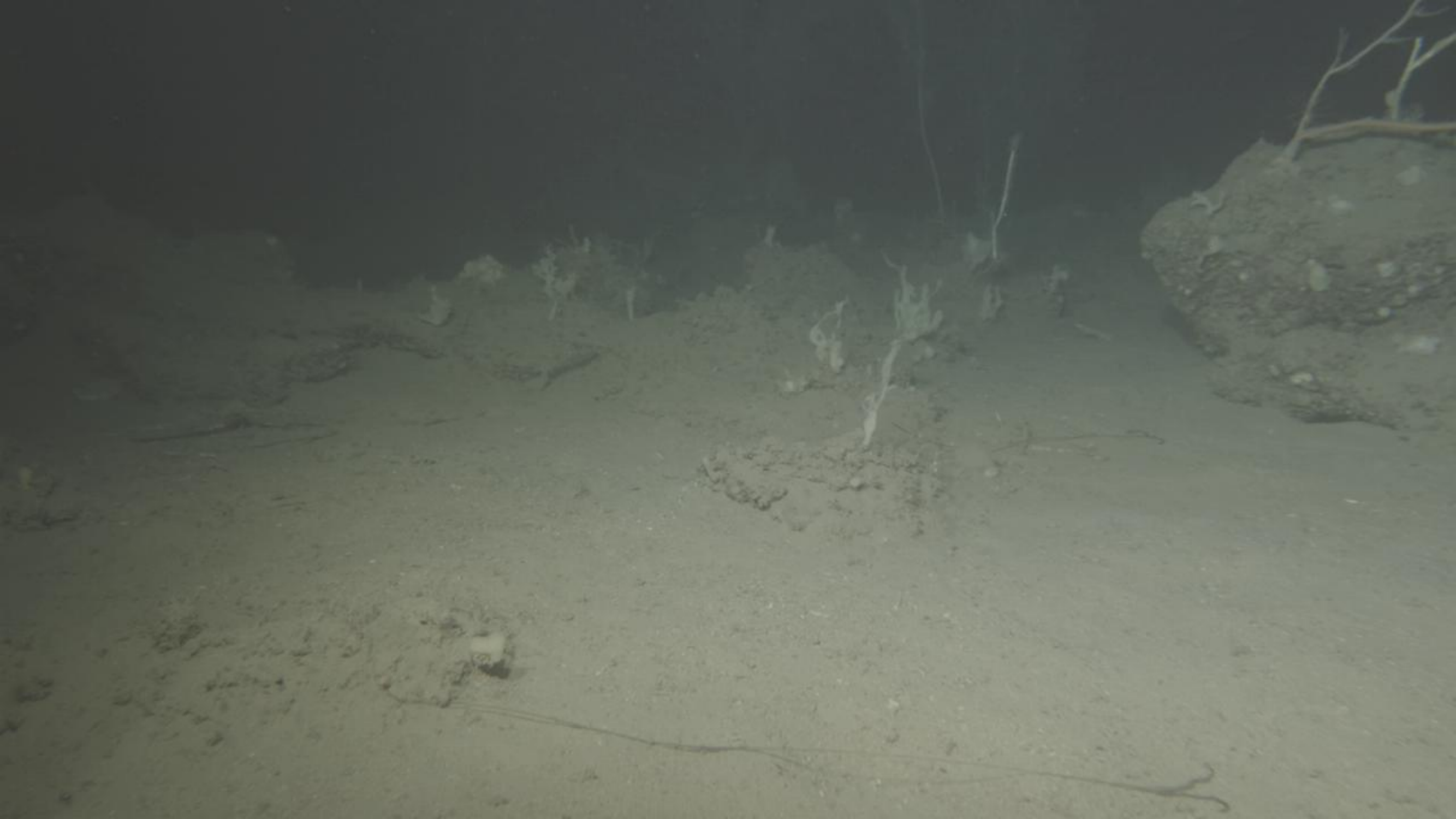} &
\includegraphics[width=\deepimgwidth,height=\deepimgheight]{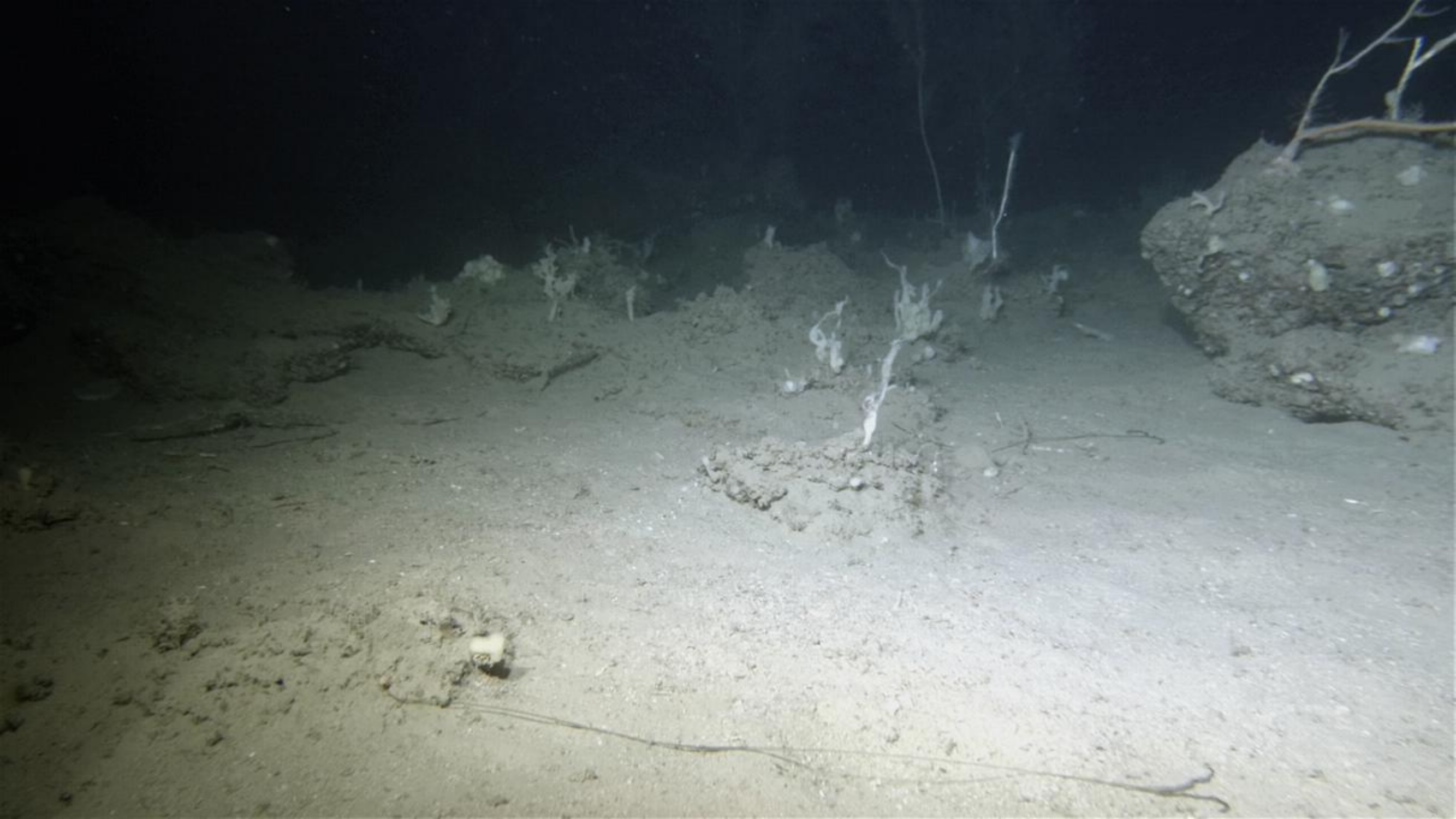} &
\includegraphics[width=\deepimgwidth,height=\deepimgheight]{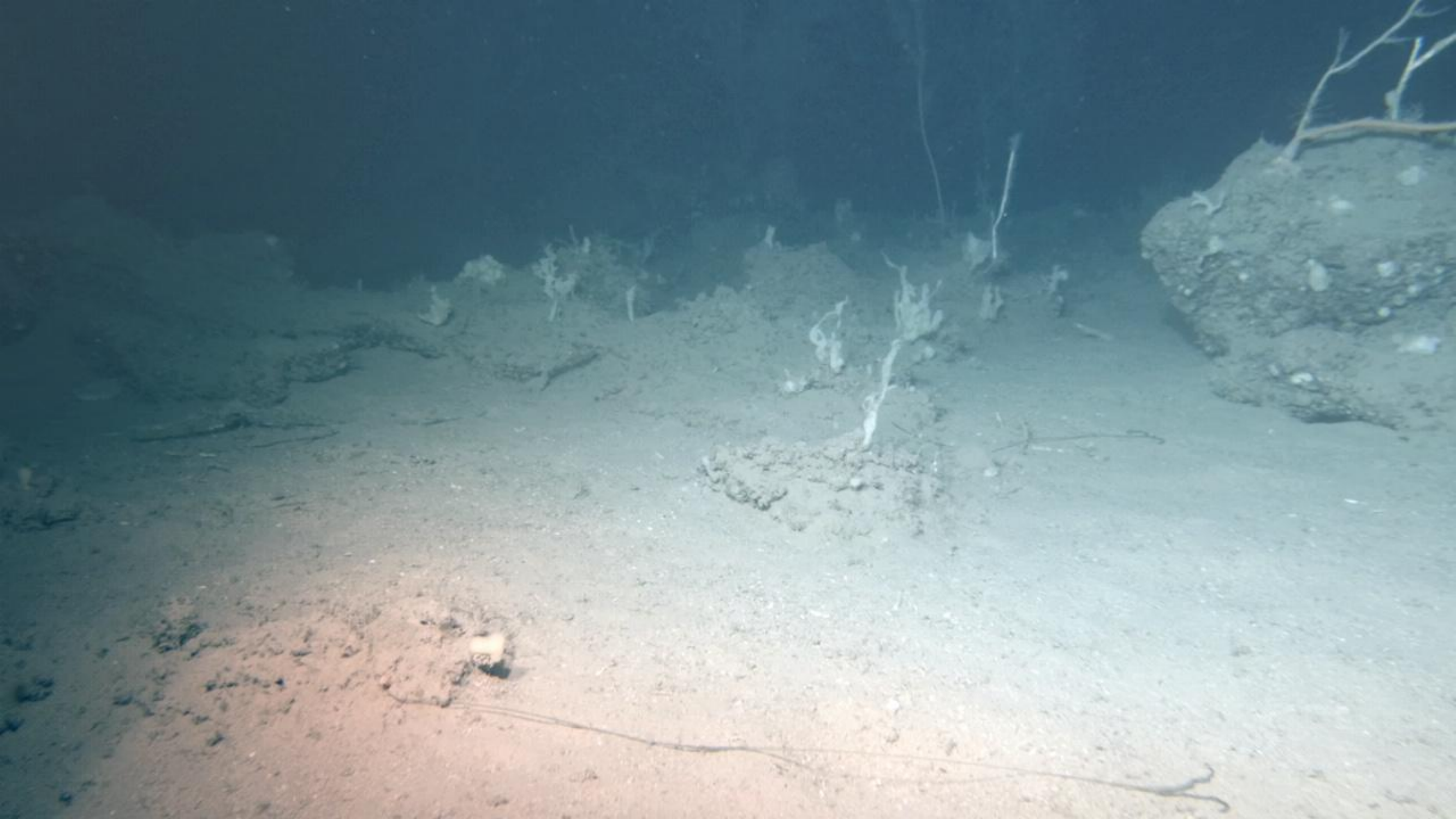} &
\includegraphics[width=\deepimgwidth,height=\deepimgheight]{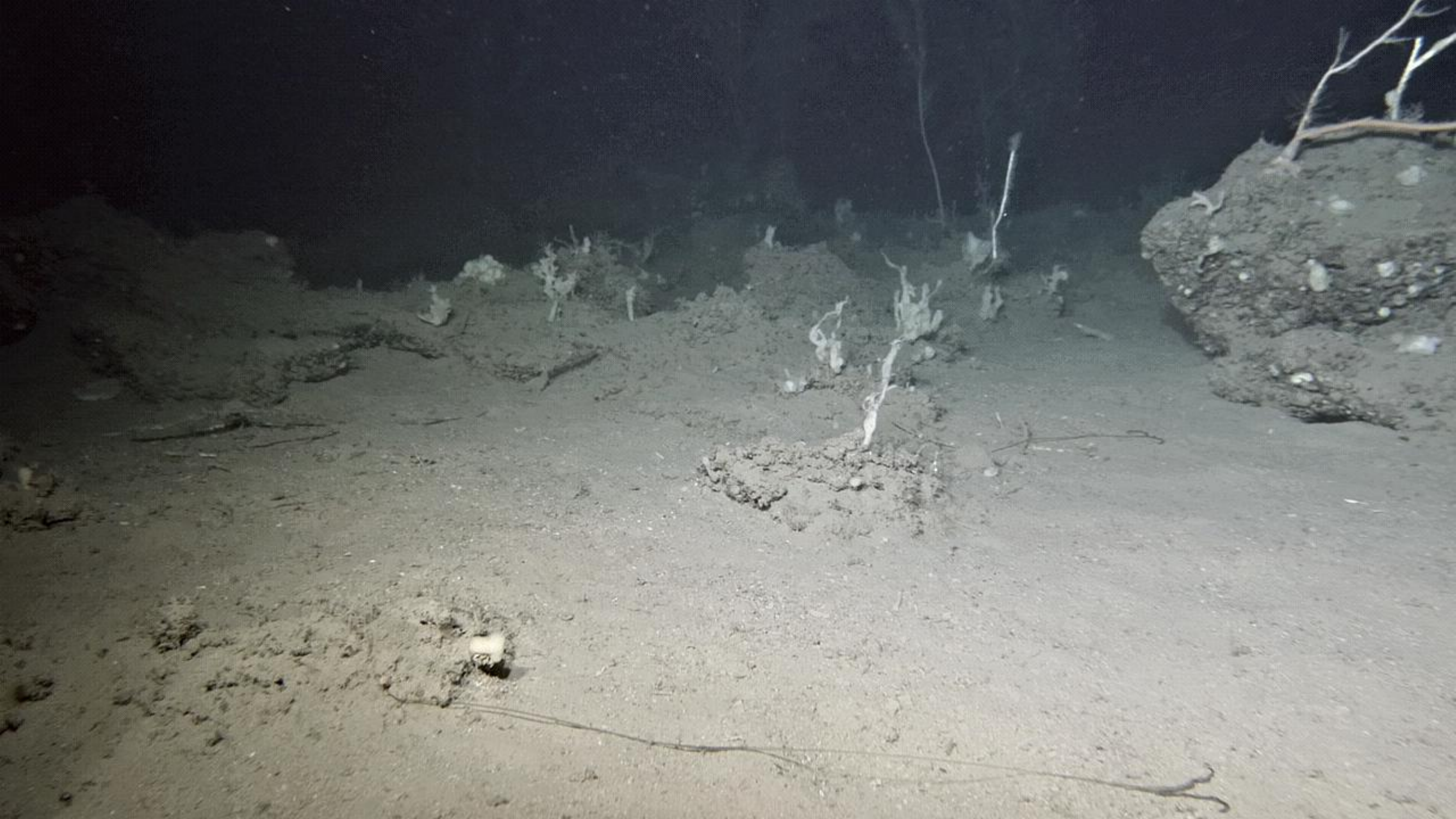} &
\includegraphics[width=\deepimgwidth,height=\deepimgheight]{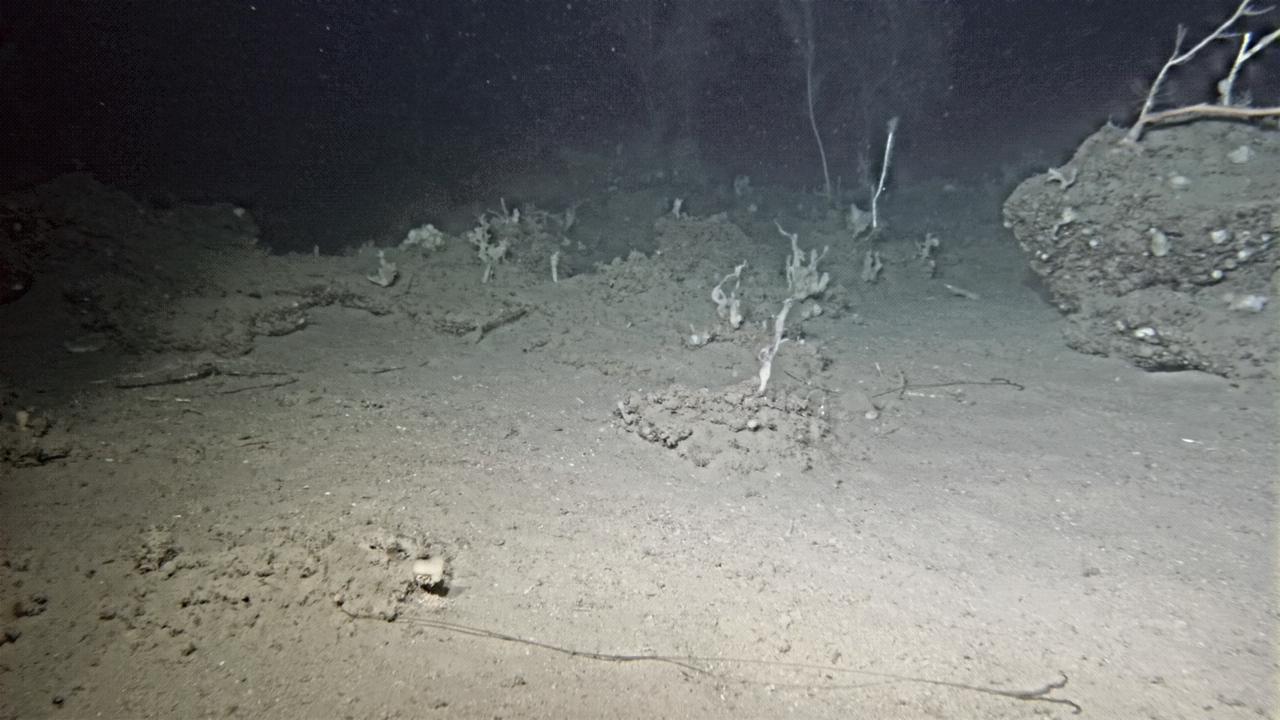} &
\includegraphics[width=\deepimgwidth,height=\deepimgheight]{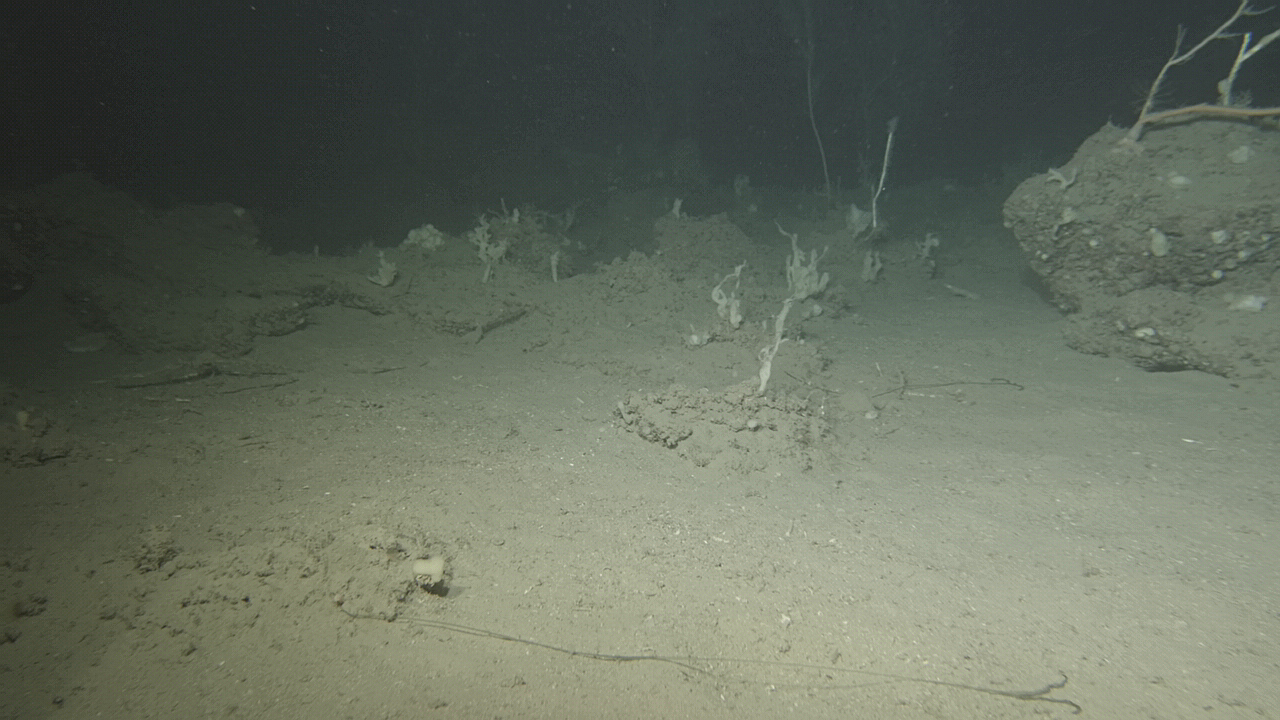} &
\includegraphics[width=\deepimgwidth,height=\deepimgheight]{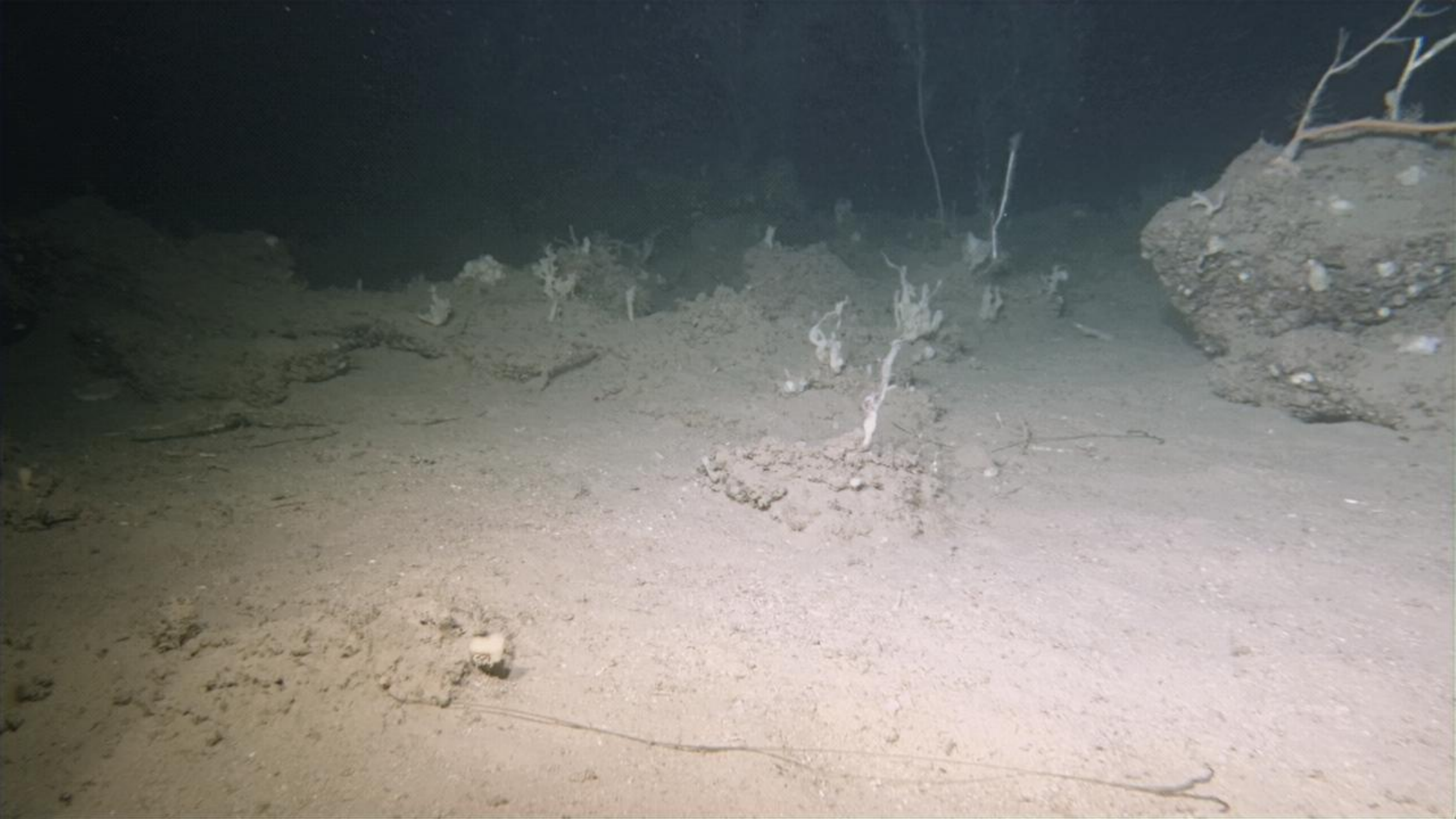} \\

\includegraphics[width=\deepimgwidth,height=\deepimgheight]{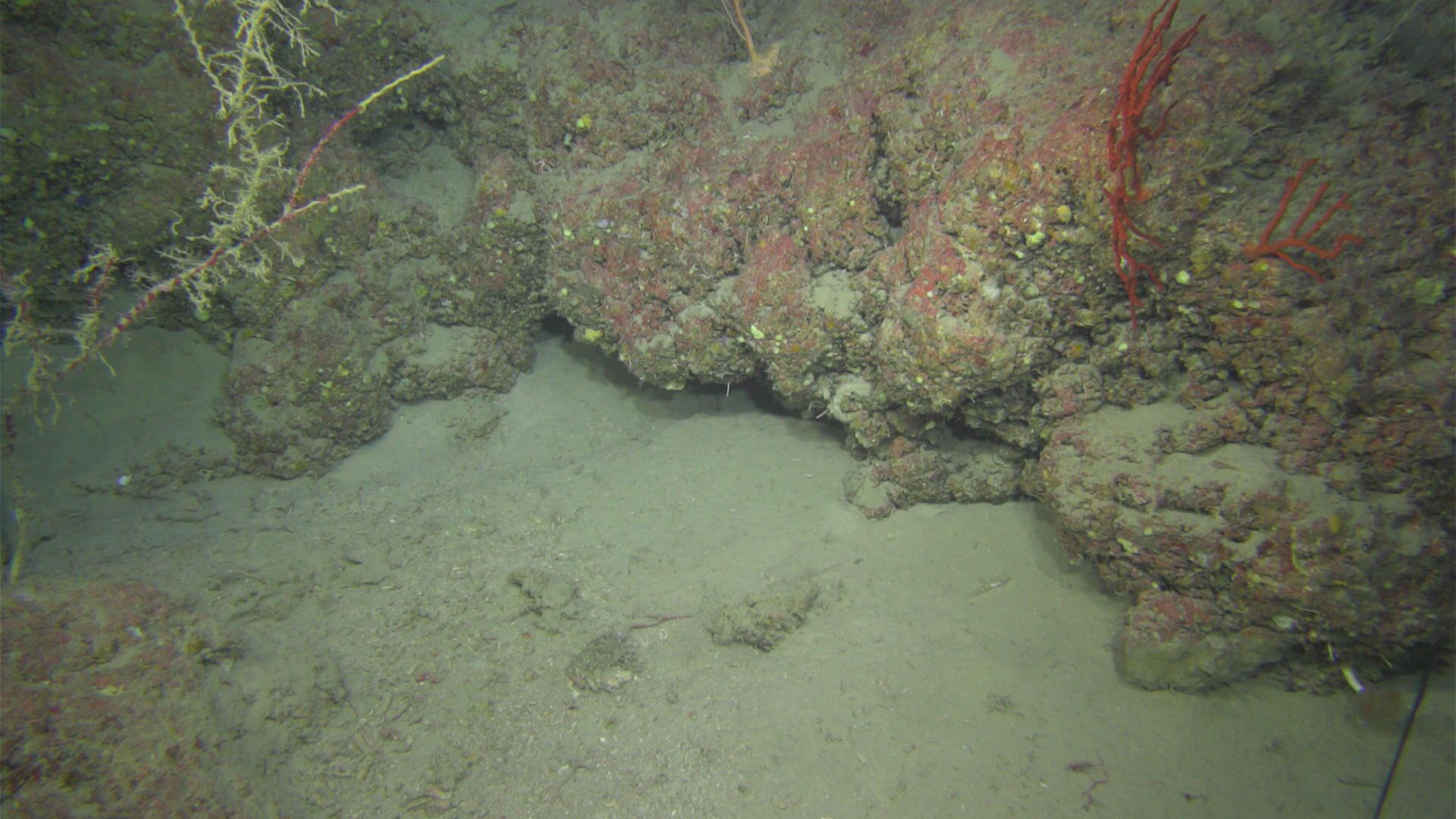} &
\includegraphics[width=\deepimgwidth,height=\deepimgheight]{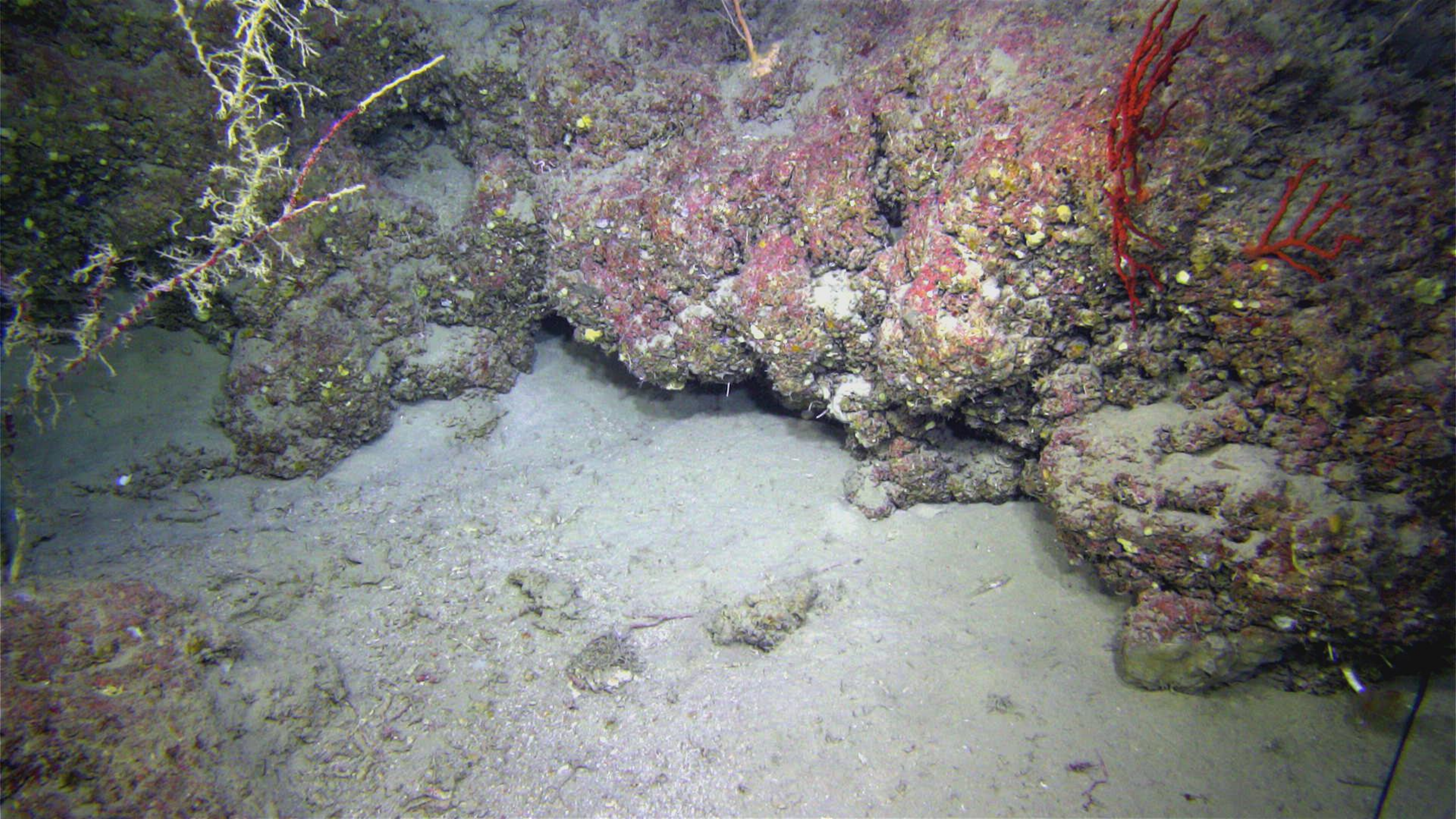} &
\includegraphics[width=\deepimgwidth,height=\deepimgheight]{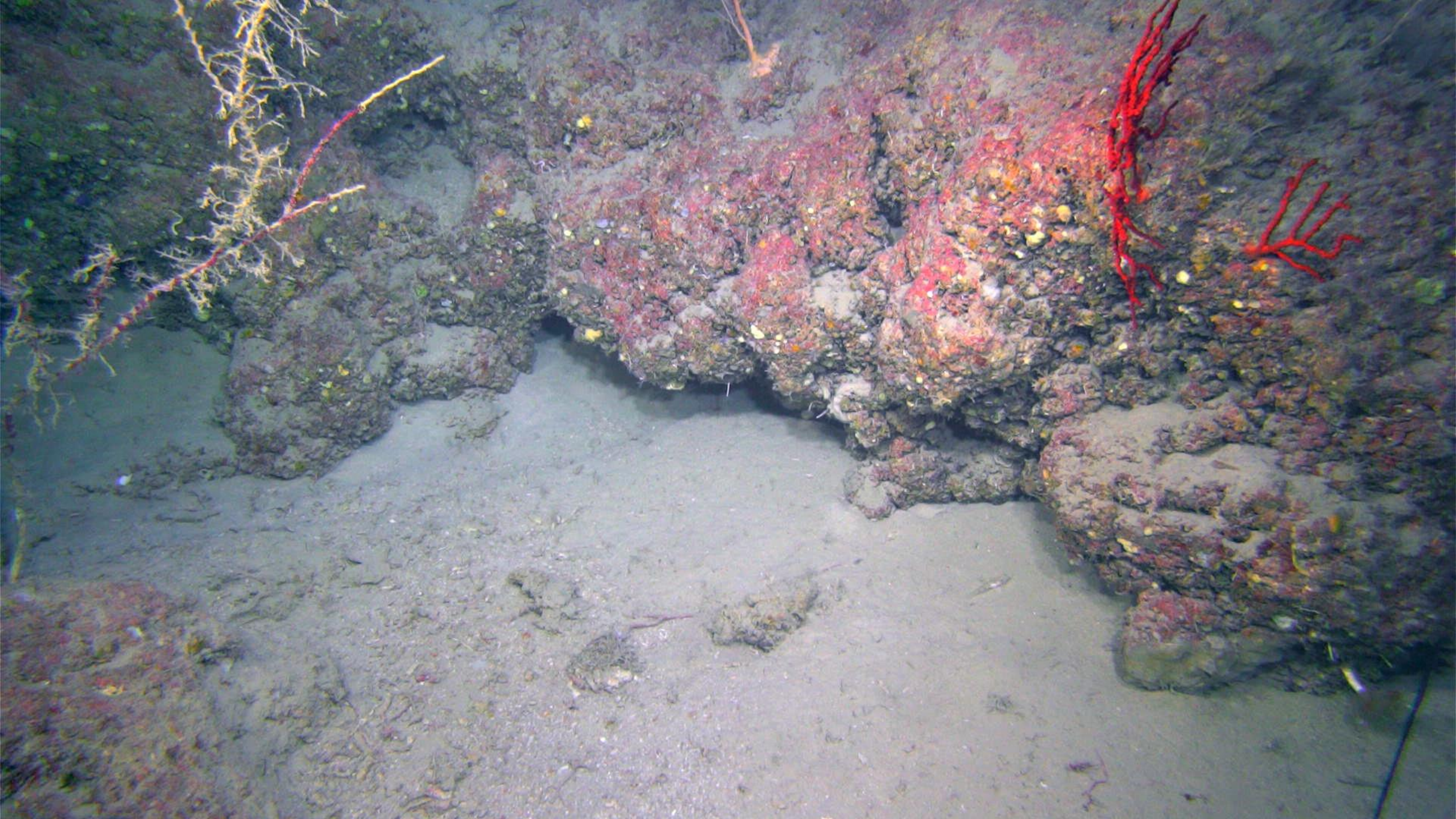} &
\includegraphics[width=\deepimgwidth,height=\deepimgheight]{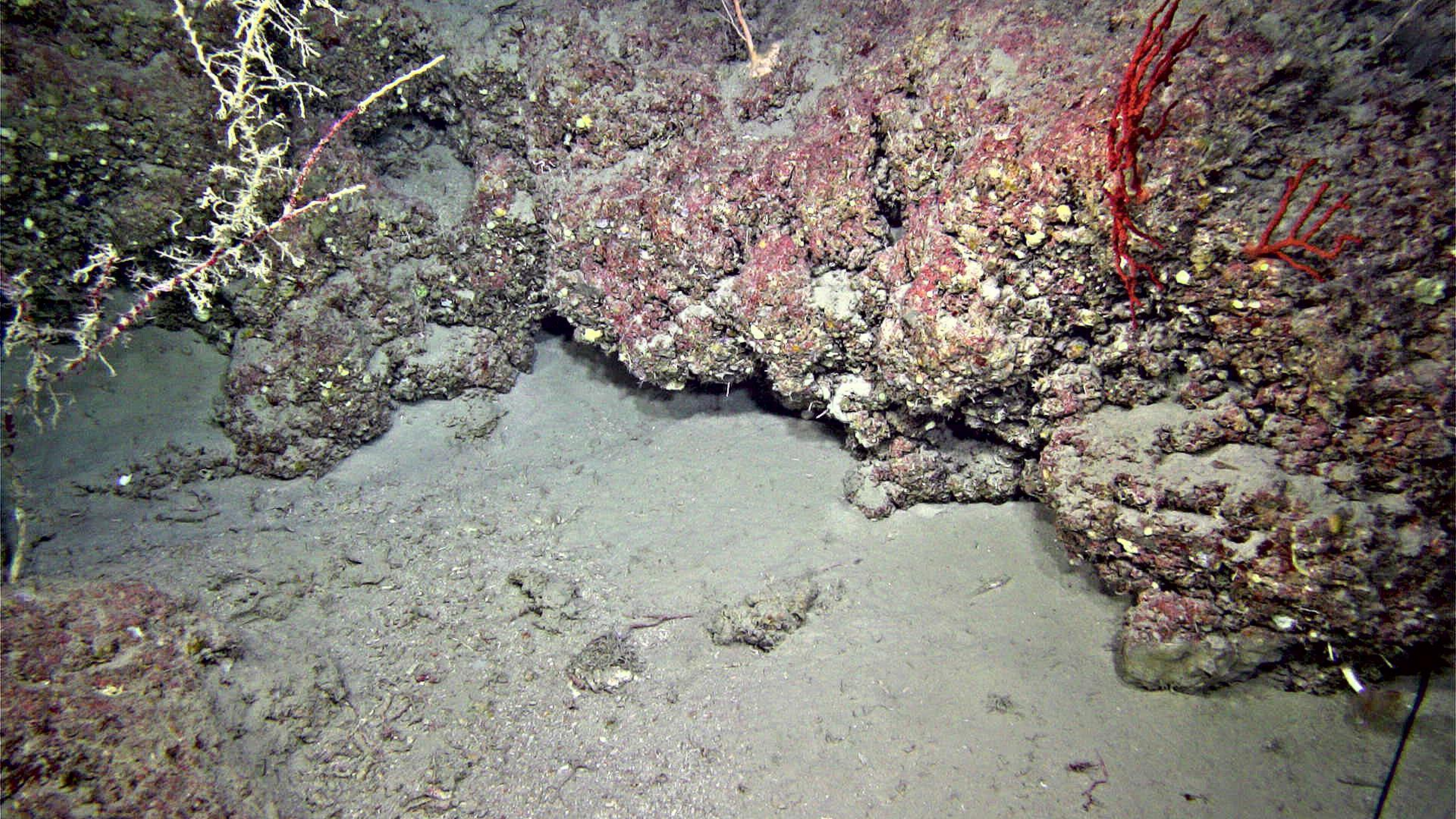} &
\includegraphics[width=\deepimgwidth,height=\deepimgheight]{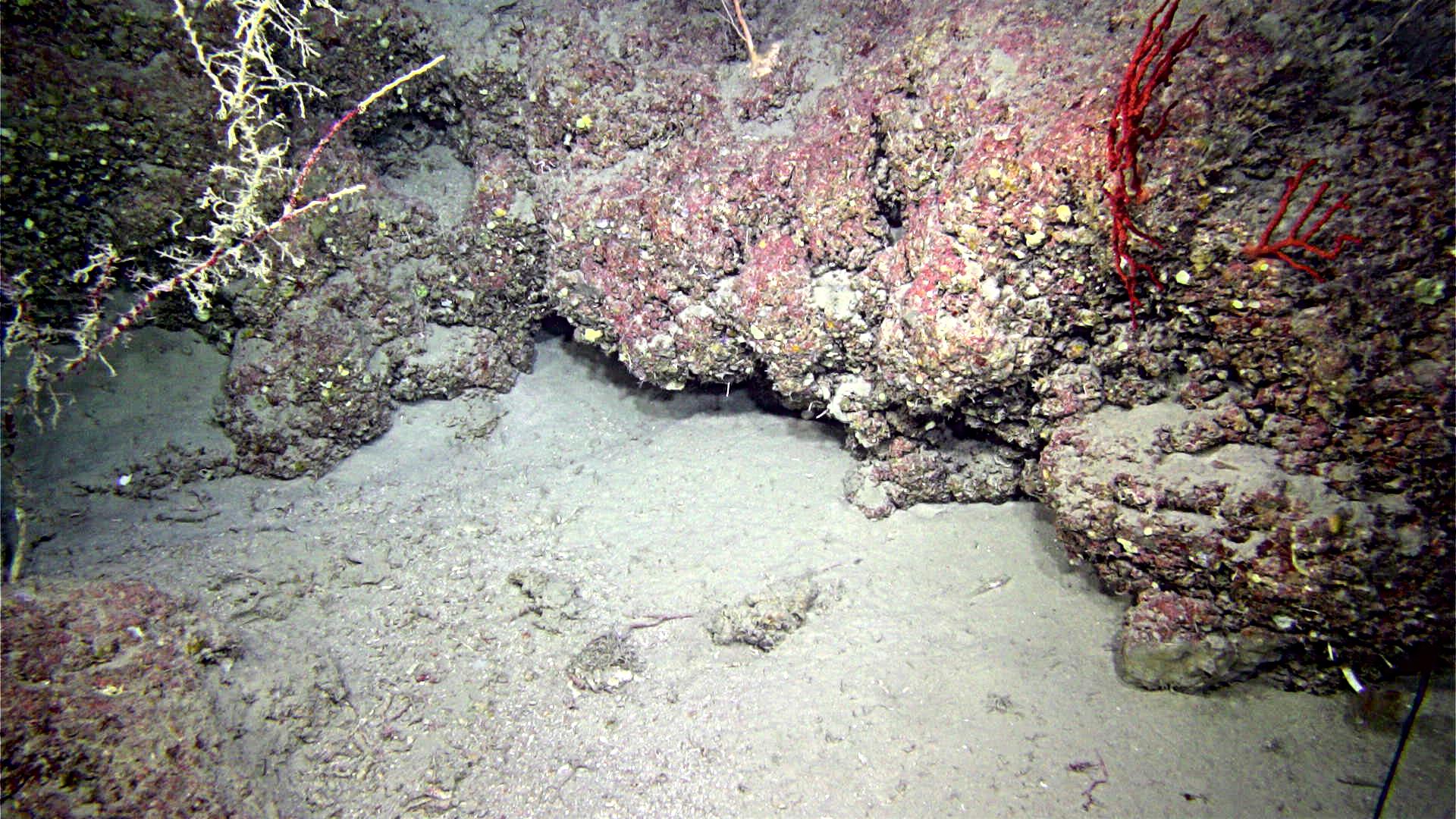} &
\includegraphics[width=\deepimgwidth,height=\deepimgheight]{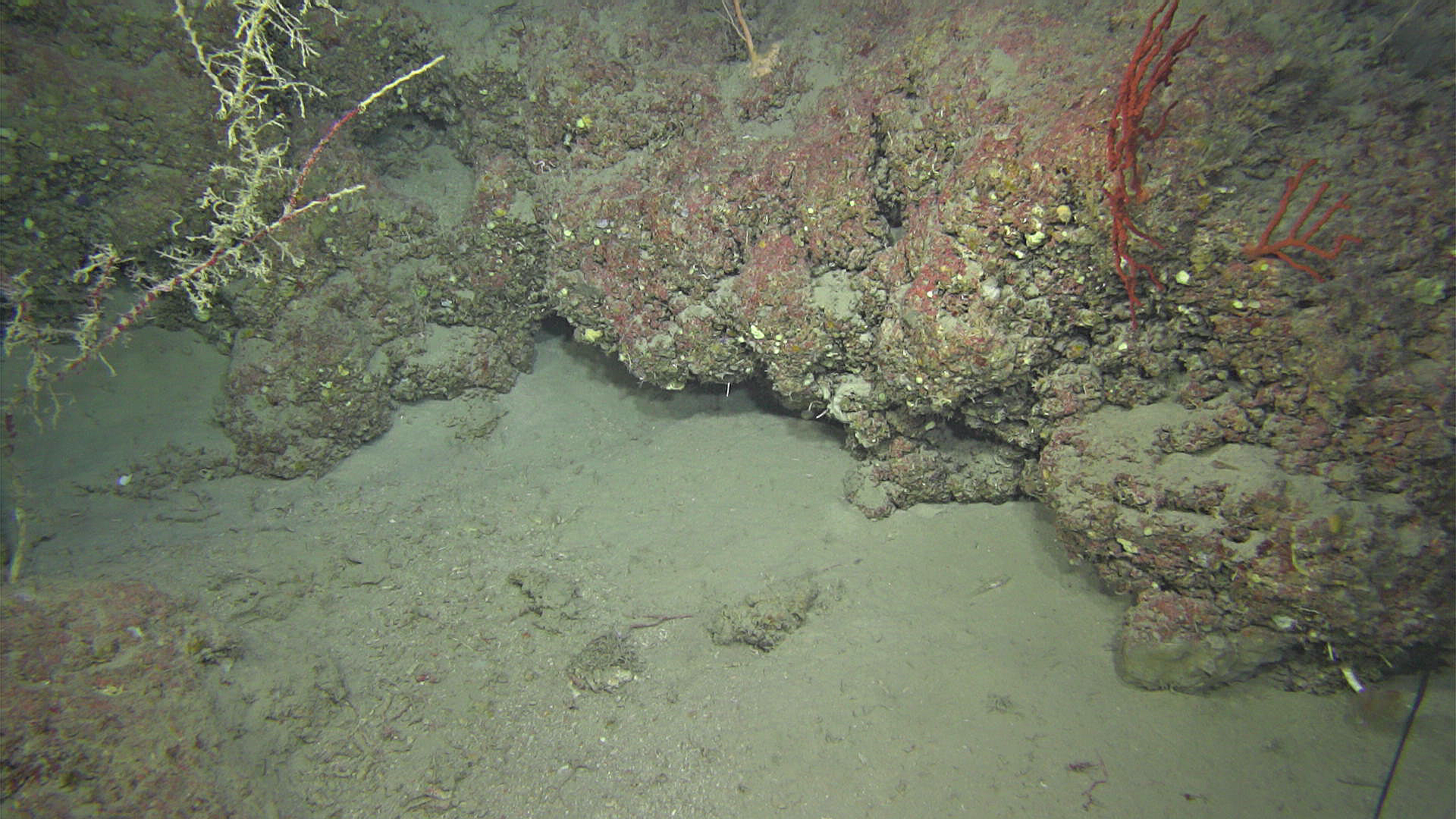} &
\includegraphics[width=\deepimgwidth,height=\deepimgheight]{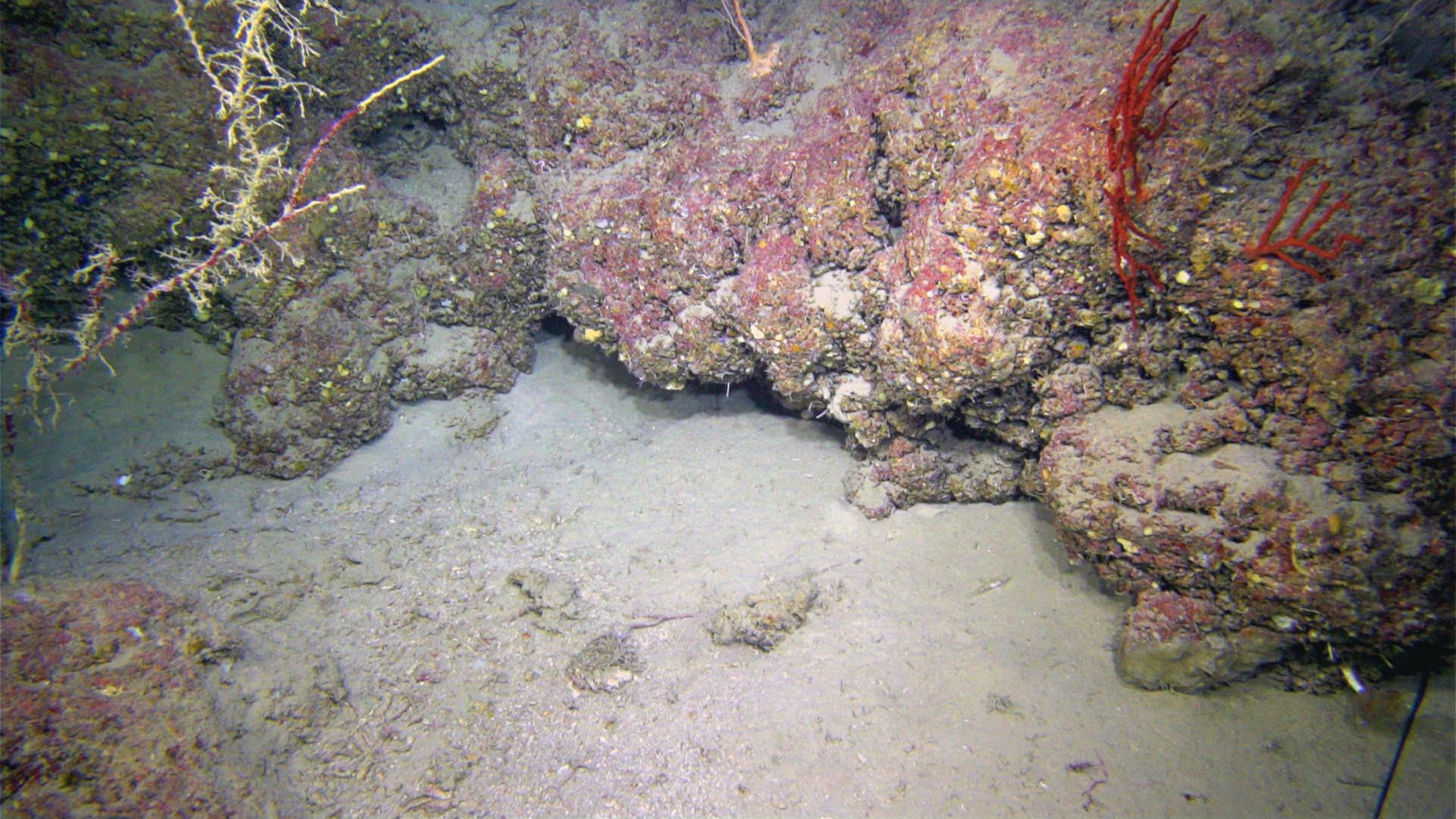} \\

\includegraphics[width=\deepimgwidth,height=\deepimgheight]{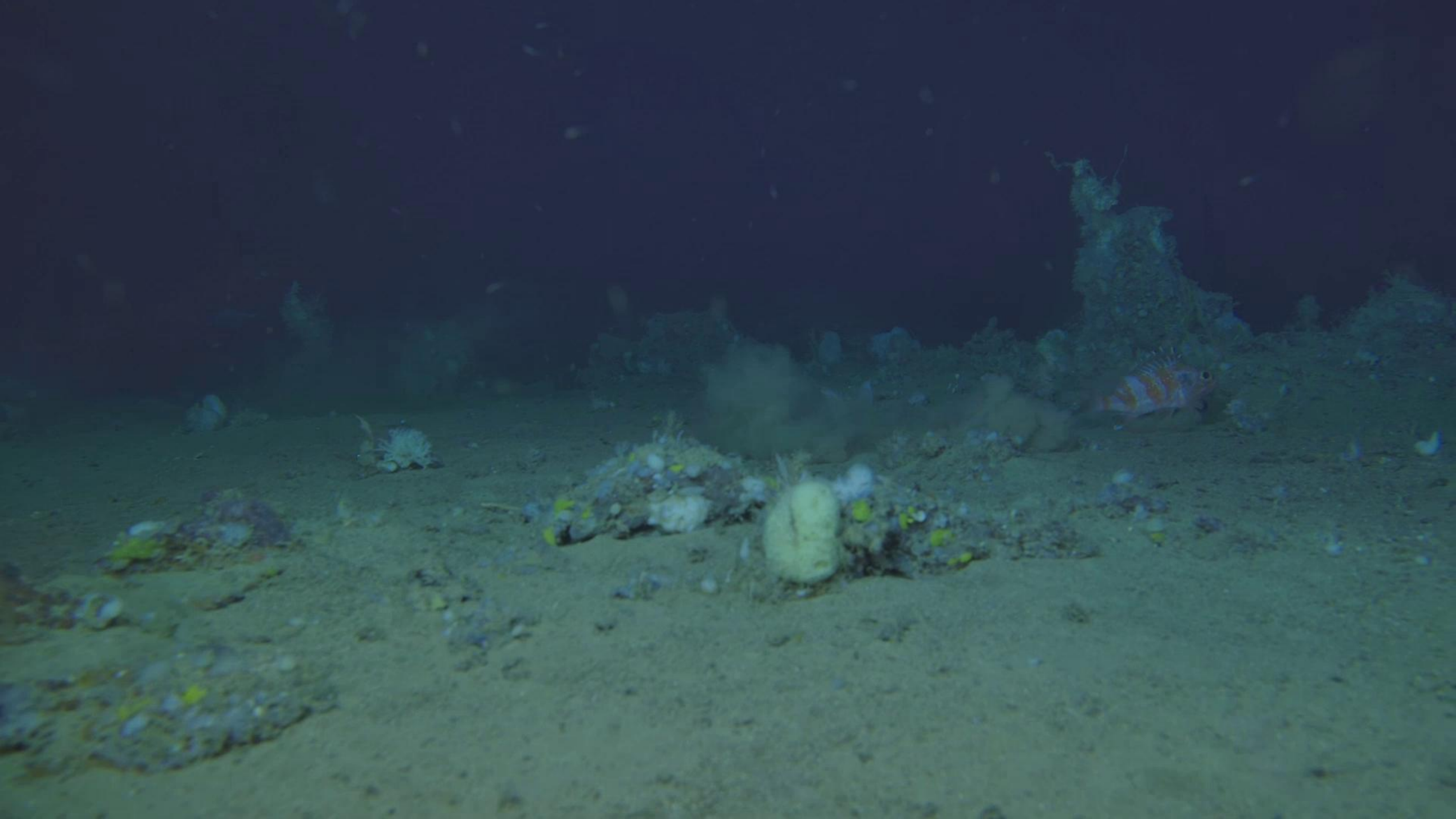} &
\includegraphics[width=\deepimgwidth,height=\deepimgheight]{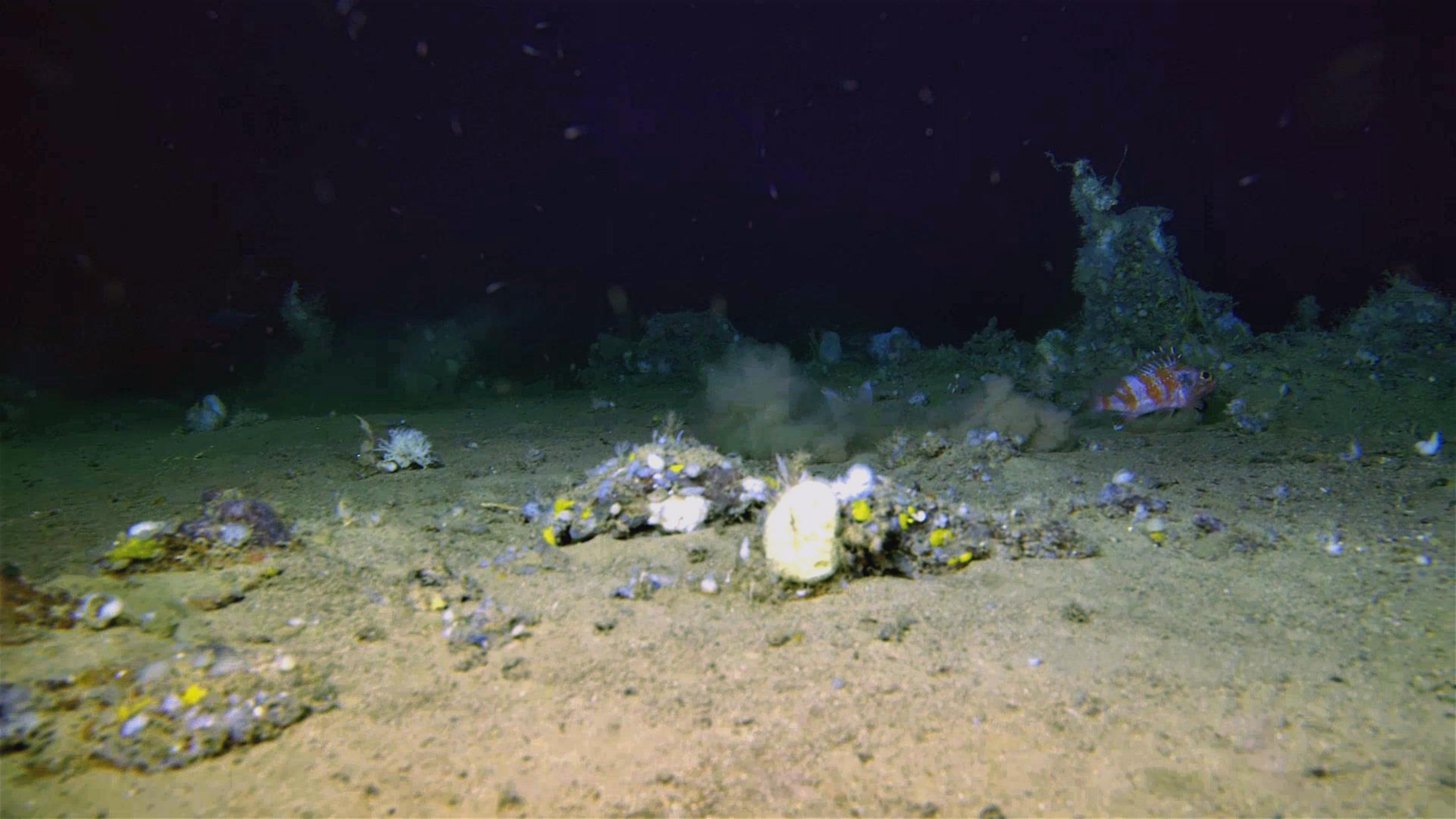} &
\includegraphics[width=\deepimgwidth,height=\deepimgheight]{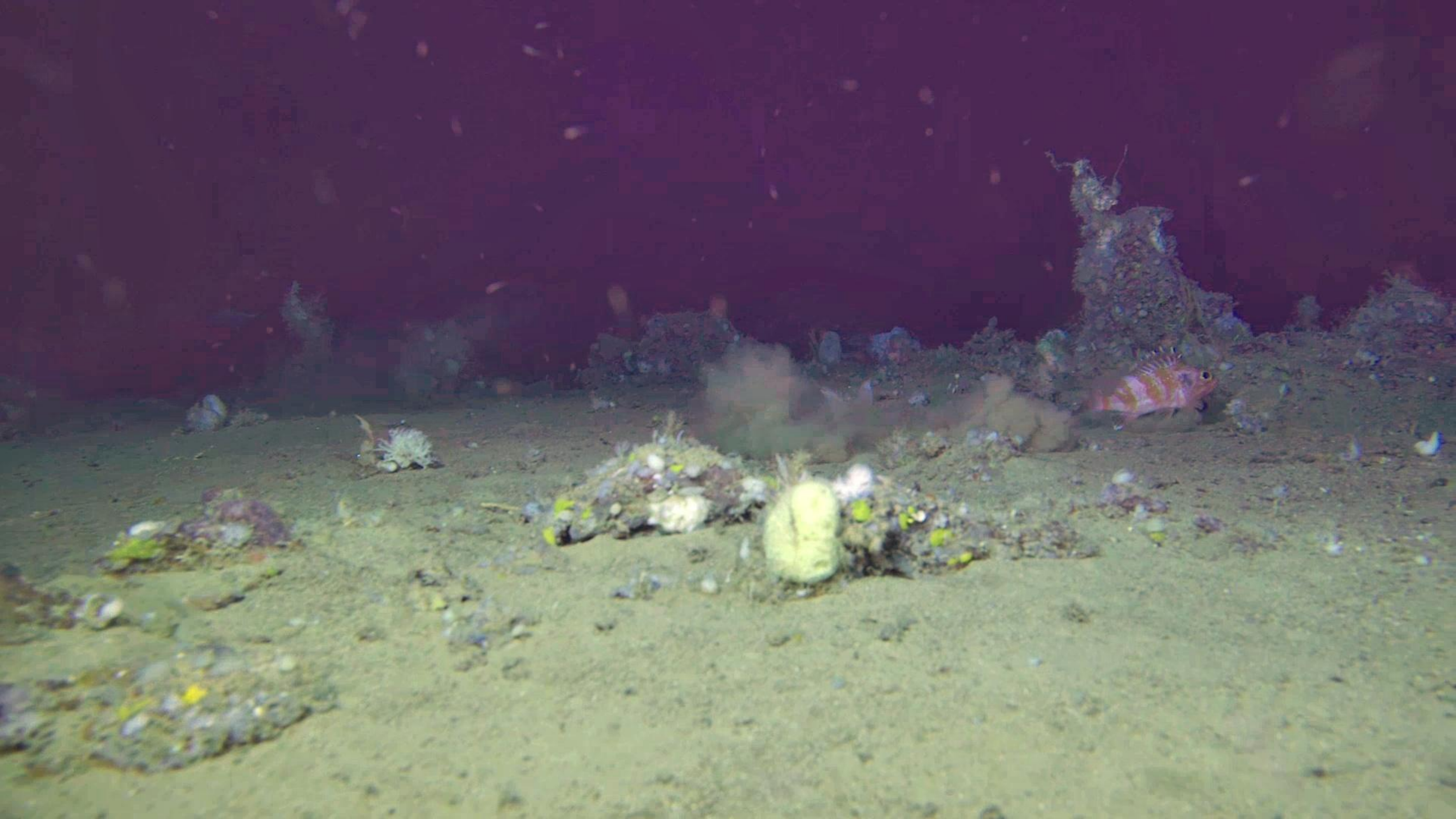} &
\includegraphics[width=\deepimgwidth,height=\deepimgheight]{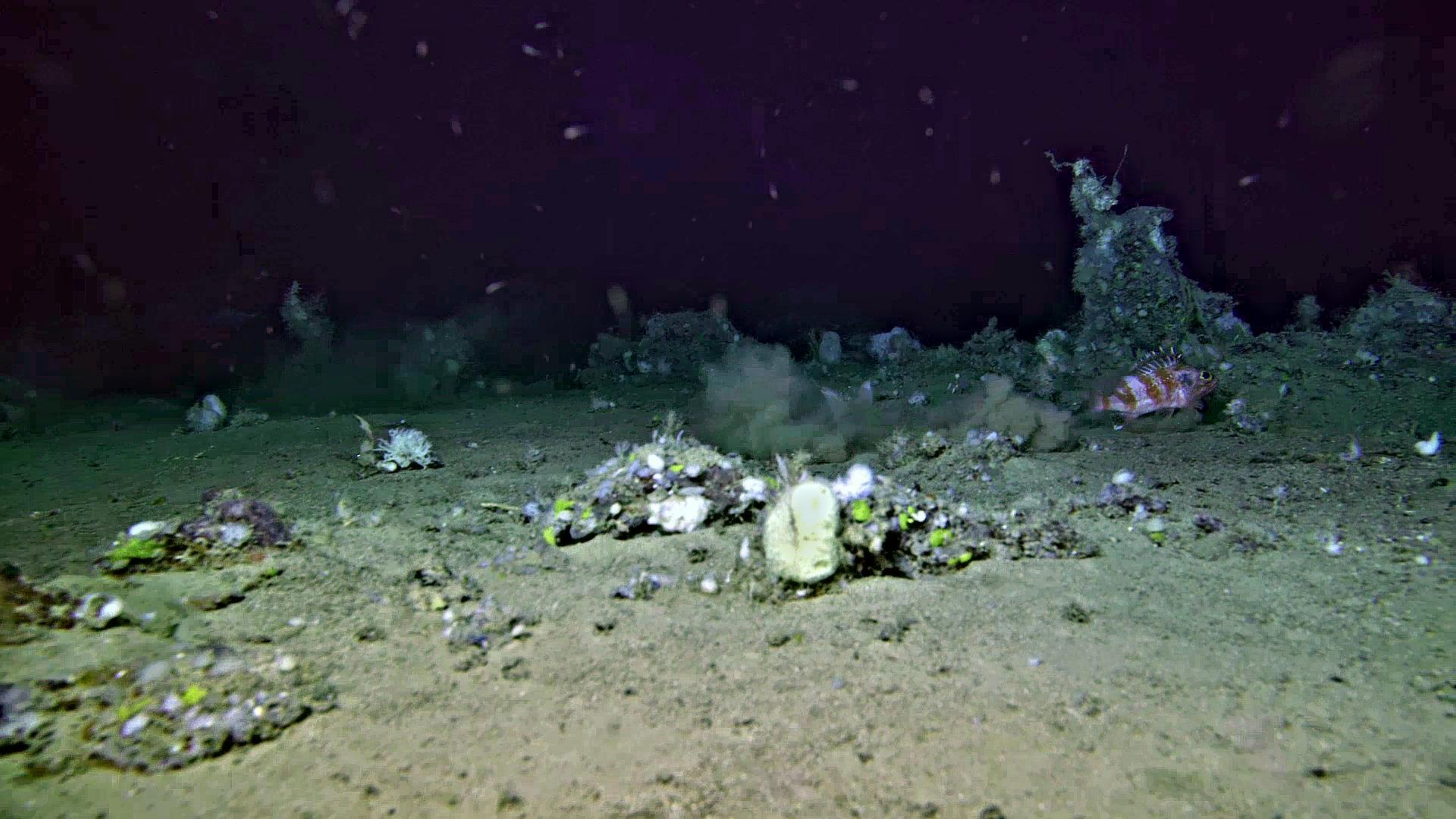} &
\includegraphics[width=\deepimgwidth,height=\deepimgheight]{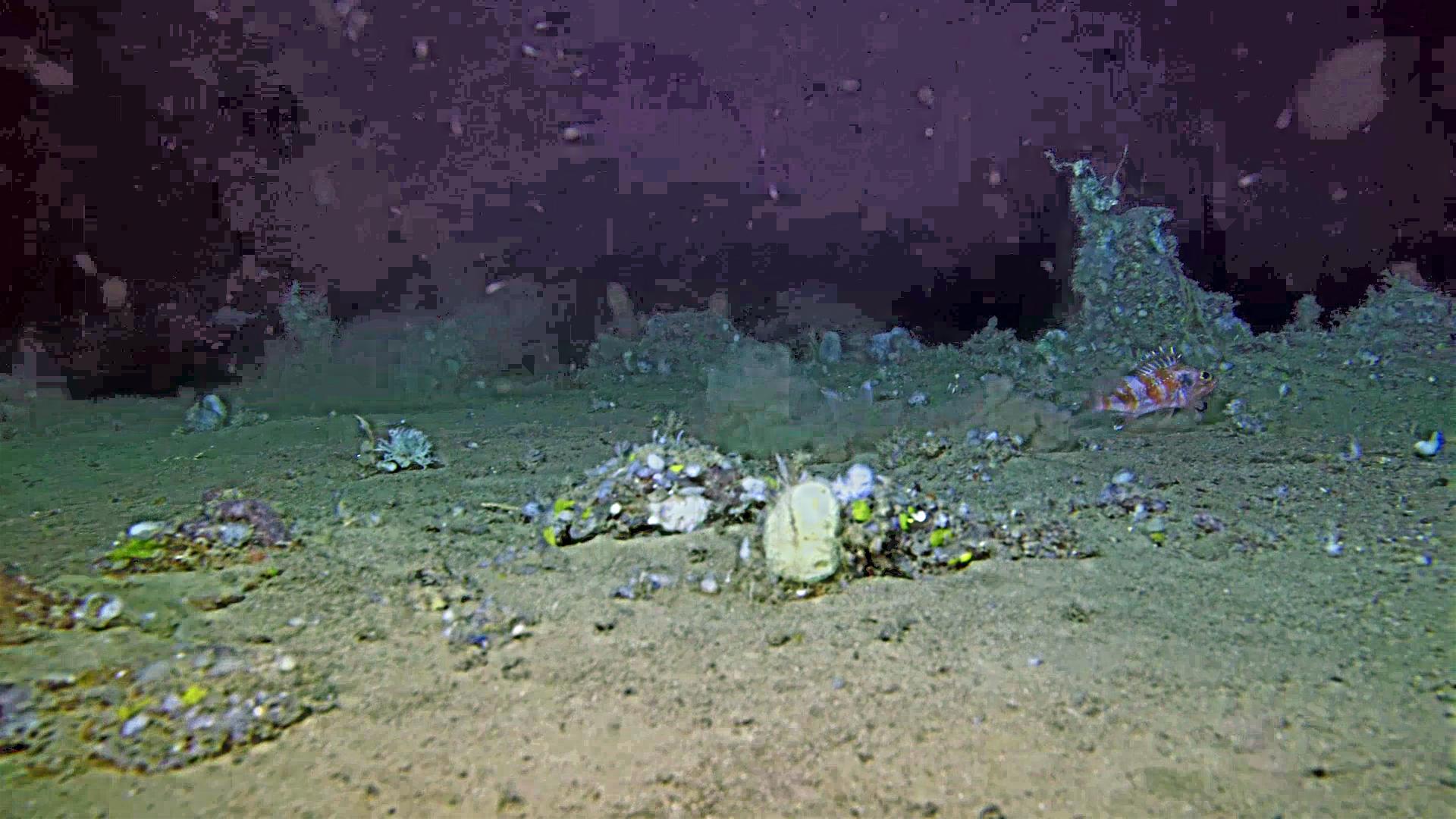} &
\includegraphics[width=\deepimgwidth,height=\deepimgheight]{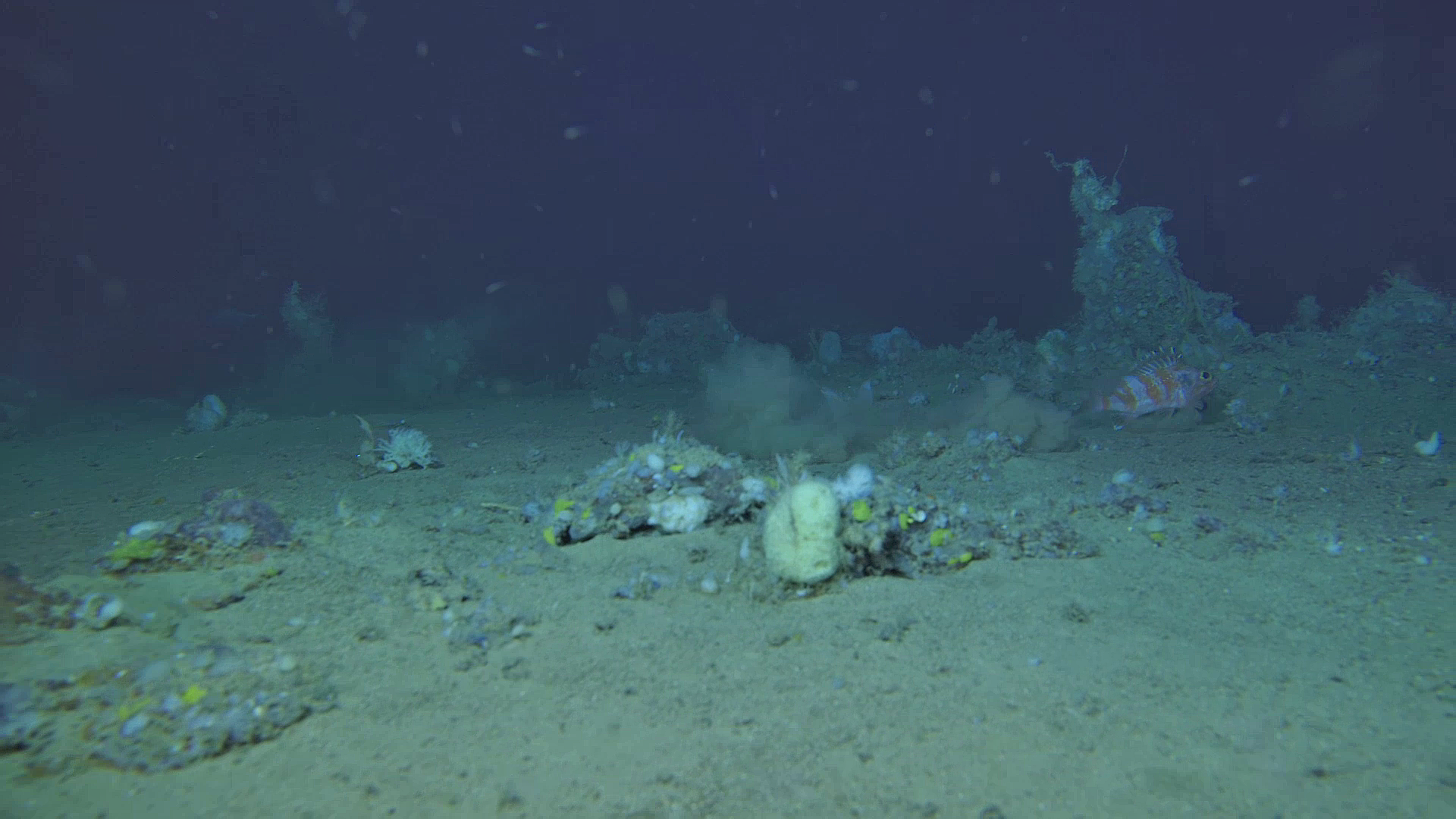} &
\includegraphics[width=\deepimgwidth,height=\deepimgheight]{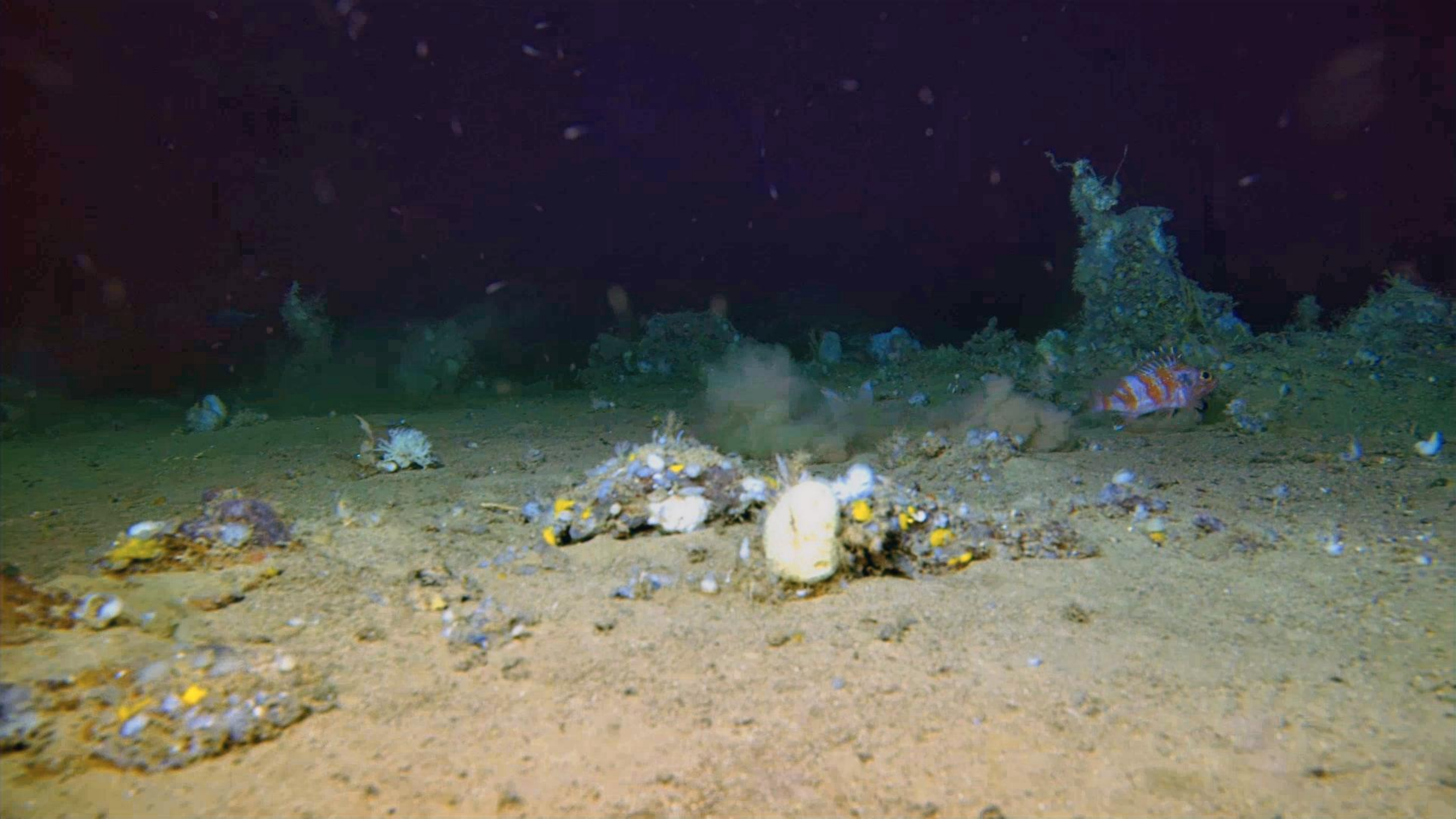} \\

\includegraphics[width=\deepimgwidth,height=\deepimgheight]{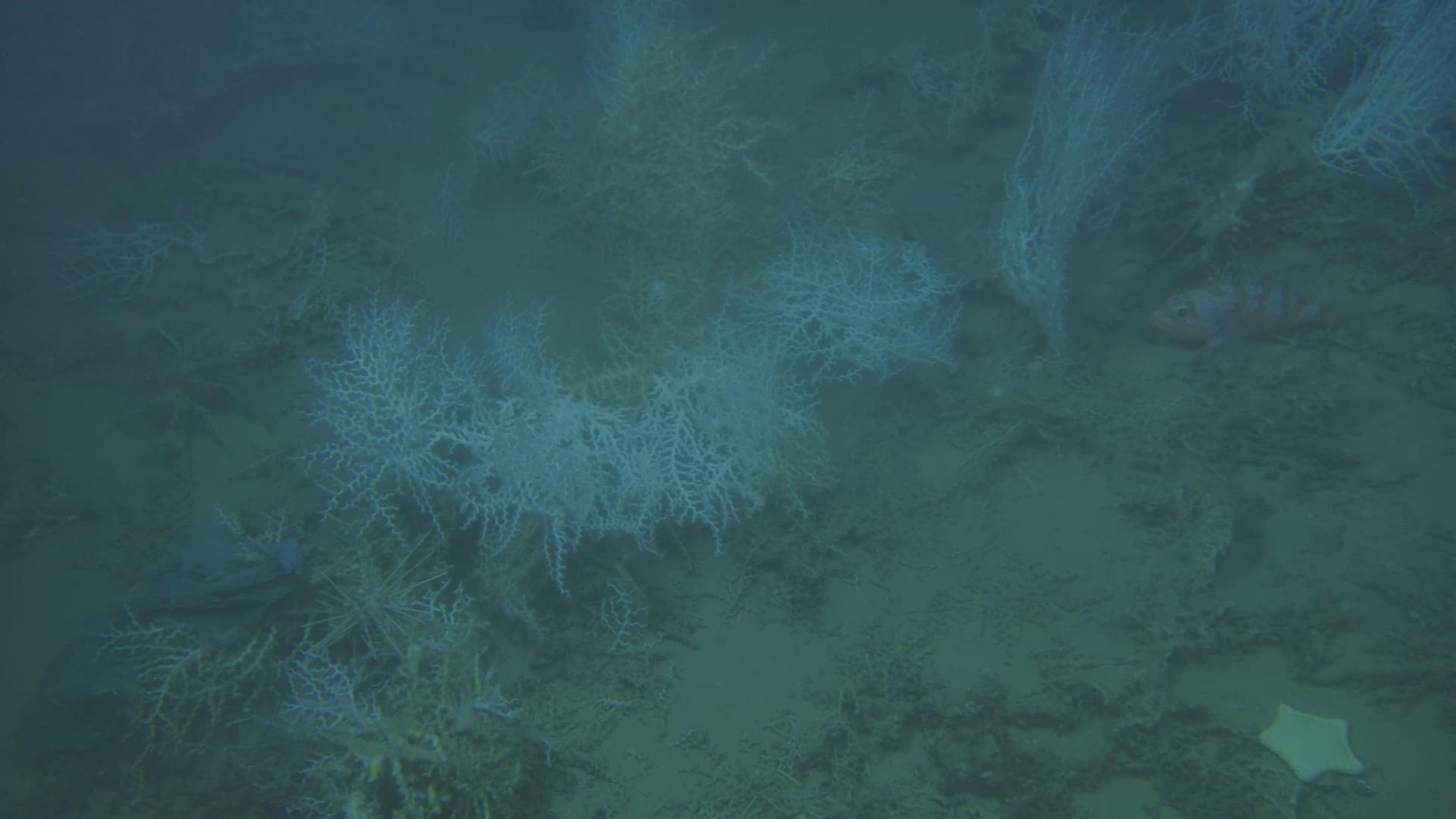} &
\includegraphics[width=\deepimgwidth,height=\deepimgheight]{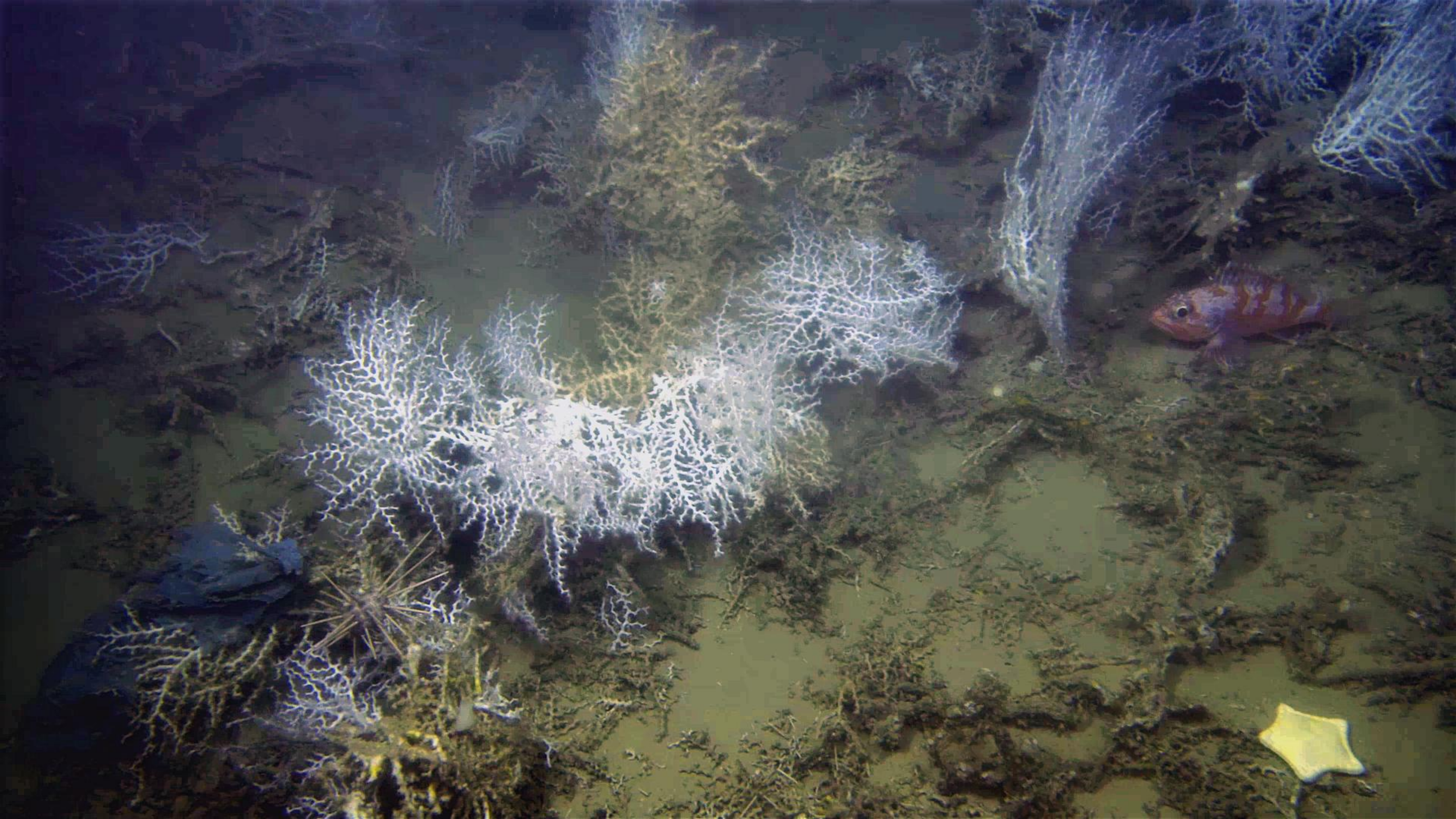} &
\includegraphics[width=\deepimgwidth,height=\deepimgheight]{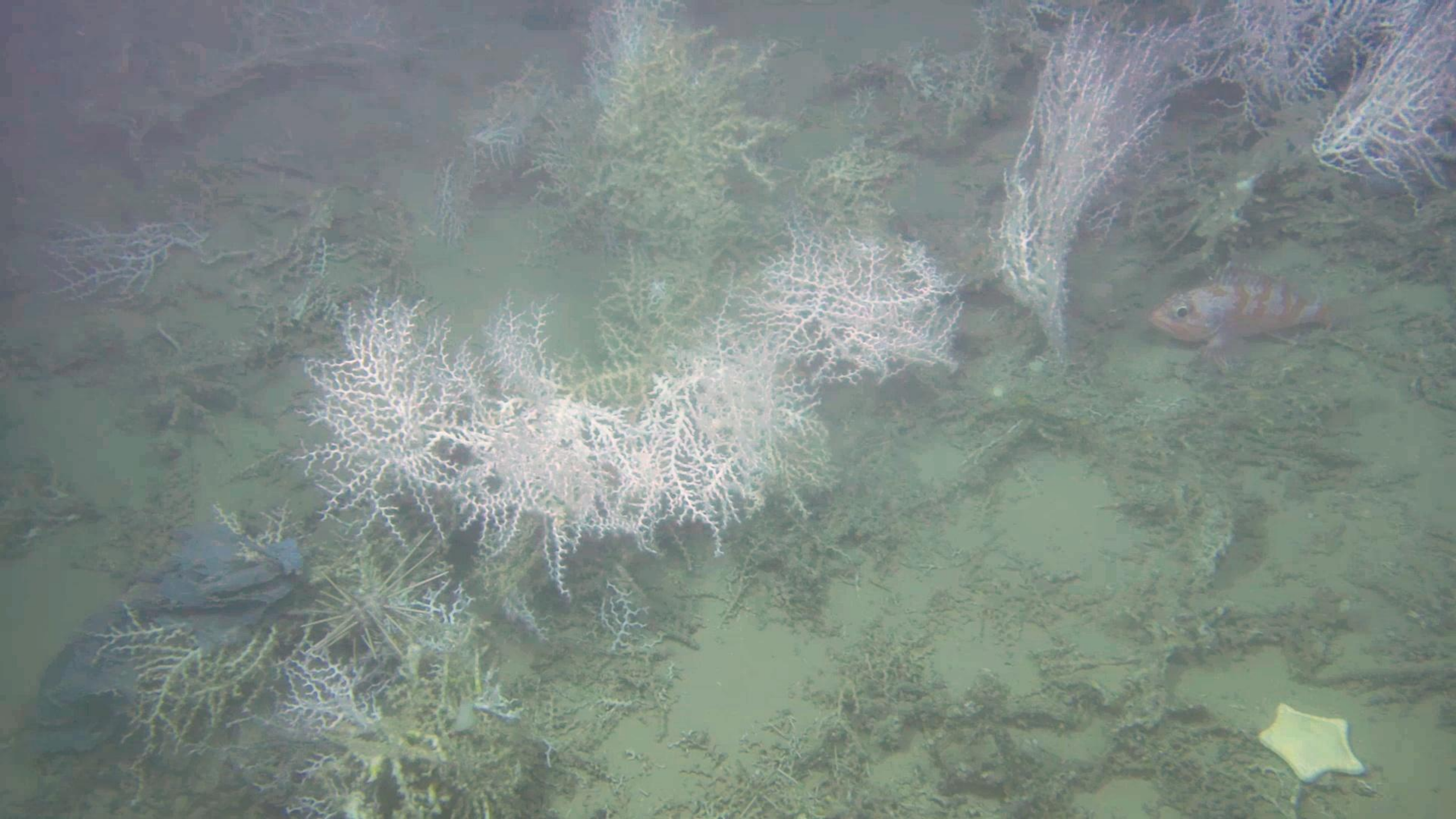} &
\includegraphics[width=\deepimgwidth,height=\deepimgheight]{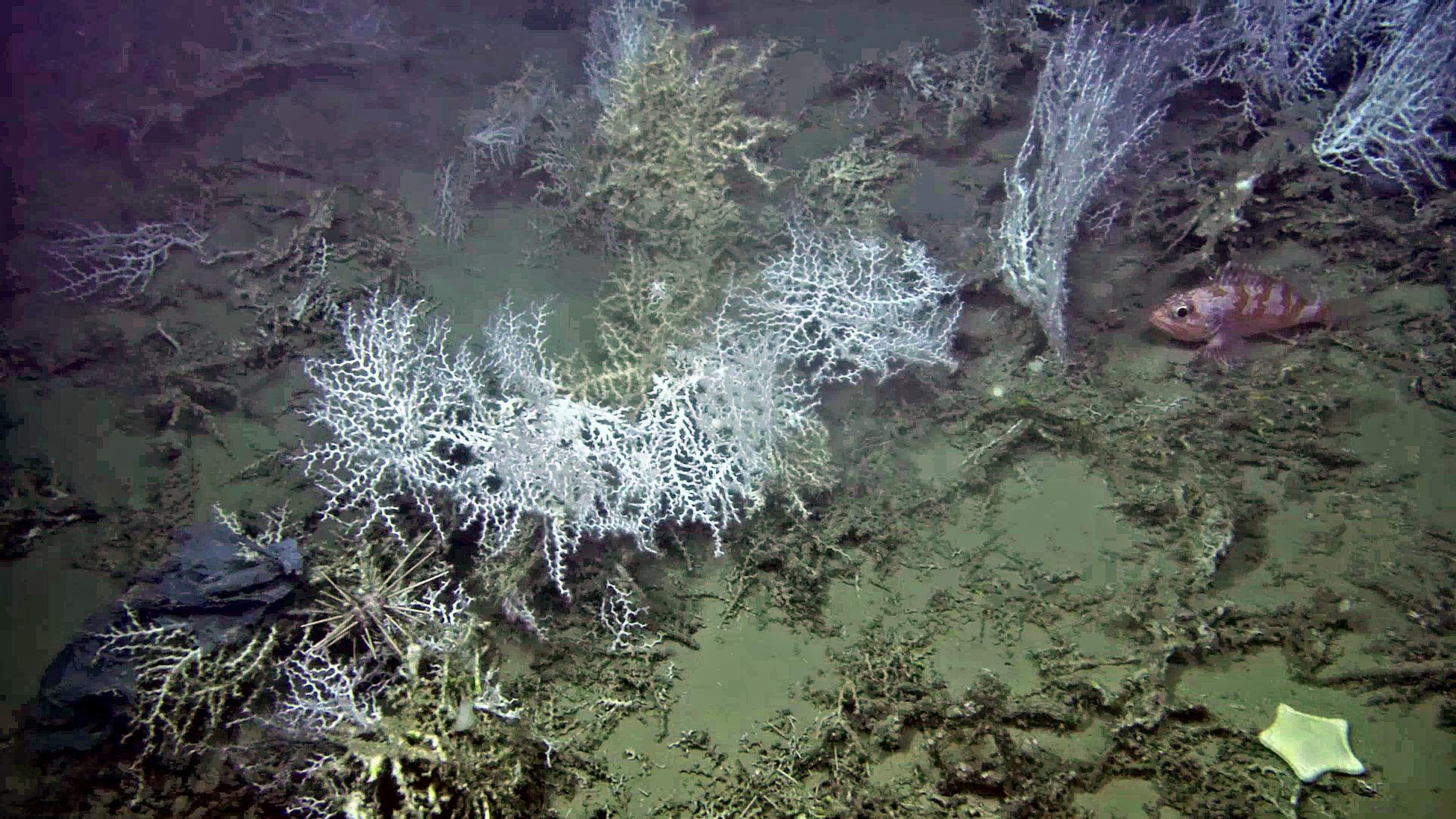} &
\includegraphics[width=\deepimgwidth,height=\deepimgheight]{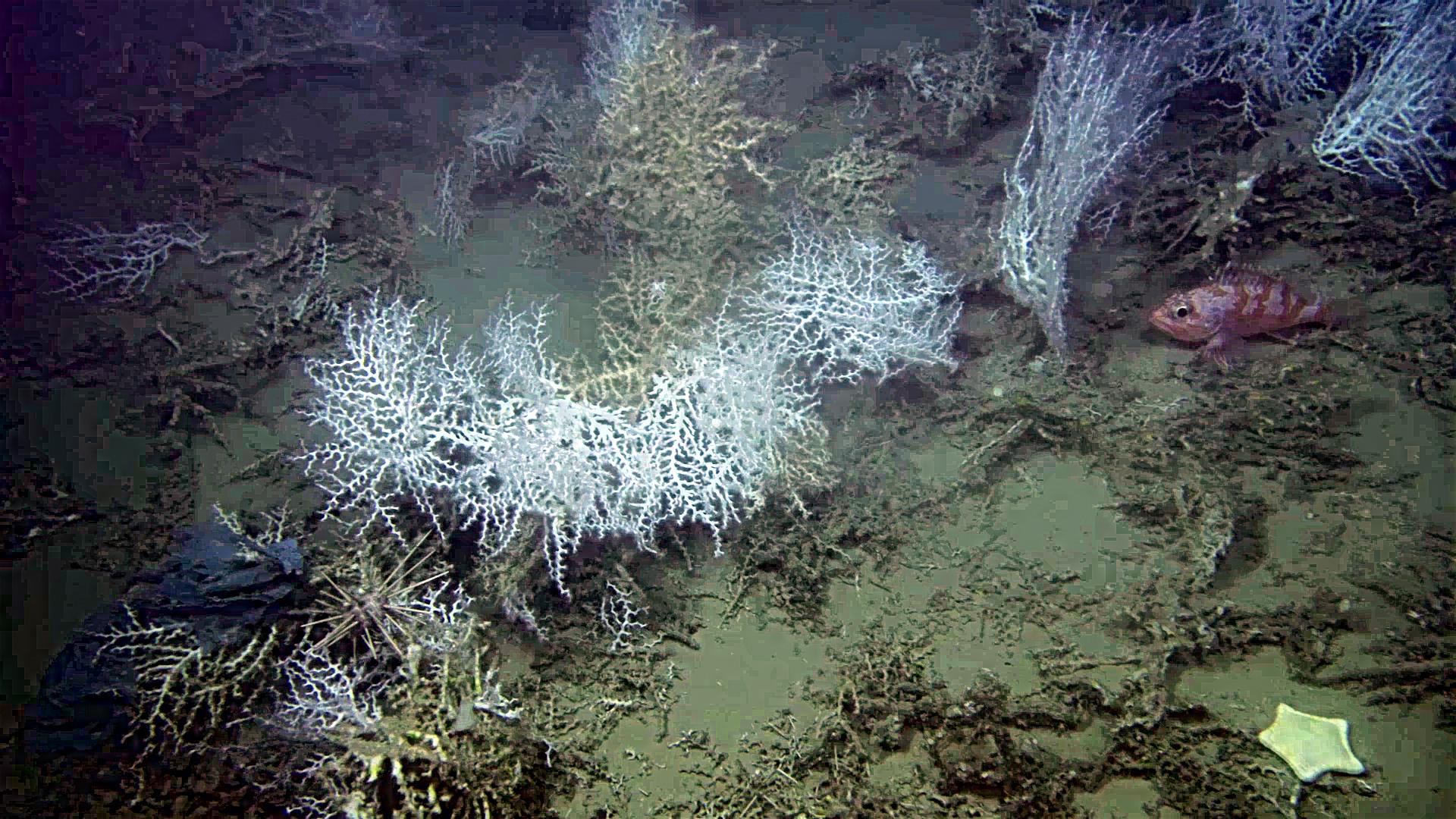} &
\includegraphics[width=\deepimgwidth,height=\deepimgheight]{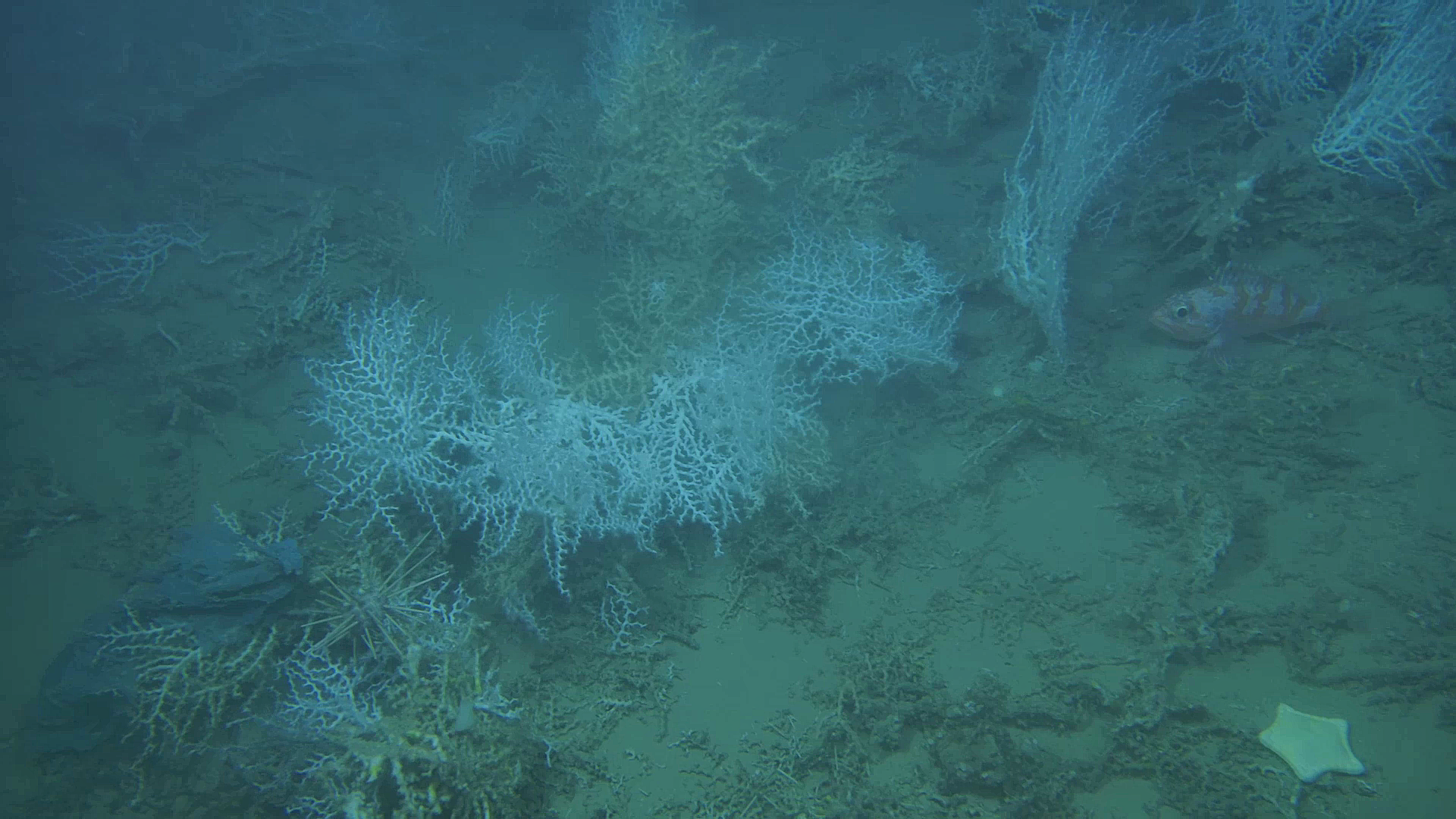} &
\includegraphics[width=\deepimgwidth,height=\deepimgheight]{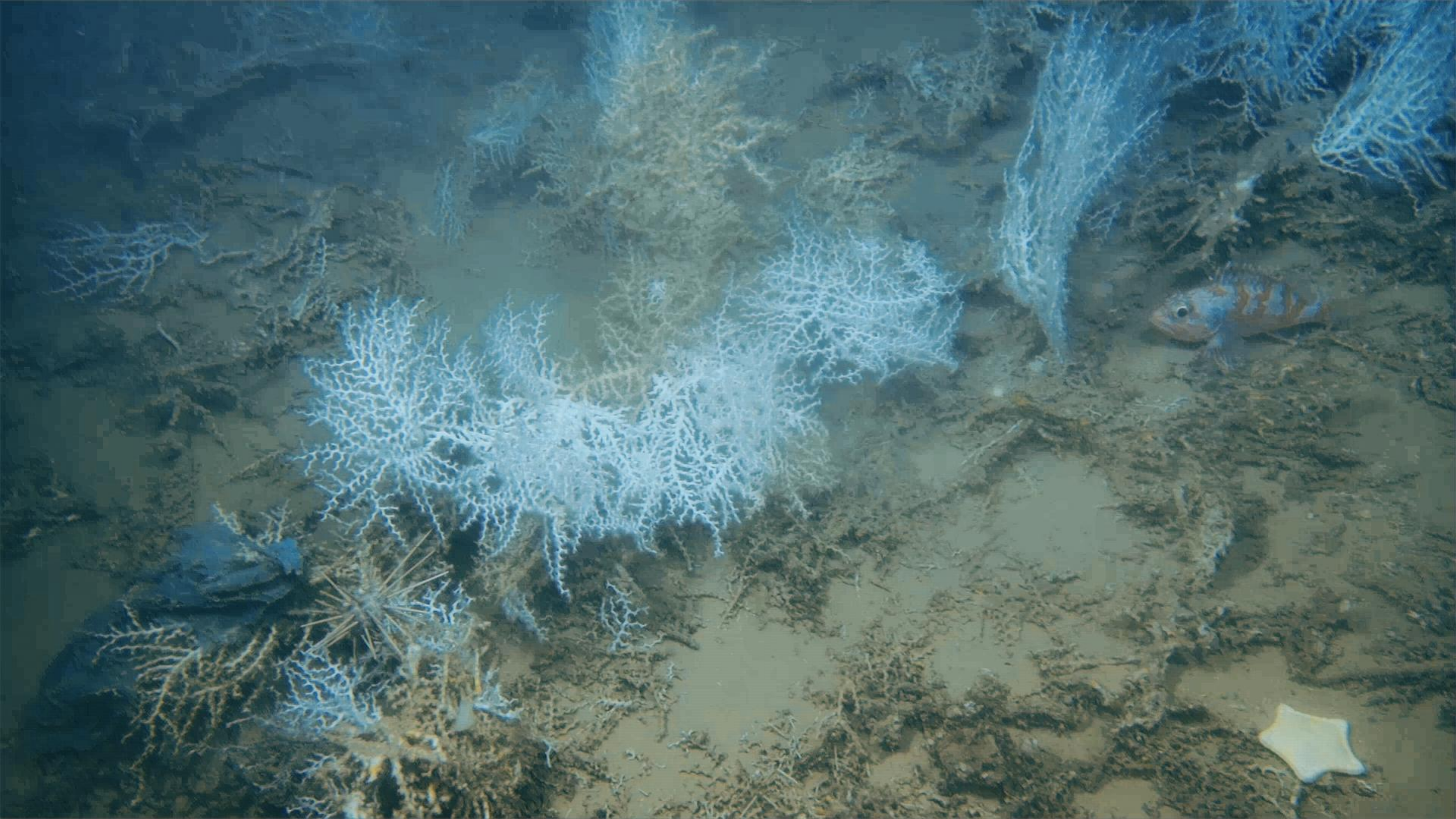} \\[5pt]

\textbf{\tiny UWCNN \cite{li2020underwater}} &
\textbf{\tiny PUIE \cite{fu2022uncertainty}} &
\textbf{\tiny TACL \cite{liu2022twin}} &
\textbf{\tiny NU2Net \cite{guo2023underwater}} &
\textbf{\tiny CCL-Net \cite{liu2024underwater}} &
\textbf{\tiny OUNet-JL \cite{wang2025optimized}} &
\textbf{\tiny AQUA-Net} \\[3pt]

\includegraphics[width=\deepimgwidth,height=\deepimgheight]{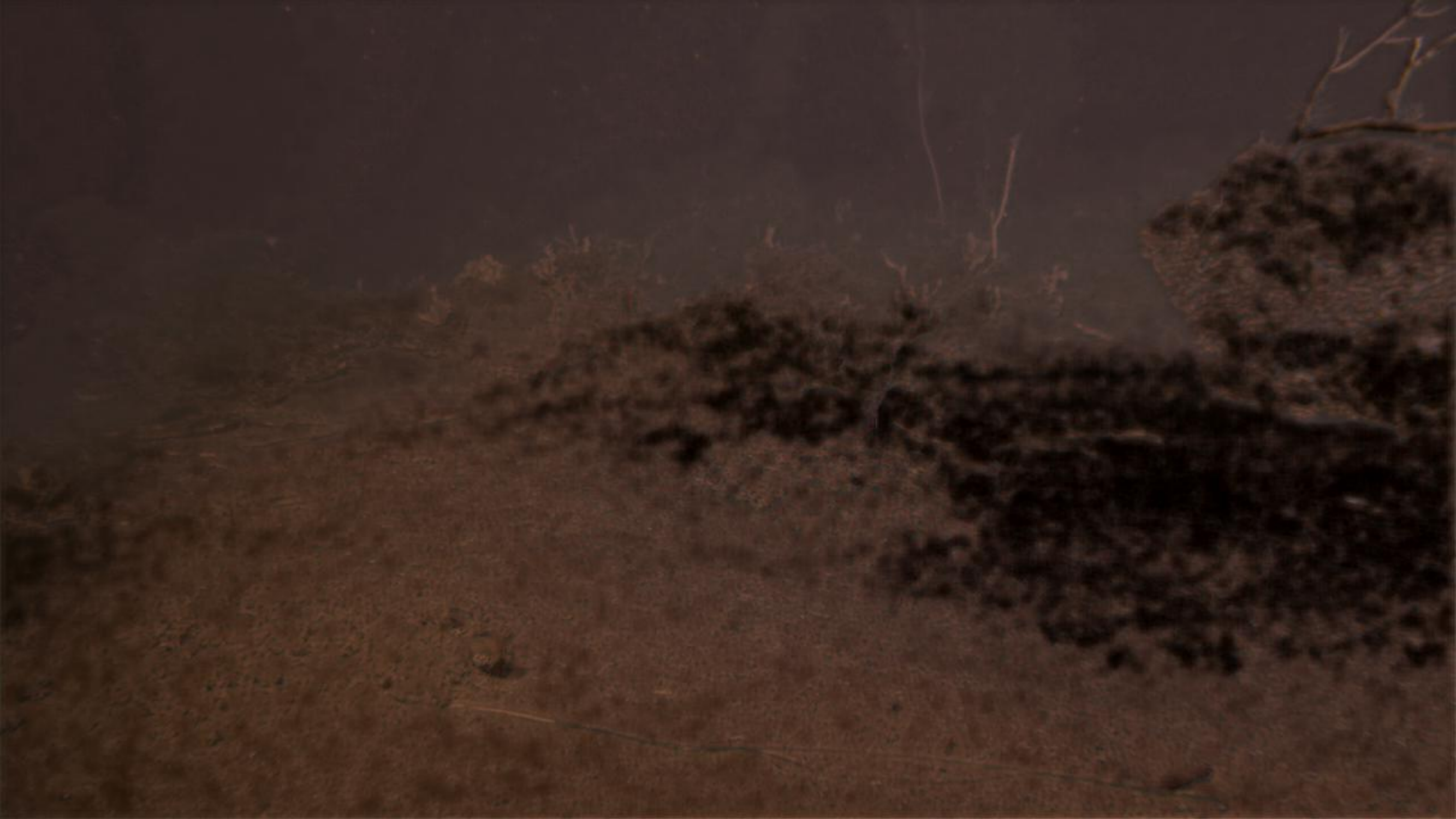} &
\includegraphics[width=\deepimgwidth,height=\deepimgheight]{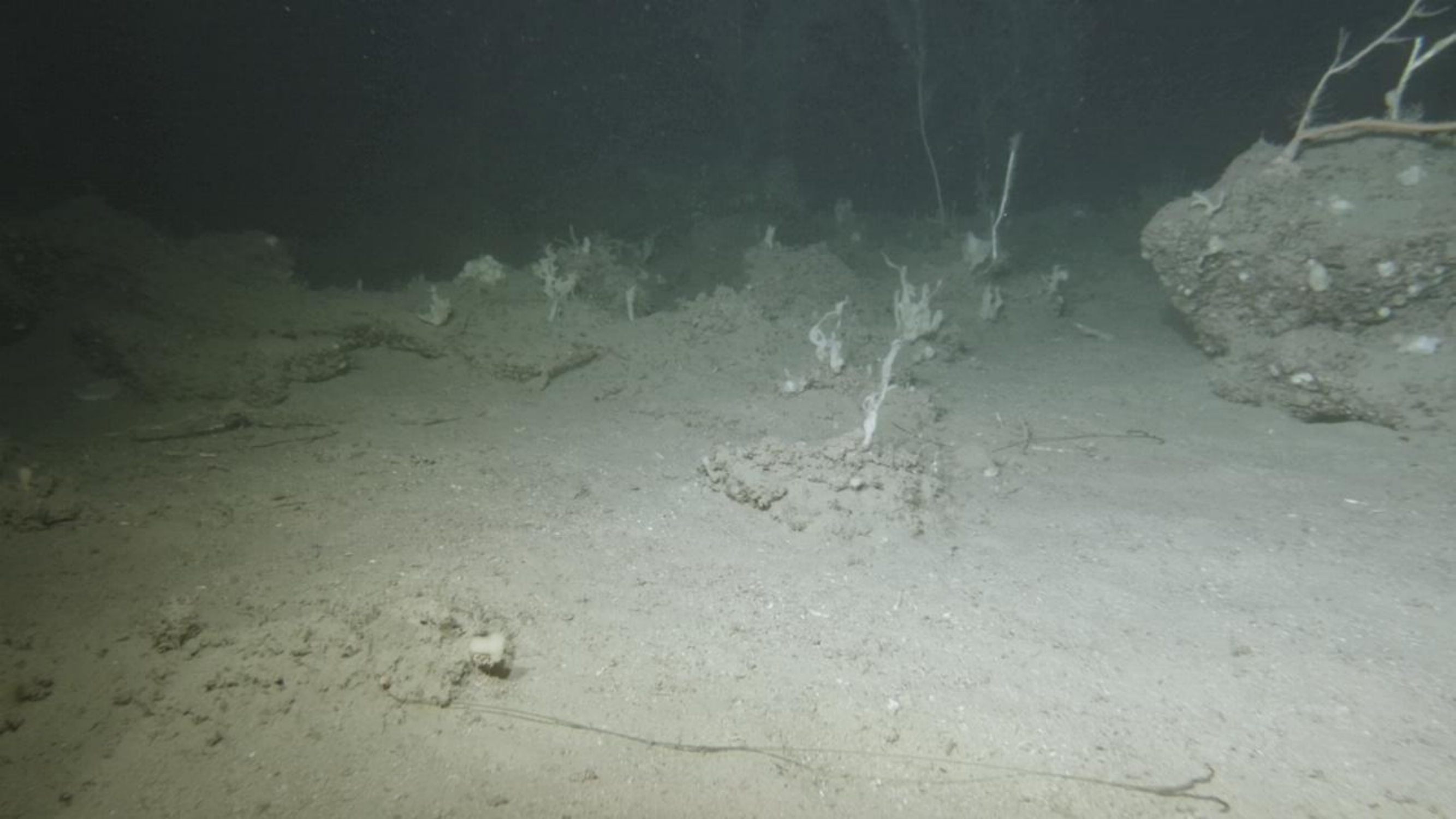} &
\includegraphics[width=\deepimgwidth,height=\deepimgheight]{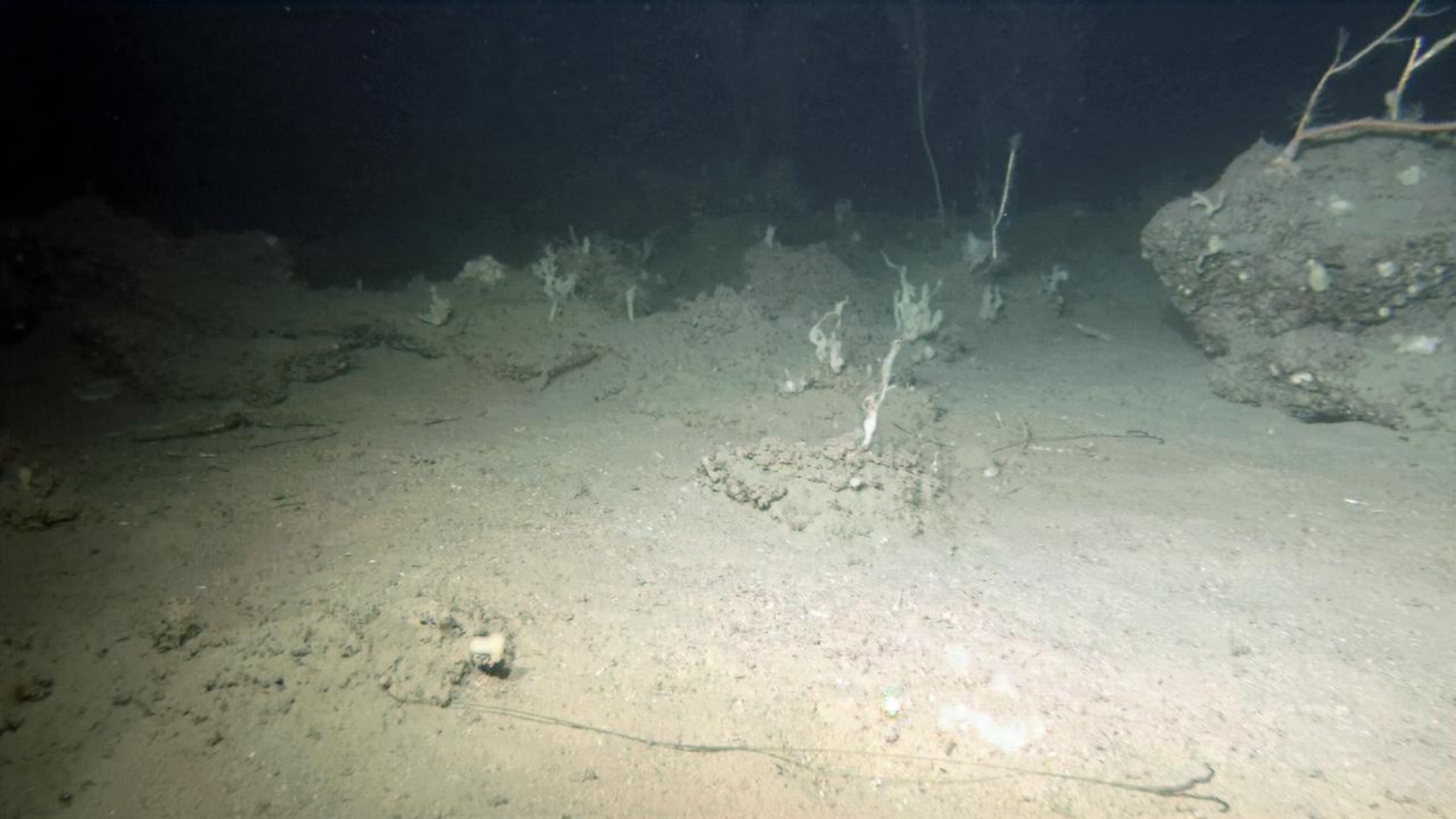} &
\includegraphics[width=\deepimgwidth,height=\deepimgheight]{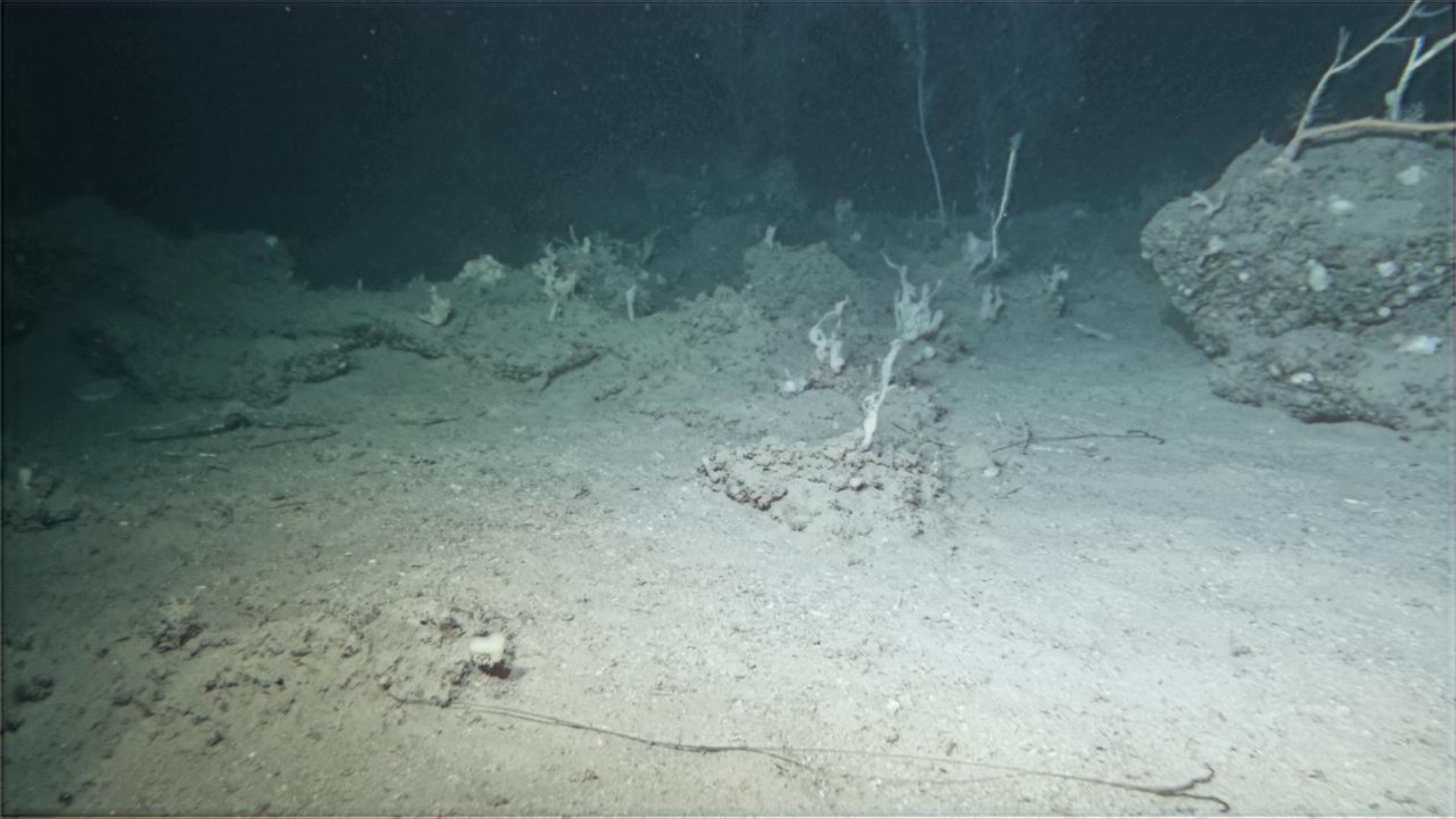} &
\includegraphics[width=\deepimgwidth,height=\deepimgheight]{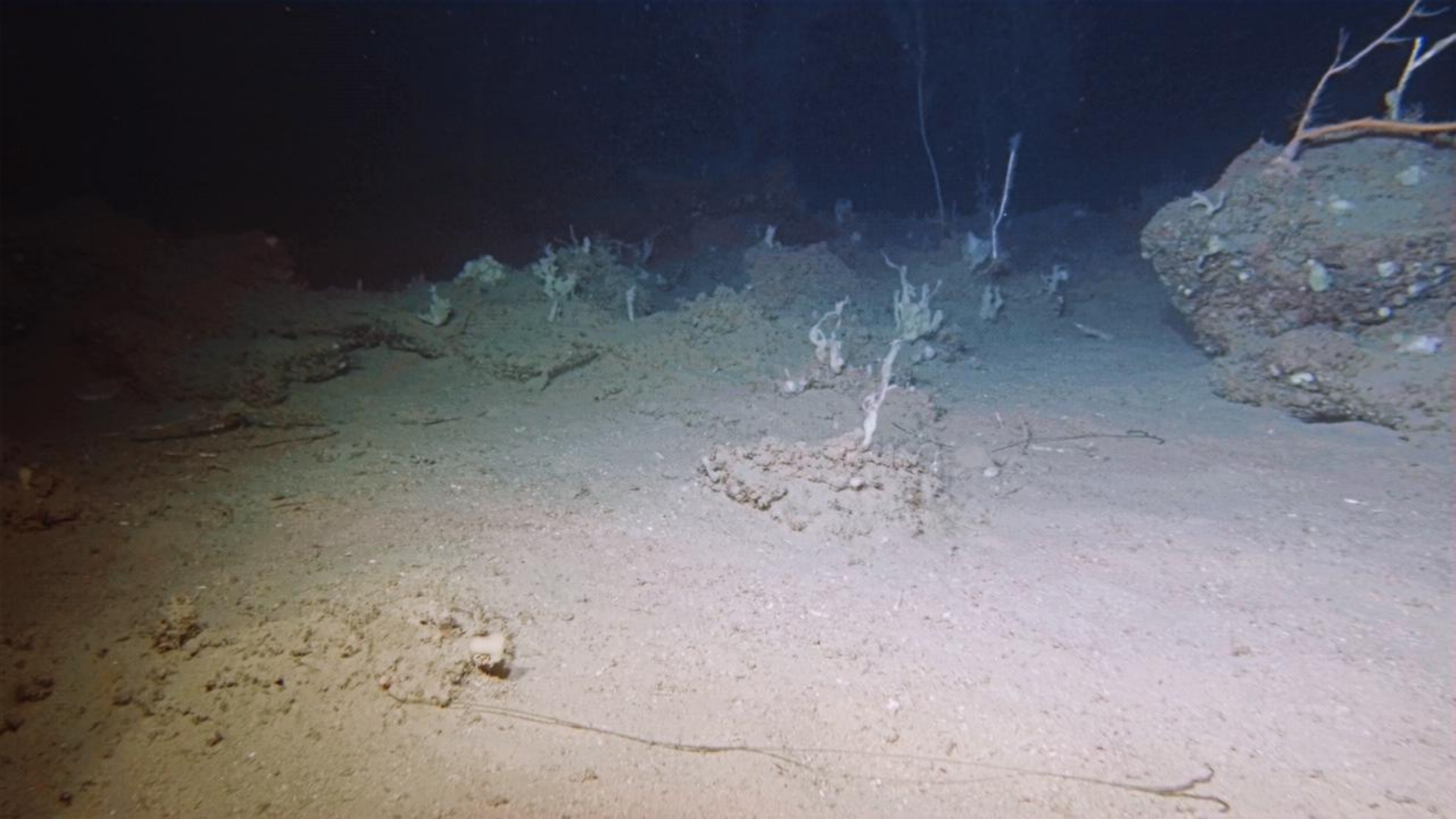} &
\includegraphics[width=\deepimgwidth,height=\deepimgheight]{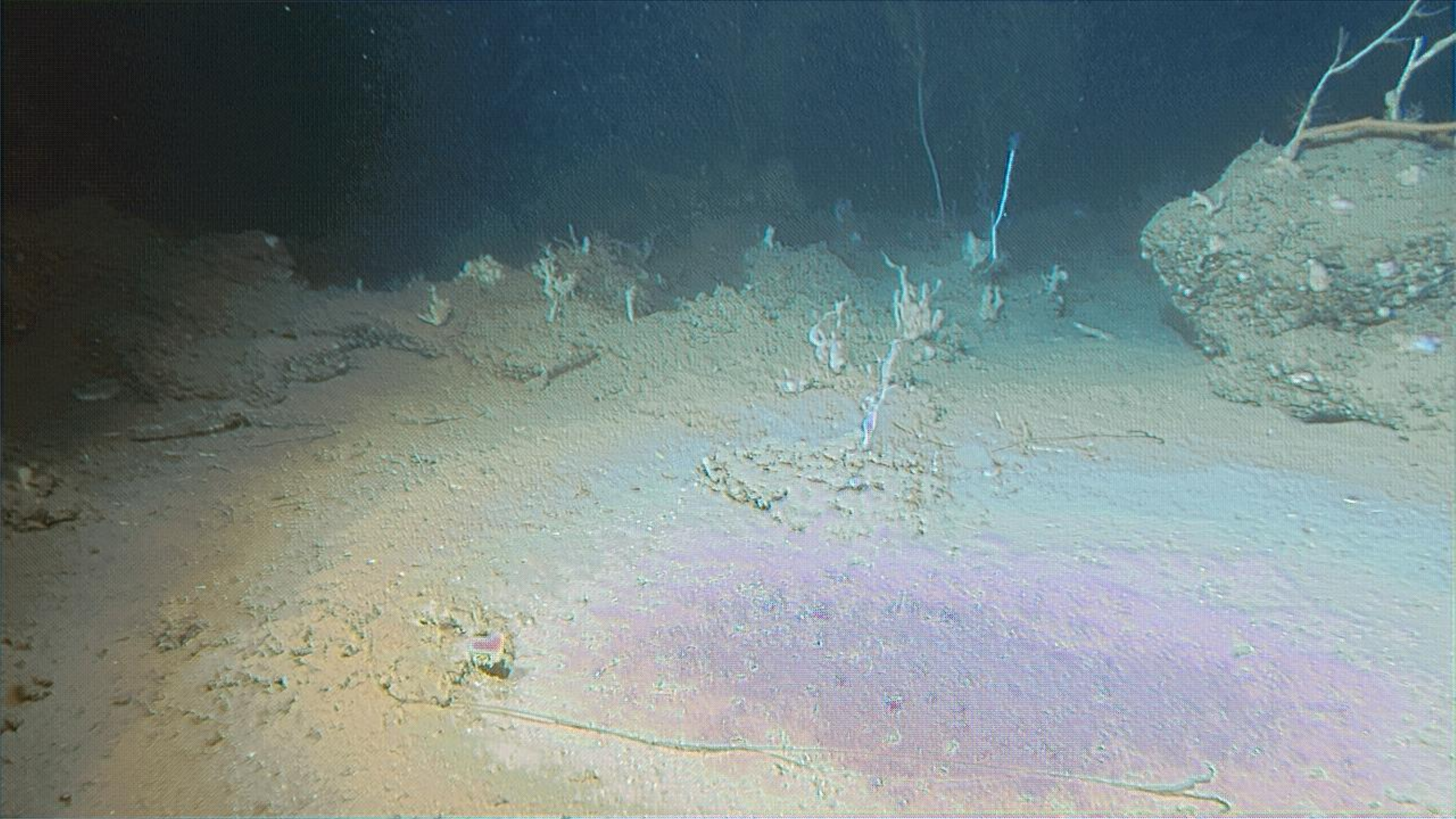} &
\includegraphics[width=\deepimgwidth,height=\deepimgheight]{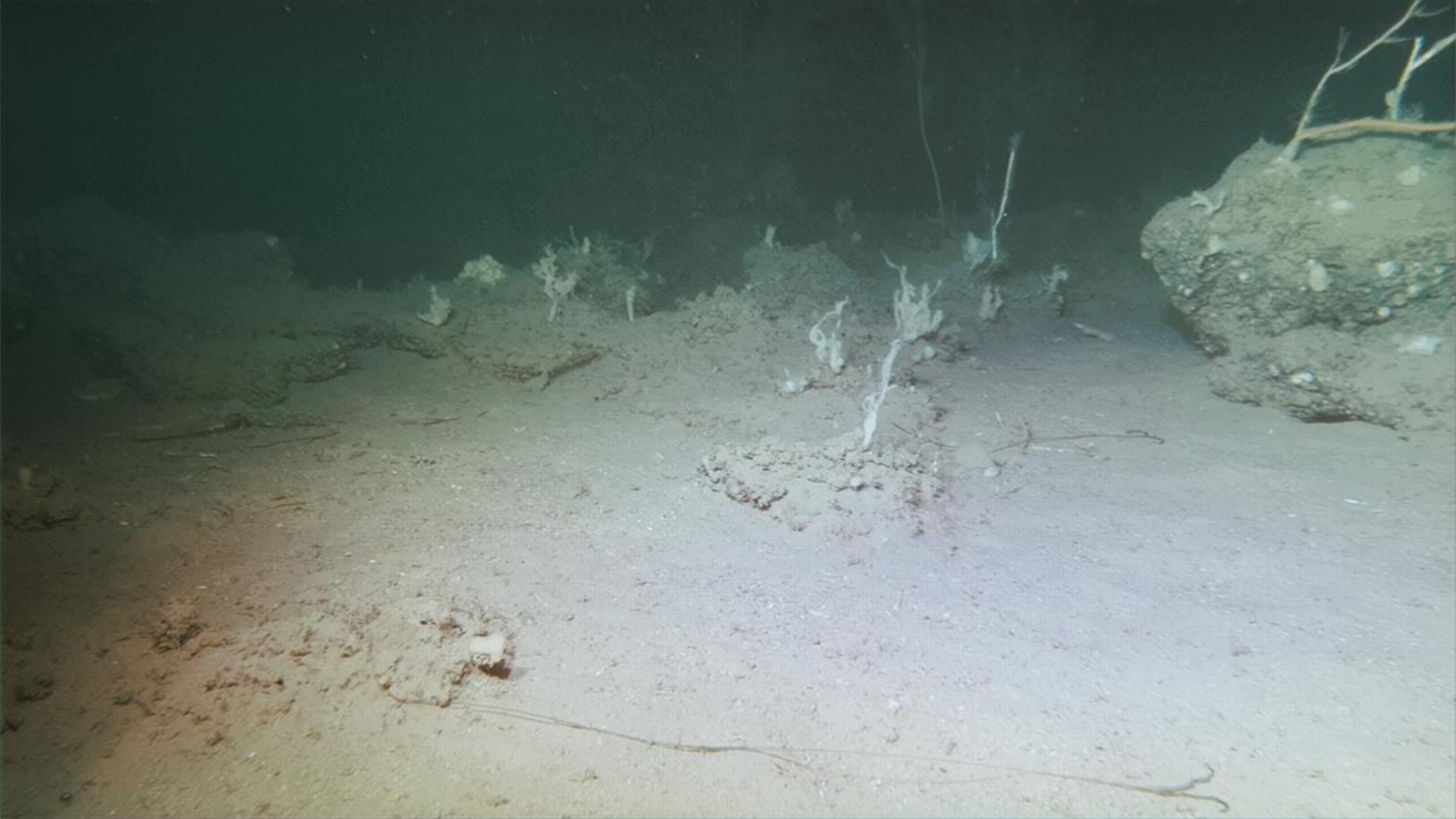} \\

\includegraphics[width=\deepimgwidth,height=\deepimgheight]{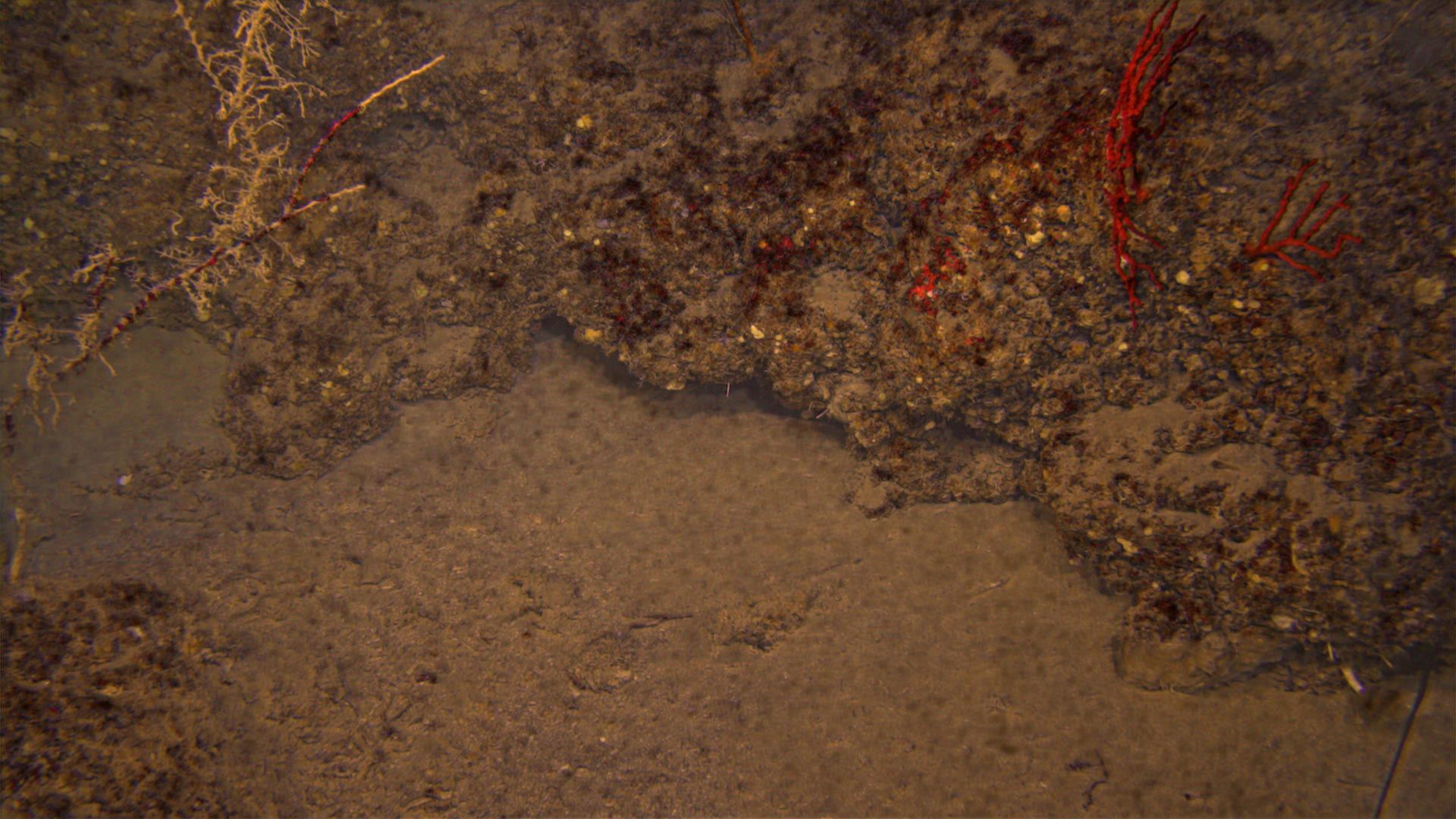} &
\includegraphics[width=\deepimgwidth,height=\deepimgheight]{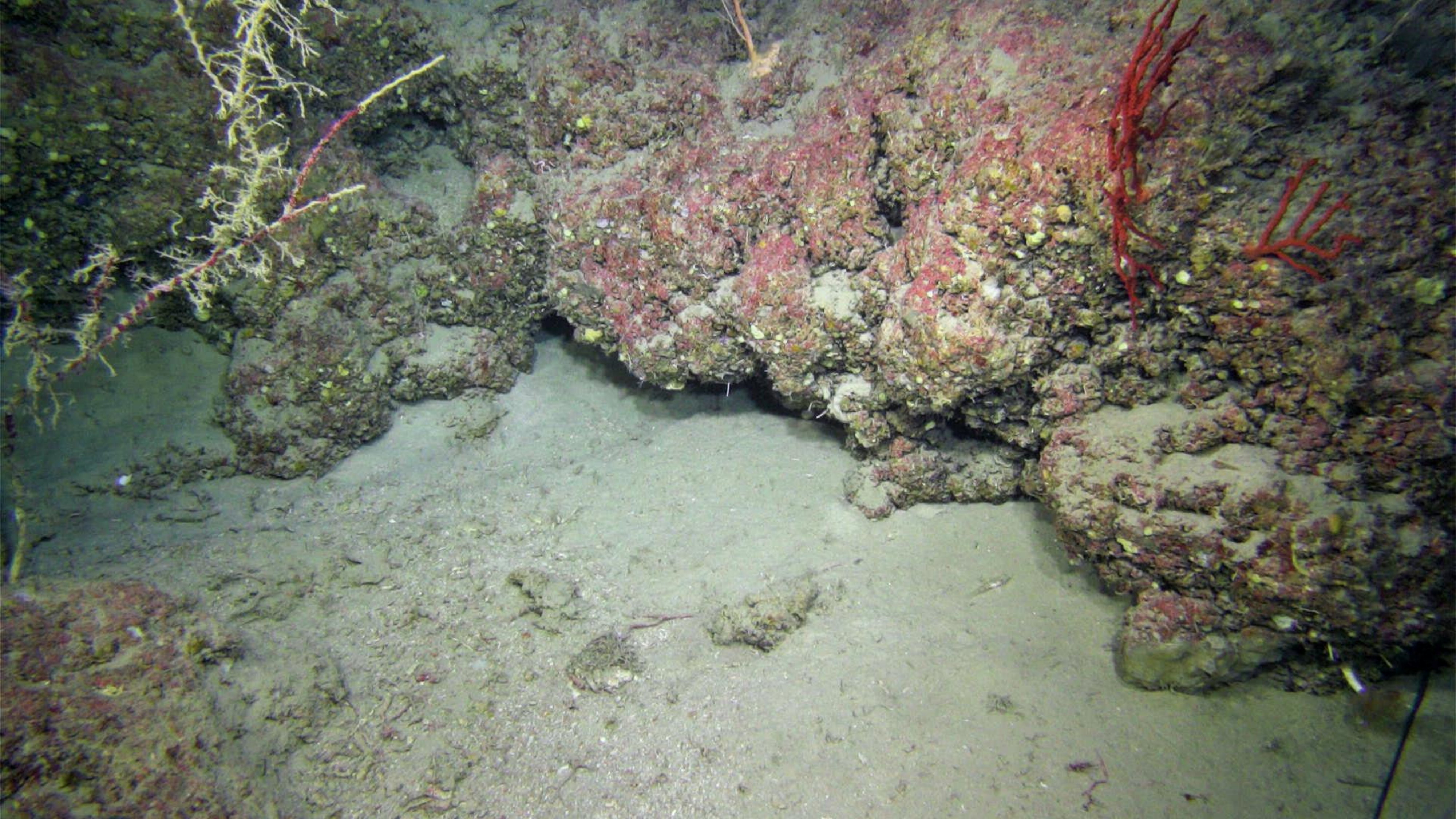} &
\includegraphics[width=\deepimgwidth,height=\deepimgheight]{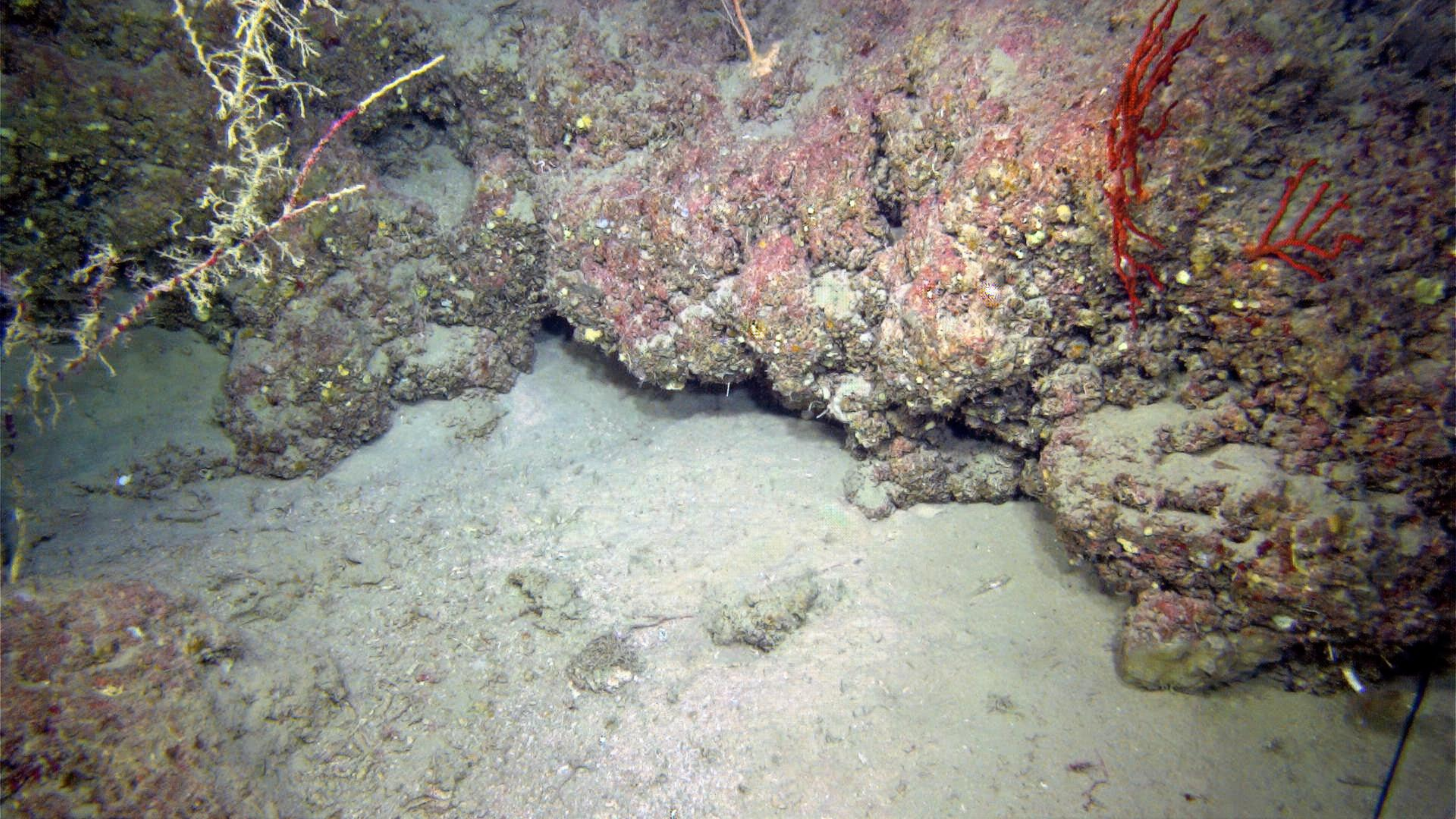} &
\includegraphics[width=\deepimgwidth,height=\deepimgheight]{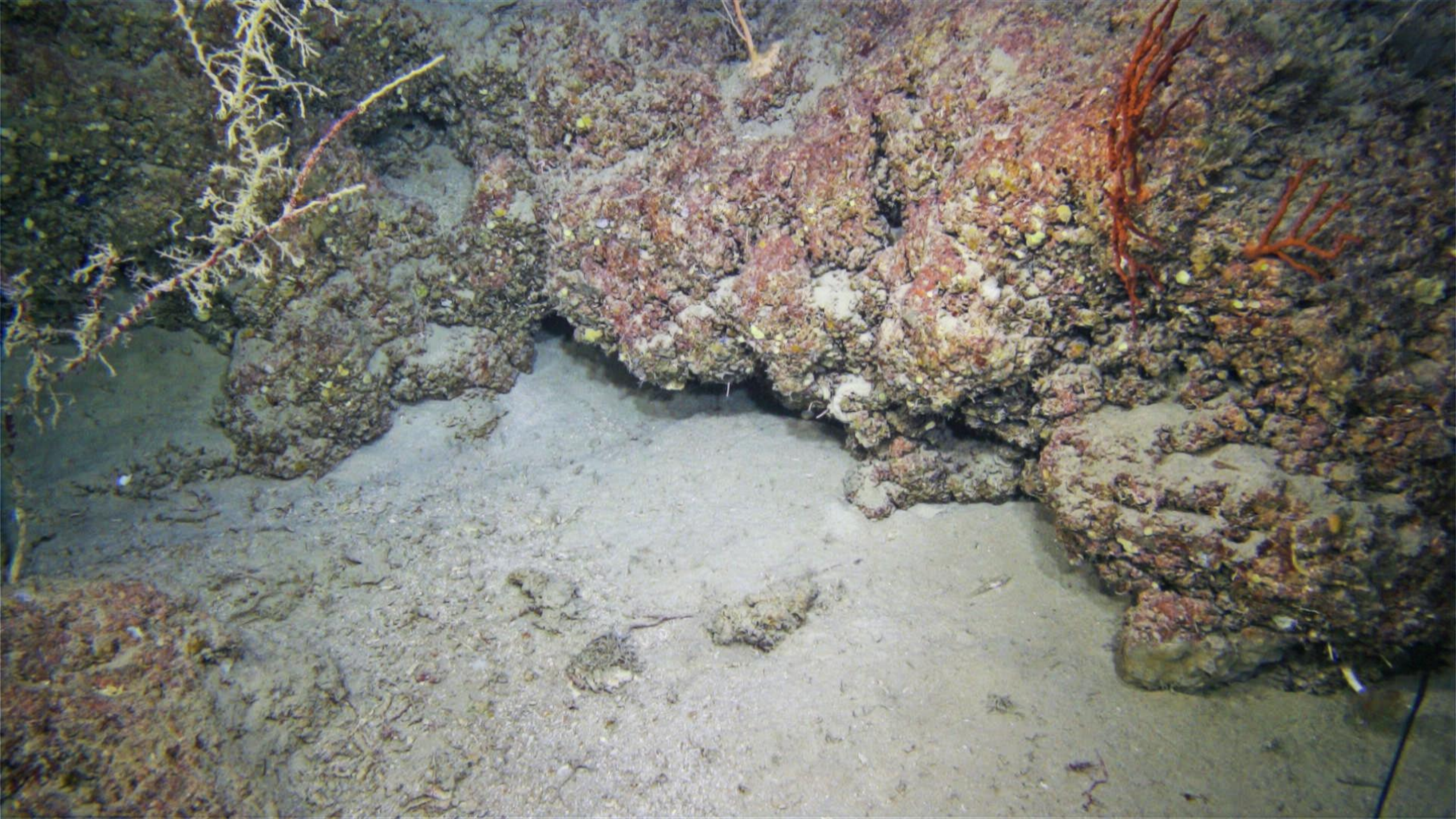} &
\includegraphics[width=\deepimgwidth,height=\deepimgheight]{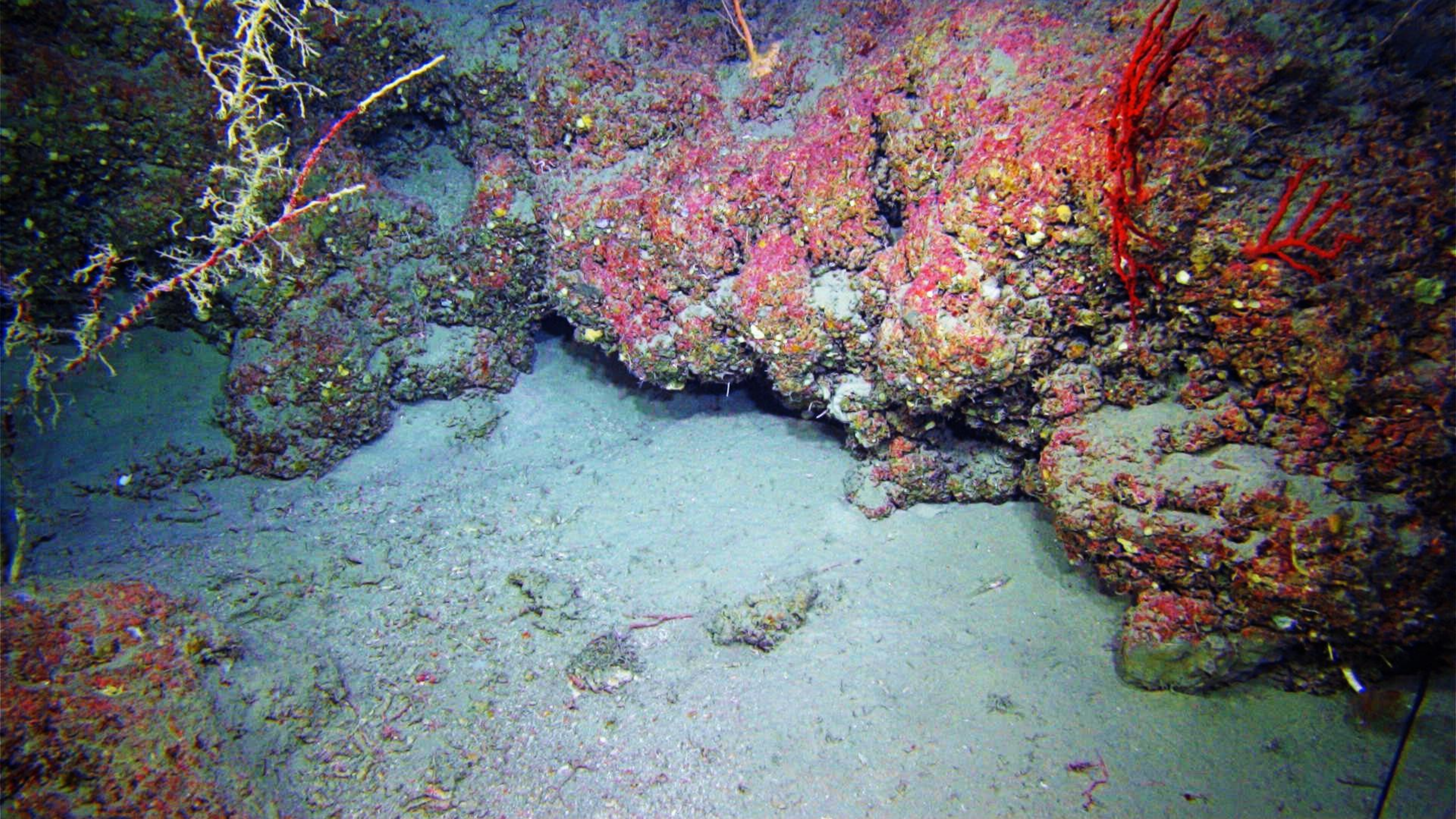} &
\includegraphics[width=\deepimgwidth,height=\deepimgheight]{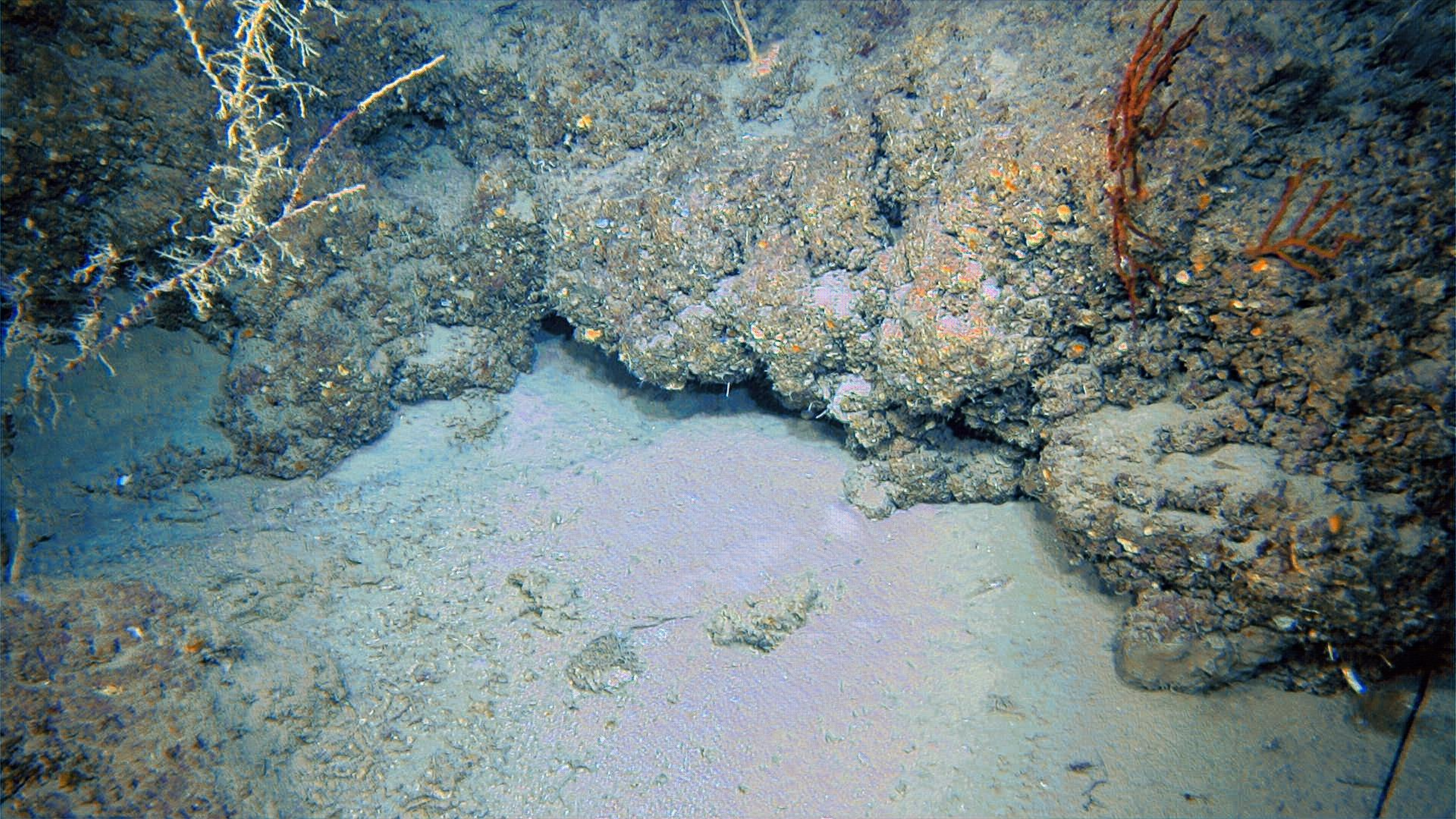} &
\includegraphics[width=\deepimgwidth,height=\deepimgheight]{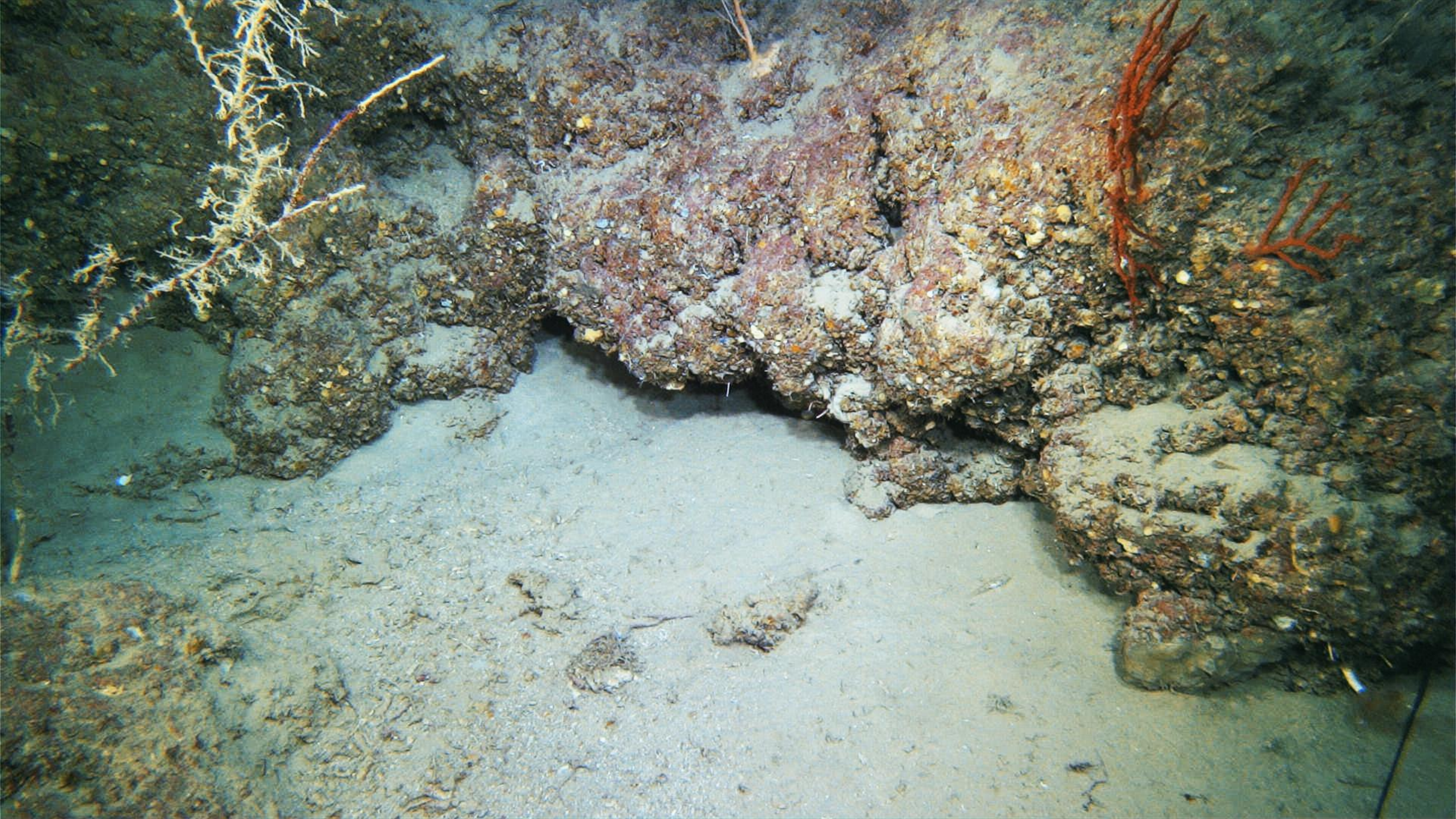} \\

\includegraphics[width=\deepimgwidth,height=\deepimgheight]{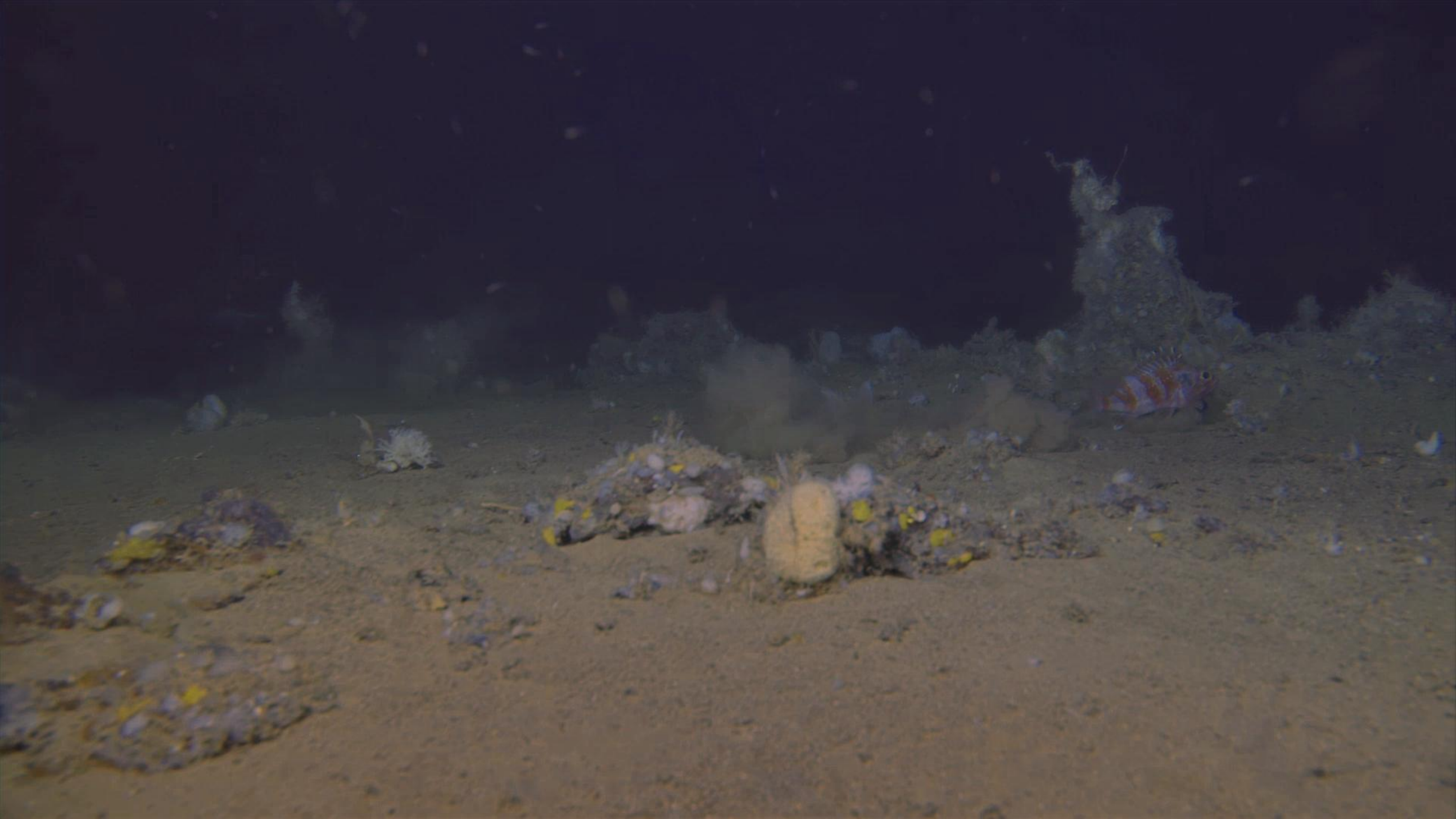} &
\includegraphics[width=\deepimgwidth,height=\deepimgheight]{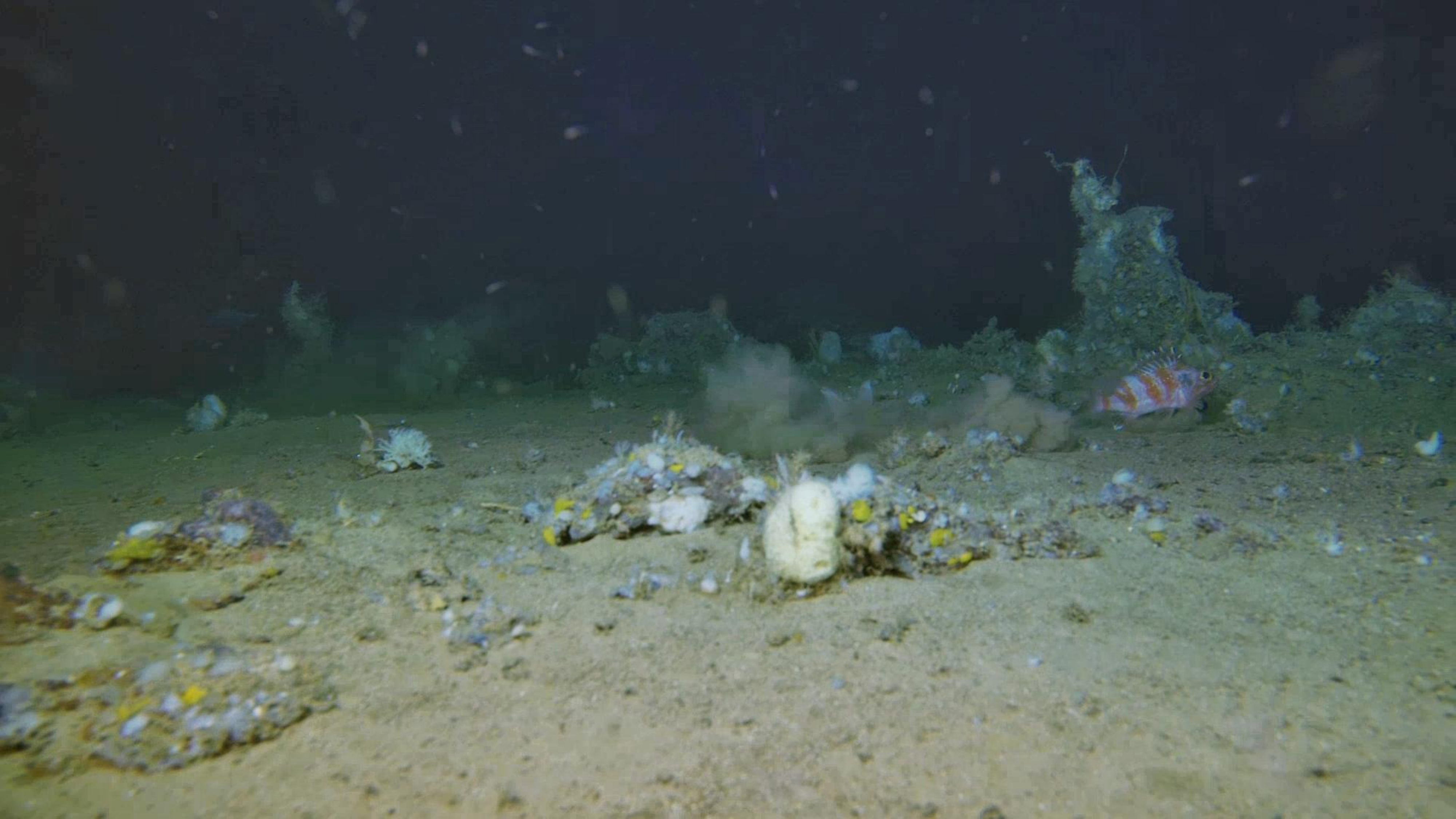} &
\includegraphics[width=\deepimgwidth,height=\deepimgheight]{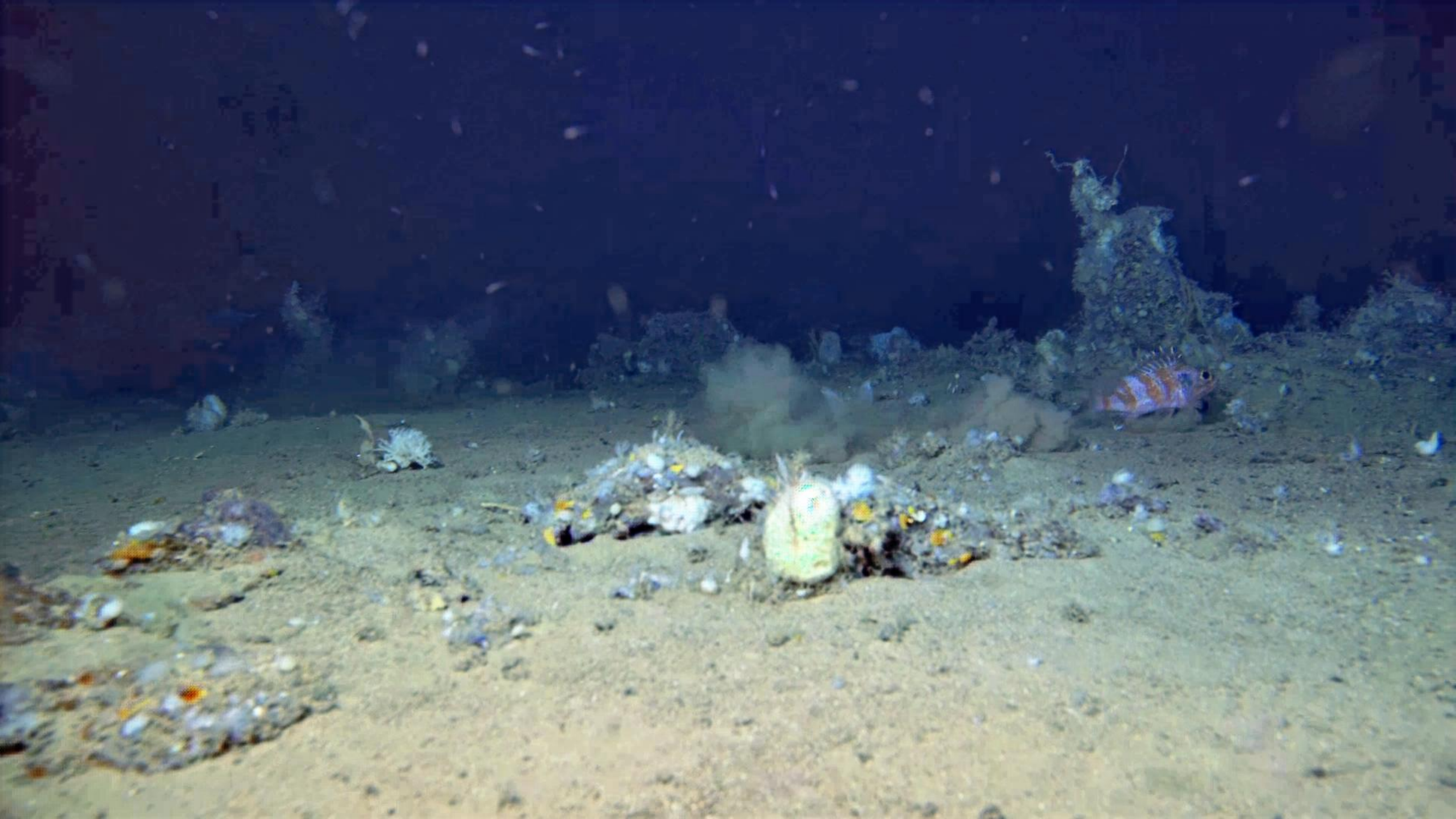} &
\includegraphics[width=\deepimgwidth,height=\deepimgheight]{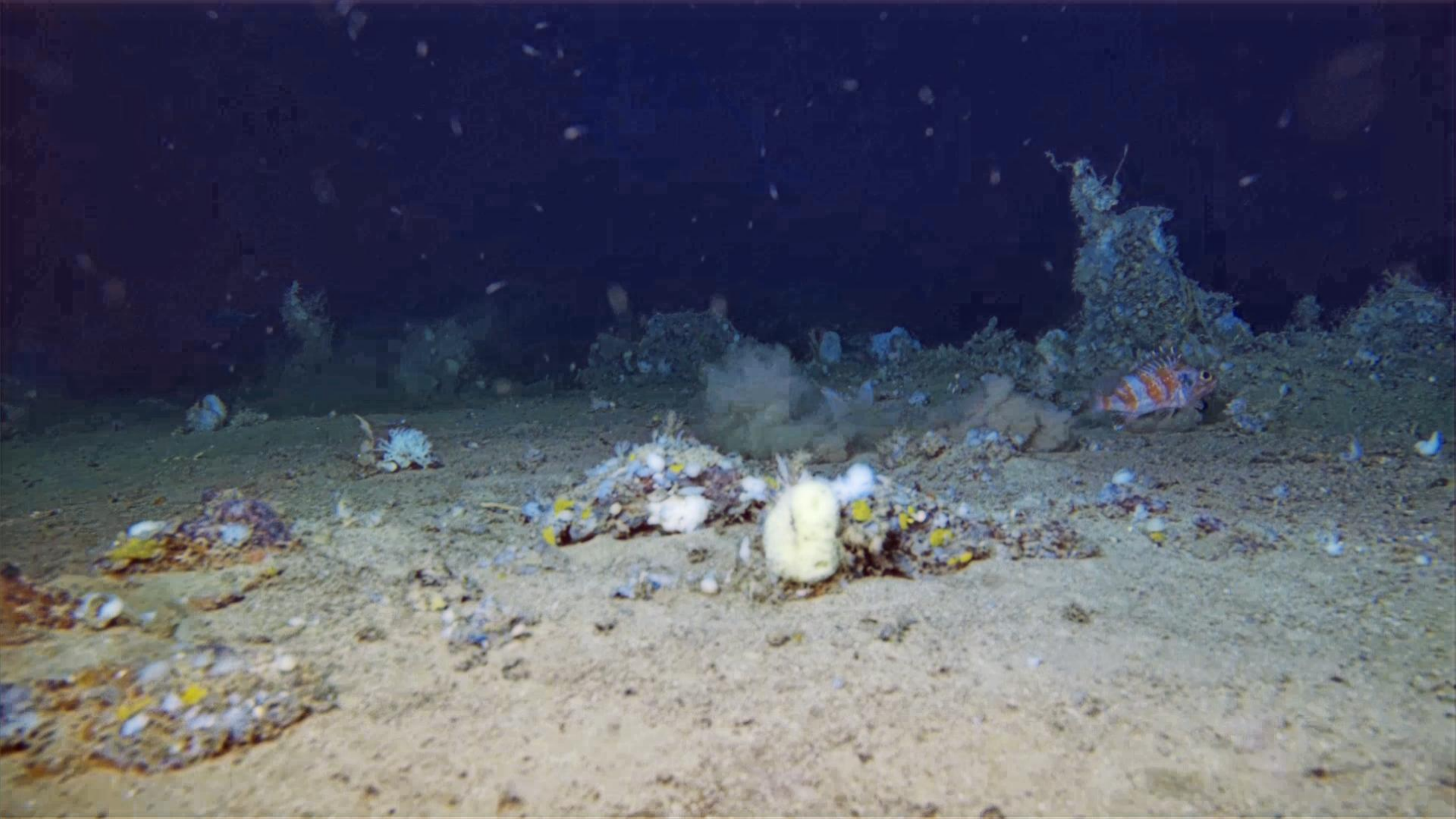} &
\includegraphics[width=\deepimgwidth,height=\deepimgheight]{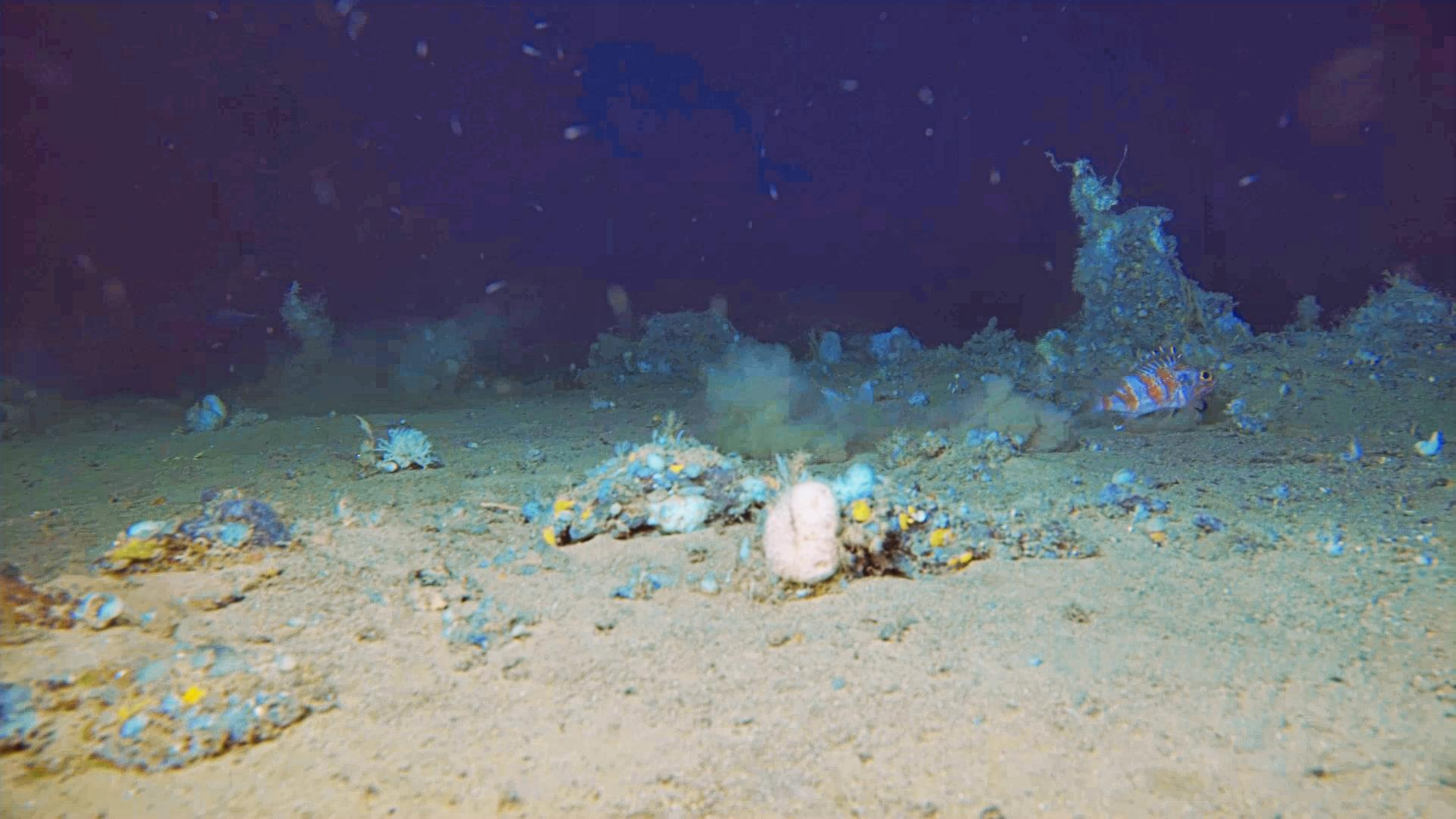} &
\includegraphics[width=\deepimgwidth,height=\deepimgheight]{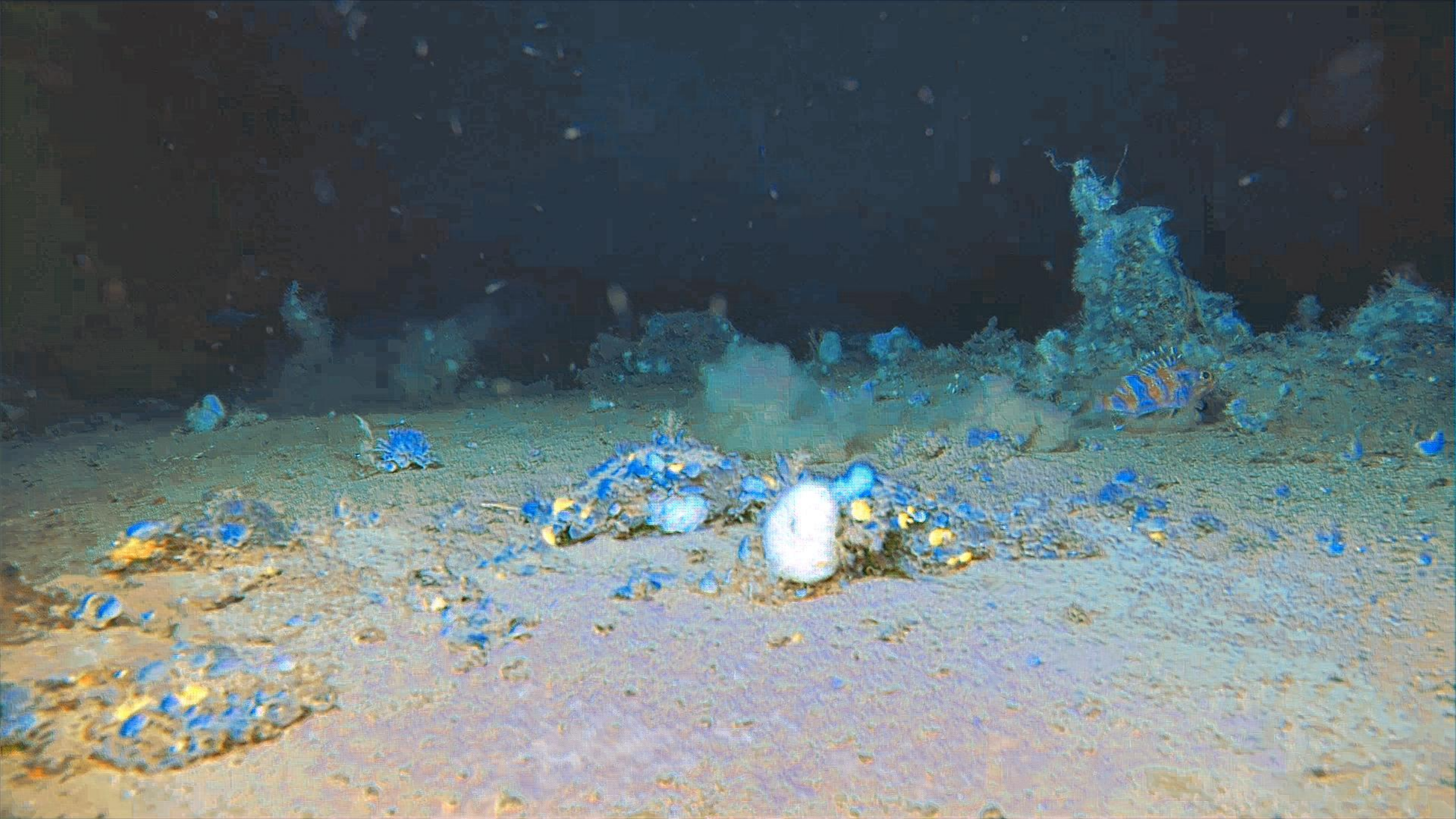} &
\includegraphics[width=\deepimgwidth,height=\deepimgheight]{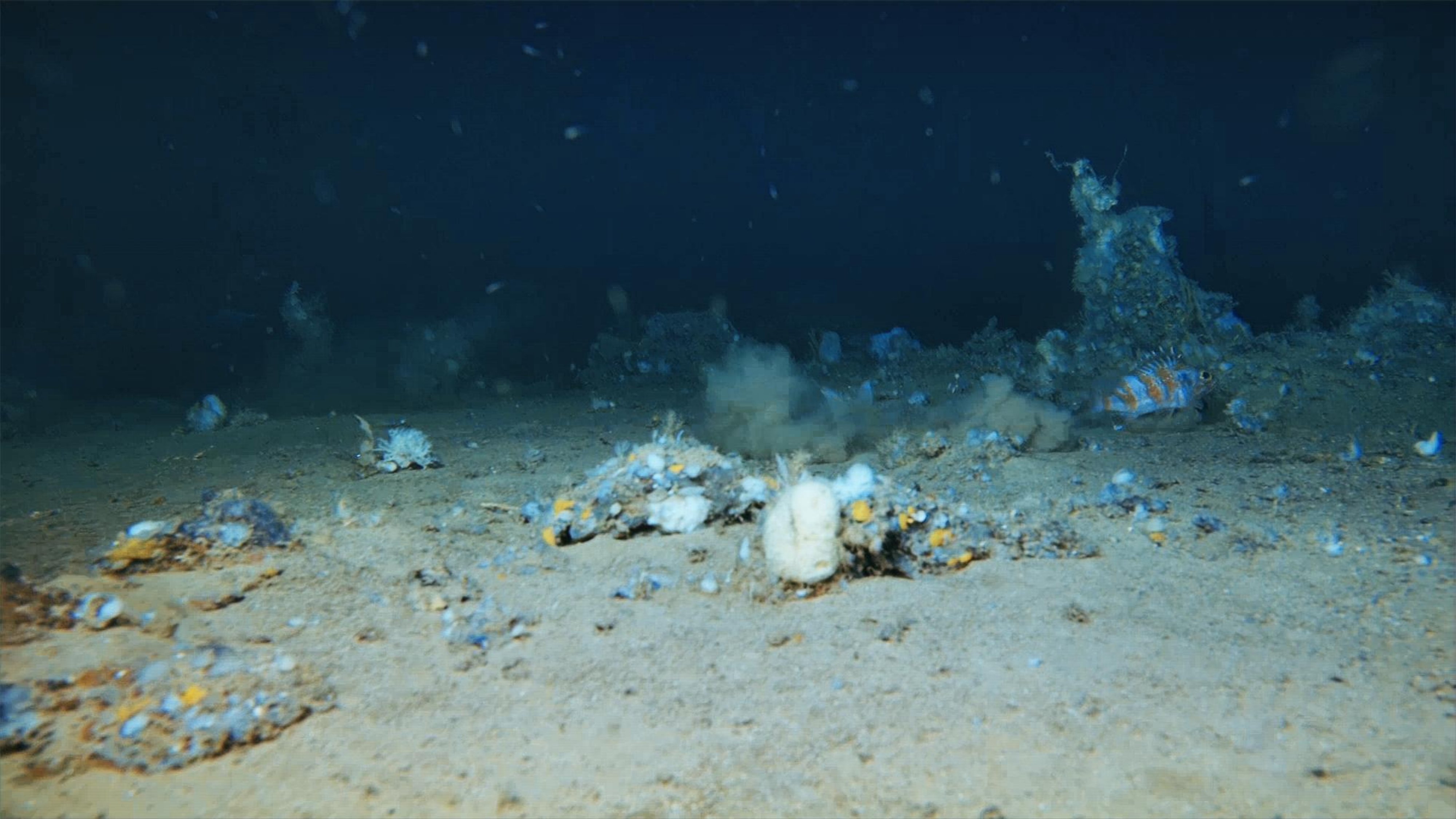} \\

\includegraphics[width=\deepimgwidth,height=\deepimgheight]{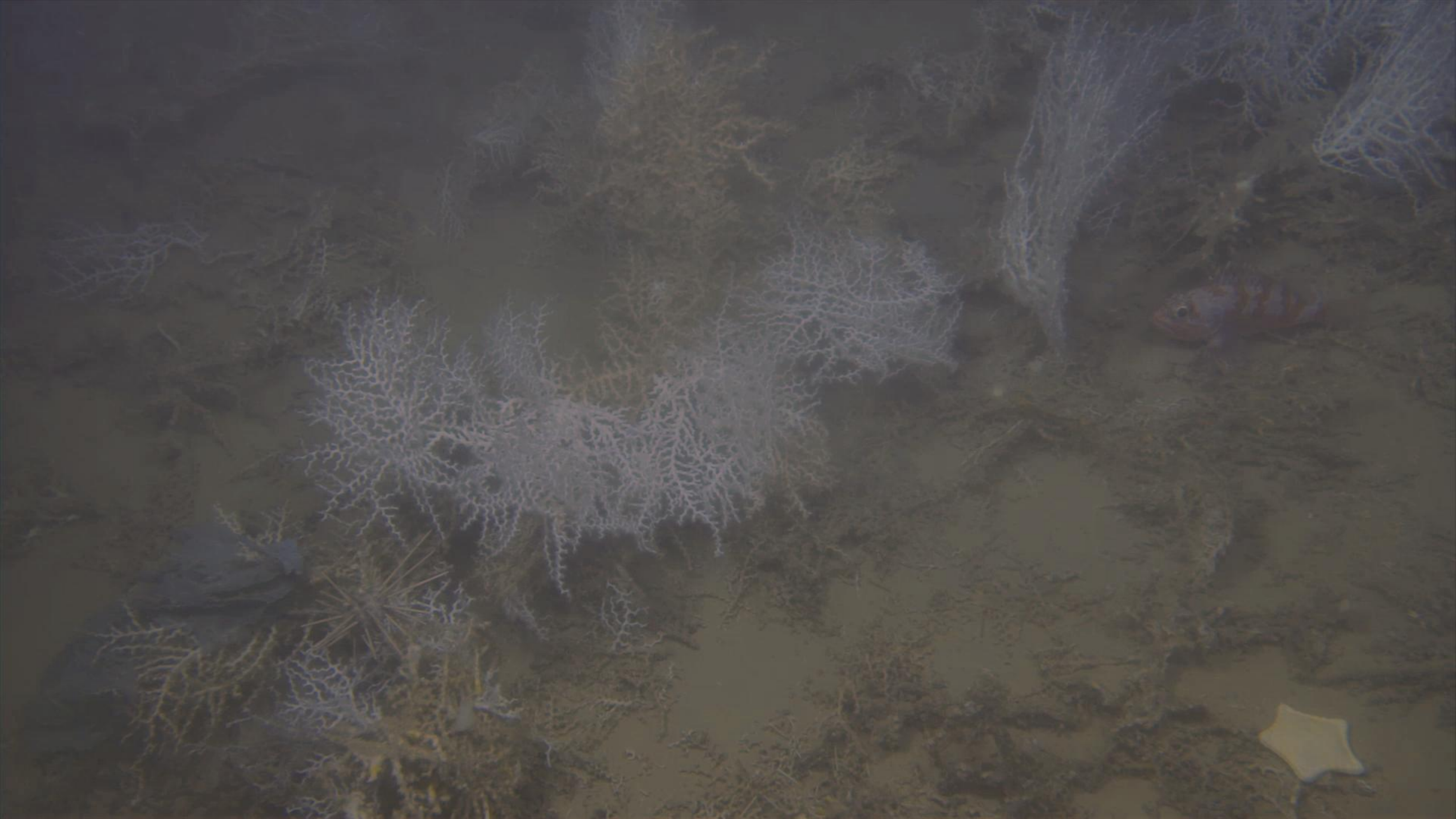} &
\includegraphics[width=\deepimgwidth,height=\deepimgheight]{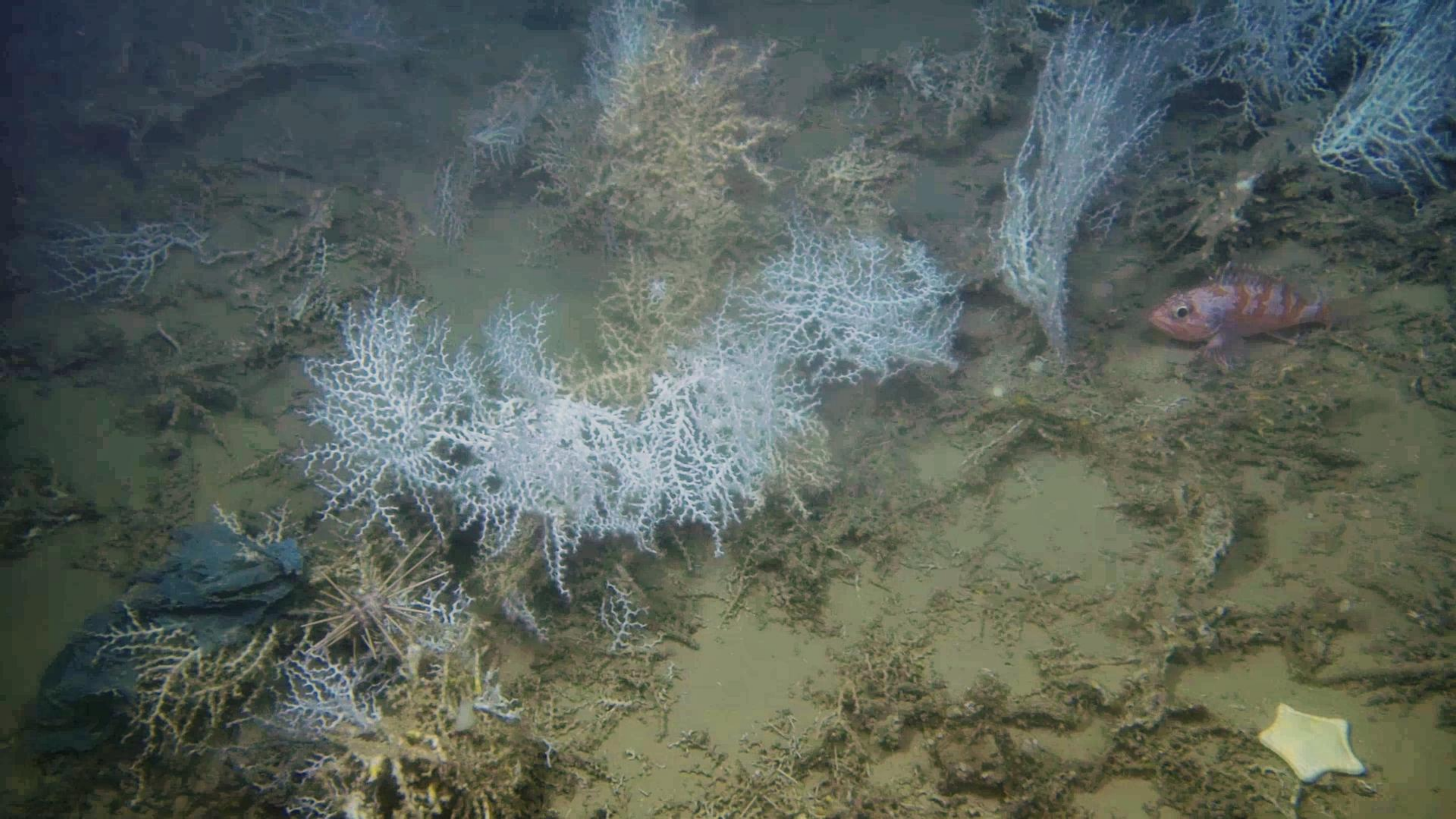} &
\includegraphics[width=\deepimgwidth,height=\deepimgheight]{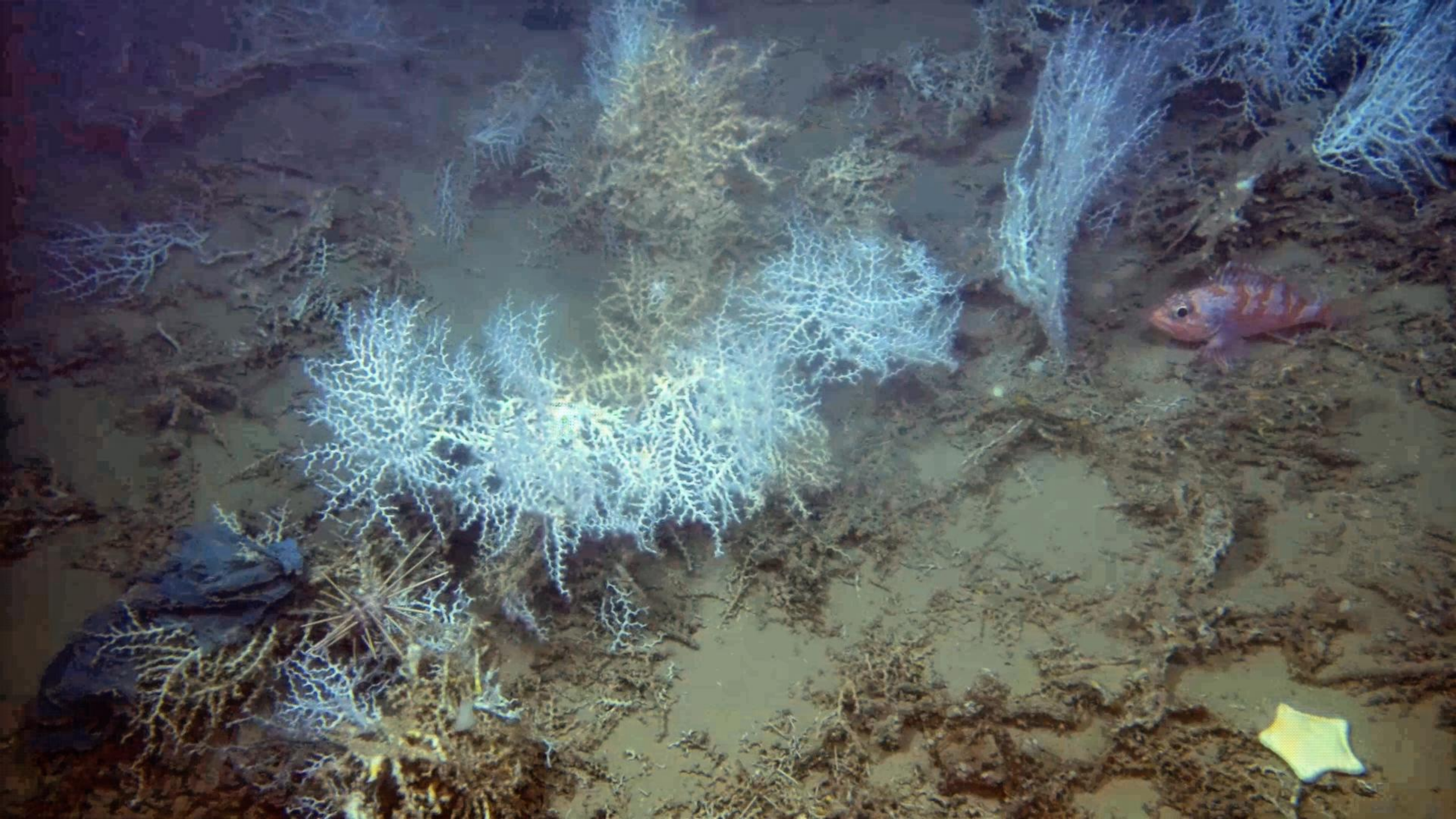} &
\includegraphics[width=\deepimgwidth,height=\deepimgheight]{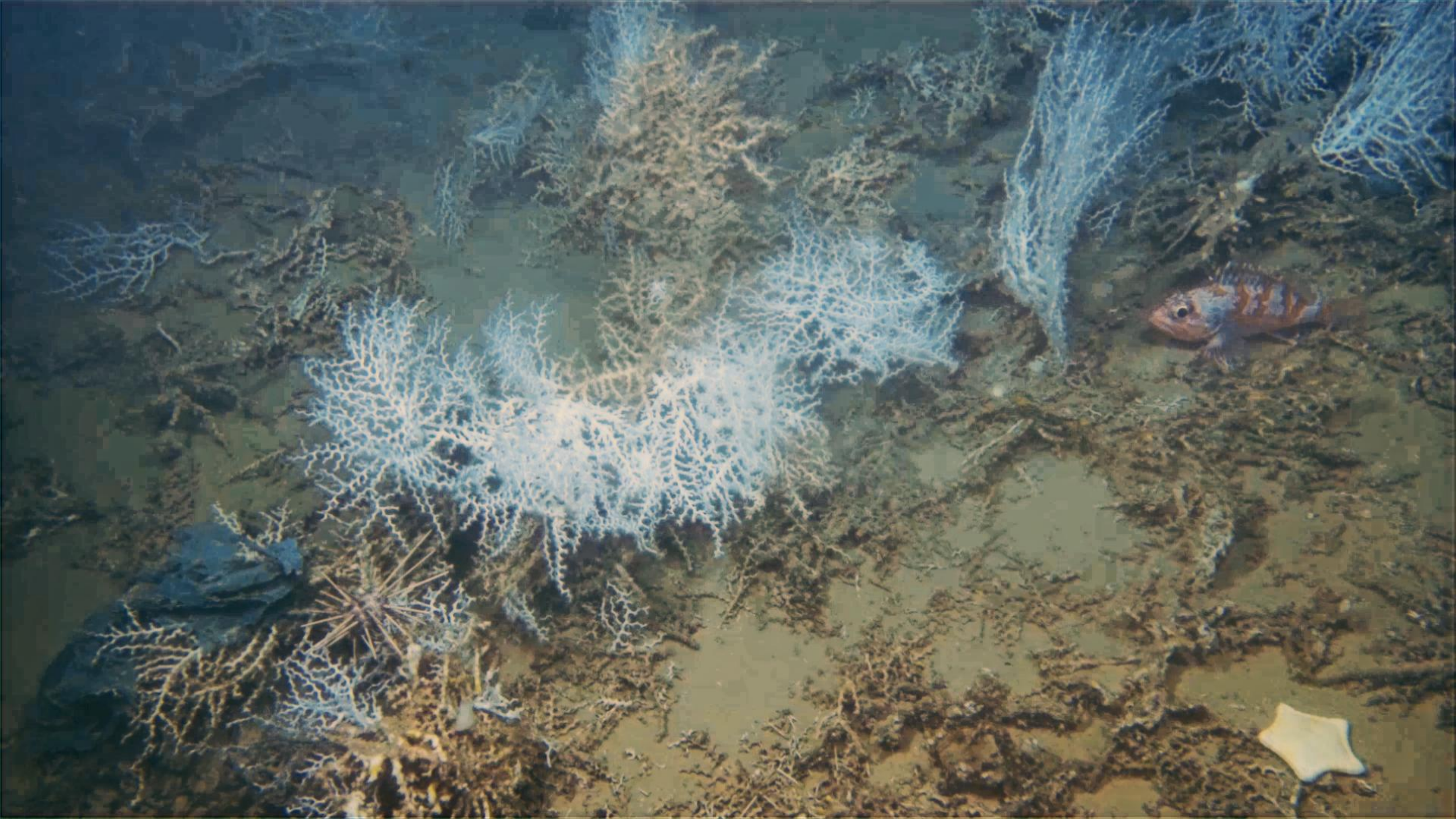} &
\includegraphics[width=\deepimgwidth,height=\deepimgheight]{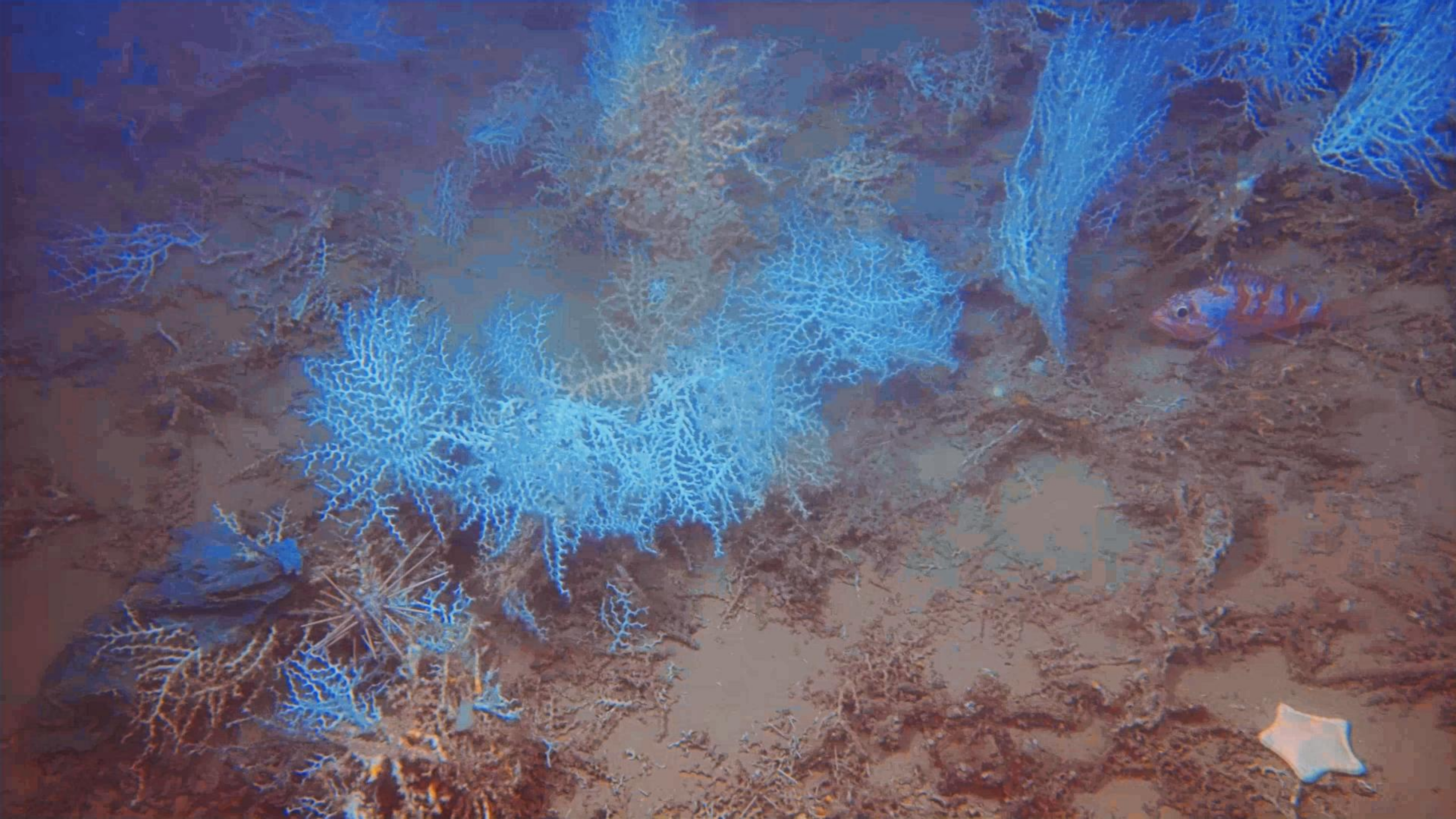} &
\includegraphics[width=\deepimgwidth,height=\deepimgheight]{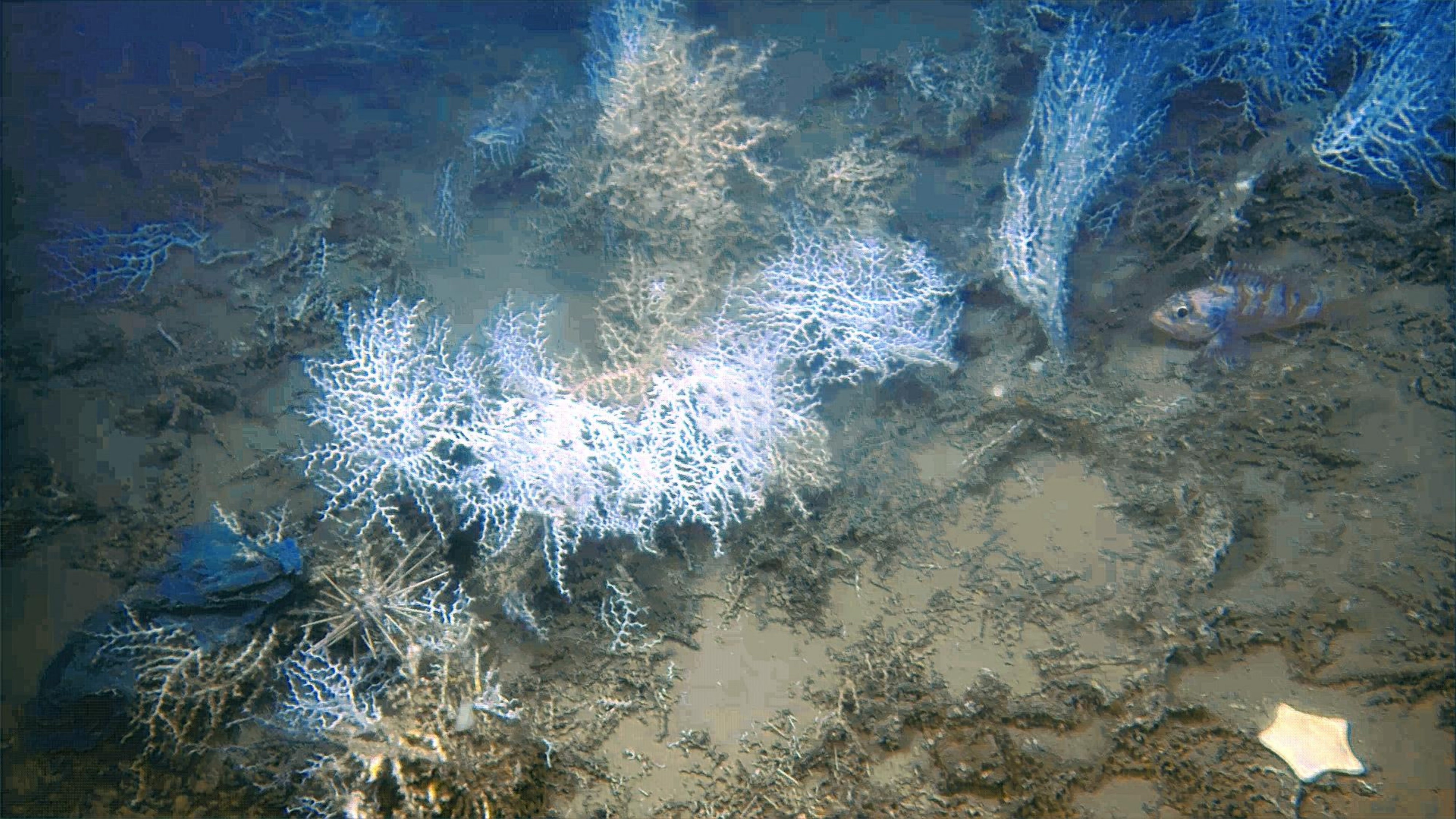} &
\includegraphics[width=\deepimgwidth,height=\deepimgheight]{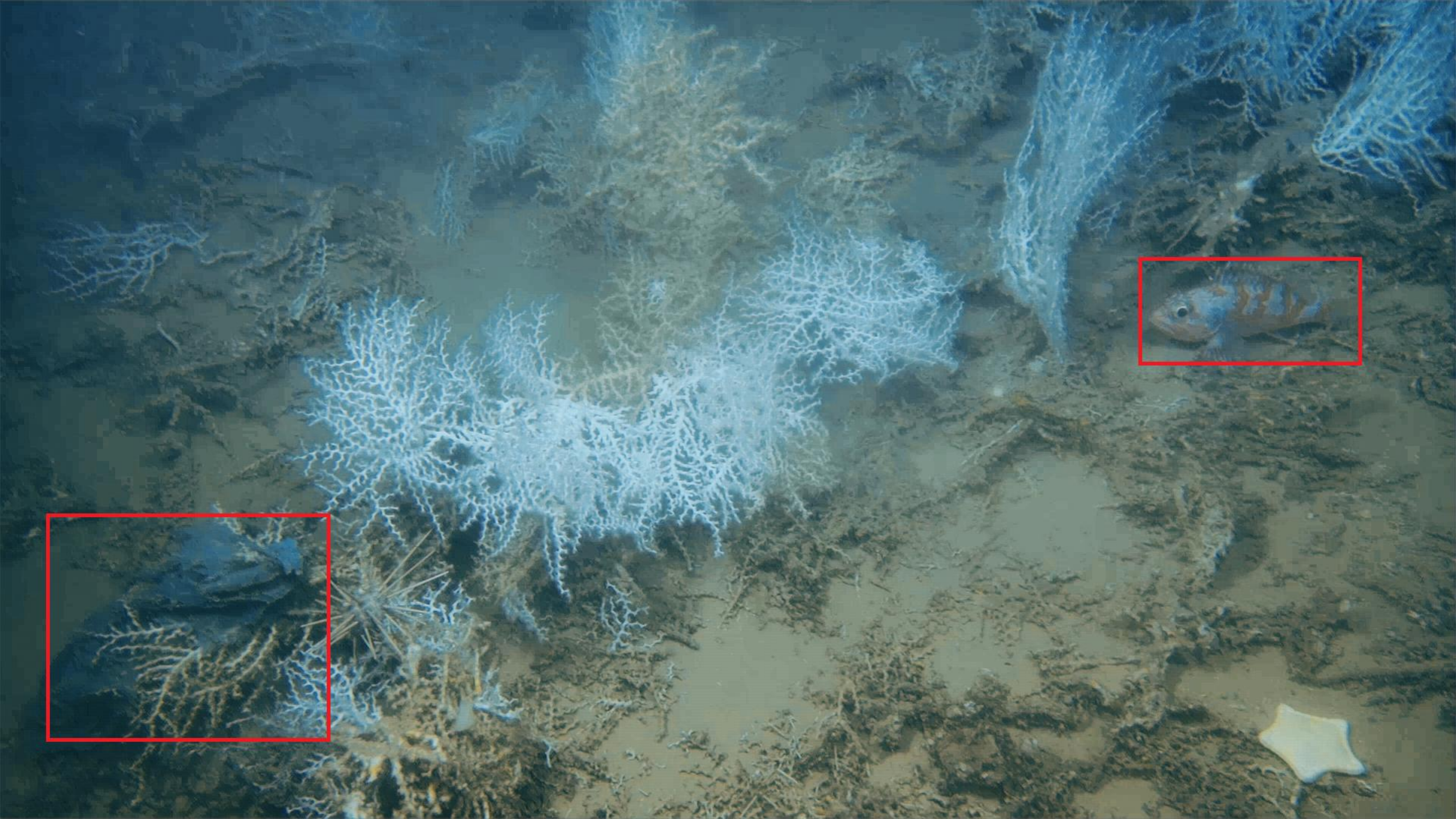} \\

\end{tabular}
\caption{Qualitative comparison on the proposed DeepSea-T80 real deep-sea dataset. Four representative images with different color casts are enhanced by classical and DL-based UIE methods, as well as the proposed AQUA-Net. In these examples, AQUA-Net better preserves color balance and illumination and reduces reddish artifacts, resulting in a clearer separation between foreground structures and the background.}
\label{fig:Deepsea_T80}
\end{figure*}

\textbf{DeepSea-T80:}
The model is also evaluated on real aquatic images, as shown in Figure \ref{fig:Deepsea_T80}. Five different images with varying color tints are presented. Our model demonstrates good visual results when compared to recent models such as CCL-Net and OUNet-JL, which tend to produce reddish images and introduce artificial color artifacts. In contrast, our model effectively preserves color balance and illumination, clearly distinguishing between background and foreground objects without introducing unnatural color shifts.

\subsection{Ablation Study}
\begin{table}[t!]
\centering
\fontsize{12.0pt}{14.0pt}\selectfont
\caption{Quantitative results of the ablation study. The table shows the contribution of each component in the AQUA-Net model. The top scores are marked in \textcolor{red}{\textbf{red}}.}
\renewcommand{\arraystretch}{1.2}
\setlength{\tabcolsep}{4pt}
\begin{tabular}{l|cccc|cc}
\hline
\rowcolor[HTML]{ADD8E6} \textbf{Modules} & \multicolumn{4}{c|}{\textbf{UIEB-T90}} & \multicolumn{2}{c}{\textbf{UIEB-C60}} \\ 
 & PSNR$\uparrow$ & SSIM$\uparrow$ & UIQM$\uparrow$ & UCIQE$\uparrow$ & UIQM$\uparrow$ & UCIQE$\uparrow$ \\ 
\hline
Raw & 16.134 & 0.748 & 2.346 & 0.362 & 1.856 & 0.359 \\
Base & 18.473 & 0.832 & 2.872 & 0.377 & 1.602 & 0.418 \\
Base $+$ Frequency & 20.614 & 0.872 & 3.089 & \textbf{\textcolor{red}{0.415}} & 2.233 & \textbf{\textcolor{red}{0.436}} \\
Base $+$ Illumination & 20.730 & 0.879 & 3.086 & 0.391 & 2.123 & 0.414 \\
Full Model & \textbf{\textcolor{red}{21.482}} & \textbf{\textcolor{red}{0.890}} & \textbf{\textcolor{red}{3.250}} & 0.397 & \textbf{\textcolor{red}{2.313}} & 0.427 \\
\hline
\end{tabular}\label{tab: Ablation}
\end{table}

\begin{figure}[htbp!]
\centering
\setlength{\tabcolsep}{1pt}       
\renewcommand{\arraystretch}{0.4} 

\newcommand{\imgwidth}{0.15\textwidth}
\newcommand{\imgheight}{0.12\textwidth}

\begin{tabular}{ccc}
\textbf{\tiny Raw} & \textbf{\tiny Reference} & \textbf{\tiny Full} \\[3pt]
\includegraphics[width=\imgwidth,height=\imgheight]{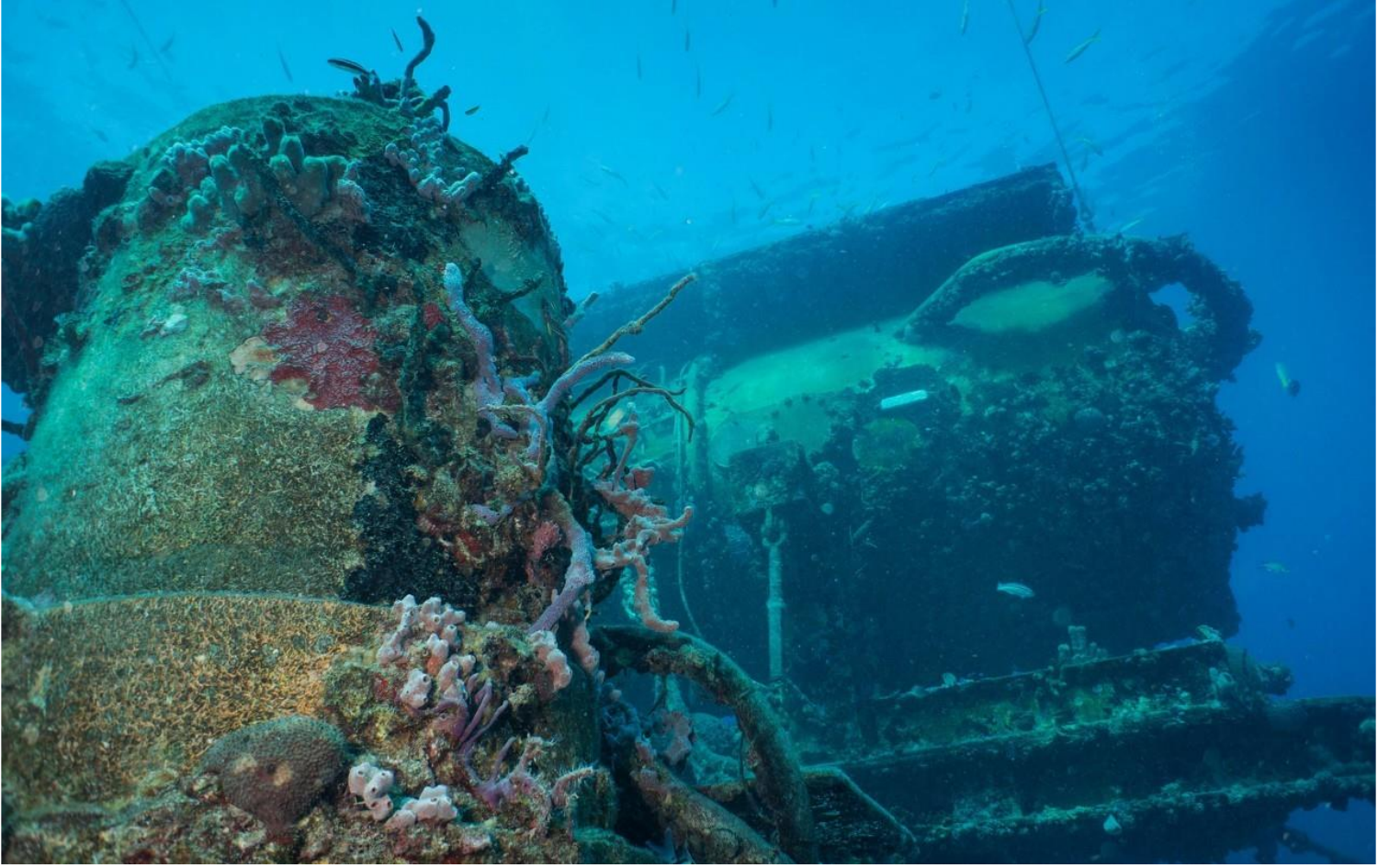} &
\includegraphics[width=\imgwidth,height=\imgheight]{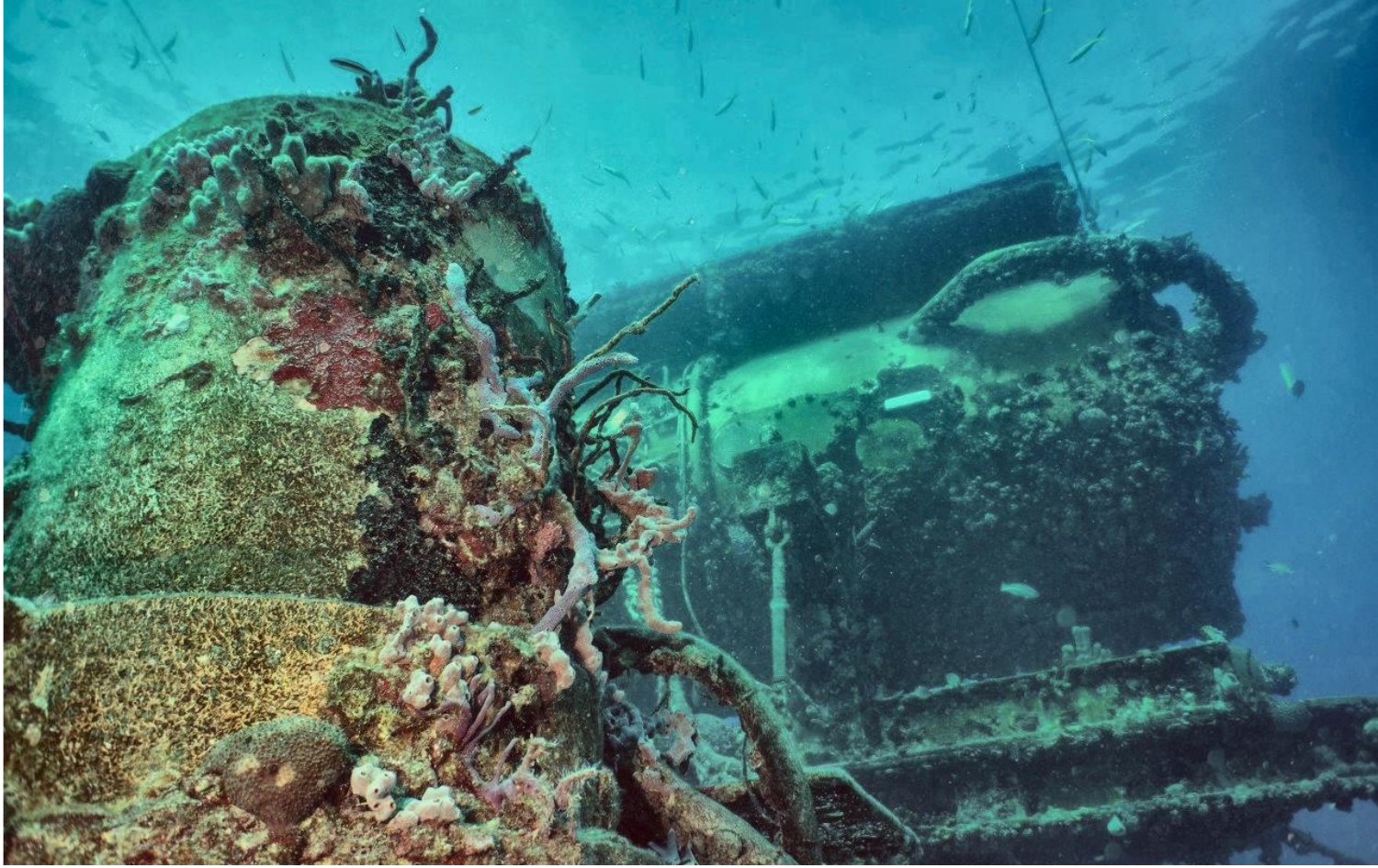} &
\includegraphics[width=\imgwidth,height=\imgheight]{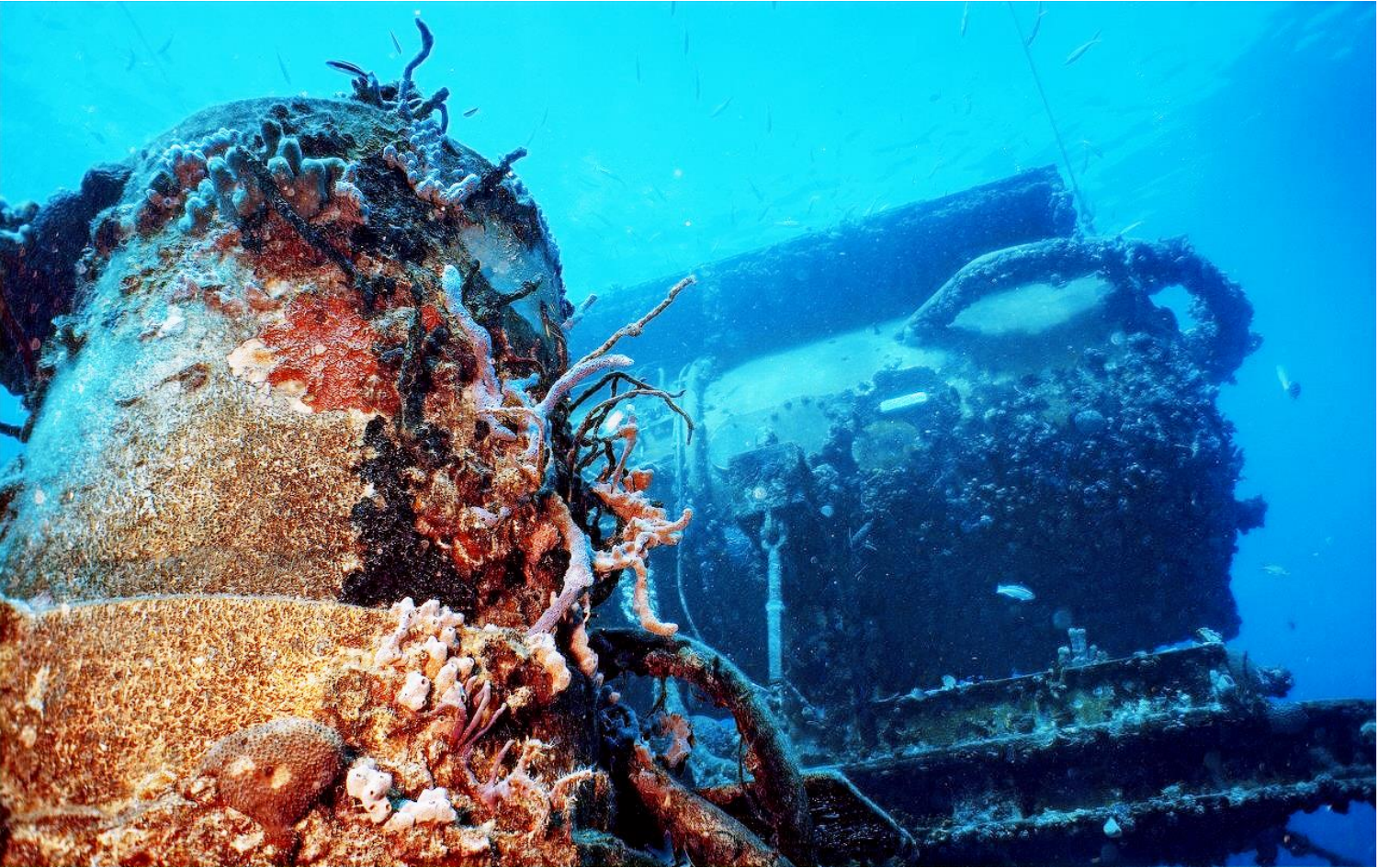} \\[5pt]

\textbf{\tiny Base (B)} & \textbf{\tiny B \texttt{+} Frequency} & \textbf{\tiny B \texttt{+} illumination} \\[3pt]
\includegraphics[width=\imgwidth,height=\imgheight]{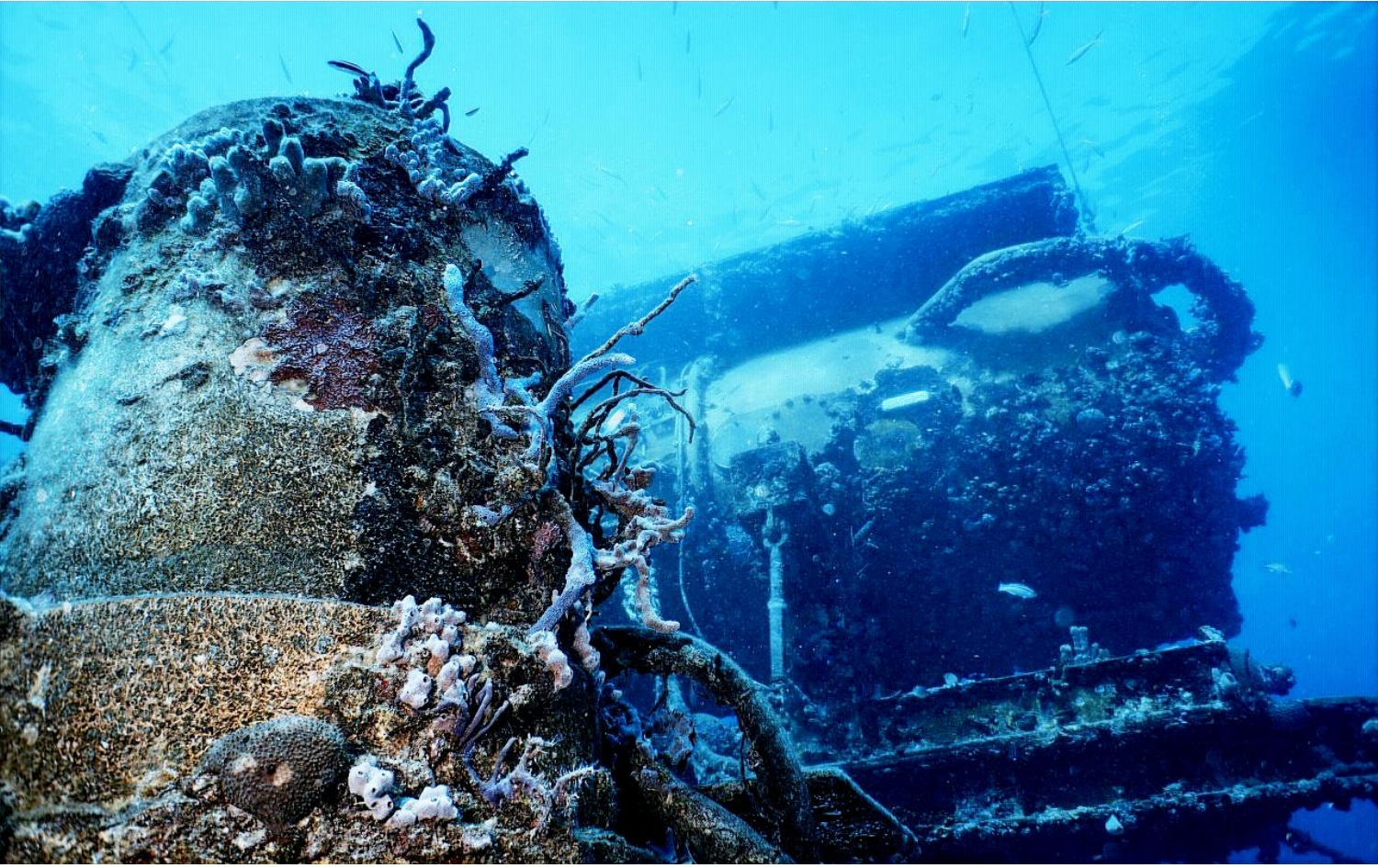} &
\includegraphics[width=\imgwidth,height=\imgheight]{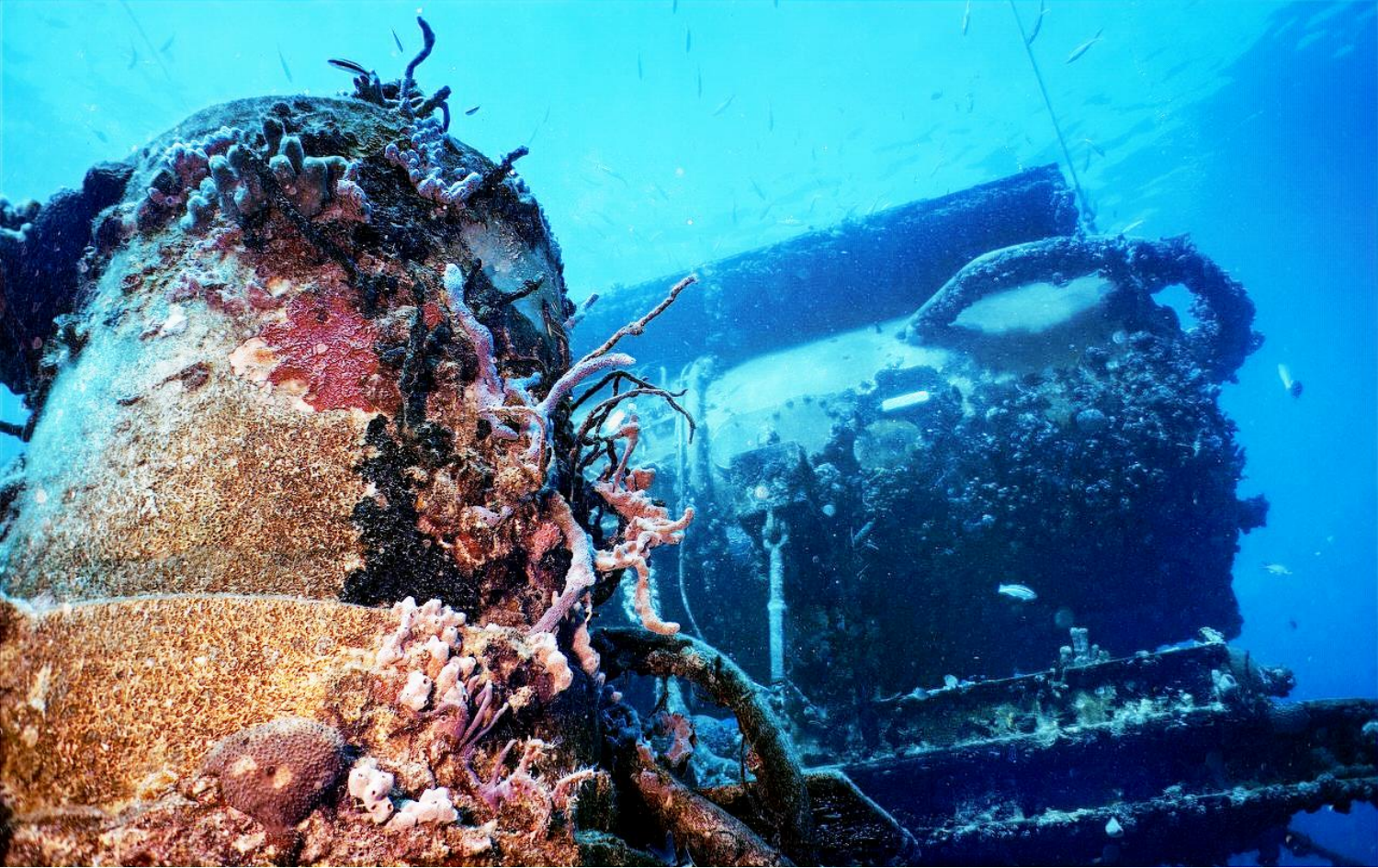} &
\includegraphics[width=\imgwidth,height=\imgheight]{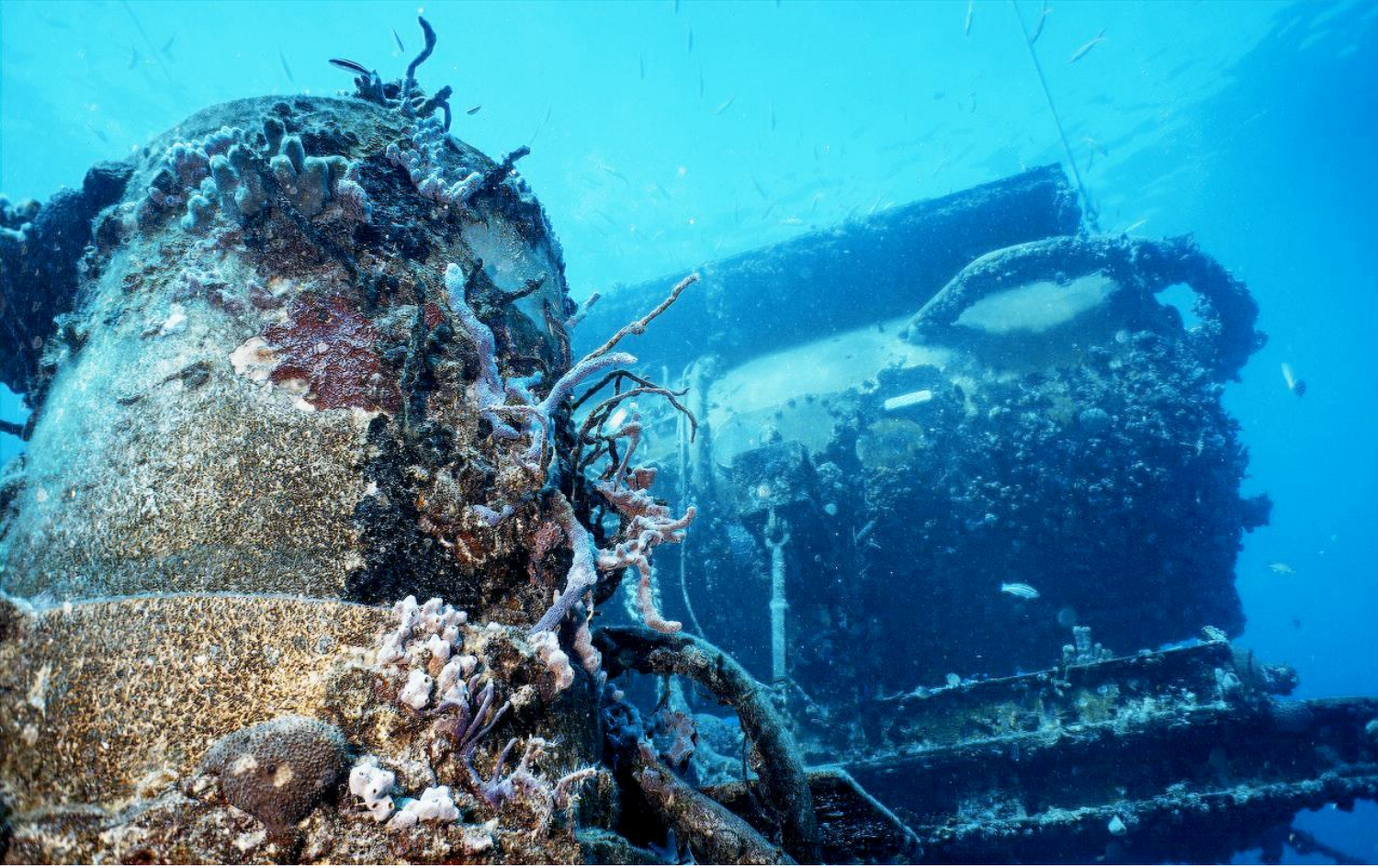} \\
\end{tabular}

\caption{Visual results of the ablation study. The frequency block improves object visibility, and the illumination block enhances overall scene clarity in the AQUA-Net model.}
\label{fig: Albation}
\end{figure}

We conduct an ablation study to evaluate the contribution of each component in the AQUA-Net model. The encoder-decoder network, referred to as the base model, serves as the starting point. We then incrementally add the frequency block and the illumination block. As shown in Table \ref{tab: Ablation}, each block improves the quantitative metrics, with the frequency block significantly enhancing UCIQE on UIEB-T90 and UIQM on UIEB-C60, and the illumination block further boosting PSNR and SSIM. The full model, which combines both blocks, achieves the best overall performance across all metrics. Visual comparisons in Figure \ref{fig: Albation} illustrate the effect of each component: the frequency block improves object visibility, and the illumination block enhances overall scene clarity, confirming the complementary benefits of the two modules.

\begin{figure*}[t!]
\centering

\setlength{\tabcolsep}{1.2pt}
\renewcommand{\arraystretch}{0.35}

\newcommand{\lusiimgwidth}{0.185\textwidth}
\newcommand{\lusiimgheight}{0.145\textwidth}

\begin{tabular}{ccccc}

\textbf{\tiny Raw} &
\textbf{\tiny Reference} &
\textbf{\tiny Fusion \cite{ancuti2017color}} &
\textbf{\tiny WWPF \cite{zhang2023underwater}} &
\textbf{\tiny MLLE \cite{zhang2022underwater}} \\[3pt]

\includegraphics[width=\lusiimgwidth,height=\lusiimgheight]{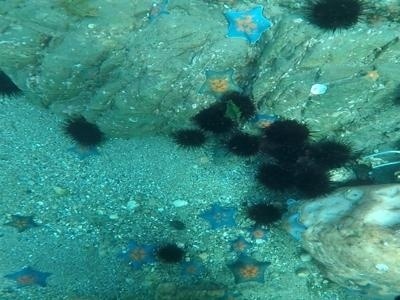} &
\includegraphics[width=\lusiimgwidth,height=\lusiimgheight]{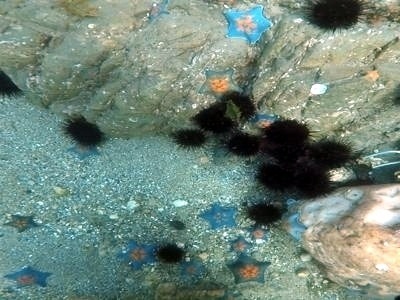} &
\includegraphics[width=\lusiimgwidth,height=\lusiimgheight]{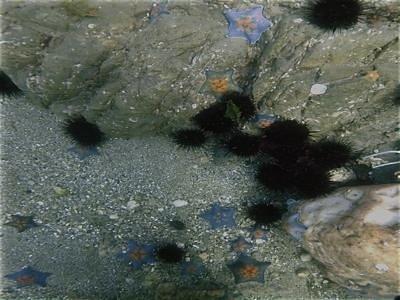} &
\includegraphics[width=\lusiimgwidth,height=\lusiimgheight]{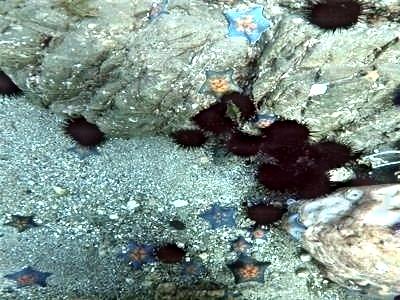} &
\includegraphics[width=\lusiimgwidth,height=\lusiimgheight]{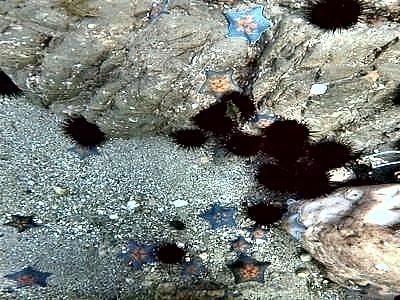} \\[5pt]

\textbf{\tiny WFAC \cite{zhang2025underwater1}} &
\textbf{\tiny WaterNet \cite{li2019underwater}} &
\textbf{\tiny LMF \cite{zhou2025lightweight}} &
\textbf{\tiny OUNet-JL \cite{wang2025optimized}} &
\textbf{\tiny AQUA-Net} \\[3pt]

\includegraphics[width=\lusiimgwidth,height=\lusiimgheight]{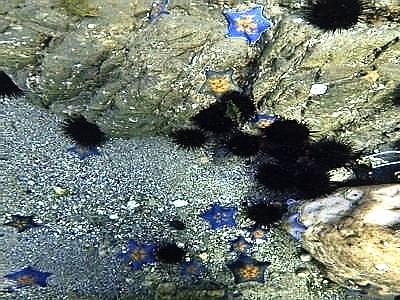} &
\includegraphics[width=\lusiimgwidth,height=\lusiimgheight]{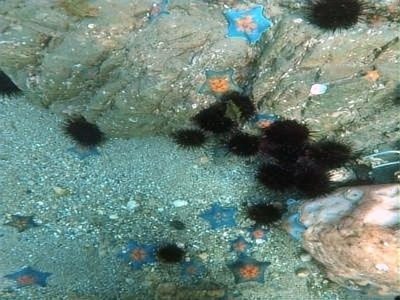} &
\includegraphics[width=\lusiimgwidth,height=\lusiimgheight]{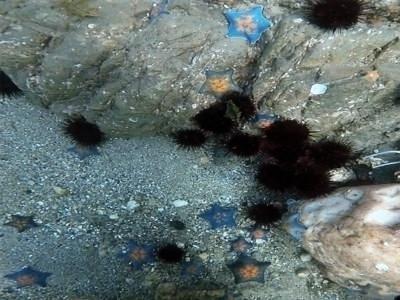} &
\includegraphics[width=\lusiimgwidth,height=\lusiimgheight]{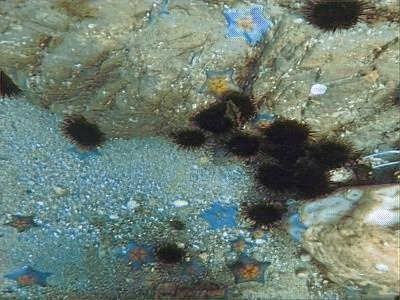} &
\includegraphics[width=\lusiimgwidth,height=\lusiimgheight]{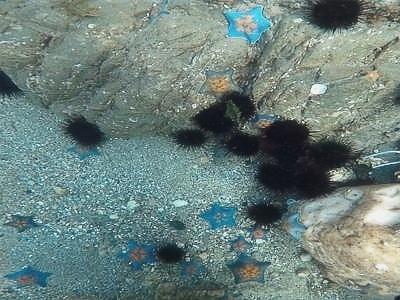} \\
\end{tabular}
\caption{Qualitative comparison on the LSUI-400 test set using different underwater image enhancement methods. The comparison includes the raw input image, reference image, and enhanced results obtained using Fusion, WWPF, MLLE, WFAC, WaterNet, LMF, OUNet-JL, and AQUA-Net.}
\label{fig:lsui_visual}
\end{figure*}

\subsection{Comparison on LSUI-400}
\label{sec:lsui_comparison}

We further evaluated AQUA-Net on the LSUI-400 test dataset using both visual and quantitative comparisons. The compared methods include Fusion~\cite{ancuti2017color}, WWPF~\cite{zhang2023underwater}, MLLE~\cite{zhang2022underwater}, WFAC~\cite{zhang2025underwater1}, WaterNet~\cite{li2019underwater}, LMF~\cite{zhou2025lightweight}, and OUNet-JL~\cite{wang2025optimized}.

As shown in Fig.~\ref{fig:lsui_visual}, the raw image contains color distortion, low contrast, and reduced visibility. Traditional methods improve contrast to some extent, but some results show over-enhancement or unnatural colors. Deep learning-based methods improve the visual quality, but slight haze, color imbalance, or weak details may still remain. In comparison, AQUA-Net produces a more balanced result with clearer details and better visibility
The quantitative results in Table~\ref{tab:lsui400_comparison} also show that AQUA-Net achieves the best performance among the compared methods, with a PSNR of 23.0322 dB and an SSIM of 0.8511. These results demonstrate that AQUA-Net provides effective underwater image enhancement on the LSUI-400 test dataset in both visual quality and numerical accuracy.

\begin{table}[t!]
\centering
\fontsize{12.0pt}{14.0pt}\selectfont
\caption{Comparison with SOTA methods on the LSUI-400 test dataset. The top three results are shown in \textbf{\textcolor{red}{red}}, \textbf{\textcolor{green}{green}}, and \textbf{\textcolor{blue}{blue}}, respectively.}
\label{tab:lsui400_comparison}
\begin{tabular}{l|cc}
\hline
\rowcolor[HTML]{ADD8E6}
\textbf{Methods} & \textbf{PSNR$\uparrow$} & \textbf{SSIM$\uparrow$} \\
\hline
Fusion~\cite{ancuti2017color}       & 20.13  & 0.8294 \\
MLLE~\cite{zhang2022underwater}     & 16.63  & 0.6491 \\
WFAC~\cite{zhang2025underwater1}    & 15.85  & 0.6665 \\
WWPF~\cite{zhang2023underwater}     & 16.97  & 0.6622 \\
WaterNet~\cite{li2019underwater}    & \textbf{\textcolor{green}{22.75}} & \textbf{\textcolor{red}{0.8583}} \\
LMF~\cite{zhou2025lightweight}      & \textbf{\textcolor{blue}{21.44}}  & \textbf{\textcolor{green}{0.8547}} \\
OUNet-JL~\cite{wang2025optimized}   & 19.56  & 0.7809 \\
\hline
\textbf{AQUA-Net}                   & \textbf{\textcolor{red}{23.03}} & \textbf{\textcolor{blue}{0.8511}} \\
\hline
\end{tabular}
\end{table}
\subsection{Application}
\label{sec:downstream_detection}

To further examine the practical usefulness of underwater image enhancement for high-level vision tasks, we evaluated object detection performance on the DUO dataset~\cite{liu2021dataset}. The DUO dataset contains four underwater object categories: holothurian, echinus, scallop, and starfish. YOLO12s~\cite{tian2025yolo12} was adopted as the detection model. In this experiment, the enhanced images generated by CCL-Net~\cite{liu2024underwater}, OUNet-JL~\cite{wang2025optimized}, and AQUA-Net were used as inputs to the detector. Since enhancement methods modify image appearance but do not change object positions, the original DUO bounding-box annotations were kept unchanged for all enhanced datasets. To ensure a fair comparison, all experiments used the same train/test split, annotation files, detector architecture, input resolution, batch size, and 100-epoch training setting.

\begin{table}[t!]
\centering
\caption{Downstream object detection comparison of different underwater image enhancement methods on the DUO dataset using YOLO12s.}
\label{tab:downstream_detection_sota}
\begin{tabular}{l|cccc}
\hline
\rowcolor[HTML]{ADD8E6} 
\textbf{Enhancement Method} & \textbf{Precision} & \textbf{Recall} & \textbf{mAP@0.5} & \textbf{mAP@0.5:0.95} \\ 
\hline
CCL-Net~\cite{liu2024underwater} 
& 0.856 
& 0.768 
& \textbf{\textcolor{red}{0.848}} 
& \textbf{\textcolor{red}{0.658}} \\

OUNet-JL~\cite{wang2025optimized} 
& \textbf{\textcolor{red}{0.857}} 
& 0.730 
& 0.823 
& 0.618 \\

AQUA-Net 
& 0.842 
& \textbf{\textcolor{red}{0.774}} 
& 0.828 
& 0.634 \\ 
\hline
\end{tabular}
\end{table}

\begin{figure*}[t!]
\centering

\setlength{\tabcolsep}{2.0pt}
\renewcommand{\arraystretch}{0.35}

\newcommand{\detimgwidth}{0.23\textwidth}
\newcommand{\detimgheight}{0.18\textwidth}

\begin{tabular}{cccc}

\textbf{\tiny Ground Truth} &
\textbf{\tiny CCL-Net \cite{liu2024underwater}} &
\textbf{\tiny OUNet-JL \cite{wang2025optimized}} &
\textbf{\tiny AQUA-Net} \\[3pt]

\includegraphics[width=\detimgwidth,height=\detimgheight]{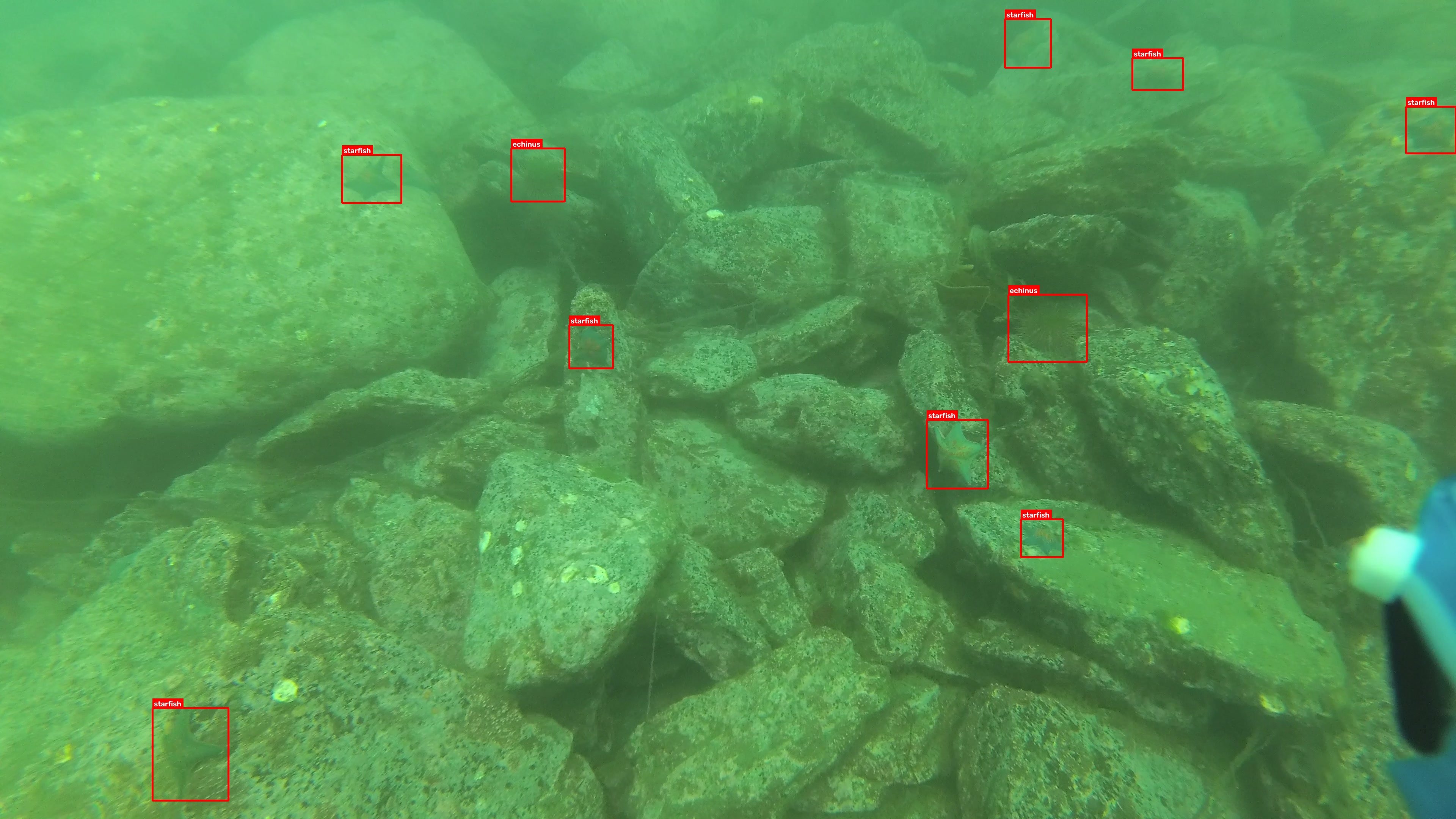} &
\includegraphics[width=\detimgwidth,height=\detimgheight]{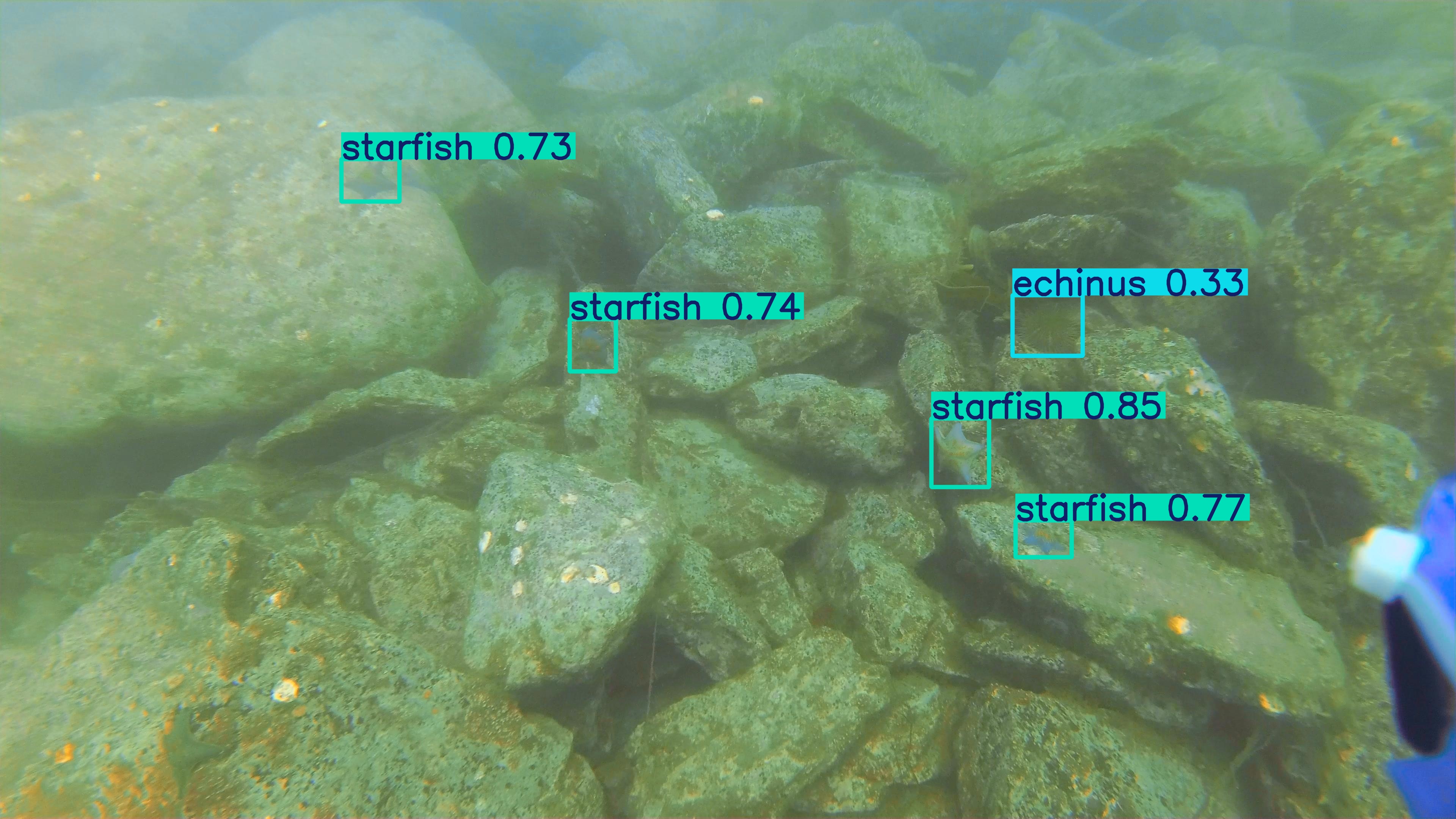} &
\includegraphics[width=\detimgwidth,height=\detimgheight]{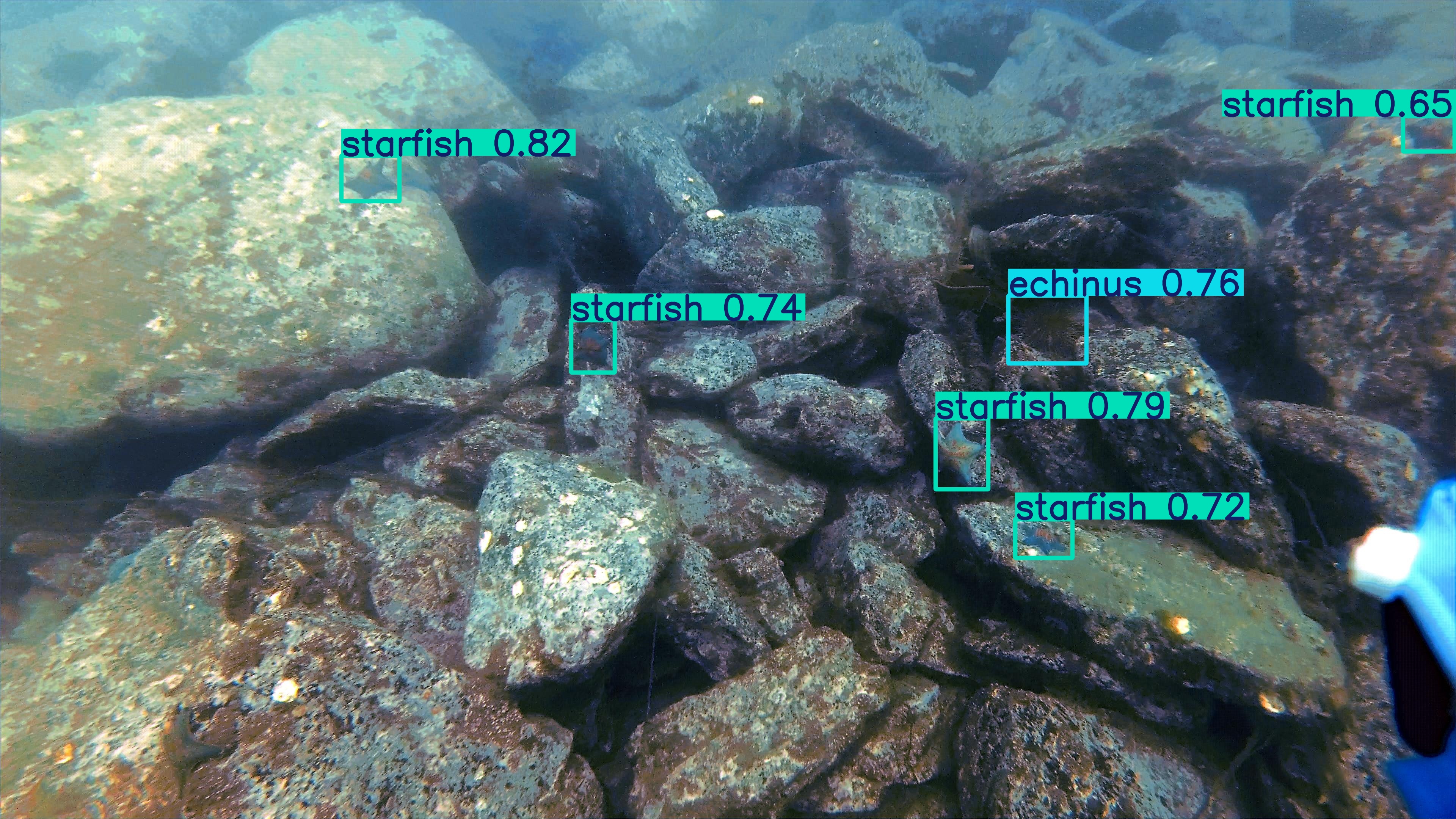} &
\includegraphics[width=\detimgwidth,height=\detimgheight]{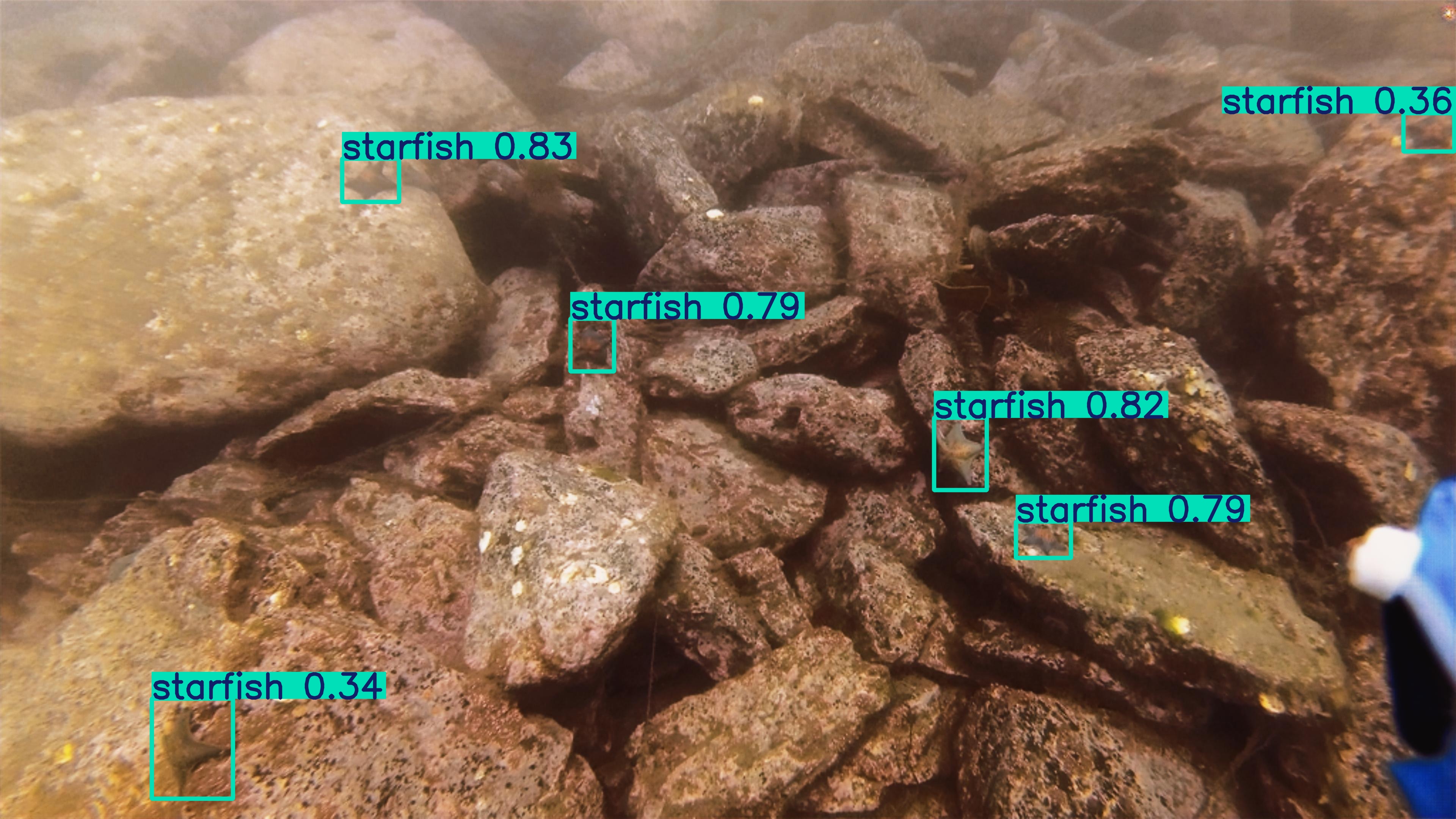} \\

\includegraphics[width=\detimgwidth,height=\detimgheight]{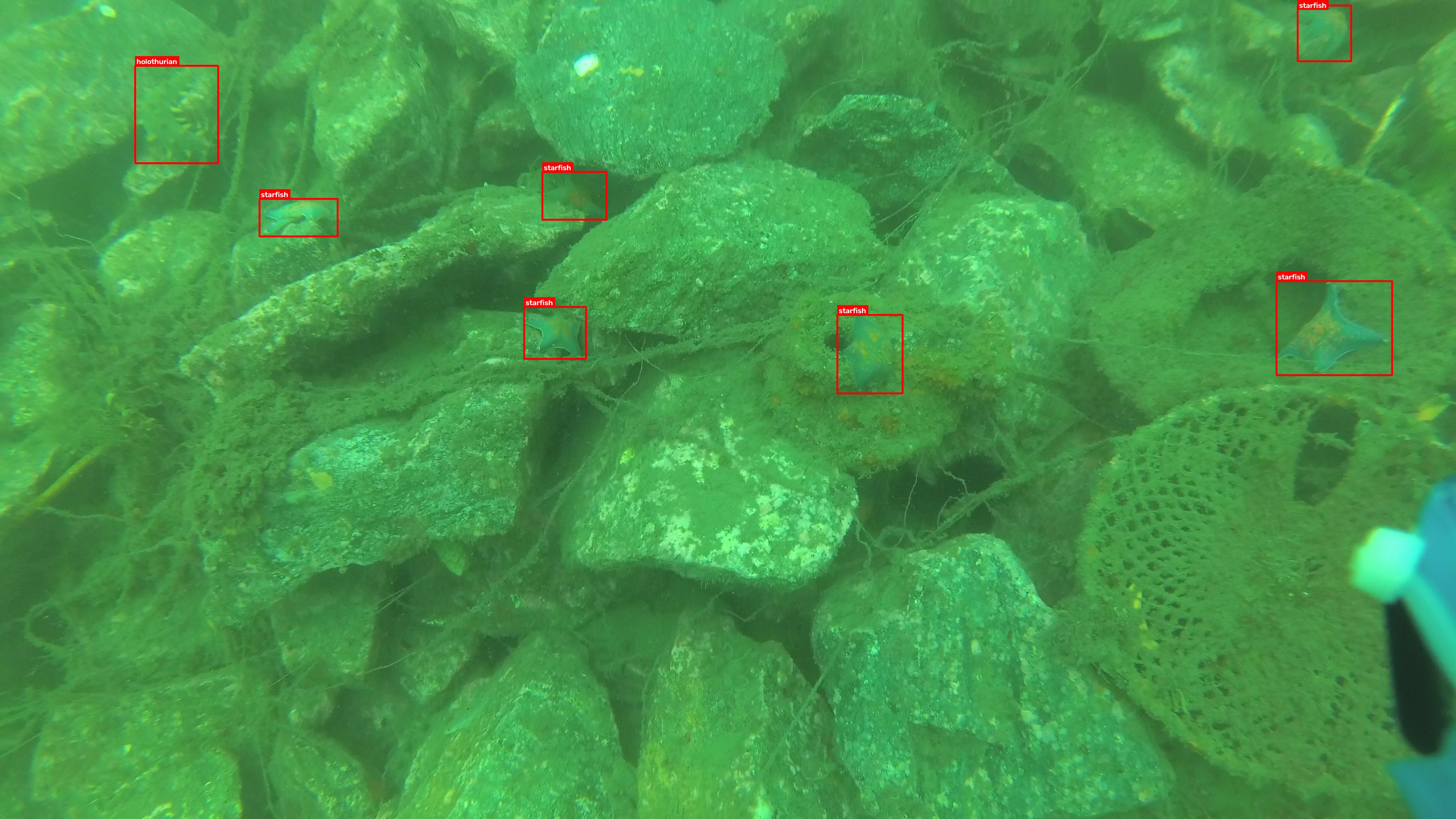} &
\includegraphics[width=\detimgwidth,height=\detimgheight]{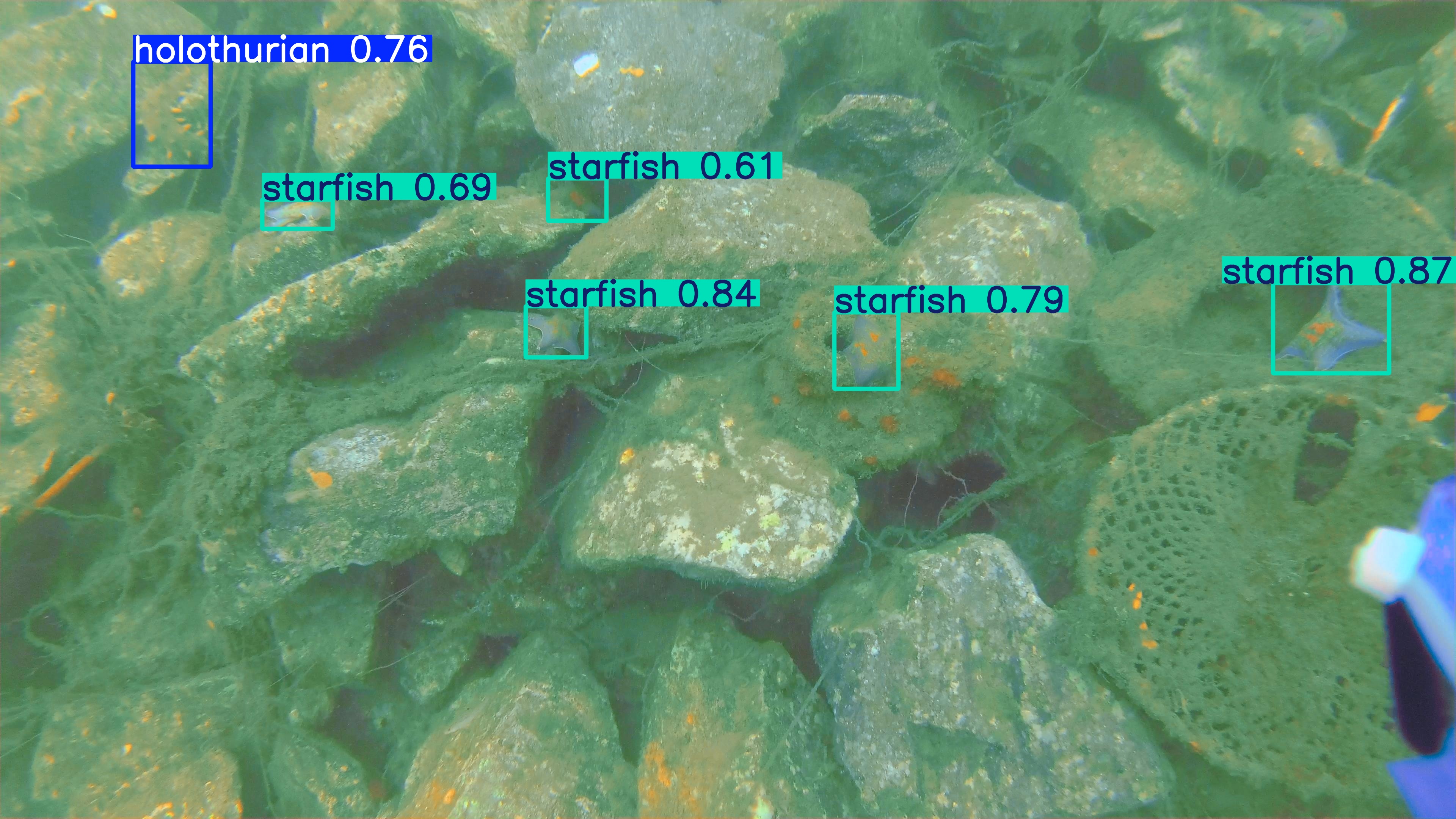} &
\includegraphics[width=\detimgwidth,height=\detimgheight]{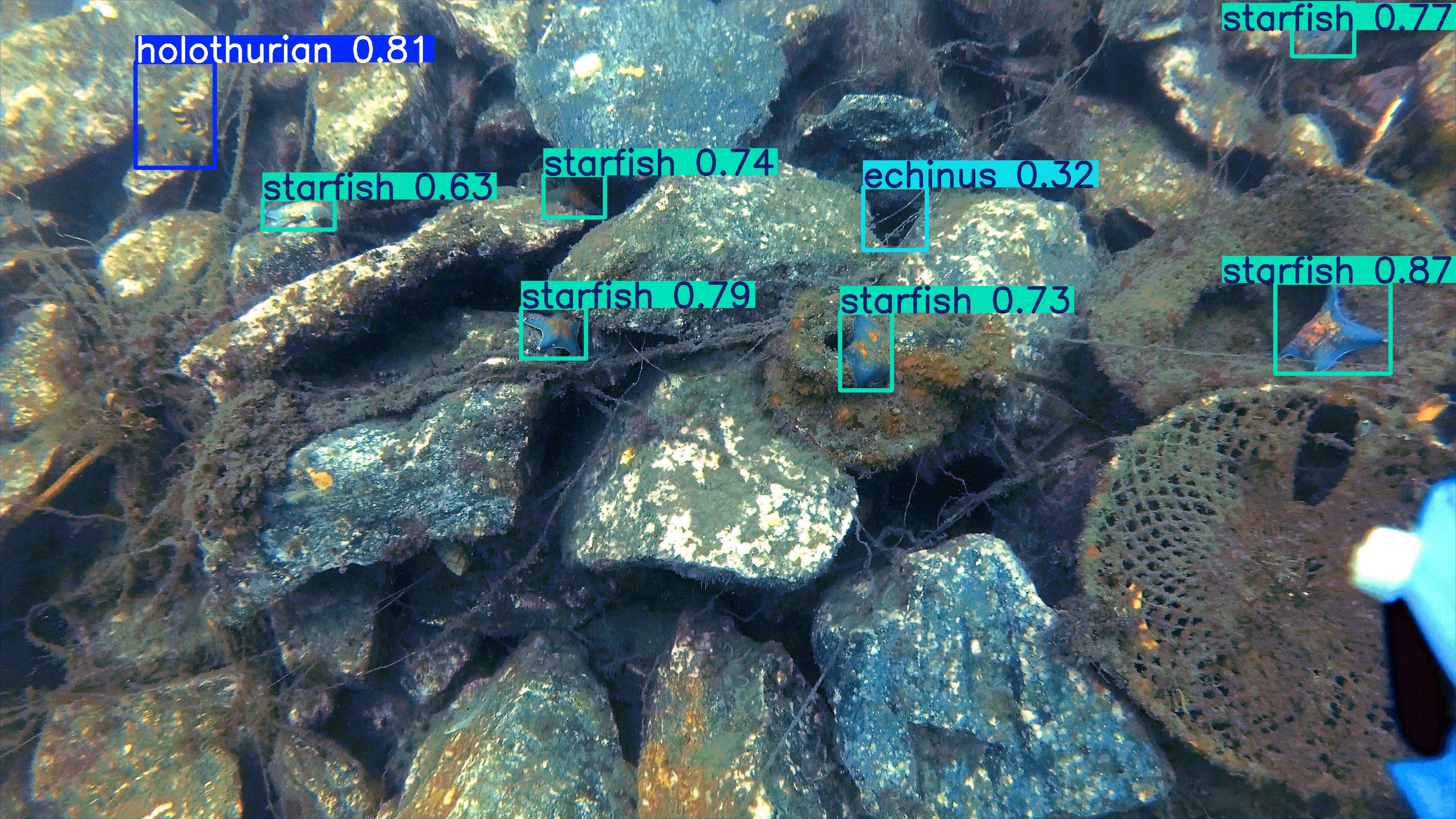} &
\includegraphics[width=\detimgwidth,height=\detimgheight]{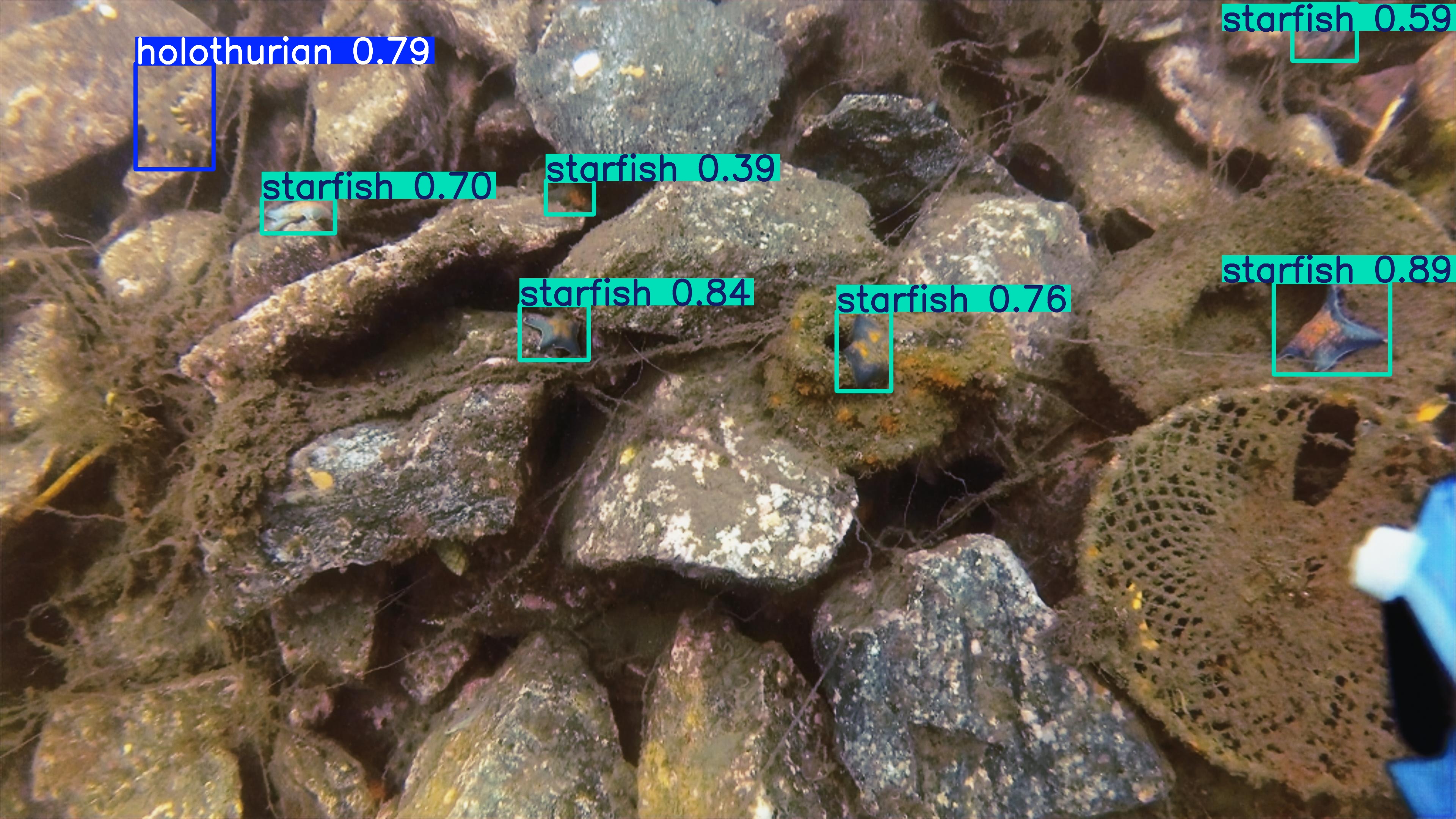} \\

\end{tabular}

\caption{Assessment of underwater object detection on the DUO dataset using different enhancement methods with the YOLO12s detection algorithm. The visual results show that AQUA-Net-enhanced images support reliable underwater object detection compared with alternative enhancement approaches.}
\label{fig:downstream_detection_visual}
\end{figure*}
As reported in Table~\ref{tab:downstream_detection_sota}, the enhancement methods show different effects on downstream object detection. CCL-Net obtains the highest mAP@0.5 and mAP@0.5:0.95, while AQUA-Net achieves the highest recall. OUNet-JL provides the highest precision but produces lower recall and lower mAP@0.5:0.95. These results indicate that different enhancement strategies influence detection behavior in different ways. CCL-Net provides stronger localization-oriented performance, whereas AQUA-Net helps YOLO12s detect more objects, leading to improved recall.

The qualitative examples in Fig.~\ref{fig:downstream_detection_visual} further support this observation. Compared with the original underwater images, the enhanced images improve object visibility and make target regions more distinguishable for the detector. CCL-Net produces clearer visual content than the degraded input, but some object regions remain weakly contrasted. OUNet-JL increases contrast strongly, which can improve object boundaries but may also introduce over-enhancement in some regions. AQUA-Net provides a more balanced enhancement effect by improving visibility while preserving natural object appearance. As a result, YOLO12s can detect multiple starfish and holothurian instances reliably in the AQUA-Net-enhanced images. These visual results are consistent with the quantitative findings, where AQUA-Net achieves the best recall among the compared methods.

\section{Conclusion and Future Work}
This paper presents AQUA-Net, a novel UIE model that integrates a hierarchical residual encoder–decoder with frequency-domain and illumination-aware branches. The proposed architecture effectively addresses key challenges in underwater imaging, including color distortion, low contrast, and scattering-induced visibility degradation. The frequency fusion branch enables the model to enhance low-frequency components, improve visibility, and refine structural details. Meanwhile, the illumination-aware branch performs adaptive color and illumination correction, which enhances visibility in UIWs. Extensive evaluations on multiple challenging datasets and also on the proposed dataset show that the proposed model achieves SOTA performance and maintains computational efficiency with fewer parameters. Ablation studies highlight the complementary contributions of the frequency and illumination branches. The results indicate the robustness and adaptability of AQUA-Net, making it suitable for practical applications in challenging underwater environments. Future studies may focus on further reduction of model complexity to enable deployment on low-power or embedded underwater devices.

\section*{Acknowledgments}
This work is supported by Stazione Zoologica Anton Dohrn, Naples, Italy.

\section*{Data Availability}
\textcolor{black}{
The datasets used in this study are publicly available. The UIEB dataset is available at \url{https://li-chongyi.github.io/proj_benchmark.html}; the LSUI dataset is available at \url{https://lintaopeng.github.io/_pages/UIE\%20Project\%20Page.html}; the EUVP dataset is described on the official University of Minnesota IRVLab webpage at \url{https://irvlab.cs.umn.edu/resources/euvp-dataset} and the RUIE dataset is available at \url{https://github.com/dlut-dimt/Realworld-Underwater-Image-Enhancement-RUIE-Benchmark}.
}
\section*{Author contributions}
M.A. and N.H. contributed to the conceptualization, methodology, model development, and experiments. M.A., N.H., L.V., D.D.B., and S.C. contributed to data analysis, validation, and review of the results. S.C. and J.S. supervised the work and contributed to manuscript review and editing. All authors reviewed and approved the final manuscript.
\section*{Competing Interests}
The authors declare no competing interests.
\bibliography{biblio}

\end{document}